%% file: main.tex
\documentclass[10pt, conference]{IEEEtran}

\usepackage{amsmath,amssymb,amsfonts}
\usepackage{graphicx}
\usepackage{pifont}

\usepackage{amsthm}
\usepackage{xcolor}
\usepackage[figuresright]{rotating}

\usepackage{graphicx}
\graphicspath{{/}{fig/}}

\usepackage{array}
\usepackage{textcomp}
\usepackage{xcolor}
\usepackage{multirow}
\usepackage{booktabs}

\usepackage{mathtools}
\usepackage{breqn}
\usepackage{float}

\usepackage{pgfplots}

\usepackage{caption}
\usepackage{subcaption}

\usepackage{multirow,tabularx}
\usepackage{hyperref}
\usepackage{flushend}
\usepackage{algorithmic}
\usepackage[vlined, ruled, shortend]{algorithm2e}

\usepackage{cleveref}

\newcommand{\yess}{\scriptsize{\ding{52}}}

\newlength\figureheight
\newlength\figurewidth
\SetAlCapNameFnt{\footnotesize}
\SetAlCapFnt{\footnotesize}

\title{
    \Huge
    Benchmarking UWB-Based Infrastructure-Free Positioning and Multi-Robot Relative Localization: Dataset and Characterization
}

\author{
    \IEEEauthorblockN{
        \vspace{1em}
        Paola Torrico Mor\'on\IEEEauthorrefmark{2},
        Sahar Salimpour\IEEEauthorrefmark{2},
        Lei Fu\IEEEauthorrefmark{2},
        Xianjia Yu\IEEEauthorrefmark{2},
        Jorge Pe\~na Queralta\IEEEauthorrefmark{2},
        Tomi Westerlund\IEEEauthorrefmark{2}
    }
    \IEEEauthorblockA{
        \normalsize
        \IEEEauthorrefmark{2}\href{https://tiers.utu.fi}{Turku Intelligent Embedded and Robotic Systems (TIERS) Lab, University of Turku, Finland}.\\
        Emails: \textsuperscript{1}\{pctomo, sahars, leifu, xianjia.yu, jopequ, tovewe\}@utu.fi\\[+6pt]
    }
}

\begin{document}

\maketitle
\thispagestyle{empty}
\pagestyle{empty}

\input{sec/00_Abstract.tex}
\IEEEpeerreviewmaketitle

\input{sec/01_Intro}

\input{sec/02_RelatedWorks}

\input{sec/03_ProblemDefinition}

\input{sec/04_Methodology}
\input{sec/05_Experiments}
\input{sec/06_Conclusion}


\section*{Acknowledgment}

This research work is supported by the Academy of Finland's AeroPolis project (Grant No. 348480) and by the R3Swarms project funded by the Secure Systems Research Center (SSRC), Technology Innovation Institute (TII).

\bibliographystyle{unsrt}
\bibliography{bibliography}

\end{document}

%% file: sec/00_Abstract.tex

\begin{abstract}%
    \label{sec:abstract}%
    Ultra-wideband (UWB) positioning has emerged as a low-cost and dependable localization solution for multiple use cases, from mobile robots to asset tracking within the Industrial IoT. The technology is mature and the scientific literature contains multiple datasets and methods for localization based on fixed UWB nodes. At the same time, research in UWB-based relative localization and infrastructure-free localization is gaining traction, further domains. tools and datasets in this domain are scarce. Therefore, we introduce in this paper a novel dataset for benchmarking infrastructure-free relative localization targeting the domain of multi-robot systems. Compared to previous datasets, we analyze the performance of different relative localization approaches for a much wider variety of scenarios with varying numbers of fixed and mobile nodes. A motion capture system provides ground truth data, are multi-modal and include inertial or odometry measurements for benchmarking sensor fusion methods. Additionally, the dataset contains measurements of ranging accuracy based on the relative orientation of antennas and a comprehensive set of measurements for ranging between a single pair of nodes. Our experimental analysis shows that high accuracy can be localization, but the variability of the ranging error is significant across different settings and setups. 
\end{abstract}

\begin{IEEEkeywords}

    UAV; GNSS; Ultra-wideband; UWB;
    VIO; Localization; MAV; UGV;
    Cooperative localization;
    Navigation;

\end{IEEEkeywords}

%% file: sec/01_Intro.tex

\section{Introduction}\label{sec:introduction}

Over the last years, autonomous robots have become more popular in society. As a result, the need for connectivity and networked collaborative systems has gained increasing importance \cite{hayat2016survey}. Their autonomous operations are usually supported by GNSS sensors such as GPS \cite{alsalam2017autonomous}. There are, however, several scenarios where robots need to operate in GNSS-denied environments. Warehouses \cite{plaksina2018development}, construction \cite{albeaino2021trends}, and mining, for example, require different approaches.

A popular approach to the localization problem in such scenarios is the use of visual sensors such as cameras. This is due to their low cost and flexibility \cite{lu2018survey}. A common solution includes visual-inertial odometry with either monocular cameras \cite{qin2018vins} or multiple sensors \cite{qin2019a}. Other solutions include onboard sensors such as IMUs\cite{qi2020cooperative} or odometry estimation from Lidars \cite{paneque2019multi}. Either of the presented solutions however has its limitations. Visual cameras are affected by the light conditions of the environment, and dust or smoke present affects visibility even more \cite{queralta2020uwb}. On the other hand, IMUs are affected by long-term drifts and LiDARs by the lack of clearly defined structures \cite{torrico2022towards}.

\begin{figure}[t]
    \centering
    \includegraphics[width=0.40\textwidth]{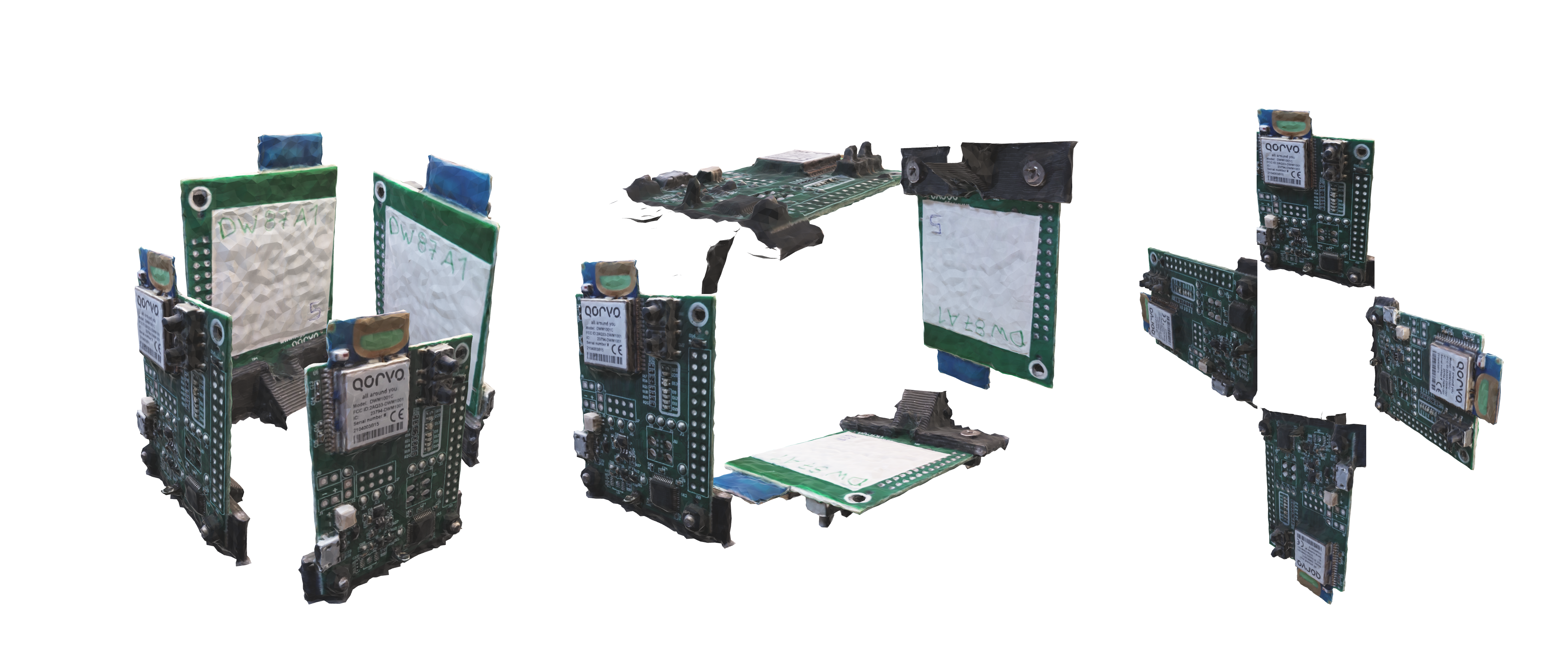}
    \caption{Different rotations for an UWB node. Yaw, pitch and roll plane of rotations respectively for UWB error characterization}
    \label{fig:rotations}
\vspace{-1.5em}
\end{figure}

\begin{figure}[t]
    \begin{subfigure}{0.23\textwidth}
        \centering
        \includegraphics[width=0.95\textwidth]{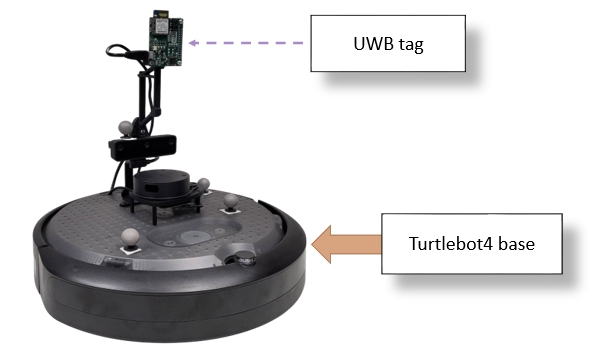}
        \caption{Turtlebot4}
        \label{fig:tb4}
    \end{subfigure}
    \hfill
    \begin{subfigure}{0.26\textwidth}
        \centering
        \includegraphics[width=0.95\textwidth]{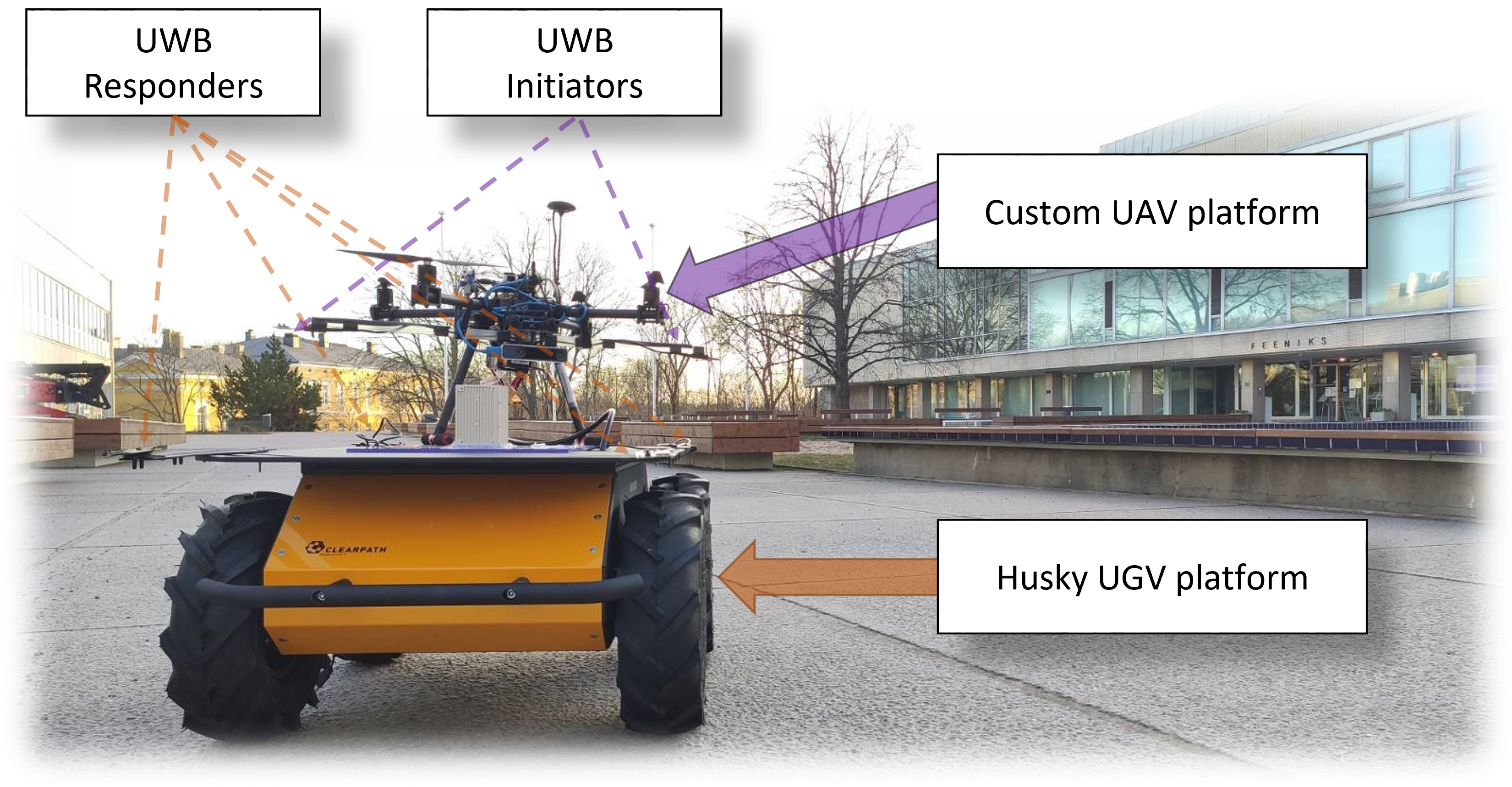}
        \caption{Husky + Holybro}
        \label{fig:husky}     
    \end{subfigure}
    \caption{Robots used for the data collection for the dataset. \Cref{fig:tb4} shows a Turtlebot4 mounted with a UWB module. \Cref{fig:husky} shows a Husky with a landing platform and a custom UAV equipped with two UWBs.}
    \label{fig:robots} 
\vspace{-2em}
\end{figure}

As a solution to overcome the challenges with visual-based localization caused by environmental conditions, ultra-wideband (UWB) technology has emerged as a promising alternative for robot localization \cite{zafari2019survey}. In recent years, there has been a significant increase in research on UWB-based relative localization in GNSS-denied environments, which aims to eliminate the requirements for stationary anchors \cite{chen2022survey}. It has been particularly useful in multi-robot systems to achieve accurate and robust position estimation without the need for external infrastructure. We will be presenting some of the existing datasets for UWB localization as background for the construction of our dataset. 

In this work, we will present the different cases of UWB data collection for error characterization and localization. We present the details of data acquisition in each case as well as some analysis of the results, both for the errors as well as the localization solutions implemented. The dataset will present different scenarios, focusing on infrastructure-free positioning and multi-robot relative localization, with the data acquired capable of being implemented in multi-robot solutions. We also present different spatial configurations.

The rest of the document is organized as follows. Section II reviews the background of UWB localization and presents existing datasets with UWB localization. Section III presents different relative localization methods. Section IV presents the methodology used for data acquisition and the subsets present in the dataset. Section V presents an analysis of the dataset, with error characterization, positioning error, and localization algorithm comparison. Finally, section VI presents the conclusions.

%% file: sec/02_RelatedWorks.tex

\section{Background} \label{sec:related_work}

\begin{table*}
    \centering
    \caption{\footnotesize{Comparison table of existing UWB localization datasets and ours. For each one we include the number of UWB nodes, if they are fixed of mobile, if the data is used for absolute or relative localization, the number of subsets given, if the UWB are inside or outside the convex envelope, the type of UWB utilized and finally if the also present other sensor information.}}
    \label{tab:datasets_comparison}
    \footnotesize	
    \begin{tabular}{@{}lccccccccc@{}}
        \toprule
                                                &\#Nodes    &\#Fixed    &\#Mobile   &\#Absolute &\#Relative &Subsets    &Convex     &UWB        &Extra\\
                                                &           &Nodes      &Nodes      &Loc.       &Loc.       &           &Envelope   &Node       &Sensors\\
        \midrule
        Bregar~\cite{bregar2018improving}(2018) &5          &4          &1          &\yess      &-          &2          &In         &DWM1000    &\\
        Li~\cite{li2018accurate}(2018)          &7          &6          &1          &\yess      &-          &1          &In         &PulsON P440&IMU\\
        Barral~\cite{barral2019multi}(2019)     &8          &7          &1          &\yess      &-          &2          &In         &Pozyx      &\\
        Raza~\cite{raza2019dataset}(2019)       &6          &4          &3          &\yess      &-          &2          &In         &DWM1001    &Bluetooth\\
        Pe\~na~\cite{queralta2020uwb}(2020)     &3-6        &3-6        &1-4        &\yess      &\yess      &12         &In/Out     &DWM1001    &\\
        Delamare~\cite{delamare2020new}(2020)   &10         &4          &6          &\yess      &-          &2          &In         &DWM1001    &\\
        Vey~\cite{vey2022indoor}(2022)          &7          &6          &1          &-          &           &1          &In         &DWM1001    &\\
        \textbf{Ours}                           &2-8        &2-8        &0-5        &\yess      &\yess      &24         &In/Out     &DWM1001    &VIO\\

        \bottomrule
    \end{tabular}
\vspace{-1.5em}
\end{table*}

Through this section, we discuss the existing works on UWB localization datasets and methods. A detailed comparison of our dataset with others is shown in~\Cref{tab:datasets_comparison}.

Bregar et al. introduced in~\cite{bregar2018improving} a method to reduce the localization error for non-line-of-sight (NLOS) conditions by utilizing convolutional neural networks (CNNs) or ranging error regression models and analyzing the Channel Impulse Response (CIR) of the UWB in unknown environments. They utilize DWM1000 modules using four as anchors and one as a tag while implementing Time of Flight (ToF) ranging. Bregar et al. build a dataset of measurements from the tag to each anchor and the known distances for training the CNN.

Li et al. employed in~\cite{li2018accurate} an Extended Kalman Filter (EKF) to fuse UWB and IMU data. Based on known distances and angles, the author implements six fixed UWB anchors using PulsON P440. Using Time of Arrival (TOA) a micro aerial vehicle (MAV) equipped with a UWB was able to autonomously track a path in the environemnt. 

Barral et al. in~\cite{barral2019multi} present the dataset to study the feasibility of utilizing UWB technology in an industrial setting for tracking forklift trucks. The authors implement DWM1000 UWB devices to record ranging measurements for their UWB gazebo simulator. Measures were taken at different distances in three different scenarios with line-of-sight (LOS) and NLOS. Their simulator then uses fixed anchors and one mobile tag implementing a Kalman Filter approach for localization.

Raza et al. presents in~\cite{raza2019dataset} a dataset for intelligent calibration of different localization technologies such as UWB, Bluetooth, and MOCAP from initial calibration. The group uses four anchors and one mobile tag with the DWM1001 modules while using Time Difference of Arrival (TDoA) for the ranging measurements.

Pe\~na~\cite{queralta2020uwb} et al. present a UWB dataset with twelve subsets of data. They implement three, four, or six anchors for robust localization. Four of the scenarios provided have the UWB anchors in the corner of their testing arena, while the others present different height, number, or placement configurations. They have a moving tag mounted in a quadrotor, moved manually around the testing area, and implement ToF ranging. They utilize DWM1001 UWB modules for all their data acquisition, for studying the localization accuracy with self-calibrating anchor positions.

Delamare~\cite{delamare2020new} et al. provide a dataset for the flow of movement in an industrial site with UWB, however, it is done with human workers instead of robots. They provide the movements of an entire assembly line with the use of four anchors with known locations and six moving nodes, one per person. Implementing DWM1001 UWB nodes, and tracking movements with the Two Way Ranging (TWR) algorithm provided by Qorvo. Their UWB localization is validated by obtaining ground truth positions using a MOCAP system.

Vey~\cite{vey2022indoor} et al. introduce a UWB dataset acquired in a UWB Based Rangin testbed for the Internet of Things. They present a scenario with six anchors also in known locations for LOS and NLOS scenarios. They implement an autonomously moving tag and implement TWR ranging. The UWB ranging and ranging errors are presented in the dataset for the enhancement of localization algorithms and protocols.












%% file: sec/03_ProblemDefinition.tex

\setlength{\abovedisplayskip}{3pt}
\setlength{\belowdisplayskip}{3pt}

\section{Relative Localization Methods}

In this section, we introduce different methods for relative localization between a series of $N$ UWB nodes without known positions, using multilateration, least squares, and gradient descent techniques.

\subsection{Multilateration}

Multilateration is a technique that estimates the position of a node based on the distances or difference of distances between multiple nodes \cite{gao2009particle}. In the case of UWB-based positioning with $N$ nodes, we can use ToF ranging for distances or TDoA for the difference of distances. Given the distances, the position of an unknown node can be determined by solving a system of equations:

\begin{equation}
    \lVert \boldsymbol{p}_i - \boldsymbol{p}_j \rVert = d_{ij}, \quad i,j = 1, 2, \dots, N
\end{equation}

where $p_i$ and $p_j$ are the positions of nodes $i$ and $j$, respectively, and $d_{ij}$ is the distance between them.

\subsection{Least Squares}

Least squares is a popular optimization technique that can be used for relative localization \cite{bregar2018improving}. Given a set of distances, we can formulate the localization problem as a least squares optimization problem:

\begin{equation}
\min_{\boldsymbol{p}_1, \boldsymbol{p}2, \dots, \boldsymbol{p}n} \displaystyle\sum_{i=1}^{n}\displaystyle\sum_{j=i+1}^{n} \left(||\boldsymbol{p}_i - \boldsymbol{p}j|| - d_{ij}\right)^2
\end{equation}

Solving this optimization problem yields the positions of the nodes that minimize the total squared error between the measured and estimated distances.

\subsection{Gradient Descent}

Gradient descent is an iterative optimization algorithm that can be employed for relative localization~\cite{ridolfi2021uwb}. In this approach, we start with an initial guess for the node positions and iteratively update the positions based on the gradient of the objective function:

\begin{equation}
\boldsymbol{p}_i^{(k+1)} = \boldsymbol{p}i^{(k)} - \alpha \nabla{\boldsymbol{p}_i} J(\boldsymbol{p}_1, \boldsymbol{p}_2, \dots, \boldsymbol{p}_n)
\end{equation}

where $\boldsymbol{p}_i^{(k)}$ is the position of node $i$ at iteration $k$, $\alpha$ is the learning rate, and $J(p_1, p_2, \dots, p_n)$ is the objective function defined in the least squares formulation.

By gradually updating the positions and moving towards the minimum of the error function, gradient descent converges to the final accurate and robust positioning that minimizes the error function.

%% file: sec/04_Methodology.tex

\section{Methodology}

This section describes the data included in the dataset, and the steps followed for acquiring the data in the different scenarios. All the data and code used for recording it has been made public in Github~\footnote{\url{https://github.com/TIERS/uwb-relative-localization-dataset}}.

\subsection{Data Acquisition}

The dataset presented in this work consists of five different cases. All of them utilize Qorvo’s DWM1001 UWB modules with custom firmware for ToF ranging. Depending on the case, the UWB modules are mounted either in Turtlebot4, a custom platform for the Husky robot or custom UAV shown in \Cref{fig:robots}, or in a known location in the testing arena. All measurements are obtained from the DWM1001 device by UART connection and interfaced with ROS2 nodes for processing and recording. The ground truth for measurements is provided by a MOCAP system of eight Optitrack cameras in a testing arena of 8$\,m\times9\,m$ except for Case I, where the area was expanded to 8$\,m\times16\,m$.

Each case consists of one or more subsets of data, which are going to be further detailed in the next section. The data acquisition for each case is obtained as follows. \textit{Case I:} is the UWB raw measurements for two UWB devices one set as initiator and one as responder. Measurements were taken for distances between 1$\,m$ to 16$\,m$ with steps of 1$\,m$, and for each of the different rotations in \Cref{fig:rotations}. A hundred measurements are taken for each distance and rotation combination. For this case, one UWB is static while the second is moved manually both for the distance and the rotations.

\textit{Case II:} four UWB modules are fixed in known locations, either in the arena corners or on the platform for the Husky robot. Two UWB modules located in the custom UAV are moved around the arena in different predefined trajectories. The UAV’s modules act as initiators while the other four nodes act as responders. The communication frequency in this case is 38Hz. \textit{Case III:} consists of eight UWB devices mounted in the Turtlebot4 and located in different predefined configurations. They communicate all-to-all, with all taking turns in being initiators and responders. \textit{Case IV:} consists on the datasets utilized for our previous work in \cite{salimpour2023exploiting}. This scenario also utilizes the all-to-all communication with the UWB on the Turtlebots, with two static Turtlebots and five moving ones. Finally, \textit{Case V:} has five Turtlebots equipped with one UWB module and a VIO camera. This case comes from \cite{yu2023loosely} and has one static node and four moving ones. For all the cases implemented in the Turtlebots, the frequency is 15Hz.

\subsection{Data Subsets}

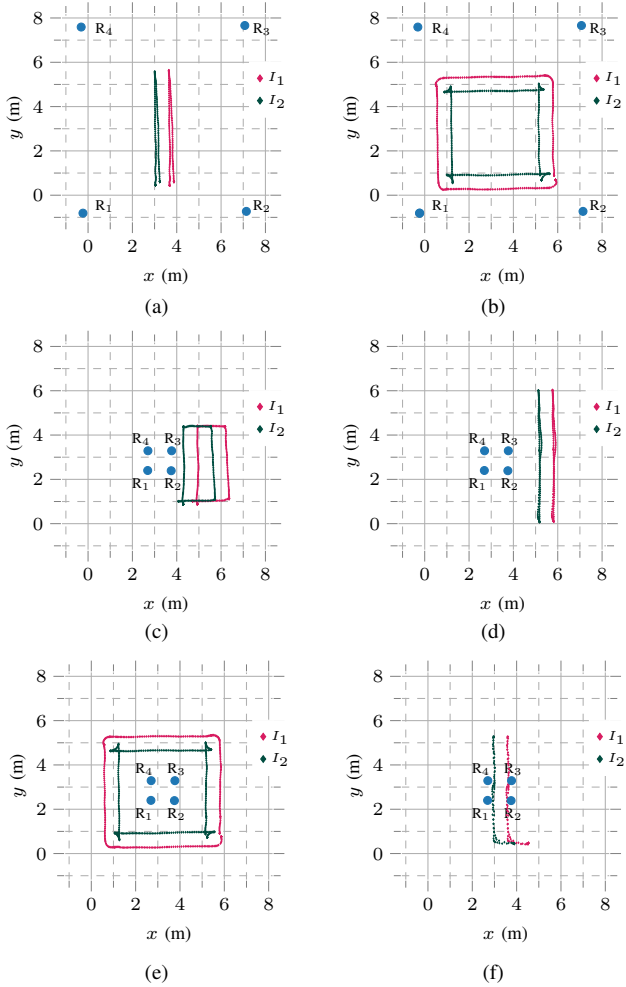
\begin{figure}

    \centering
    \begin{subfigure}[t]{0.24\textwidth}
        \centering
        \setlength\figureheight{\textwidth}
        \setlength\figurewidth{\textwidth}
        \scriptsize{\input{tex/data_characterization/inside_opti_pose_line_2023-03-24-19-02-56.bag}}
        \vspace{-.42em}
        \caption{}
        \label{fig:in_lines}
    \end{subfigure}
    \begin{subfigure}[t]{0.24\textwidth}
        \centering
        \setlength\figureheight{\textwidth}
        \setlength\figurewidth{\textwidth}     
        \scriptsize{\input{tex/data_characterization/inside_opti_pose_square_rotation_2023-03-24-19-08-41.bag}}
        \vspace{-.42em}
        \caption{}
        \label{fig:in_square}
    \end{subfigure}

    \vspace{.42em}
    
    \begin{subfigure}[t]{0.24\textwidth}
        \centering
        \setlength\figureheight{\textwidth}
        \setlength\figurewidth{\textwidth} 
        \scriptsize{\input{tex/data_characterization/outside_opti_pose_outside_square_no_rotation_2023-03-24-18-08-05.bag}}
        \vspace{-.6em}
        \caption{}
        \label{fig:out_square}
    \end{subfigure}
    \begin{subfigure}[t]{0.24\textwidth}
        \centering
        \setlength\figureheight{\textwidth}
        \setlength\figurewidth{\textwidth}
        \scriptsize{\input{tex/data_characterization/outside_opti_pose_perpendicular_line_2023-03-24-17-46-14.bag}}
        \vspace{-.6em}
        \caption{}
        \label{fig:out_lines}
    \end{subfigure}

    \vspace{.6em}
    
    \begin{subfigure}[t]{0.24\textwidth}
        \centering
        \setlength\figureheight{\textwidth}
        \setlength\figurewidth{\textwidth}
        \scriptsize{\input{tex/data_characterization/outside_opti_pose_square_rotation_2023-03-24-18-00-34.bag}}
        \caption{}
        \label{fig:out_square_around}
    \end{subfigure}
    \begin{subfigure}[t]{0.24\textwidth}
        \centering
        \setlength\figureheight{\textwidth}
        \setlength\figurewidth{\textwidth} 
        \scriptsize{\input{tex/data_characterization/outside_opti_pose_over_perpendicular_2023-03-24-18-16-39.bag}}
        \caption{}
        \label{fig:in_out}
    \end{subfigure}  

    \caption{A a subset of the different scenarios for Case II. It presents both the scenarios where the initiators move inside the convex envelope, \Cref{fig:in_lines} and \Cref{fig:in_square}, and outside the envelope, \Cref{fig:out_lines}, \Cref{fig:out_square} and \Cref{fig:out_square_around}.}
    \label{fig:experiment_results}

\vspace{-1.5em}
    
\end{figure}

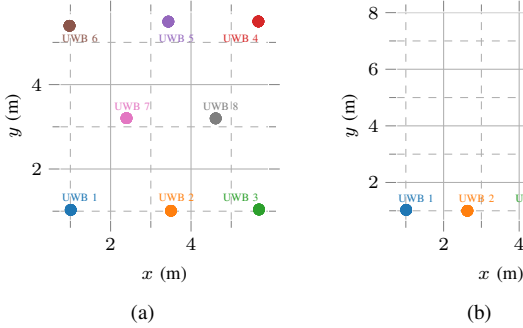
\begin{figure}

    \centering
    \begin{subfigure}[t]{0.24\textwidth}
        \centering
        \setlength\figureheight{\textwidth}
        \setlength\figurewidth{\textwidth}\scriptsize{\input{tex/data_characterization/all_to_all_opti_pose_army_2023-03-24-20-09-29_v2.bag}}
        \caption{}
        \label{fig:army}
    \end{subfigure}
    \begin{subfigure}[t]{0.24\textwidth}
        \centering
        \setlength\figureheight{\textwidth}
        \setlength\figurewidth{\textwidth}        \scriptsize{\input{tex/data_characterization/all_to_all_opti_pose_corner_2023-03-24-20-18-25_v2.bag}}
        \caption{}
        \label{fig:corner}
    \end{subfigure}
    
    \caption{Different Case III configurations; eight nodes all-to-all communication.}
    \label{fig:configurations}

\vspace{-1.5em}
    
\end{figure}

Each of the five cases mentioned before consists of one or more subsets of data. \textit{Case I:} has only one subset with the twelve rotations for each of the distances from 1\,m$ to 16\,m$. \text{Case II:} consists of fourteen subsets of data, each with a different responder localization or the initiator trajectory. The initiators in the UAV can be moving either inside the convex envelope created by the four responders or outside it. This will impact the localization implemented for each scenario, with a worse performance when outside the envelope. Some of the scenarios presented in this case can be appreciated in \Cref{fig:experiment_results}.

\textit{Case III}: has a total of three different configurations for the eight Turtlebot robots, with two being shown in \Cref{fig:configurations}. The configurations were chosen because they represent some edge cases for localization and are challenging to solve. In this paper, we present two methods using the solutions from \cite{salimpour2023exploiting}. \textit{Case IV:} consists of four subsets, with the robots moving in the same pattern for each one, and finally \textit{Case V:} has two subsets with the Turtlebots also moving in the same patterns, but this time also with VIO data. A summary table of all the different scenarios for the dataset for each of the cases can be seen in \Cref{tab:dataset_summary}

\begin{table}[b]
    \centering
    \vspace{.42em}
    \caption{\footnotesize{Description of the Data Subsets for each of the cases presented.}} 
    \label{tab:dataset_summary}
    \footnotesize	
    \resizebox{0.5\textwidth}{!}{
    \begin{tabular}{@{}lcccccccc@{}}
        \toprule
         & \#UWB & \#Fixed & \#Mobile & \#Resp & \#Init & \#Freq. & \#Sub. &  \\
         &  Nodes & Nodes & Nodes &Nodes &Nodes &[Hz] &  \\
        \midrule
        Case I  & 2       & 1 & 1 & 1 & 1 & 45        & 1                    \\
        Case II  & 6       & 4 & 2 & 4 & 2 & 38        & 14                   \\
        Case III  & 8       & 8 & 0 & 8 & 8 & 15        & 3                   \\
        Case IV  & 7       & 2 & 5 & 7 & 7 & 15        & 4                   \\
        Case V  & 5       & 1 & 4 & 5 & 5 & 7        & 2                  \\
        \bottomrule
    \end{tabular}}
\vspace{-1.5em}
\end{table}

%% file: tex/data_characterization/inside_opti_pose_line_2023-03-24-19-02-56.bag.tex
\begin{tikzpicture}

\definecolor{crimson2143940}{RGB}{214,39,40}
\definecolor{darkgray176}{RGB}{176,176,176}
\definecolor{darkorange25512714}{RGB}{255,127,14}
\definecolor{forestgreen4416044}{RGB}{44,160,44}
\definecolor{lightgray204}{RGB}{204,204,204}
\definecolor{darkorange25512714}{RGB}{216, 27, 96}
\definecolor{forestgreen4416044}{RGB}{0, 77, 64}
\definecolor{steelblue31119180}{RGB}{31,119,180}

\begin{axis}[
    height=\figureheight,
    width=\figurewidth,
    axis background/.style={fill=white},
    axis line style={white},
    legend cell align={left},
    legend style={
      fill opacity=1,
      draw opacity=1,
      text opacity=1,
      at={(0.9,0.5)},
      anchor=south west,
      draw=white,
      legend columns=1,
      font=\tiny
    },
    legend image post style={scale=5.5},
    tick align=outside,
    x grid style={white!69.0196078431373!black},
    xlabel={\(\displaystyle x\) (m)},
    minor tick num = 1,
    minor grid style={dashed},
    xmajorgrids,
    xmajorgrids,
    y grid style={white!69.0196078431373!black},
    ylabel={\(\displaystyle y\) (m)}, 
    xminorgrids,
    xminorgrids=true,
    ymajorgrids,
    ymajorticks=true,
    yminorgrids,
    yminorgrids=true,
    xmin=-1.2, xmax=8.2,
    ymin=-1.2, ymax=8.2,
    ylabel near ticks, 
    ylabel shift={-1pt},
]
\addplot [draw=darkorange25512714, fill=darkorange25512714, mark=diamond*, only marks, mark size=0.23pt]
table{%
x  y
3.67971897125244 0.435290962457657
3.67947959899902 0.435153514146805
3.67961573600769 0.435351520776749
3.67964458465576 0.434741973876953
3.67993235588074 0.43583869934082
3.67998695373535 0.435786992311478
3.67966914176941 0.435692340135574
3.67984008789062 0.436200141906738
3.67961192131042 0.436281353235245
3.67906308174133 0.446624904870987
3.67557644844055 0.463252753019333
3.66996836662292 0.493506193161011
3.675616979599 0.534788429737091
3.66701006889343 0.584239959716797
3.67886161804199 0.618399322032928
3.69162368774414 0.661783993244171
3.69635391235352 0.718053042888641
3.69596767425537 0.757090985774994
3.6970100402832 0.7853963971138
3.69170641899109 0.833718061447144
3.68162083625793 0.902310192584991
3.67668342590332 0.976717710494995
3.6725161075592 1.06228697299957
3.67096424102783 1.14653372764587
3.67468953132629 1.23570430278778
3.68066430091858 1.31717669963837
3.68647646903992 1.38800489902496
3.69105052947998 1.45090615749359
3.69570231437683 1.51843881607056
3.69389081001282 1.58642721176147
3.69499111175537 1.65585851669312
3.69722557067871 1.73161423206329
3.70018291473389 1.82136285305023
3.70081305503845 1.90846920013428
3.69908738136292 1.98643445968628
3.69756603240967 2.06308102607727
3.69539546966553 2.13347482681274
3.6932361125946 2.20581936836243
3.69150948524475 2.28504538536072
3.69061279296875 2.37010216712952
3.68629908561707 2.4502124786377
3.68106293678284 2.52807021141052
3.67461657524109 2.62164354324341
3.67083048820496 2.71529769897461
3.66926574707031 2.80182862281799
3.66594696044922 2.89673924446106
3.66254591941833 2.99117875099182
3.66045689582825 3.0796422958374
3.65706372261047 3.16373896598816
3.65468120574951 3.25218319892883
3.65473079681396 3.34434413909912
3.65716195106506 3.42943477630615
3.66077589988708 3.51528763771057
3.6608943939209 3.59028077125549
3.66328048706055 3.66637802124023
3.66425776481628 3.75053024291992
3.66180801391602 3.84207797050476
3.65889477729797 3.92815852165222
3.6580913066864 4.03045415878296
3.65666794776917 4.13564491271973
3.65675759315491 4.22897577285767
3.65709614753723 4.31574296951294
3.65846824645996 4.40834093093872
3.65723967552185 4.50175666809082
3.65539073944092 4.59249925613403
3.65398406982422 4.68507242202759
3.65474247932434 4.77691841125488
3.65201115608215 4.85484457015991
3.65135264396667 4.93460893630981
3.65110301971436 5.01441049575806
3.65170121192932 5.09467649459839
3.653000831604 5.17179298400879
3.65511512756348 5.25659990310669
3.65574741363525 5.33898973464966
3.65523242950439 5.42185926437378
3.65184593200684 5.49361801147461
3.65035605430603 5.55639982223511
3.64782333374023 5.60596704483032
3.64559864997864 5.63920545578003
3.64324116706848 5.63158226013184
3.64395976066589 5.63012838363647
3.64336013793945 5.61638307571411
3.64426493644714 5.59771776199341
3.64535975456238 5.57221412658691
3.64710307121277 5.54084062576294
3.64867758750916 5.50500631332397
3.65171360969543 5.45544004440308
3.65533065795898 5.40764427185059
3.66283392906189 5.35147285461426
3.67058372497559 5.29733657836914
3.67684173583984 5.26496124267578
3.68250846862793 5.21787214279175
3.69454193115234 5.16322135925293
3.69799447059631 5.10735082626343
3.70443272590637 5.04746055603027
3.70596241950989 4.97571897506714
3.70930218696594 4.89479684829712
3.71142506599426 4.80827951431274
3.71592211723328 4.72921943664551
3.71946144104004 4.666015625
3.72431063652039 4.59484815597534
3.72602391242981 4.51630258560181
3.73359894752502 4.43994474411011
3.74233531951904 4.35589790344238
3.74519753456116 4.27800273895264
3.74666500091553 4.19508409500122
3.74841904640198 4.10482835769653
3.75169539451599 4.01804113388062
3.75765633583069 3.92388939857483
3.76167750358582 3.83646845817566
3.76485180854797 3.75594401359558
3.76538348197937 3.68265891075134
3.76991677284241 3.60348510742188
3.77620387077332 3.52143549919128
3.78186869621277 3.44943165779114
3.78591704368591 3.36348271369934
3.79313063621521 3.26530146598816
3.8006956577301 3.16848874092102
3.80794143676758 3.07815861701965
3.80634140968323 2.99465584754944
3.805997133255 2.90479922294617
3.80525517463684 2.82231259346008
3.80826759338379 2.74794960021973
3.81051087379456 2.65024662017822
3.81360173225403 2.54322743415833
3.81508755683899 2.45483899116516
3.81286120414734 2.37729239463806
3.81547737121582 2.29682540893555
3.81454491615295 2.22225737571716
3.81478309631348 2.15974116325378
3.81195259094238 2.0910484790802
3.81327271461487 2.00433039665222
3.81158351898193 1.92459106445312
3.8094208240509 1.85859286785126
3.8090546131134 1.79545021057129
3.81142234802246 1.73114490509033
3.81295228004456 1.66596233844757
3.82170653343201 1.58632385730743
3.82702851295471 1.50970733165741
3.83147835731506 1.43744897842407
3.83443760871887 1.36291611194611
3.83545470237732 1.29084813594818
3.83512687683105 1.22444450855255
3.83904337882996 1.15371930599213
3.84642624855042 1.07533013820648
3.85384750366211 0.981216609477997
3.8572142124176 0.896789789199829
3.86282992362976 0.824152648448944
3.8610200881958 0.755477726459503
3.86580872535706 0.685403048992157
3.87246441841125 0.63545435667038
3.87314414978027 0.605923235416412
3.87543797492981 0.603141307830811
3.87512040138245 0.60616534948349
3.87528014183044 0.606956303119659
3.87525820732117 0.606831848621368
3.87516570091248 0.60663914680481
3.87536406517029 0.60694432258606
3.8752338886261 0.606453657150269
3.87528681755066 0.606410264968872
3.87527060508728 0.606534659862518
3.87560248374939 0.606436312198639
3.87524151802063 0.606275260448456
3.87527847290039 0.606278479099274
3.87537384033203 0.606137454509735
};
\addlegendentry{$I_1$}
\addplot [draw=forestgreen4416044, fill=forestgreen4416044, mark=diamond*, only marks, mark size=0.23pt]
table{%
x  y
3.04251050949097 0.450169086456299
3.04237103462219 0.449911087751389
3.04246211051941 0.450169712305069
3.04161167144775 0.449426263570786
3.04269886016846 0.450554728507996
3.04280066490173 0.450450956821442
3.04244756698608 0.450330764055252
3.04262661933899 0.450822174549103
3.04241299629211 0.450809597969055
3.04218721389771 0.460061520338058
3.03822636604309 0.474835515022278
3.0314724445343 0.496105372905731
3.03774189949036 0.537903904914856
3.02905535697937 0.577214300632477
3.04122018814087 0.602325737476349
3.05381846427917 0.65644359588623
3.05768823623657 0.716843783855438
3.05830073356628 0.774851560592651
3.05864834785461 0.804778456687927
3.05445551872253 0.847842335700989
3.04432535171509 0.904611468315125
3.03921866416931 0.972652733325958
3.03503799438477 1.0582515001297
3.03321409225464 1.14208817481995
3.03716015815735 1.22456395626068
3.04288744926453 1.30162858963013
3.04927206039429 1.36809980869293
3.05372595787048 1.43277168273926
3.05838084220886 1.50174844264984
3.05690813064575 1.56305944919586
3.05795097351074 1.63713145256042
3.05981636047363 1.72104609012604
3.06296038627625 1.81739723682404
3.06357431411743 1.89950525760651
3.06155920028687 1.97676980495453
3.06039214134216 2.05687952041626
3.05777406692505 2.13112664222717
3.05584979057312 2.20443272590637
3.05442142486572 2.28508329391479
3.05344915390015 2.36840891838074
3.04917168617249 2.44043636322021
3.04393362998962 2.5088574886322
3.03754949569702 2.60209798812866
3.03383231163025 2.70065999031067
3.03229761123657 2.78855395317078
3.02923965454102 2.87820839881897
3.02580213546753 2.97134518623352
3.02402567863464 3.05423498153687
3.02027893066406 3.13499712944031
3.01822543144226 3.22710490226746
3.01787686347961 3.32868361473083
3.02025270462036 3.42022514343262
3.0234112739563 3.50531315803528
3.02359080314636 3.57791042327881
3.02633571624756 3.64838790893555
3.02764821052551 3.7282292842865
3.02530550956726 3.81063532829285
3.02243185043335 3.89732480049133
3.0220103263855 3.99771285057068
3.02057862281799 4.10054063796997
3.01973938941956 4.19892835617065
3.02045822143555 4.28495597839355
3.02168893814087 4.380784034729
3.02082347869873 4.46713495254517
3.01894402503967 4.55570650100708
3.0176694393158 4.64992570877075
3.01842498779297 4.73613548278809
3.01612448692322 4.81178522109985
3.01524806022644 4.89433193206787
3.01444435119629 4.98789644241333
3.01486253738403 5.07081651687622
3.01633024215698 5.15490579605103
3.0179283618927 5.23837184906006
3.01918315887451 5.31525611877441
3.01912951469421 5.38600921630859
3.01544880867004 5.45510578155518
3.01472449302673 5.50777530670166
3.01282525062561 5.55395555496216
3.01116967201233 5.58042526245117
3.0091814994812 5.57076740264893
3.00976228713989 5.57046508789062
3.00928115844727 5.55617713928223
3.01000642776489 5.5379810333252
3.01119613647461 5.51328182220459
3.01203489303589 5.49121570587158
3.0129930973053 5.46199798583984
3.01520943641663 5.4224157333374
3.01896595954895 5.37429857254028
3.02649927139282 5.320481300354
3.03426456451416 5.26786661148071
3.0405285358429 5.23542213439941
3.04593968391418 5.19765949249268
3.05734658241272 5.16172695159912
3.06095361709595 5.11244869232178
3.06742405891418 5.05866289138794
3.06910538673401 4.98988628387451
3.07239437103271 4.91211843490601
3.07434272766113 4.82145977020264
3.07923984527588 4.74271774291992
3.08294343948364 4.67428779602051
3.08745718002319 4.60329389572144
3.08958959579468 4.52862787246704
3.09683847427368 4.46186208724976
3.10586190223694 4.38640594482422
3.10851836204529 4.30278968811035
3.10975289344788 4.20660543441772
3.11200761795044 4.10947132110596
3.11501359939575 4.0259838104248
3.12139940261841 3.94593000411987
3.12582755088806 3.86630964279175
3.12881851196289 3.78472352027893
3.1290922164917 3.70076656341553
3.13304805755615 3.61215209960938
3.13895058631897 3.52714276313782
3.14561700820923 3.45730328559875
3.14928197860718 3.37888765335083
3.15707492828369 3.29200100898743
3.16566276550293 3.20915031433105
3.17311596870422 3.1226692199707
3.17039108276367 3.02415752410889
3.16991209983826 2.92194008827209
3.16910409927368 2.83838272094727
3.17247867584229 2.77125668525696
3.1749472618103 2.68028545379639
3.17833805084229 2.57835602760315
3.17949032783508 2.48647546768188
3.17680263519287 2.39667367935181
3.17952823638916 2.31546831130981
3.17891383171082 2.24762797355652
3.17933416366577 2.19061756134033
3.17672491073608 2.12447166442871
3.17803764343262 2.0408718585968
3.17584872245789 1.95176100730896
3.17338562011719 1.87054944038391
3.17287921905518 1.79680955410004
3.17495632171631 1.72982370853424
3.17679142951965 1.66898095607758
3.1857705116272 1.60255992412567
3.19148206710815 1.53163111209869
3.19552993774414 1.45703101158142
3.19841861724854 1.37584459781647
3.19909119606018 1.28957486152649
3.19900631904602 1.21575629711151
3.20294880867004 1.14789998531342
3.20920872688293 1.07877540588379
3.21788859367371 0.99837452173233
3.22168731689453 0.918491125106812
3.22683906555176 0.837976336479187
3.22467947006226 0.754084527492523
3.22964286804199 0.68457293510437
3.23659753799438 0.636798977851868
3.23742938041687 0.605339467525482
3.23952603340149 0.604576289653778
3.2392885684967 0.607455492019653
3.23961448669434 0.608331620693207
3.23770332336426 0.607955634593964
3.23922395706177 0.607820689678192
3.2395715713501 0.608254194259644
3.23941850662231 0.60773104429245
3.23938059806824 0.607638657093048
3.2393913269043 0.607763409614563
3.23976492881775 0.607776999473572
3.23939847946167 0.607531607151031
3.23960256576538 0.60765141248703
3.23955988883972 0.60738730430603
};
\addlegendentry{$I_2$}
\addplot [draw=steelblue31119180, fill=steelblue31119180, mark=*, only marks,mark size=1.5pt]
table{%
x  y
-0.23 -0.81
};
\addplot [draw=steelblue31119180, fill=steelblue31119180, mark=*, only marks,mark size=1.5pt]
table{%
x  y
7.14 -0.73
};
\addplot [draw=steelblue31119180, fill=steelblue31119180, mark=*, only marks,mark size=1.5pt]
table{%
x  y
7.07 7.66
};
\addplot [draw=steelblue31119180, fill=steelblue31119180, mark=*, only marks,mark size=1.5pt]
table{%
x  y
-0.31 7.59
};
\draw (axis cs:0.1,-0.5) node[
  scale=0.8,
  anchor=west,
  text=black,
  rotate=0.0
]{$\text{R}_1$};
\draw (axis cs:7.2,-0.5) node[
  scale=0.8,
  anchor=west,
  text=black,
  rotate=0.0
]{$\text{R}_2$};
\draw (axis cs:7.2,7.5) node[
  scale=0.8,
  anchor=west,
  text=black,
  rotate=0.0
]{$\text{R}_3$};
\draw (axis cs:0.1,7.5) node[
  scale=0.8,
  anchor=west,
  text=black,
  rotate=0.0
]{$\text{R}_4$};
\end{axis}

\end{tikzpicture}

%% file: tex/data_characterization/inside_opti_pose_square_rotation_2023-03-24-19-08-41.bag.tex
\begin{tikzpicture}

\definecolor{crimson2143940}{RGB}{214,39,40}
\definecolor{darkgray176}{RGB}{176,176,176}
\definecolor{darkorange25512714}{RGB}{255,127,14}
\definecolor{forestgreen4416044}{RGB}{44,160,44}
\definecolor{lightgray204}{RGB}{204,204,204}
\definecolor{mediumpurple148103189}{RGB}{216, 27, 96}
\definecolor{sienna1408675}{RGB}{0, 77, 64}
\definecolor{steelblue31119180}{RGB}{31,119,180}

\begin{axis}[
    height=\figureheight,
    width=\figurewidth,
    axis background/.style={fill=white},
    axis line style={white},
    legend cell align={left},
    legend image post style={scale=5.5},
    legend style={
      fill opacity=1,
      draw opacity=1,
      text opacity=1,
      at={(0.9,0.5)},
      anchor=south west,
      draw=white,
      legend columns=1,
      font=\tiny
    },
    tick align=outside,
    x grid style={white!69.0196078431373!black},
    xlabel={\(\displaystyle x\) (m)},
    minor tick num = 1,
    minor grid style={dashed},
    xmajorgrids,
    xmajorgrids,
    y grid style={white!69.0196078431373!black},
    ylabel={\(\displaystyle y\) (m)}, 
    xminorgrids,
    xminorgrids=true,
    ymajorgrids,
    ymajorticks=true,
    yminorgrids,
    yminorgrids=true,
    xmin=-1.2, xmax=8.2,
    ymin=-1.2, ymax=8.2,
    ylabel near ticks, 
    ylabel shift={-1pt},
]
\addplot [draw=mediumpurple148103189, fill=mediumpurple148103189, mark=diamond*, only marks,mark size=0.23pt]
table{%
x  y
5.83387136459351 0.891956865787506
5.83429098129272 0.892139732837677
5.83445882797241 0.89239913225174
5.83443117141724 0.892365634441376
5.83468103408813 0.894041538238525
5.83461236953735 0.894409358501434
5.83248090744019 0.898139297962189
5.8285346031189 0.925285041332245
5.83033227920532 0.969742059707642
5.83641195297241 1.04513370990753
5.83563947677612 1.13844418525696
5.82477569580078 1.23174774646759
5.81205987930298 1.32632732391357
5.80499792098999 1.42503368854523
5.7960319519043 1.50718307495117
5.78820466995239 1.59189879894257
5.77841901779175 1.68221604824066
5.77646207809448 1.76315343379974
5.77496147155762 1.83964502811432
5.77791213989258 1.91006660461426
5.78180265426636 1.9860543012619
5.7873387336731 2.07678532600403
5.78659439086914 2.15654349327087
5.78934335708618 2.24420619010925
5.79251527786255 2.33383584022522
5.79380655288696 2.41720080375671
5.79313611984253 2.50324630737305
5.79301595687866 2.58546137809753
5.79201936721802 2.68924784660339
5.78607559204102 2.79101204872131
5.77915191650391 2.89338111877441
5.77105188369751 3.00768184661865
5.76789808273315 3.10552334785461
5.76191473007202 3.19323134422302
5.75629663467407 3.28540682792664
5.75588941574097 3.38038063049316
5.75852823257446 3.45941352844238
5.76302528381348 3.53838157653809
5.76657485961914 3.61664962768555
5.77281427383423 3.69415092468262
5.77701950073242 3.76015257835388
5.78180074691772 3.82529187202454
5.78430604934692 3.89479923248291
5.78704595565796 3.97107100486755
5.78891324996948 4.05031156539917
5.78801488876343 4.1216835975647
5.7877402305603 4.20080423355103
5.7854905128479 4.27868556976318
5.78484869003296 4.35471534729004
5.7847728729248 4.42429304122925
5.78358697891235 4.48381280899048
5.78326940536499 4.55833196640015
5.78161382675171 4.63892936706543
5.78041315078735 4.71599626541138
5.7795581817627 4.78989362716675
5.78229856491089 4.86817979812622
5.78254079818726 4.9252495765686
5.78309440612793 4.96777486801147
5.78320837020874 5.01005935668945
5.78424453735352 5.05700922012329
5.78146505355835 5.08271503448486
5.77535963058472 5.12236356735229
5.76563787460327 5.16509675979614
5.74889516830444 5.20452785491943
5.72653818130493 5.24299240112305
5.69520998001099 5.28093004226685
5.65857648849487 5.32192039489746
5.6149582862854 5.35100555419922
5.57226943969727 5.37635183334351
5.54175567626953 5.38902807235718
5.5134015083313 5.39178133010864
5.48731851577759 5.39598703384399
5.46071863174438 5.39919996261597
5.43367052078247 5.40181112289429
5.42318677902222 5.40262794494629
5.4232964515686 5.40186262130737
5.40790224075317 5.40543985366821
5.36624670028687 5.40081453323364
5.29578638076782 5.39382553100586
5.23382949829102 5.38968133926392
5.16841459274292 5.37932062149048
5.09270524978638 5.37309455871582
5.01590204238892 5.36674880981445
4.9392557144165 5.36478424072266
4.87582731246948 5.36361455917358
4.79719924926758 5.36520051956177
4.71809816360474 5.36475992202759
4.64126777648926 5.36741256713867
4.57174110412598 5.3658332824707
4.4980525970459 5.36118841171265
4.4097580909729 5.35725927352905
4.31123781204224 5.35423707962036
4.21458292007446 5.34721326828003
4.12886953353882 5.3422966003418
4.04949426651001 5.33679437637329
3.97503089904785 5.33428049087524
3.91257548332214 5.3305139541626
3.83707547187805 5.33152151107788
3.76124787330627 5.33343267440796
3.68099403381348 5.33515739440918
3.60730004310608 5.33748006820679
3.53070473670959 5.33995962142944
3.44472861289978 5.34055852890015
3.36839962005615 5.33875417709351
3.30820941925049 5.33926010131836
3.25397610664368 5.3404221534729
3.21233057975769 5.33987998962402
3.15241456031799 5.34408807754517
3.08417868614197 5.34415435791016
2.99201202392578 5.34594774246216
2.89290881156921 5.34481048583984
2.80083847045898 5.34481811523438
2.71639037132263 5.34549474716187
2.63613128662109 5.34291696548462
2.5596923828125 5.34164571762085
2.48243832588196 5.33925580978394
2.399005651474 5.33609008789062
2.31110644340515 5.33296060562134
2.23893332481384 5.32738494873047
2.1708128452301 5.32659769058228
2.11054730415344 5.32173347473145
2.04002928733826 5.31720113754272
1.95946109294891 5.31490278244019
1.88156890869141 5.31351566314697
1.79327356815338 5.31292867660522
1.69934952259064 5.31149435043335
1.61341059207916 5.3074836730957
1.53061819076538 5.30673265457153
1.45410668849945 5.30685806274414
1.37258780002594 5.30666542053223
1.28453433513641 5.30517101287842
1.20247268676758 5.30596971511841
1.13549077510834 5.30487585067749
1.06487560272217 5.29931497573853
0.987060785293579 5.29919385910034
0.913959741592407 5.29474878311157
0.862355768680573 5.29250001907349
0.833509922027588 5.28952169418335
0.784430682659149 5.27859497070312
0.708225309848785 5.25896501541138
0.655542016029358 5.2381739616394
0.607792258262634 5.20078897476196
0.548135340213776 5.1341381072998
0.510437905788422 5.07238149642944
0.496571451425552 5.03440141677856
0.499010235071182 5.00599908828735
0.501659274101257 4.97876882553101
0.507097661495209 4.95287036895752
0.515923678874969 4.93084096908569
0.527425229549408 4.92001533508301
0.537388861179352 4.91625261306763
0.542519152164459 4.91315317153931
0.540454804897308 4.90884160995483
0.53946715593338 4.89550971984863
0.544520258903503 4.83836555480957
0.547369480133057 4.80341529846191
0.543466150760651 4.78009653091431
0.545365333557129 4.73329591751099
0.549937307834625 4.67538785934448
0.550513505935669 4.61638021469116
0.55484664440155 4.56359434127808
0.556904137134552 4.51326942443848
0.557633638381958 4.45181035995483
0.564165055751801 4.38127088546753
0.567398726940155 4.29603576660156
0.568709015846252 4.21622323989868
0.568756282329559 4.13670110702515
0.565668165683746 4.06487512588501
0.563206255435944 3.98979544639587
0.561224520206451 3.90747904777527
0.559685945510864 3.81977844238281
0.562239527702332 3.73647594451904
0.561796903610229 3.6463725566864
0.558766305446625 3.55568599700928
0.557652175426483 3.46945953369141
0.55642169713974 3.38741278648376
0.557034730911255 3.31266498565674
0.558654606342316 3.22911953926086
0.56210058927536 3.14960765838623
0.561882555484772 3.06104636192322
0.563513100147247 2.98154544830322
0.566600978374481 2.89479565620422
0.570127904415131 2.81644177436829
0.572860062122345 2.74107193946838
0.578338325023651 2.68142819404602
0.578436136245728 2.62662982940674
0.579445123672485 2.57480645179749
0.580678284168243 2.51906299591064
0.580188035964966 2.44668579101562
0.580510258674622 2.37297940254211
0.582425773143768 2.29974341392517
0.58207643032074 2.2294557094574
0.582971751689911 2.1595721244812
0.583323001861572 2.0885488986969
0.583033263683319 2.0140552520752
0.583221614360809 1.93247377872467
0.583467960357666 1.85032761096954
0.583529770374298 1.76117074489594
0.586945593357086 1.6717392206192
0.589783728122711 1.58722734451294
0.593315064907074 1.5125766992569
0.595396995544434 1.44741094112396
0.59668630361557 1.38129580020905
0.597160995006561 1.31714737415314
0.596125245094299 1.23438656330109
0.595337808132172 1.15175068378448
0.59598183631897 1.07154929637909
0.59663313627243 1.00085473060608
0.599161386489868 0.935509204864502
0.601416230201721 0.868170559406281
0.605493545532227 0.803251028060913
0.605959296226501 0.732801377773285
0.613675177097321 0.672718286514282
0.616296231746674 0.620668172836304
0.619315624237061 0.587120831012726
0.621543407440186 0.553159236907959
0.622941434383392 0.552238345146179
0.625000953674316 0.523685991764069
0.645902931690216 0.456405490636826
0.665266573429108 0.399802654981613
0.686195552349091 0.361207008361816
0.728060722351074 0.308185428380966
0.780484735965729 0.269947439432144
0.8162602186203 0.25026997923851
0.842388391494751 0.24802029132843
0.900161385536194 0.255281120538712
0.939981520175934 0.257376819849014
0.978806972503662 0.27041557431221
0.990441620349884 0.279243469238281
0.996153354644775 0.282025694847107
1.00124847888947 0.278821915388107
1.03282535076141 0.269012838602066
1.08181488513947 0.270390301942825
1.15601599216461 0.273446708917618
1.22187530994415 0.272802144289017
1.29463529586792 0.27526643872261
1.36726558208466 0.276718139648438
1.43479347229004 0.278389602899551
1.50704324245453 0.279140561819077
1.58647775650024 0.279178023338318
1.66571199893951 0.281096965074539
1.74811553955078 0.281699299812317
1.83689725399017 0.283365309238434
1.93157708644867 0.279530793428421
1.99755084514618 0.281769245862961
2.06526970863342 0.276922255754471
2.1419951915741 0.273536682128906
2.22862434387207 0.270261436700821
2.31011557579041 0.263537019491196
2.39826083183289 0.258387446403503
2.49139356613159 0.260893315076828
2.59024500846863 0.258094936609268
2.6812105178833 0.258160829544067
2.76329302787781 0.259823054075241
2.82853722572327 0.26058304309845
2.8947274684906 0.260613113641739
2.97473216056824 0.261558026075363
3.07049179077148 0.262518793344498
3.17594456672668 0.264777272939682
3.28540730476379 0.26778319478035
3.38143920898438 0.270305603742599
3.47298741340637 0.275200664997101
3.56575703620911 0.276960581541061
3.66466689109802 0.279768913984299
3.76472282409668 0.283988475799561
3.85432815551758 0.288977384567261
3.93565964698792 0.293206959962845
4.00651025772095 0.297549098730087
4.07805442810059 0.302318423986435
4.15518522262573 0.304903835058212
4.23336935043335 0.311281561851501
4.32431364059448 0.315870523452759
4.41933345794678 0.316323012113571
4.50826644897461 0.320111006498337
4.58935785293579 0.322750359773636
4.67172479629517 0.321623086929321
4.75422477722168 0.320444911718369
4.84549140930176 0.318710148334503
4.93232583999634 0.316757947206497
5.01968955993652 0.316866636276245
5.10134267807007 0.315828055143356
5.17695188522339 0.318984121084213
5.25907182693481 0.319077223539352
5.34052038192749 0.321541130542755
5.41690111160278 0.322640955448151
5.48334121704102 0.325722694396973
5.53982830047607 0.329812973737717
5.59585523605347 0.329820215702057
5.62831878662109 0.330659598112106
5.632568359375 0.33240008354187
5.70127487182617 0.351462215185165
5.78785562515259 0.388632535934448
5.86305236816406 0.443993538618088
5.91544961929321 0.498483210802078
5.94460725784302 0.558950960636139
5.9382209777832 0.603883922100067
5.91863489151001 0.641533136367798
5.91055059432983 0.671860039234161
5.89959573745728 0.688371181488037
5.89800310134888 0.693874657154083
5.8982834815979 0.692348659038544
5.89763879776001 0.692107141017914
5.89750671386719 0.692273139953613
5.8979287147522 0.692171573638916
5.89798498153687 0.692187786102295
5.89703130722046 0.691707909107208
5.89774322509766 0.692142963409424
5.89769983291626 0.692198932170868
};
\addlegendentry{$I_1$}
\addplot [draw=sienna1408675, fill=sienna1408675, mark=diamond*, only marks,mark size=0.23pt]
table{%
x  y
5.20008945465088 0.916648149490356
5.20046806335449 0.916883528232574
5.20066261291504 0.917321622371674
5.20064401626587 0.91712749004364
5.20045709609985 0.918688654899597
5.20053625106812 0.919409036636353
5.19843435287476 0.921175479888916
5.1945424079895 0.947716891765594
5.19621849060059 0.990075886249542
5.20180511474609 1.05070841312408
5.20090246200562 1.12955093383789
5.19023752212524 1.21275341510773
5.17815780639648 1.29468524456024
5.17107772827148 1.38986873626709
5.16171646118164 1.47500574588776
5.15397882461548 1.56509280204773
5.14374685287476 1.65481185913086
5.14148092269897 1.74121069908142
5.13970470428467 1.8192538022995
5.14254188537598 1.89266049861908
5.14679336547852 1.96266829967499
5.15246248245239 2.0460467338562
5.15236186981201 2.11420917510986
5.15469217300415 2.209468126297
5.15779399871826 2.3016209602356
5.15904188156128 2.38621211051941
5.15889024734497 2.46643137931824
5.15870475769043 2.54974579811096
5.15768432617188 2.65117025375366
5.15204191207886 2.75162625312805
5.14505100250244 2.85532689094543
5.13705348968506 2.96958351135254
5.13336420059204 3.07302975654602
5.12753009796143 3.15712189674377
5.12201929092407 3.24812984466553
5.121497631073 3.34519100189209
5.12386035919189 3.43126487731934
5.12816333770752 3.51205730438232
5.13160419464111 3.59762978553772
5.13778448104858 3.67915725708008
5.14193630218506 3.74091339111328
5.14673280715942 3.8068106174469
5.14943790435791 3.87430763244629
5.1521201133728 3.94919538497925
5.15420055389404 4.0258960723877
5.15333557128906 4.09623622894287
5.15303039550781 4.18003940582275
5.15062999725342 4.26011753082275
5.14999961853027 4.34068632125854
5.14976119995117 4.41432619094849
5.14863443374634 4.47681570053101
5.14803647994995 4.54925394058228
5.14673805236816 4.62171792984009
5.14554643630981 4.69214725494385
5.14468622207642 4.77081298828125
5.14712810516357 4.85455656051636
5.14730644226074 4.9132833480835
5.14811611175537 4.95614862442017
5.14830017089844 4.99380016326904
5.14956378936768 5.02527666091919
5.14877319335938 5.02370452880859
5.14981889724731 5.01198196411133
5.15233993530273 5.0001335144043
5.15777826309204 4.97225952148438
5.16358327865601 4.94901418685913
5.17327880859375 4.9191780090332
5.18731594085693 4.89603805541992
5.20499038696289 4.86618185043335
5.22745513916016 4.84282255172729
5.25078582763672 4.82492065429688
5.27758169174194 4.80216121673584
5.30255031585693 4.78883361816406
5.3282904624939 4.77869081497192
5.35700178146362 4.77190208435059
5.36538982391357 4.77094411849976
5.36364793777466 4.77025604248047
5.34613227844238 4.77383089065552
5.32077836990356 4.76815462112427
5.27192401885986 4.75981664657593
5.21434879302979 4.75566434860229
5.13908576965332 4.74555349349976
5.04783487319946 4.74026679992676
4.973388671875 4.73341035842896
4.90415906906128 4.7316198348999
4.84378623962402 4.72974395751953
4.77178144454956 4.73088026046753
4.69401550292969 4.73085927963257
4.61928653717041 4.73387384414673
4.54670524597168 4.73245859146118
4.47259712219238 4.72750186920166
4.38370513916016 4.72371959686279
4.2874436378479 4.7208309173584
4.19816637039185 4.71340847015381
4.11627054214478 4.70680665969849
4.03594255447388 4.70220804214478
3.96367692947388 4.69929027557373
3.89488363265991 4.69646024703979
3.82216429710388 4.6971173286438
3.74846076965332 4.69878387451172
3.67597985267639 4.70050668716431
3.60201954841614 4.70221996307373
3.52420520782471 4.70453691482544
3.43502616882324 4.70507383346558
3.36059284210205 4.70378351211548
3.29380416870117 4.70436191558838
3.23793697357178 4.70556831359863
3.19297504425049 4.70517539978027
3.13163924217224 4.7091178894043
3.06030869483948 4.70955562591553
2.97333240509033 4.71117305755615
2.87677383422852 4.70923328399658
2.7851185798645 4.71008634567261
2.6967306137085 4.71088266372681
2.61015748977661 4.7086353302002
2.53375911712646 4.70750999450684
2.45856332778931 4.70505332946777
2.38059449195862 4.70178127288818
2.3016836643219 4.69844675064087
2.22923040390015 4.6931004524231
2.15564250946045 4.69214057922363
2.08711552619934 4.68774795532227
2.00636982917786 4.68344831466675
1.9199470281601 4.68166923522949
1.83804261684418 4.68048191070557
1.75614869594574 4.67978668212891
1.67235517501831 4.67751216888428
1.58655047416687 4.67366409301758
1.50074565410614 4.67300701141357
1.42043340206146 4.67301559448242
1.34139680862427 4.6723837852478
1.25648379325867 4.670973777771
1.17874836921692 4.67110109329224
1.11560809612274 4.67051887512207
1.04925847053528 4.66456985473633
0.973773956298828 4.66454315185547
0.908800899982452 4.65965747833252
0.880559265613556 4.65769863128662
0.877766966819763 4.65599918365479
0.893282532691956 4.65306901931763
0.910732567310333 4.65766716003418
0.92907851934433 4.66490173339844
0.96154111623764 4.67432022094727
0.989039182662964 4.67822122573853
1.01443731784821 4.68735790252686
1.0376216173172 4.70312643051147
1.06962084770203 4.72820138931274
1.09850335121155 4.76180076599121
1.12120485305786 4.79183959960938
1.14165472984314 4.82411861419678
1.15961885452271 4.86245632171631
1.17167639732361 4.88920640945435
1.17750906944275 4.89894676208496
1.17560493946075 4.8923773765564
1.17444467544556 4.88052558898926
1.17950630187988 4.85576200485229
1.18115508556366 4.84323072433472
1.17778635025024 4.81273126602173
1.17904531955719 4.776780128479
1.1836633682251 4.71563053131104
1.1852593421936 4.63866138458252
1.18976044654846 4.57477951049805
1.19207036495209 4.51363849639893
1.19291353225708 4.44671869277954
1.19941568374634 4.38197374343872
1.20254862308502 4.31271266937256
1.20328032970428 4.24766492843628
1.20293855667114 4.17130088806152
1.20069301128387 4.08498001098633
1.19842600822449 3.99821734428406
1.19663643836975 3.91486501693726
1.19503998756409 3.83177614212036
1.19752633571625 3.75595378875732
1.19698190689087 3.66726636886597
1.19396388530731 3.57561087608337
1.19307267665863 3.48379397392273
1.19190382957458 3.39557671546936
1.19253718852997 3.3173463344574
1.19414961338043 3.23986911773682
1.1975599527359 3.1570360660553
1.19739723205566 3.0672881603241
1.19895195960999 2.99265289306641
1.20204293727875 2.91429805755615
1.20560204982758 2.83269476890564
1.20792770385742 2.76671981811523
1.21299695968628 2.71080732345581
1.21313762664795 2.65824437141418
1.21414422988892 2.60721182823181
1.21516478061676 2.5529773235321
1.21484708786011 2.48027610778809
1.21543657779694 2.40497732162476
1.21773934364319 2.32218313217163
1.21773040294647 2.24348831176758
1.21892213821411 2.16588068008423
1.21915245056152 2.09145903587341
1.21881473064423 2.01448726654053
1.21872341632843 1.93137335777283
1.21887266635895 1.84856688976288
1.21912515163422 1.76299595832825
1.22206890583038 1.67307841777802
1.22502076625824 1.58949840068817
1.22832310199738 1.51804614067078
1.23020017147064 1.45528972148895
1.2315491437912 1.39136385917664
1.23217189311981 1.32154452800751
1.23124706745148 1.2331520318985
1.23020720481873 1.14353263378143
1.23071265220642 1.05805695056915
1.23116481304169 0.98553079366684
1.23412048816681 0.926666676998138
1.23637425899506 0.865185618400574
1.24021983146667 0.807732224464417
1.242311835289 0.744045853614807
1.24875795841217 0.679184079170227
1.25086188316345 0.621935486793518
1.2540717124939 0.585715532302856
1.25650858879089 0.55893611907959
1.25818490982056 0.557893931865692
1.25811064243317 0.560325026512146
1.26634645462036 0.584992170333862
1.26810121536255 0.600152790546417
1.26524114608765 0.620404660701752
1.26029980182648 0.653215050697327
1.24799203872681 0.699189364910126
1.22368085384369 0.736863493919373
1.1901832818985 0.778951942920685
1.13908076286316 0.837302386760712
1.0968120098114 0.870399177074432
1.04813492298126 0.896702826023102
1.00921177864075 0.912320971488953
0.990795075893402 0.917579650878906
0.999224245548248 0.915867447853088
1.01918482780457 0.907452881336212
1.04033088684082 0.902836322784424
1.08233773708344 0.907442569732666
1.14544594287872 0.903951048851013
1.21912682056427 0.908592879772186
1.31296324729919 0.911730587482452
1.40099990367889 0.913601696491241
1.48182785511017 0.914132416248322
1.5559184551239 0.915193498134613
1.63443756103516 0.916736304759979
1.71451759338379 0.916128516197205
1.7924542427063 0.918952226638794
1.88569474220276 0.916986107826233
1.95624911785126 0.918361902236938
2.02983140945435 0.912853002548218
2.10687208175659 0.911274254322052
2.19848251342773 0.908018231391907
2.28437089920044 0.900813043117523
2.36560010910034 0.895959496498108
2.4465491771698 0.896579384803772
2.54197573661804 0.894763469696045
2.63813161849976 0.893903911113739
2.72786593437195 0.893751800060272
2.80368304252625 0.896991789340973
2.87427711486816 0.896644711494446
2.95546340942383 0.896839141845703
3.04782867431641 0.89723414182663
3.14899492263794 0.898884773254395
3.25493192672729 0.902239501476288
3.35458326339722 0.905264854431152
3.45052433013916 0.908552706241608
3.54139184951782 0.91027045249939
3.62898564338684 0.914014339447021
3.730637550354 0.917075753211975
3.82956838607788 0.921956479549408
3.9218418598175 0.926340460777283
4.00292587280273 0.930806756019592
4.0789041519165 0.935919642448425
4.16014099121094 0.93720668554306
4.2354736328125 0.943627238273621
4.32244396209717 0.947551667690277
4.4154691696167 0.95069545507431
4.51452589035034 0.952998638153076
4.60248374938965 0.956199109554291
4.68723058700562 0.955578446388245
4.76482486724854 0.954711496829987
4.85053682327271 0.952127814292908
4.9357533454895 0.951562762260437
5.01970529556274 0.952236831188202
5.10475540161133 0.949893832206726
5.18298196792603 0.951385378837585
5.26372051239014 0.952252864837646
5.33629703521729 0.956452250480652
5.4152250289917 0.955807685852051
5.48489570617676 0.958818972110748
5.54791927337646 0.962761044502258
5.60935211181641 0.96253377199173
5.63831424713135 0.965555250644684
5.63968324661255 0.966062486171722
5.62062358856201 0.974185883998871
5.58315467834473 0.985412776470184
5.55029106140137 0.982568919658661
5.50259399414062 0.967897474765778
5.45486831665039 0.938813924789429
5.39864253997803 0.893780946731567
5.33533191680908 0.82529628276825
5.2937445640564 0.768241286277771
5.27552509307861 0.719617486000061
5.26499366760254 0.688046753406525
5.26302337646484 0.675946652889252
5.26362323760986 0.679300665855408
5.2631254196167 0.679130613803864
5.26297187805176 0.679017007350922
5.26358222961426 0.678900241851807
5.26306915283203 0.678712487220764
5.26276826858521 0.678577959537506
5.26302671432495 0.678651928901672
5.26330709457397 0.678888857364655
};
\addlegendentry{$I_2$}
\addplot [draw=steelblue31119180, fill=steelblue31119180, mark=*, only marks,mark size=1.5pt]
table{%
x  y
-0.23 -0.81
-0.23 -0.81
};
\addplot [draw=steelblue31119180, fill=steelblue31119180, mark=*, only marks,mark size=1.5pt]
table{%
x  y
7.14 -0.73
};
\addplot [draw=steelblue31119180, fill=steelblue31119180, mark=*, only marks,mark size=1.5pt]
table{%
x  y
7.07 7.66
};
\addplot [draw=steelblue31119180, fill=steelblue31119180, mark=*, only marks,mark size=1.5pt]
table{%
x  y
-0.31 7.59
};
\draw (axis cs:0.1,-0.5) node[
  scale=0.8,
  anchor=west,
  text=black,
  rotate=0.0
]{$\text{R}_1$};
\draw (axis cs:7.2,-0.5) node[
  scale=0.8,
  anchor=west,
  text=black,
  rotate=0.0
]{$\text{R}_2$};
\draw (axis cs:7.2,7.5) node[
  scale=0.8,
  anchor=west,
  text=black,
  rotate=0.0
]{$\text{R}_3$};
\draw (axis cs:0.1,7.5) node[
  scale=0.8,
  anchor=west,
  text=black,
  rotate=0.0
]{$\text{R}_4$};
\end{axis}

\end{tikzpicture}

%% file: tex/data_characterization/outside_opti_pose_outside_square_no_rotation_2023-03-24-18-08-05.bag.tex
\begin{tikzpicture}

\definecolor{crimson2143940}{RGB}{214,39,40}
\definecolor{darkgray176}{RGB}{176,176,176}
\definecolor{darkorange25512714}{RGB}{255,127,14}
\definecolor{forestgreen4416044}{RGB}{44,160,44}
\definecolor{lightgray204}{RGB}{204,204,204}
\definecolor{mediumpurple148103189}{RGB}{216, 27, 96}
\definecolor{sienna1408675}{RGB}{0, 77, 64}
\definecolor{steelblue31119180}{RGB}{31,119,180}

\begin{axis}[
    height=\figureheight,
    width=\figurewidth,
    axis background/.style={fill=white},
    axis line style={white},
    legend cell align={left},
    legend image post style={scale=5.5},
    legend style={
      fill opacity=1,
      draw opacity=1,
      text opacity=1,
      at={(0.9,0.5)},
      anchor=south west,
      draw=white,
      legend columns=1,
      font=\tiny
    },
    tick align=outside,
    x grid style={white!69.0196078431373!black},
    xlabel={\(\displaystyle x\) (m)},
    minor tick num = 1,
    minor grid style={dashed},
    xmajorgrids,
    xmajorgrids,
    y grid style={white!69.0196078431373!black},
    ylabel={\(\displaystyle y\) (m)}, 
    xminorgrids,
    xminorgrids=true,
    ymajorgrids,
    ymajorticks=true,
    yminorgrids,
    yminorgrids=true,
    xmin=-1.2, xmax=8.2,
    ymin=-1.2, ymax=8.2,
    ylabel near ticks, 
    ylabel shift={-1pt},
]
\addplot [draw=mediumpurple148103189, fill=mediumpurple148103189, mark=diamond*, only marks,mark size=0.23pt]
table{%
x  y
4.92934942245483 0.874320805072784
4.92929458618164 0.874542891979218
4.92939853668213 0.874314606189728
4.92928981781006 0.874343574047089
4.92932939529419 0.874718248844147
4.92975759506226 0.876277148723602
4.93150949478149 0.879769861698151
4.93315935134888 0.884076416492462
4.9336724281311 0.886350631713867
4.93307161331177 0.886553287506104
4.92785263061523 0.877651512622833
4.92832326889038 0.877804100513458
4.928138256073 0.878456830978394
4.92843580245972 0.878974854946136
4.93008279800415 0.882365882396698
4.9281177520752 0.878020107746124
4.92813348770142 0.878057718276978
4.9282374382019 0.878579556941986
4.9281210899353 0.878611147403717
4.92957830429077 0.882294356822968
4.93226480484009 0.892939865589142
4.93092584609985 0.914842784404755
4.93016862869263 0.946422576904297
4.90862703323364 0.986341953277588
4.89667701721191 1.02367627620697
4.89195871353149 1.05532431602478
4.89399766921997 1.0849072933197
4.89326143264771 1.12569499015808
4.89946842193604 1.18625557422638
4.90606880187988 1.25198638439178
4.9128737449646 1.3161598443985
4.91957759857178 1.37844598293304
4.92373895645142 1.43134558200836
4.93046140670776 1.48922777175903
4.93568563461304 1.54813611507416
4.94010543823242 1.60348951816559
4.94446611404419 1.66097605228424
4.94881105422974 1.72242081165314
4.95166969299316 1.78499567508698
4.95433187484741 1.84949111938477
4.95448064804077 1.92263507843018
4.9550347328186 1.98940980434418
4.95684862136841 2.0545449256897
4.96126317977905 2.11957240104675
4.96379327774048 2.17987084388733
4.96391677856445 2.23834753036499
4.96755266189575 2.30503129959106
4.97066402435303 2.37410545349121
4.9750714302063 2.43992972373962
4.97802209854126 2.49149894714355
4.97958040237427 2.53898167610168
4.97794103622437 2.58791422843933
4.97762632369995 2.63855147361755
4.97260522842407 2.70206713676453
4.97042894363403 2.77249884605408
4.96774435043335 2.83580780029297
4.96361684799194 2.89400792121887
4.96289491653442 2.95288753509521
4.95751523971558 3.01582074165344
4.95707130432129 3.08332824707031
4.95199584960938 3.1438672542572
4.94715261459351 3.2041130065918
4.94330453872681 3.26019787788391
4.93979787826538 3.30599284172058
4.94106245040894 3.36038208007812
4.93582344055176 3.41115975379944
4.93200874328613 3.4560432434082
4.93141031265259 3.51059174537659
4.92902040481567 3.56417012214661
4.92831373214722 3.61874222755432
4.9292688369751 3.67203545570374
4.92945289611816 3.71785378456116
4.93009424209595 3.76178479194641
4.93135261535645 3.81525874137878
4.9321551322937 3.87361431121826
4.93275499343872 3.92353916168213
4.93278312683105 3.9818058013916
4.93314695358276 4.04105949401855
4.93339586257935 4.10078430175781
4.93543863296509 4.15933847427368
4.93754386901855 4.21273803710938
4.93776893615723 4.26132154464722
4.93952226638794 4.30071306228638
4.94090414047241 4.33909606933594
4.9427227973938 4.37541961669922
4.94099569320679 4.40376853942871
4.94043874740601 4.40924978256226
4.93574476242065 4.39599800109863
4.93585729598999 4.39617919921875
4.93601131439209 4.39664125442505
4.9360671043396 4.3970103263855
4.93624925613403 4.39716577529907
4.93625497817993 4.39701652526855
4.93575239181519 4.39479303359985
4.93644142150879 4.39648771286011
4.93755960464478 4.39850091934204
4.93800449371338 4.39995527267456
4.935631275177 4.39060640335083
4.9367847442627 4.39296388626099
4.94370079040527 4.39428853988647
4.96716451644897 4.39873743057251
4.9854474067688 4.38689565658569
5.01112699508667 4.38176965713501
5.06955194473267 4.39111566543579
5.1197829246521 4.39647626876831
5.16531801223755 4.40102529525757
5.20824337005615 4.40709257125854
5.25988578796387 4.40864896774292
5.30492639541626 4.41079568862915
5.35575723648071 4.41576528549194
5.4028377532959 4.41552734375
5.45384836196899 4.41315841674805
5.51168823242188 4.41219139099121
5.56373596191406 4.40742540359497
5.61483240127563 4.40616846084595
5.66700601577759 4.40616798400879
5.72915124893188 4.40417575836182
5.80042839050293 4.40159463882446
5.86322641372681 4.40078163146973
5.92806386947632 4.39773511886597
5.99438858032227 4.39606666564941
6.04825592041016 4.3936595916748
6.09632921218872 4.38979148864746
6.13191080093384 4.38759660720825
6.16583776473999 4.38645124435425
6.18611288070679 4.3855152130127
6.18110799789429 4.38394021987915
6.18191146850586 4.38478183746338
6.18181610107422 4.38470506668091
6.18160104751587 4.38420677185059
6.18158483505249 4.38437414169312
6.18207025527954 4.37905693054199
6.18295478820801 4.36031675338745
6.17984390258789 4.32393503189087
6.2010326385498 4.28530025482178
6.21441268920898 4.22622346878052
6.22405433654785 4.15531778335571
6.22325372695923 4.07291221618652
6.21838235855103 3.99044799804688
6.21229410171509 3.91783213615417
6.21074056625366 3.84162497520447
6.21563959121704 3.7678747177124
6.21735620498657 3.70976638793945
6.22090864181519 3.65596199035645
6.22217893600464 3.5942325592041
6.22480726242065 3.52902579307556
6.2304253578186 3.46199321746826
6.23427200317383 3.39908623695374
6.24156188964844 3.33660697937012
6.24920892715454 3.2730724811554
6.25662994384766 3.2091178894043
6.26609754562378 3.13784599304199
6.27081632614136 3.06960415840149
6.27026987075806 2.99524021148682
6.27010869979858 2.91473865509033
6.26940393447876 2.83830571174622
6.27206563949585 2.76737022399902
6.27448987960815 2.69729018211365
6.28022813796997 2.62132120132446
6.28671455383301 2.54004454612732
6.28926610946655 2.46111536026001
6.29215478897095 2.39169597625732
6.29395151138306 2.32661700248718
6.29773473739624 2.26029920578003
6.30382919311523 2.18537330627441
6.31074380874634 2.10963034629822
6.3174147605896 2.02236938476562
6.32469177246094 1.94511842727661
6.33023071289062 1.86858546733856
6.33173799514771 1.80300199985504
6.33289909362793 1.74486482143402
6.33819341659546 1.68314254283905
6.34583902359009 1.61365985870361
6.35087585449219 1.5511440038681
6.35185098648071 1.49596118927002
6.35079050064087 1.44114601612091
6.35197305679321 1.38354647159576
6.35515594482422 1.32554185390472
6.35666465759277 1.27054965496063
6.36332130432129 1.21945869922638
6.36317110061646 1.17685401439667
6.36208963394165 1.14606416225433
6.36210632324219 1.12508547306061
6.36215019226074 1.12022125720978
6.36199617385864 1.12071442604065
6.362464427948 1.12255096435547
6.36222267150879 1.12284815311432
6.36245584487915 1.12238168716431
6.36182594299316 1.12140262126923
6.35863065719604 1.11923241615295
6.34430122375488 1.11788415908813
6.31813383102417 1.12134301662445
6.27391815185547 1.10859310626984
6.21100902557373 1.07926833629608
6.14244508743286 1.05589377880096
6.07716989517212 1.04509484767914
6.00263118743896 1.04017424583435
5.92484569549561 1.03894460201263
5.8552188873291 1.03705024719238
5.79265260696411 1.03696489334106
5.73007583618164 1.03521144390106
5.66502523422241 1.03673183917999
5.5994701385498 1.03169620037079
5.52520895004272 1.03035032749176
5.44713926315308 1.02676999568939
5.36899709701538 1.02481424808502
5.29817628860474 1.02119171619415
5.22961330413818 1.01931059360504
5.16191101074219 1.0190247297287
5.09593534469604 1.01577377319336
5.03667545318604 1.02144265174866
4.97092580795288 1.02534973621368
4.90645360946655 1.02298724651337
4.84943294525146 1.0238983631134
4.80031681060791 1.01934099197388
4.76244401931763 1.02008831501007
4.73096942901611 1.01854526996613
4.71246194839478 1.0186265707016
4.71247625350952 1.01937830448151
4.71447515487671 1.01863491535187
4.71469116210938 1.01835966110229
4.71454572677612 1.01836955547333
4.71467256546021 1.01854145526886
4.71453523635864 1.01811635494232
4.71524333953857 1.01878428459167
4.71493005752563 1.01831245422363
4.71501874923706 1.01823723316193
4.71493482589722 1.01830363273621
};
\addlegendentry{$I_1$}
\addplot [draw=sienna1408675, fill=sienna1408675, mark=diamond*, only marks,mark size=0.23pt]
table{%
x  y
4.29369783401489 0.86186957359314
4.29365634918213 0.861983835697174
4.29369688034058 0.861887216567993
4.29358530044556 0.861897051334381
4.29354524612427 0.862409651279449
4.29395008087158 0.864008605480194
4.29548931121826 0.868499755859375
4.29720401763916 0.873305439949036
4.29752826690674 0.877015233039856
4.29690980911255 0.878730893135071
4.29188966751099 0.869636058807373
4.29226875305176 0.869990170001984
4.29216480255127 0.870058953762054
4.29224109649658 0.870772242546082
4.29394912719727 0.874425530433655
4.29202842712402 0.869997382164001
4.29199457168579 0.870077848434448
4.29222679138184 0.870146870613098
4.2920618057251 0.870169222354889
4.29332685470581 0.87434709072113
4.29601860046387 0.886906147003174
4.29467535018921 0.908667504787445
4.29384136199951 0.939167201519012
4.27283525466919 0.961041986942291
4.26077938079834 0.994989633560181
4.25580883026123 1.02955687046051
4.25751161575317 1.07740449905396
4.25688600540161 1.12368702888489
4.26331472396851 1.18176698684692
4.26983785629272 1.24406170845032
4.27708148956299 1.30230069160461
4.28387928009033 1.36167252063751
4.28817844390869 1.40996897220612
4.29456806182861 1.4701806306839
4.29978370666504 1.52924716472626
4.30429649353027 1.58795416355133
4.30826187133789 1.64771199226379
4.31267070770264 1.70761406421661
4.31593608856201 1.76759612560272
4.31878709793091 1.82760465145111
4.31865692138672 1.89833271503448
4.319655418396 1.96613562107086
4.32146644592285 2.03030776977539
4.32577753067017 2.09205842018127
4.32806348800659 2.15141296386719
4.32869577407837 2.20698142051697
4.33178901672363 2.27618765830994
4.33519649505615 2.34744572639465
4.33894109725952 2.42297911643982
4.34142351150513 2.47785925865173
4.3429913520813 2.52684736251831
4.34171342849731 2.57130694389343
4.34089612960815 2.61892080307007
4.33663606643677 2.67549610137939
4.3343620300293 2.74537086486816
4.33170318603516 2.8117208480835
4.3276686668396 2.87237501144409
4.32586669921875 2.93507647514343
4.32123565673828 2.99509024620056
4.32034969329834 3.05867338180542
4.31639003753662 3.11505556106567
4.31117725372314 3.17426800727844
4.30697202682495 3.2318172454834
4.30343532562256 3.2744288444519
4.30510234832764 3.32880115509033
4.29934930801392 3.38048338890076
4.29561853408813 3.42543745040894
4.29482364654541 3.47687745094299
4.2931547164917 3.53196477890015
4.29228162765503 3.58911609649658
4.29256868362427 3.64533138275146
4.29254102706909 3.69681477546692
4.29388904571533 3.73975944519043
4.29528141021729 3.7858464717865
4.2967095375061 3.83649492263794
4.29729413986206 3.88752603530884
4.29718399047852 3.94773602485657
4.29735565185547 4.01545810699463
4.29748964309692 4.08618450164795
4.29931497573853 4.14573764801025
4.30075931549072 4.19916772842407
4.30188655853271 4.23978519439697
4.30439949035645 4.27156209945679
4.30557298660278 4.31324863433838
4.30685758590698 4.34886837005615
4.30576944351196 4.38266611099243
4.30518054962158 4.38924646377563
4.30055332183838 4.37638282775879
4.30061721801758 4.37666606903076
4.30077171325684 4.37704372406006
4.30084466934204 4.37735509872437
4.30102109909058 4.37750816345215
4.30101728439331 4.3772873878479
4.30071592330933 4.37541437149048
4.30117988586426 4.37698698043823
4.30177307128906 4.37889385223389
4.30221652984619 4.37981700897217
4.29989957809448 4.37109136581421
4.30109930038452 4.37284088134766
4.30806875228882 4.37341976165771
4.33132982254028 4.3752269744873
4.34905958175659 4.35987138748169
4.37533473968506 4.35123443603516
4.4336519241333 4.36742687225342
4.48194026947021 4.38458061218262
4.5278787612915 4.39453649520874
4.57015228271484 4.40542554855347
4.62249898910522 4.40947818756104
4.66842699050903 4.41194725036621
4.71775007247925 4.41321325302124
4.76556253433228 4.40549755096436
4.81710195541382 4.39871025085449
4.87507200241089 4.39618015289307
4.9267430305481 4.39302635192871
4.9779896736145 4.39456653594971
5.02994918823242 4.39791250228882
5.09164476394653 4.39910411834717
5.16318273544312 4.39722347259521
5.22615242004395 4.39284658432007
5.2908992767334 4.38634061813354
5.35675764083862 4.38496065139771
5.41062116622925 4.38357257843018
5.45862770080566 4.38295602798462
5.49426746368408 4.38345336914062
5.52824640274048 4.38501024246216
5.54865217208862 4.38764095306396
5.5436840057373 4.38523483276367
5.54449987411499 4.38600921630859
5.54439401626587 4.38568496704102
5.54411792755127 4.38547229766846
5.54410934448242 4.38547945022583
5.54455852508545 4.38159847259521
5.54537153244019 4.36168622970581
5.54233264923096 4.32346105575562
5.56374549865723 4.27329158782959
5.57722043991089 4.21172571182251
5.58685779571533 4.15756225585938
5.58586263656616 4.07973051071167
5.58106470108032 3.99774718284607
5.57499217987061 3.91887927055359
5.57335233688354 3.84196281433105
5.57820320129395 3.76800203323364
5.58012771606445 3.70203256607056
5.58367919921875 3.64334535598755
5.58511829376221 3.57768154144287
5.58786296844482 3.50442481040955
5.59371566772461 3.43304228782654
5.59787845611572 3.36396861076355
5.60497999191284 3.30514717102051
5.61250400543213 3.24663686752319
5.61971855163574 3.18770027160645
5.62892150878906 3.13108444213867
5.63346481323242 3.06434917449951
5.63301753997803 2.98730278015137
5.63275051116943 2.90536117553711
5.63218259811401 2.82509088516235
5.63463687896729 2.75805401802063
5.6371955871582 2.69234299659729
5.64294147491455 2.62394070625305
5.64927577972412 2.54234766960144
5.65209865570068 2.46073508262634
5.65491008758545 2.38645768165588
5.65704727172852 2.31136155128479
5.66128063201904 2.23691463470459
5.66778564453125 2.15935134887695
5.67486000061035 2.08310985565186
5.68127536773682 1.99944996833801
5.68842077255249 1.91991138458252
5.69445896148682 1.84077131748199
5.69647884368896 1.77191412448883
5.69792747497559 1.70784211158752
5.70296573638916 1.647984623909
5.7094841003418 1.59281027317047
5.71478939056396 1.53481388092041
5.71704769134521 1.47429299354553
5.71553897857666 1.41349017620087
5.71662902832031 1.35582339763641
5.71963214874268 1.30175495147705
5.72059345245361 1.25036203861237
5.72691869735718 1.20730841159821
5.72678184509277 1.16738319396973
5.72600269317627 1.1342990398407
5.72569513320923 1.11234819889069
5.72552013397217 1.10407698154449
5.72561836242676 1.10472619533539
5.72586393356323 1.10732400417328
5.72583770751953 1.1072541475296
5.72605133056641 1.10713446140289
5.72554206848145 1.10639989376068
5.72304677963257 1.10348749160767
5.71015787124634 1.08831644058228
5.68705606460571 1.08254158496857
5.64502906799316 1.06818985939026
5.58271837234497 1.05297982692719
5.51185417175293 1.04089152812958
5.44725608825684 1.03783845901489
5.37374782562256 1.03710770606995
5.29515981674194 1.03573727607727
5.22579765319824 1.03528583049774
5.16272258758545 1.03797090053558
5.09990072250366 1.03840184211731
5.0354905128479 1.03987073898315
4.97011661529541 1.03701794147491
4.89699172973633 1.03742659091949
4.81892013549805 1.03778731822968
4.74118518829346 1.03806173801422
4.66886901855469 1.0376170873642
4.60065221786499 1.03726422786713
4.53274965286255 1.03963208198547
4.46631908416748 1.03422725200653
4.40742683410645 1.03258049488068
4.34109926223755 1.02687108516693
4.2755823135376 1.0189163684845
4.21886396408081 1.01614451408386
4.16821813583374 1.00984621047974
4.12938451766968 1.00843393802643
4.0974178314209 1.00561571121216
4.07675457000732 1.00555646419525
4.07602787017822 1.00648462772369
4.07771110534668 1.00609993934631
4.07816123962402 1.00611388683319
4.07807922363281 1.00610625743866
4.07808017730713 1.00630843639374
4.07803916931152 1.00590801239014
4.07862567901611 1.00653636455536
4.07830762863159 1.006103515625
4.07848024368286 1.00601089000702
4.07839202880859 1.00608241558075
};
\addlegendentry{$I_2$}
\addplot [draw=steelblue31119180, fill=steelblue31119180, mark=*, only marks, mark size=1.5pt]
table{%
x  y
2.69 2.4
};
\addplot [draw=steelblue31119180, fill=steelblue31119180, mark=*, only marks, mark size=1.5pt]
table{%
x  y
3.75 2.39
};
\addplot [draw=steelblue31119180, fill=steelblue31119180, mark=*, only marks, mark size=1.5pt]
table{%
x  y
3.77 3.29
};
\addplot [draw=steelblue31119180, fill=steelblue31119180, mark=*, only marks, mark size=1.5pt]
table{%
x  y
2.7 3.29
};
\draw (axis cs:1.75,1.8) node[
  scale=0.8,
  anchor=west,
  text=black,
  rotate=0.0
]{$\text{R}_1$};
\draw (axis cs:3.2,1.8) node[
  scale=0.8,
  anchor=west,
  text=black,
  rotate=0.0
]{$\text{R}_2$};
\draw (axis cs:3.2,3.8) node[
  scale=0.8,
  anchor=west,
  text=black,
  rotate=0.0
]{$\text{R}_3$};
\draw (axis cs:1.75,3.8) node[
  scale=0.8,
  anchor=west,
  text=black,
  rotate=0.0
]{$\text{R}_4$};
\end{axis}

\end{tikzpicture}

%% file: tex/data_characterization/outside_opti_pose_perpendicular_line_2023-03-24-17-46-14.bag.tex
\begin{tikzpicture}

\definecolor{crimson2143940}{RGB}{214,39,40}
\definecolor{darkgray176}{RGB}{176,176,176}
\definecolor{darkorange25512714}{RGB}{255,127,14}
\definecolor{forestgreen4416044}{RGB}{44,160,44}
\definecolor{lightgray204}{RGB}{204,204,204}
\definecolor{mediumpurple148103189}{RGB}{216, 27, 96}
\definecolor{sienna1408675}{RGB}{0, 77, 64}
\definecolor{steelblue31119180}{RGB}{31,119,180}

\begin{axis}[
    height=\figureheight,
    width=\figurewidth,
    axis background/.style={fill=white},
    axis line style={white},
    legend cell align={left},
    legend image post style={scale=5.5},
    legend style={
      fill opacity=1,
      draw opacity=1,
      text opacity=1,
      at={(0.9,0.5)},
      anchor=south west,
      draw=white,
      legend columns=1,
      font=\tiny
    },
    tick align=outside,
    x grid style={white!69.0196078431373!black},
    xlabel={\(\displaystyle x\) (m)},
    minor tick num = 1,
    minor grid style={dashed},
    xmajorgrids,
    xmajorgrids,
    y grid style={white!69.0196078431373!black},
    ylabel={\(\displaystyle y\) (m)}, 
    xminorgrids,
    xminorgrids=true,
    ymajorgrids,
    ymajorticks=true,
    yminorgrids,
    yminorgrids=true,
    xmin=-1.2, xmax=8.2,
    ymin=-1.2, ymax=8.2,
    ylabel near ticks, 
    ylabel shift={-1pt},
]
\addplot [draw=mediumpurple148103189, fill=mediumpurple148103189, mark=diamond*, only marks, mark size=0.23pt]
table{%
x  y
5.78352069854736 0.203395485877991
5.78292369842529 0.20333394408226
5.78297519683838 0.203392386436462
5.78330993652344 0.200992450118065
5.78402948379517 0.201696887612343
5.78301429748535 0.20381835103035
5.78479766845703 0.205708488821983
5.78579950332642 0.210479155182838
5.78591299057007 0.209605529904366
5.78376150131226 0.207625091075897
5.78068161010742 0.217507585883141
5.76902961730957 0.24541537463665
5.74959850311279 0.290723294019699
5.74621820449829 0.284847289323807
5.74788618087769 0.292916387319565
5.74366521835327 0.319518983364105
5.73731088638306 0.367247760295868
5.73686599731445 0.438268184661865
5.74441862106323 0.535562992095947
5.75179052352905 0.658790230751038
5.76265096664429 0.782320022583008
5.77061223983765 0.910764694213867
5.78021574020386 1.04382634162903
5.78777551651001 1.19141495227814
5.78976058959961 1.33303141593933
5.7974009513855 1.46695744991302
5.80079174041748 1.56166470050812
5.8056960105896 1.65472638607025
5.81303024291992 1.7676500082016
5.81731557846069 1.87321889400482
5.81986379623413 1.9669109582901
5.82402753829956 2.06187224388123
5.82851552963257 2.1563937664032
5.83638429641724 2.26053214073181
5.83850717544556 2.35762858390808
5.83710813522339 2.45536780357361
5.83382844924927 2.56304574012756
5.82859945297241 2.67319941520691
5.82353353500366 2.7731568813324
5.82062530517578 2.86594223976135
5.81542253494263 2.96422076225281
5.81447601318359 3.06736159324646
5.81163454055786 3.14537620544434
5.81177139282227 3.22376465797424
5.81046056747437 3.29886507987976
5.80696868896484 3.34281635284424
5.80760431289673 3.34666323661804
5.80462026596069 3.34057998657227
5.80676460266113 3.35515522956848
5.80434083938599 3.39074611663818
5.80585479736328 3.45021414756775
5.81037378311157 3.53761506080627
5.80806875228882 3.64376831054688
5.8076229095459 3.74971580505371
5.80568695068359 3.84416127204895
5.80491018295288 3.94026255607605
5.80131340026855 4.03278684616089
5.79522180557251 4.12714099884033
5.78947973251343 4.22088241577148
5.78293800354004 4.31443357467651
5.77788019180298 4.40866327285767
5.77673864364624 4.49970769882202
5.77124547958374 4.59647035598755
5.76629114151001 4.69031476974487
5.76384305953979 4.77475023269653
5.76221466064453 4.85609436035156
5.76283693313599 4.93532371520996
5.76646041870117 5.01817464828491
5.77273416519165 5.10989570617676
5.77594757080078 5.20634651184082
5.77910184860229 5.30758190155029
5.78403854370117 5.40624761581421
5.78682470321655 5.51551866531372
5.78804063796997 5.61732721328735
5.78822946548462 5.70934104919434
5.78741216659546 5.79821252822876
5.7794976234436 5.87354850769043
5.77509832382202 5.94369840621948
5.76850652694702 6.00147390365601
5.76478290557861 6.02686357498169
5.76467561721802 6.02635383605957
5.76492881774902 6.02658128738403
5.76439428329468 6.02623510360718
5.76410150527954 6.01998281478882
5.76495599746704 5.99922943115234
5.76800537109375 5.95586204528809
5.77081918716431 5.91135406494141
5.77660512924194 5.86840867996216
5.77605247497559 5.83501052856445
5.77901649475098 5.79678201675415
5.77770280838013 5.76390075683594
5.78811883926392 5.72650527954102
5.78278303146362 5.69013214111328
5.78213119506836 5.63809061050415
5.78600645065308 5.58054351806641
5.79015588760376 5.51597452163696
5.79538583755493 5.44930601119995
5.80243539810181 5.38624715805054
5.80840539932251 5.33252763748169
5.81386709213257 5.27301359176636
5.81866025924683 5.21078443527222
5.81423950195312 5.15811347961426
5.81046533584595 5.09371900558472
5.80352592468262 5.0417628288269
5.80139112472534 4.99221420288086
5.7987961769104 4.93212747573853
5.80252313613892 4.8664231300354
5.80949449539185 4.80116510391235
5.81819772720337 4.73521614074707
5.82586669921875 4.66463232040405
5.83004522323608 4.59834337234497
5.83155107498169 4.53389978408813
5.8371901512146 4.45930814743042
5.84028053283691 4.38121652603149
5.85079717636108 4.29031753540039
5.86288833618164 4.19753885269165
5.87246942520142 4.12112426757812
5.88077926635742 4.04877710342407
5.8873724937439 3.9642608165741
5.89353942871094 3.88206744194031
5.89724111557007 3.80891346931458
5.90361785888672 3.73572516441345
5.91000604629517 3.65640330314636
5.90682077407837 3.59063339233398
5.9053783416748 3.52079272270203
5.90283918380737 3.44554996490479
5.90012121200562 3.36676287651062
5.89193344116211 3.2891833782196
5.88107299804688 3.20009922981262
5.86742401123047 3.1113920211792
5.8576340675354 3.03435587882996
5.84485101699829 2.95849919319153
5.83520126342773 2.87261199951172
5.82434511184692 2.7930953502655
5.82171630859375 2.69182562828064
5.82324600219727 2.58571410179138
5.82354307174683 2.49306130409241
5.82085275650024 2.40886473655701
5.81848287582397 2.32588362693787
5.81747817993164 2.24464702606201
5.82477521896362 2.14847588539124
5.82652902603149 2.05633854866028
5.82435035705566 1.97579717636108
5.82413053512573 1.89969539642334
5.8254828453064 1.81860983371735
5.83054113388062 1.72910642623901
5.83373117446899 1.63429737091064
5.83924436569214 1.53910958766937
5.83785200119019 1.45312643051147
5.83553886413574 1.36874151229858
5.83228778839111 1.27975273132324
5.82725143432617 1.18813037872314
5.82469844818115 1.08627879619598
5.8265814781189 0.981933653354645
5.82463359832764 0.886925160884857
5.81705570220947 0.786757230758667
5.81079292297363 0.687608063220978
5.80838966369629 0.598890721797943
5.81185007095337 0.501563787460327
5.81415224075317 0.411856055259705
5.82219839096069 0.31985530257225
5.81509923934937 0.248640045523643
5.80833625793457 0.180076718330383
5.81057786941528 0.117107480764389
5.81482839584351 0.0922192186117172
5.81513357162476 0.100030444562435
5.81433343887329 0.0996915549039841
5.81491136550903 0.100325427949429
5.81499528884888 0.100675195455551
5.815101146698 0.100782752037048
5.81513690948486 0.100872956216335
5.81511831283569 0.100746661424637
5.81520938873291 0.101063698530197
5.81498003005981 0.100845701992512
5.81506013870239 0.100896701216698
};
\addlegendentry{$I_1$}
\addplot [draw=sienna1408675, fill=sienna1408675, mark=diamond*, only marks, mark size=0.23pt]
table{%
x  y
5.14738702774048 0.182399362325668
5.14725542068481 0.182025790214539
5.14734220504761 0.181956812739372
5.14736890792847 0.182258516550064
5.14745712280273 0.182151049375534
5.14737462997437 0.182650074362755
5.14853382110596 0.184258460998535
5.15016222000122 0.189834147691727
5.15026950836182 0.188488155603409
5.14747714996338 0.189071282744408
5.14555072784424 0.202911749482155
5.13324117660522 0.238432392477989
5.11311149597168 0.278757959604263
5.10910081863403 0.273366034030914
5.11196041107178 0.281027019023895
5.1070990562439 0.298207432031631
5.10048341751099 0.351055562496185
5.10067892074585 0.430731922388077
5.10893392562866 0.528626561164856
5.11558198928833 0.63280987739563
5.12689590454102 0.761355102062225
5.13443470001221 0.895999312400818
5.14269113540649 1.03470039367676
5.15084218978882 1.1671553850174
5.15328741073608 1.29639232158661
5.16087865829468 1.43400239944458
5.16398143768311 1.53722727298737
5.16901922225952 1.64162647724152
5.17596197128296 1.74653947353363
5.18046426773071 1.84159815311432
5.18366718292236 1.93265557289124
5.18740844726562 2.03179836273193
5.19129705429077 2.12769341468811
5.20000123977661 2.2330493927002
5.20154333114624 2.32647275924683
5.20129299163818 2.41626715660095
5.19808292388916 2.52977919578552
5.19238471984863 2.64107370376587
5.1878457069397 2.73925232887268
5.18476438522339 2.83079218864441
5.17936992645264 2.92389297485352
5.17740297317505 3.02556419372559
5.17532396316528 3.10846447944641
5.17631483078003 3.19016051292419
5.17457914352417 3.26024794578552
5.17117547988892 3.30056858062744
5.17305707931519 3.30183076858521
5.16991186141968 3.29432725906372
5.17171096801758 3.31286978721619
5.1696252822876 3.35020446777344
5.16956186294556 3.42113852500916
5.17339658737183 3.51634216308594
5.17171812057495 3.61361074447632
5.17189264297485 3.70547485351562
5.17016458511353 3.79999303817749
5.16939878463745 3.90986251831055
5.16570663452148 4.00412368774414
5.16011524200439 4.10109233856201
5.15431356430054 4.19160270690918
5.14756536483765 4.28226327896118
5.14272308349609 4.37841463088989
5.14182376861572 4.47213983535767
5.13624572753906 4.56181335449219
5.13151264190674 4.64972114562988
5.12889862060547 4.7351131439209
5.12662220001221 4.82143068313599
5.12759590148926 4.89996957778931
5.13063716888428 4.98853302001953
5.13683700561523 5.08842658996582
5.13969898223877 5.18146753311157
5.14265775680542 5.2756462097168
5.14768934249878 5.38161945343018
5.15121030807495 5.48706865310669
5.15268802642822 5.58883094787598
5.1526460647583 5.68389225006104
5.1520299911499 5.7806339263916
5.14430475234985 5.85348796844482
5.14008760452271 5.92464065551758
5.13368463516235 5.98306131362915
5.12902450561523 6.01379013061523
5.12950563430786 6.01358652114868
5.12967538833618 6.01349449157715
5.12924957275391 6.01387119293213
5.12887525558472 6.00720357894897
5.12957143783569 5.98240280151367
5.13288021087646 5.93851470947266
5.13564586639404 5.89679002761841
5.14118194580078 5.86296272277832
5.14090824127197 5.82572460174561
5.14292764663696 5.79338932037354
5.14272546768188 5.75393772125244
5.15138244628906 5.71451568603516
5.14947271347046 5.65414762496948
5.14756155014038 5.59775495529175
5.15108346939087 5.54348134994507
5.15425872802734 5.48616600036621
5.15966129302979 5.4265193939209
5.16677331924438 5.37204170227051
5.17272281646729 5.31566190719604
5.17925119400024 5.246178150177
5.18375205993652 5.18170690536499
5.17961025238037 5.13068819046021
5.17501497268677 5.08619260787964
5.16880083084106 5.02787399291992
5.16561460494995 4.96783685684204
5.16410064697266 4.89420032501221
5.1682596206665 4.82253932952881
5.1745810508728 4.76159763336182
5.18181180953979 4.71077394485474
5.18913602828979 4.65266370773315
5.19342041015625 4.58783721923828
5.19622707366943 4.51882743835449
5.2005763053894 4.44080257415771
5.2050576210022 4.35050201416016
5.21390533447266 4.26709699630737
5.22523593902588 4.18893146514893
5.23518133163452 4.11959409713745
5.24329042434692 4.04469299316406
5.25113391876221 3.95326614379883
5.25646305084229 3.87177634239197
5.26052761077881 3.80256009101868
5.26753425598145 3.73753547668457
5.27295112609863 3.66859769821167
5.27066516876221 3.60006356239319
5.26947069168091 3.52667999267578
5.26661491394043 3.44928908348083
5.26459789276123 3.3777003288269
5.2570972442627 3.30269551277161
5.24678421020508 3.21949625015259
5.23156833648682 3.13361740112305
5.2238917350769 3.05702137947083
5.21009635925293 2.9631462097168
5.19925308227539 2.86659812927246
5.18945693969727 2.77751445770264
5.18576526641846 2.68164849281311
5.18720006942749 2.58442568778992
5.18764019012451 2.49209642410278
5.18510580062866 2.39606904983521
5.18323945999146 2.29991817474365
5.18163013458252 2.22248792648315
5.18865919113159 2.14594125747681
5.19010162353516 2.05752754211426
5.18931674957275 1.96795535087585
5.18781280517578 1.886234998703
5.19006252288818 1.80505275726318
5.19435119628906 1.71850216388702
5.19703531265259 1.62930750846863
5.20368385314941 1.54653239250183
5.20151662826538 1.46076881885529
5.19941473007202 1.3712443113327
5.19564437866211 1.27154779434204
5.19314670562744 1.17237901687622
5.18949937820435 1.07879328727722
5.19023704528809 0.986447036266327
5.18935251235962 0.892152309417725
5.1812744140625 0.779674708843231
5.17719602584839 0.669159531593323
5.17325496673584 0.580056250095367
5.17352342605591 0.496898502111435
5.17738628387451 0.414127498865128
5.18475484848022 0.335473865270615
5.1810622215271 0.246668696403503
5.17485904693604 0.165440157055855
5.17221784591675 0.118189640343189
5.17647647857666 0.090655118227005
5.17748641967773 0.0800270363688469
5.17825651168823 0.0813314467668533
5.17915058135986 0.0838045552372932
5.17874097824097 0.0831111893057823
5.17907428741455 0.084830217063427
5.17917490005493 0.0851372256875038
5.17916870117188 0.0852191299200058
5.17922639846802 0.0853320211172104
5.17913198471069 0.0850479304790497
5.17933368682861 0.0854382589459419
5.1790976524353 0.0851843655109406
5.17916440963745 0.0852207988500595
5.17926692962646 0.085483193397522
};
\addlegendentry{$I_2$}
\addplot [draw=steelblue31119180, fill=steelblue31119180, mark=*, only marks, mark size=1.5pt]
table{%
x  y
2.69 2.4
};
\addplot [draw=steelblue31119180, fill=steelblue31119180, mark=*, only marks, mark size=1.5pt]
table{%
x  y
3.75 2.39
};
\addplot [draw=steelblue31119180, fill=steelblue31119180, mark=*, only marks, mark size=1.5pt]
table{%
x  y
3.77 3.29
};
\addplot [draw=steelblue31119180, fill=steelblue31119180, mark=*, only marks, mark size=1.5pt]
table{%
x  y
2.7 3.29
};
\draw (axis cs:1.75,1.8) node[
  scale=0.8,
  anchor=west,
  text=black,
  rotate=0.0
]{$\text{R}_1$};
\draw (axis cs:3.2,1.8) node[
  scale=0.8,
  anchor=west,
  text=black,
  rotate=0.0
]{$\text{R}_2$};
\draw (axis cs:3.2,3.8) node[
  scale=0.8,
  anchor=west,
  text=black,
  rotate=0.0
]{$\text{R}_3$};
\draw (axis cs:1.75,3.8) node[
  scale=0.8,
  anchor=west,
  text=black,
  rotate=0.0
]{$\text{R}_4$};
\end{axis}

\end{tikzpicture}

%% file: tex/data_characterization/outside_opti_pose_square_rotation_2023-03-24-18-00-34.bag.tex
\begin{tikzpicture}

\definecolor{crimson2143940}{RGB}{214,39,40}
\definecolor{darkgray176}{RGB}{176,176,176}
\definecolor{darkorange25512714}{RGB}{255,127,14}
\definecolor{forestgreen4416044}{RGB}{44,160,44}
\definecolor{lightgray204}{RGB}{204,204,204}
\definecolor{mediumpurple148103189}{RGB}{216, 27, 96}
\definecolor{sienna1408675}{RGB}{0, 77, 64}
\definecolor{steelblue31119180}{RGB}{31,119,180}

\begin{axis}[
    height=\figureheight,
    width=\figurewidth,
    axis background/.style={fill=white},
    axis line style={white},
    legend cell align={left},
    legend image post style={scale=5.5},
    legend style={
      fill opacity=1,
      draw opacity=1,
      text opacity=1,
      at={(0.9,0.5)},
      anchor=south west,
      draw=white,
      legend columns=1,
      font=\tiny
    },
    tick align=outside,
    x grid style={white!69.0196078431373!black},
    xlabel={\(\displaystyle x\) (m)},
    minor tick num = 1,
    minor grid style={dashed},
    xmajorgrids,
    xmajorgrids,
    y grid style={white!69.0196078431373!black},
    ylabel={\(\displaystyle y\) (m)}, 
    xminorgrids,
    xminorgrids=true,
    ymajorgrids,
    ymajorticks=true,
    yminorgrids,
    yminorgrids=true,
    xmin=-1.2, xmax=8.2,
    ymin=-1.2, ymax=8.2,
    ylabel near ticks, 
    ylabel shift={-1pt},
]
\addplot [draw=mediumpurple148103189, fill=mediumpurple148103189, mark=diamond*, only marks,mark size=0.23pt]
table{%
x  y
5.78716659545898 0.903215110301971
5.78713989257812 0.905052661895752
5.78748941421509 0.90825742483139
5.78920364379883 0.916987955570221
5.79094076156616 0.935624897480011
5.79262638092041 0.969326853752136
5.78866910934448 1.00969171524048
5.77695798873901 1.05424642562866
5.78405237197876 1.10317695140839
5.78112077713013 1.14610815048218
5.78465032577515 1.1883031129837
5.79053974151611 1.23315262794495
5.79950714111328 1.29035103321075
5.81010961532593 1.35334491729736
5.81808090209961 1.41865599155426
5.82374382019043 1.49531018733978
5.82617044448853 1.57284784317017
5.82340383529663 1.64083778858185
5.82109785079956 1.66149318218231
5.82047128677368 1.6580935716629
5.82026147842407 1.65993404388428
5.82193803787231 1.68015277385712
5.81987142562866 1.73690211772919
5.81340789794922 1.80426895618439
5.80907297134399 1.8796626329422
5.80431604385376 1.95629370212555
5.798180103302 2.0326783657074
5.79664468765259 2.11148476600647
5.7980580329895 2.18892312049866
5.7999324798584 2.25720834732056
5.8027606010437 2.33116841316223
5.80565786361694 2.41592884063721
5.80561447143555 2.50120425224304
5.80803298950195 2.58370757102966
5.8116192817688 2.66370892524719
5.81375646591187 2.7413547039032
5.81254911422729 2.81360793113708
5.81309747695923 2.8878448009491
5.81311464309692 2.96105265617371
5.81501436233521 3.03285503387451
5.81394243240356 3.08497333526611
5.81620264053345 3.1386821269989
5.81587982177734 3.19574522972107
5.81377553939819 3.26423192024231
5.81225347518921 3.34062075614929
5.81909418106079 3.4215772151947
5.8252100944519 3.50333118438721
5.83159971237183 3.5857937335968
5.83642721176147 3.6644446849823
5.84228134155273 3.73881268501282
5.84757614135742 3.80445551872253
5.85245275497437 3.86983203887939
5.84931802749634 3.92892551422119
5.84684562683105 3.99476361274719
5.83979749679565 4.06994819641113
5.83083391189575 4.15075635910034
5.82511520385742 4.23615884780884
5.82222366333008 4.31321382522583
5.8205361366272 4.38555383682251
5.81867218017578 4.45614624023438
5.81415319442749 4.51931190490723
5.8105206489563 4.58135414123535
5.80634260177612 4.65076971054077
5.80382680892944 4.72016000747681
5.80185556411743 4.78924512863159
5.79748153686523 4.86269569396973
5.79760503768921 4.92975854873657
5.79504156112671 4.99157524108887
5.79226112365723 5.04797744750977
5.79028272628784 5.09484672546387
5.78531408309937 5.119713306427
5.78298330307007 5.12909412384033
5.77662229537964 5.14913940429688
5.7668399810791 5.1702618598938
5.75698280334473 5.19639587402344
5.73807573318481 5.22700834274292
5.71857881546021 5.24796390533447
5.70129060745239 5.26347398757935
5.66831731796265 5.28694009780884
5.63579177856445 5.30580997467041
5.61363983154297 5.31643724441528
5.57915639877319 5.33090400695801
5.53240776062012 5.34289216995239
5.50438833236694 5.34604787826538
5.48221206665039 5.34282112121582
5.45017194747925 5.33855009078979
5.42736625671387 5.33085489273071
5.41578245162964 5.3302206993103
5.40514135360718 5.33055686950684
5.38457536697388 5.33425188064575
5.36913299560547 5.32840967178345
5.34195184707642 5.31941747665405
5.29193258285522 5.31418657302856
5.23981237411499 5.29670667648315
5.18803548812866 5.29198551177979
5.12990188598633 5.28883075714111
5.0609450340271 5.28944158554077
4.9976692199707 5.28766393661499
4.92458391189575 5.28740358352661
4.85240697860718 5.28584003448486
4.77920961380005 5.28679084777832
4.70002794265747 5.28827619552612
4.62518072128296 5.29162549972534
4.55714845657349 5.29376411437988
4.49694108963013 5.29382705688477
4.41971206665039 5.29062700271606
4.33080434799194 5.28988122940063
4.24522638320923 5.28791856765747
4.16211938858032 5.28670310974121
4.08463001251221 5.28241300582886
4.00489473342896 5.27924203872681
3.92863345146179 5.28241205215454
3.86083483695984 5.28446674346924
3.78721165657043 5.28923177719116
3.71312260627747 5.29051351547241
3.63719153404236 5.29135370254517
3.55204510688782 5.29606342315674
3.46928381919861 5.29797124862671
3.39701652526855 5.29597187042236
3.32090878486633 5.296959400177
3.23630976676941 5.29789590835571
3.1482675075531 5.29889345169067
3.06694626808167 5.29590940475464
2.98393511772156 5.29595899581909
2.90367245674133 5.29576301574707
2.82398676872253 5.29692316055298
2.74339771270752 5.29704236984253
2.66317105293274 5.29362678527832
2.58364009857178 5.28910827636719
2.49630928039551 5.28551959991455
2.41124558448792 5.28304290771484
2.32280898094177 5.28061151504517
2.24054336547852 5.28026723861694
2.16770029067993 5.27907800674438
2.08951354026794 5.27058839797974
1.99823760986328 5.26578187942505
1.90721905231476 5.26149892807007
1.82999563217163 5.25921249389648
1.74636650085449 5.25786209106445
1.66671812534332 5.25752544403076
1.59019327163696 5.2567310333252
1.513840675354 5.25785493850708
1.43998658657074 5.26026487350464
1.36993837356567 5.26539850234985
1.30998885631561 5.26380729675293
1.24638617038727 5.26420879364014
1.18183708190918 5.26738166809082
1.1353052854538 5.26980447769165
1.08292198181152 5.2708158493042
1.01571476459503 5.27394819259644
0.949092209339142 5.27265691757202
0.90157550573349 5.27252626419067
0.868746280670166 5.27385711669922
0.840433418750763 5.27365684509277
0.833911240100861 5.27166938781738
0.811263740062714 5.26885175704956
0.759739220142365 5.26040840148926
0.716747760772705 5.24983978271484
0.669444739818573 5.22353887557983
0.637417256832123 5.19684410095215
0.614155948162079 5.17394828796387
0.580557823181152 5.13306045532227
0.554784953594208 5.08563995361328
0.547207832336426 5.04955434799194
0.543528974056244 5.003990650177
0.545760810375214 4.9579963684082
0.55724710226059 4.93821144104004
0.572329938411713 4.9213981628418
0.589449942111969 4.90901899337769
0.59418398141861 4.91238832473755
0.591986119747162 4.90833711624146
0.585757911205292 4.89992952346802
0.580473005771637 4.88935613632202
0.576532781124115 4.87065076828003
0.579547047615051 4.83234262466431
0.589492857456207 4.79035186767578
0.592551529407501 4.74174928665161
0.603247046470642 4.68702459335327
0.609357595443726 4.62384653091431
0.611190497875214 4.55735874176025
0.611070215702057 4.48964548110962
0.61309427022934 4.41698932647705
0.613680601119995 4.34861516952515
0.614714682102203 4.27897024154663
0.61045378446579 4.21514272689819
0.605552434921265 4.15363645553589
0.599112570285797 4.08446073532104
0.591926097869873 4.00697994232178
0.587360799312592 3.92514061927795
0.582703769207001 3.84066081047058
0.579336166381836 3.74933934211731
0.578466176986694 3.65562510490417
0.579610109329224 3.565922498703
0.582948207855225 3.47061252593994
0.58510160446167 3.38319659233093
0.589103877544403 3.29662394523621
0.595376193523407 3.2035276889801
0.602171003818512 3.11391186714172
0.604554533958435 3.02484965324402
0.606402397155762 2.93534755706787
0.610603272914886 2.84398579597473
0.612972855567932 2.76628136634827
0.610957682132721 2.6856153011322
0.609209477901459 2.60858416557312
0.606081962585449 2.53464698791504
0.602334976196289 2.46492099761963
0.598780333995819 2.38694024085999
0.598079681396484 2.30976986885071
0.598177492618561 2.23159241676331
0.602945148944855 2.15343284606934
0.606519520282745 2.08406186103821
0.608874976634979 2.01966023445129
0.613517224788666 1.94977939128876
0.617002248764038 1.87078511714935
0.621731996536255 1.78925430774689
0.623758375644684 1.70461118221283
0.625389277935028 1.62664794921875
0.624443113803864 1.55195331573486
0.625889539718628 1.47925698757172
0.623761177062988 1.39696133136749
0.625881731510162 1.31037151813507
0.622773110866547 1.22483396530151
0.619612276554108 1.1448723077774
0.615790724754333 1.05540525913239
0.613956034183502 0.970143556594849
0.612817704677582 0.887712776660919
0.613994121551514 0.811649084091187
0.614877223968506 0.750697314739227
0.621261835098267 0.678529024124146
0.622971296310425 0.620304226875305
0.630679190158844 0.573597252368927
0.634994328022003 0.556340396404266
0.64203667640686 0.51688414812088
0.66168475151062 0.449998885393143
0.686427354812622 0.394725382328033
0.733079850673676 0.344523549079895
0.776490926742554 0.310975402593613
0.809685409069061 0.297599107027054
0.86203670501709 0.285465687513351
0.922390758991241 0.283298254013062
0.953095197677612 0.283452272415161
0.984378635883331 0.28391832113266
1.02634084224701 0.288469552993774
1.04588043689728 0.286561220884323
1.04688310623169 0.286467254161835
1.0556001663208 0.285445153713226
1.07395708560944 0.285981208086014
1.10625088214874 0.295444488525391
1.15164196491241 0.299771755933762
1.19549906253815 0.301948875188828
1.24377632141113 0.300617307424545
1.28067684173584 0.289377510547638
1.33296346664429 0.277978897094727
1.39505767822266 0.272022873163223
1.45416009426117 0.269920915365219
1.51174163818359 0.269836992025375
1.57395780086517 0.269800931215286
1.63689267635345 0.271596401929855
1.70031344890594 0.274476855993271
1.77247035503387 0.275502115488052
1.84198093414307 0.277543902397156
1.91666698455811 0.278751522302628
1.97683048248291 0.281464278697968
2.04186844825745 0.279271572828293
2.10643362998962 0.279074519872665
2.17817807197571 0.27917206287384
2.2541139125824 0.279201507568359
2.32460689544678 0.278129577636719
2.39693856239319 0.275646686553955
2.46959018707275 0.275488525629044
2.54571199417114 0.27365443110466
2.623291015625 0.272718101739883
2.70038771629333 0.27219220995903
2.77136969566345 0.272069543600082
2.83877348899841 0.271782845258713
2.90572834014893 0.271927714347839
2.98066639900208 0.274581164121628
3.06025791168213 0.276797205209732
3.13983511924744 0.280622631311417
3.21726489067078 0.285323292016983
3.30322813987732 0.28969469666481
3.38274192810059 0.293961435556412
3.45589661598206 0.296041995286942
3.53394412994385 0.298162639141083
3.62107539176941 0.301754087209702
3.70883774757385 0.306878358125687
3.79994416236877 0.307743310928345
3.88533115386963 0.312188357114792
3.97594618797302 0.312705248594284
4.06623983383179 0.316457152366638
4.16042900085449 0.319972068071365
4.24404859542847 0.324040204286575
4.31771850585938 0.325961828231812
4.39296007156372 0.324494510889053
4.46636343002319 0.323896139860153
4.54600095748901 0.325816363096237
4.62860631942749 0.326730489730835
4.71421527862549 0.327498435974121
4.792151927948 0.327745944261551
4.86923170089722 0.328919142484665
4.94973516464233 0.32833468914032
5.02866554260254 0.3291075527668
5.10063314437866 0.330562025308609
5.16606187820435 0.329858779907227
5.23058366775513 0.328044921159744
5.28355169296265 0.327030181884766
5.34177350997925 0.327513426542282
5.40676927566528 0.328318953514099
5.47371673583984 0.331804782152176
5.53843927383423 0.335063189268112
5.5869927406311 0.33920830488205
5.63031673431396 0.345489978790283
5.68410301208496 0.357984155416489
5.73912000656128 0.380803495645523
5.82796382904053 0.452092587947845
5.85547780990601 0.51226419210434
5.87294912338257 0.560073673725128
5.87857294082642 0.618649423122406
5.87133550643921 0.65601247549057
5.86880302429199 0.683456897735596
5.86417007446289 0.723236799240112
5.86550331115723 0.756694614887238
5.86621713638306 0.765011310577393
5.86565208435059 0.764802932739258
5.8652868270874 0.765771746635437
5.8650951385498 0.765531003475189
5.86527490615845 0.765708565711975
5.86523056030273 0.76607346534729
5.8651442527771 0.76602691411972
5.86515045166016 0.7660271525383
5.86517143249512 0.765764534473419
5.86505174636841 0.766054093837738
5.86501884460449 0.765906155109406
};
\addlegendentry{$I_1$}
\addplot [draw=sienna1408675, fill=sienna1408675, mark=diamond*, only marks,mark size=0.23pt]
table{%
x  y
5.14993238449097 0.875897109508514
5.15021419525146 0.877256691455841
5.15064430236816 0.881117761135101
5.15194511413574 0.892841100692749
5.15351915359497 0.915536880493164
5.15553569793701 0.951215744018555
5.15099048614502 0.990748286247253
5.13952684402466 1.0343371629715
5.1475944519043 1.06717336177826
5.14415884017944 1.11362385749817
5.14734697341919 1.15762329101562
5.15294504165649 1.19867610931396
5.16250801086426 1.26015269756317
5.17274236679077 1.32954347133636
5.18021202087402 1.40036308765411
5.18587017059326 1.48829972743988
5.18836307525635 1.57539582252502
5.18517351150513 1.64324223995209
5.18313217163086 1.66168797016144
5.18227481842041 1.65870451927185
5.18221759796143 1.66037464141846
5.18411922454834 1.68175554275513
5.18145275115967 1.73266279697418
5.17490434646606 1.79439234733582
5.17056798934937 1.86724102497101
5.16554403305054 1.94543695449829
5.15990161895752 2.01674580574036
5.15830516815186 2.09734106063843
5.15957927703857 2.18017816543579
5.1615514755249 2.24784588813782
5.16452550888062 2.31866359710693
5.16753482818604 2.39979672431946
5.16764450073242 2.48102140426636
5.169921875 2.56441402435303
5.17354536056519 2.64972305297852
5.17555236816406 2.72756266593933
5.17433977127075 2.79719734191895
5.17486572265625 2.87260866165161
5.17525720596313 2.94176149368286
5.17720699310303 3.01078128814697
5.17613697052002 3.06326293945312
5.17851877212524 3.11149287223816
5.17877006530762 3.1622838973999
5.17707204818726 3.22057747840881
5.17573118209839 3.29353523254395
5.18209838867188 3.38264131546021
5.18783664703369 3.47065019607544
5.19387102127075 3.5561146736145
5.19885110855103 3.63067626953125
5.20472526550293 3.70508575439453
5.2098650932312 3.78129911422729
5.21450138092041 3.85619068145752
5.21151542663574 3.91544413566589
5.20890617370605 3.97343635559082
5.20259523391724 4.04009199142456
5.19411516189575 4.11415863037109
5.18818807601929 4.20315790176392
5.18439865112305 4.28952074050903
5.18263483047485 4.36929225921631
5.18129110336304 4.44164800643921
5.17634010314941 4.49715518951416
5.173171043396 4.557053565979
5.1685676574707 4.62565517425537
5.16614198684692 4.69577360153198
5.16428661346436 4.76213836669922
5.16119813919067 4.81612157821655
5.16211700439453 4.86988019943237
5.16031837463379 4.92332792282104
5.15909242630005 4.96817493438721
5.15904808044434 4.99990367889404
5.15746688842773 5.00512933731079
5.15743446350098 5.00371026992798
5.15807437896729 4.99391794204712
5.15896606445312 4.97727537155151
5.16422367095947 4.96063947677612
5.1719536781311 4.93230962753296
5.17781257629395 4.90925168991089
5.18499135971069 4.88843822479248
5.20050716400146 4.85240793228149
5.21839046478271 4.82327032089233
5.22933578491211 4.80711364746094
5.25384902954102 4.78204917907715
5.29193878173828 4.75165033340454
5.31804180145264 4.7354793548584
5.34587287902832 4.71852922439575
5.38059711456299 4.70322179794312
5.40782833099365 4.69270133972168
5.41236066818237 4.69154977798462
5.41127347946167 4.69190216064453
5.39984035491943 4.69589138031006
5.39557456970215 4.69035911560059
5.37406730651855 4.68136215209961
5.32161998748779 4.67600679397583
5.27125549316406 4.65885066986084
5.1950511932373 4.65354776382446
5.12074327468872 4.65046405792236
5.04673480987549 4.65180063247681
4.98771667480469 4.64918661117554
4.92020463943481 4.64888715744019
4.85682535171509 4.64731788635254
4.78889131546021 4.64838886260986
4.71041345596313 4.649986743927
4.62949132919312 4.65301704406738
4.55996561050415 4.65563297271729
4.49503993988037 4.65604448318481
4.41372871398926 4.6531925201416
4.32098531723022 4.65223455429077
4.23696708679199 4.65065002441406
4.15932130813599 4.64915561676025
4.08435344696045 4.64457941055298
4.00338459014893 4.64176225662231
3.9297149181366 4.64419841766357
3.85930919647217 4.64641380310059
3.78476715087891 4.65086936950684
3.71435308456421 4.65249872207642
3.64395809173584 4.65336465835571
3.5596034526825 4.65751075744629
3.47376132011414 4.65898609161377
3.39872694015503 4.65720844268799
3.32508206367493 4.65841197967529
3.2433750629425 4.65974617004395
3.16018509864807 4.66059732437134
3.08220553398132 4.65774345397949
2.99456262588501 4.65770530700684
2.90503120422363 4.65754413604736
2.82347178459167 4.65873765945435
2.74245285987854 4.6588716506958
2.6697633266449 4.65540504455566
2.59933352470398 4.65113353729248
2.52538204193115 4.64803552627563
2.43901491165161 4.64531898498535
2.35276007652283 4.64286422729492
2.27181005477905 4.64330148696899
2.19514751434326 4.64138984680176
2.11537575721741 4.63344383239746
2.01875281333923 4.62828779220581
1.92910885810852 4.62405443191528
1.8459974527359 4.62181806564331
1.76757407188416 4.62000751495361
1.69283521175385 4.62014532089233
1.6083756685257 4.61866474151611
1.52345049381256 4.61971473693848
1.44879198074341 4.6218147277832
1.38323414325714 4.62695074081421
1.32502007484436 4.62538480758667
1.25798428058624 4.62567710876465
1.18566393852234 4.62949466705322
1.13023960590363 4.6309552192688
1.07180392742157 4.63265943527222
1.00494801998138 4.6355299949646
0.944538593292236 4.63425350189209
0.902672648429871 4.63436603546143
0.873590111732483 4.63511800765991
0.852870106697083 4.63512086868286
0.855406820774078 4.63340377807617
0.86111980676651 4.63219928741455
0.876556396484375 4.63296842575073
0.896540522575378 4.63770437240601
0.94405460357666 4.64788484573364
0.973459541797638 4.65506172180176
0.99731183052063 4.66454219818115
1.02542865276337 4.67625617980957
1.05641579627991 4.6938738822937
1.08987200260162 4.71517515182495
1.12502610683441 4.7427191734314
1.15834701061249 4.78199291229248
1.18482899665833 4.8272180557251
1.20881080627441 4.88235855102539
1.22658360004425 4.93554019927979
1.23104476928711 4.94495964050293
1.22887063026428 4.93947267532349
1.22302722930908 4.92174053192139
1.21785366535187 4.90692663192749
1.21392238140106 4.89037275314331
1.21539878845215 4.8804726600647
1.2249801158905 4.84699058532715
1.22821879386902 4.79577970504761
1.23965728282928 4.73153305053711
1.2465136051178 4.6562352180481
1.24869108200073 4.57782888412476
1.24886405467987 4.50573015213013
1.25071740150452 4.44021320343018
1.25148749351501 4.37535905838013
1.25207793712616 4.30791997909546
1.24796390533447 4.24093723297119
1.24328756332397 4.17072916030884
1.2371734380722 4.09220314025879
1.22990727424622 4.00693702697754
1.22530376911163 3.93370723724365
1.22074007987976 3.84796380996704
1.21731460094452 3.75282454490662
1.21654915809631 3.65824627876282
1.21778225898743 3.56856536865234
1.22096490859985 3.48245358467102
1.22279214859009 3.40384531021118
1.2264815568924 3.32568264007568
1.23257839679718 3.23627233505249
1.23940193653107 3.14308738708496
1.24211716651917 3.05185747146606
1.24401462078094 2.9643235206604
1.24832391738892 2.87288451194763
1.25056421756744 2.79547739028931
1.24870026111603 2.71178889274597
1.24671721458435 2.63269782066345
1.24386811256409 2.55360889434814
1.24035310745239 2.47737646102905
1.23678040504456 2.40378284454346
1.23593330383301 2.33442568778992
1.236172914505 2.26001977920532
1.24079275131226 2.17863583564758
1.2443014383316 2.1054048538208
1.24661231040955 2.03630638122559
1.25141191482544 1.97176563739777
1.25511956214905 1.8927515745163
1.25963366031647 1.8157913684845
1.2615133523941 1.73706614971161
1.26274108886719 1.65503132343292
1.26207613945007 1.57661139965057
1.2632063627243 1.5036107301712
1.26107275485992 1.42764914035797
1.26229155063629 1.35032141208649
1.25971269607544 1.26240110397339
1.25701761245728 1.17357981204987
1.25339508056641 1.09624779224396
1.2512104511261 1.01490795612335
1.2498117685318 0.942965805530548
1.24911975860596 0.87687736749649
1.25190591812134 0.814629375934601
1.25561308860779 0.749782860279083
1.26097774505615 0.684898793697357
1.26273202896118 0.637018382549286
1.26821434497833 0.614755928516388
1.26901519298553 0.614896059036255
1.26987743377686 0.627318263053894
1.26543188095093 0.650280594825745
1.25543451309204 0.677618086338043
1.2456693649292 0.701882183551788
1.23152685165405 0.741091191768646
1.20588481426239 0.781323313713074
1.18797218799591 0.807996511459351
1.16605770587921 0.837097704410553
1.13437211513519 0.874496042728424
1.10959756374359 0.899272441864014
1.09545838832855 0.907762706279755
1.06651532649994 0.920693039894104
1.05294752120972 0.927069365978241
1.0545037984848 0.925244033336639
1.05586123466492 0.925460457801819
1.06700015068054 0.920852780342102
1.07031834125519 0.929487943649292
1.08792901039124 0.935800075531006
1.1273900270462 0.937807738780975
1.18034684658051 0.936833024024963
1.21849095821381 0.934668362140656
1.27895557880402 0.923042178153992
1.3524261713028 0.91323459148407
1.40627205371857 0.910312116146088
1.46186506748199 0.906312704086304
1.52245771884918 0.908427000045776
1.5914386510849 0.909666538238525
1.65423595905304 0.913447320461273
1.72447538375854 0.912609457969666
1.78974008560181 0.915247738361359
1.85335385799408 0.917749285697937
1.92194211483002 0.916172027587891
1.98140156269073 0.919200301170349
2.04337596893311 0.918093919754028
2.11038875579834 0.916589438915253
2.17506146430969 0.91712498664856
2.2453293800354 0.915371835231781
2.31591534614563 0.91470855474472
2.39272665977478 0.911784291267395
2.46785068511963 0.912311911582947
2.54193186759949 0.910355567932129
2.61365604400635 0.910031378269196
2.68528819084167 0.909369885921478
2.75063967704773 0.909296274185181
2.82755088806152 0.910058319568634
2.90884470939636 0.911373734474182
2.99137663841248 0.914100646972656
3.06734466552734 0.916059792041779
3.1384379863739 0.919718861579895
3.21462440490723 0.924205482006073
3.3032968044281 0.928283751010895
3.38508105278015 0.931211113929749
3.45394659042358 0.933646321296692
3.53014492988586 0.936451017856598
3.61328649520874 0.939599514007568
3.70108938217163 0.942829370498657
3.79307246208191 0.945616960525513
3.88352656364441 0.949201107025146
3.97277116775513 0.95177161693573
4.06872129440308 0.95495617389679
4.15587091445923 0.959954500198364
4.2541823387146 0.962073981761932
4.34078741073608 0.964745819568634
4.41675996780396 0.96306574344635
4.49800205230713 0.962984800338745
4.57866334915161 0.963971138000488
4.65586137771606 0.967141449451447
4.73690700531006 0.96591591835022
4.82325506210327 0.965918242931366
4.90309238433838 0.966570913791656
4.97787570953369 0.967307388782501
5.05076885223389 0.967600345611572
5.12329816818237 0.967879951000214
5.19731187820435 0.967542052268982
5.26030731201172 0.966892778873444
5.32143211364746 0.964940965175629
5.38310766220093 0.964131355285645
5.4453125 0.966566503047943
5.50703811645508 0.96931254863739
5.54848384857178 0.976960957050323
5.56704330444336 0.979039549827576
5.55029630661011 0.979581117630005
5.52759647369385 0.977395951747894
5.50149726867676 0.975151300430298
5.46660041809082 0.969875693321228
5.42351579666138 0.956267893314362
5.38584136962891 0.941937923431396
5.35824394226074 0.925264894962311
5.31252956390381 0.886641383171082
5.27052640914917 0.844865441322327
5.25003814697266 0.821371972560883
5.23555612564087 0.791849732398987
5.22551012039185 0.753872334957123
5.23093366622925 0.736323356628418
5.22792339324951 0.728870868682861
5.22804641723633 0.729189455509186
5.22848892211914 0.731137990951538
5.2285213470459 0.731383085250854
5.22836971282959 0.731148719787598
5.22856760025024 0.731010556221008
5.22836256027222 0.731312155723572
5.22842693328857 0.731219708919525
5.22854280471802 0.731497347354889
5.22849941253662 0.731233060359955
};
\addlegendentry{$I_2$}
\addplot [draw=steelblue31119180, fill=steelblue31119180, mark=*, only marks,mark size=1.5pt]
table{%
x  y
2.69 2.4
};
\addplot [draw=steelblue31119180, fill=steelblue31119180, mark=*, only marks,mark size=1.5pt]
table{%
x  y
3.75 2.39
};
\addplot [draw=steelblue31119180, fill=steelblue31119180, mark=*, only marks,mark size=1.5pt]
table{%
x  y
3.77 3.29
};
\addplot [draw=steelblue31119180, fill=steelblue31119180, mark=*, only marks,mark size=1.5pt]
table{%
x  y
2.7 3.29
};
\draw (axis cs:1.75,1.8) node[
  scale=0.8,
  anchor=west,
  text=black,
  rotate=0.0
]{$\text{R}_1$};
\draw (axis cs:3.2,1.8) node[
  scale=0.8,
  anchor=west,
  text=black,
  rotate=0.0
]{$\text{R}_2$};
\draw (axis cs:3.2,3.8) node[
  scale=0.8,
  anchor=west,
  text=black,
  rotate=0.0
]{$\text{R}_3$};
\draw (axis cs:1.75,3.8) node[
  scale=0.8,
  anchor=west,
  text=black,
  rotate=0.0
]{$\text{R}_4$};
\end{axis}

\end{tikzpicture}

%% file: tex/data_characterization/outside_opti_pose_over_perpendicular_2023-03-24-18-16-39.bag.tex
\begin{tikzpicture}

\definecolor{crimson2143940}{RGB}{214,39,40}
\definecolor{darkgray176}{RGB}{176,176,176}
\definecolor{darkorange25512714}{RGB}{255,127,14}
\definecolor{forestgreen4416044}{RGB}{44,160,44}
\definecolor{lightgray204}{RGB}{204,204,204}
\definecolor{mediumpurple148103189}{RGB}{216, 27, 96}
\definecolor{sienna1408675}{RGB}{0, 77, 64}
\definecolor{steelblue31119180}{RGB}{31,119,180}

\begin{axis}[
    height=\figureheight,
    width=\figurewidth,
    axis background/.style={fill=white},
    axis line style={white},
    legend cell align={left},
    legend image post style={scale=5.5},
    legend style={
      fill opacity=1,
      draw opacity=1,
      text opacity=1,
      at={(0.9,0.5)},
      anchor=south west,
      draw=white,
      legend columns=1,
      font=\tiny
    },
    tick align=outside,
    x grid style={white!69.0196078431373!black},
    xlabel={\(\displaystyle x\) (m)},
    minor tick num = 1,
    minor grid style={dashed},
    xmajorgrids,
    xmajorgrids,
    y grid style={white!69.0196078431373!black},
    ylabel={\(\displaystyle y\) (m)}, 
    xminorgrids,
    xminorgrids=true,
    ymajorgrids,
    ymajorticks=true,
    yminorgrids,
    yminorgrids=true,
    xmin=-1.2, xmax=8.2,
    ymin=-1.2, ymax=8.2,
    ylabel near ticks, 
    ylabel shift={-1pt},
]
\addplot [draw=mediumpurple148103189, fill=mediumpurple148103189, mark=diamond*, only marks,mark size=0.23pt]
table{%
x  y
4.53123807907104 0.476680964231491
4.53130960464478 0.476729780435562
4.53135776519775 0.476923793554306
4.53141689300537 0.477614402770996
4.53116846084595 0.481192469596863
4.52879953384399 0.473044067621231
4.44965505599976 0.450955301523209
4.19759559631348 0.432783603668213
3.98580861091614 0.477409988641739
3.85679340362549 0.534402549266815
3.76090979576111 0.547614395618439
3.66383457183838 0.545901238918304
3.62357449531555 0.555013716220856
3.6080493927002 0.593126356601715
3.59568619728088 0.667035341262817
3.61124873161316 0.788333415985107
3.62584495544434 0.960292994976044
3.61226010322571 1.14001035690308
3.5875346660614 1.32621204853058
3.59562945365906 1.51465511322021
3.62104296684265 1.71353662014008
3.61555171012878 1.89185464382172
3.57656002044678 2.05025053024292
3.56108665466309 2.20027422904968
3.56845188140869 2.36771559715271
3.56344819068909 2.51741480827332
3.54110336303711 2.63987350463867
3.53497576713562 2.74777913093567
3.54219460487366 2.86554980278015
3.5492103099823 2.91968607902527
3.55478096008301 2.9689028263092
3.57274222373962 3.03112411499023
3.60708546638489 3.12003588676453
3.63508868217468 3.22555160522461
3.63268637657166 3.38630294799805
3.6353452205658 3.5334415435791
3.65250897407532 3.67286515235901
3.65928649902344 3.81179523468018
3.64194655418396 3.98560643196106
3.62555384635925 4.14642286300659
3.6307258605957 4.2932333946228
3.63362145423889 4.43708372116089
3.60702800750732 4.56434011459351
3.57767415046692 4.69335079193115
3.57830619812012 4.81133127212524
3.59333777427673 4.95094156265259
3.59194254875183 5.08349418640137
3.57524180412292 5.22020149230957
3.58616471290588 5.27795219421387
3.61143040657043 5.2713770866394
3.61390423774719 5.1922550201416
3.58469080924988 5.04502248764038
3.56550526618958 4.89199876785278
3.57858538627625 4.69157552719116
3.59687113761902 4.47804975509644
3.59286379814148 4.26273488998413
3.61118912696838 4.06074142456055
3.63887524604797 3.82384133338928
3.63288378715515 3.60540127754211
3.62452030181885 3.42654204368591
3.64535927772522 3.24613881111145
3.64898180961609 3.10739517211914
3.6362521648407 3.04314231872559
3.63080310821533 3.01562309265137
3.6264181137085 2.99086380004883
3.63205528259277 2.95784592628479
3.63563632965088 2.88594150543213
3.61124038696289 2.73760986328125
3.5772979259491 2.53374671936035
3.57911705970764 2.34616041183472
3.59481620788574 2.1495578289032
3.57370471954346 1.94810450077057
3.55023384094238 1.73612844944
3.5794506072998 1.53457832336426
3.60481548309326 1.33597898483276
3.59661412239075 1.13200581073761
3.61753678321838 0.981961250305176
3.66735816001892 0.856005489826202
3.70011925697327 0.740287959575653
3.73252868652344 0.649645745754242
3.84730219841003 0.575802087783813
4.08238649368286 0.488240718841553
4.27088689804077 0.446028023958206
4.38632965087891 0.417887777090073
4.47439622879028 0.415982991456985
4.51482343673706 0.437128931283951
4.52137422561646 0.440175771713257
4.51988077163696 0.445397526025772
4.5197548866272 0.445513606071472
4.51940870285034 0.445036977529526
4.51929616928101 0.445395916700363
4.51941251754761 0.445118069648743
4.51920127868652 0.445400089025497
4.51930618286133 0.445106595754623
4.51919507980347 0.445110946893692
4.51936531066895 0.445215731859207
};
\addlegendentry{$I_1$}
\addplot [draw=sienna1408675, fill=sienna1408675, mark=diamond*, only marks,mark size=0.23pt]
table{%
x  y
3.89428901672363 0.438491523265839
3.89429330825806 0.438607484102249
3.89433288574219 0.438497304916382
3.89428329467773 0.439595222473145
3.89426374435425 0.444755613803864
3.89159107208252 0.444466561079025
3.81462478637695 0.464451789855957
3.55999326705933 0.450391888618469
3.34723734855652 0.448800593614578
3.22105956077576 0.470320910215378
3.1257164478302 0.499471068382263
3.02824711799622 0.519835591316223
2.98683929443359 0.549190998077393
2.97037291526794 0.596063792705536
2.95751357078552 0.681653738021851
2.97324395179749 0.802724838256836
2.98712062835693 0.952223777770996
2.9741632938385 1.13604128360748
2.94910836219788 1.34097504615784
2.95711040496826 1.53731143474579
2.98309731483459 1.73207199573517
2.97724413871765 1.89987850189209
2.93807744979858 2.06824469566345
2.92240238189697 2.21939897537231
2.92983627319336 2.37875580787659
2.92491006851196 2.52315807342529
2.90339422225952 2.67613744735718
2.8991117477417 2.80739903450012
2.90451002120972 2.89671945571899
2.91210031509399 2.95573496818542
2.9167492389679 2.99615573883057
2.93509340286255 3.06476259231567
2.96964883804321 3.1624059677124
2.99756932258606 3.26861691474915
2.99395489692688 3.40381193161011
2.99683880805969 3.55108284950256
3.01627445220947 3.71932554244995
3.02396655082703 3.86994457244873
3.00421786308289 4.01729679107666
2.98626756668091 4.17657804489136
2.99428272247314 4.33787441253662
2.99828553199768 4.486403465271
2.96984815597534 4.60125350952148
2.94026732444763 4.7243185043335
2.94132256507874 4.84982967376709
2.95753502845764 4.99722766876221
2.95395827293396 5.11449432373047
2.93604445457458 5.22315979003906
2.94789528846741 5.28493070602417
2.97392630577087 5.28651857376099
2.97820782661438 5.23974800109863
2.94875955581665 5.09750127792358
2.9285843372345 4.92575788497925
2.94180226325989 4.73253536224365
2.96274781227112 4.53946685791016
2.95720076560974 4.31307792663574
2.97411775588989 4.0858154296875
3.00281953811646 3.8710823059082
2.99956703186035 3.67575550079346
2.98767900466919 3.46899604797363
3.00756502151489 3.30505394935608
3.01564764976501 3.18683362007141
2.99965357780457 3.10135459899902
2.99361228942871 3.05830550193787
2.9891881942749 3.02867007255554
2.99560022354126 2.99061322212219
2.99770498275757 2.91483521461487
2.97325730323792 2.7692813873291
2.94062829017639 2.58372378349304
2.94174599647522 2.38456130027771
2.95835828781128 2.17508697509766
2.93666839599609 1.97636950016022
2.91199970245361 1.75380611419678
2.94308042526245 1.53428506851196
2.96765327453613 1.33894884586334
2.95849418640137 1.1516660451889
2.97837018966675 0.986454963684082
3.03090214729309 0.821786761283875
3.06313514709473 0.712145090103149
3.09434533119202 0.638133645057678
3.20834922790527 0.578664779663086
3.44842267036438 0.511504590511322
3.63870620727539 0.477685749530792
3.75520014762878 0.483151525259018
3.83837032318115 0.460572749376297
3.87708282470703 0.449584305286407
3.88413834571838 0.452114760875702
3.88198828697205 0.454470157623291
3.88170218467712 0.455345988273621
3.881591796875 0.455278843641281
3.88135075569153 0.455795735120773
3.88133716583252 0.455595791339874
3.88126277923584 0.455698788166046
3.8813624382019 0.455376118421555
3.88129234313965 0.455496847629547
3.88128519058228 0.455707937479019
};
\addlegendentry{$I_2$}
\addplot [draw=steelblue31119180, fill=steelblue31119180, mark=*, only marks,mark size=1.5pt]
table{%
x  y
2.69 2.4
};
\addplot [draw=steelblue31119180, fill=steelblue31119180, mark=*, only marks,mark size=1.5pt]
table{%
x  y
3.75 2.39
};
\addplot [draw=steelblue31119180, fill=steelblue31119180, mark=*, only marks,mark size=1.5pt]
table{%
x  y
3.77 3.29
};
\addplot [draw=steelblue31119180, fill=steelblue31119180, mark=*, only marks,mark size=1.5pt]
table{%
x  y
2.7 3.29
};
\draw (axis cs:1.75,1.8) node[
  scale=0.8,
  anchor=west,
  text=black,
  rotate=0.0
]{$\text{R}_1$};
\draw (axis cs:3.2,1.8) node[
  scale=0.8,
  anchor=west,
  text=black,
  rotate=0.0
]{$\text{R}_2$};
\draw (axis cs:3.2,3.8) node[
  scale=0.8,
  anchor=west,
  text=black,
  rotate=0.0
]{$\text{R}_3$};
\draw (axis cs:1.75,3.8) node[
  scale=0.8,
  anchor=west,
  text=black,
  rotate=0.0
]{$\text{R}_4$};
\end{axis}

\end{tikzpicture}

%% file: tex/data_characterization/all_to_all_opti_pose_army_2023-03-24-20-09-29_v2.bag.tex
\begin{tikzpicture}

\definecolor{crimson2143940}{RGB}{214,39,40}
\definecolor{darkgray176}{RGB}{176,176,176}
\definecolor{darkorange25512714}{RGB}{255,127,14}
\definecolor{forestgreen4416044}{RGB}{44,160,44}
\definecolor{gray127}{RGB}{127,127,127}
\definecolor{lightgray204}{RGB}{204,204,204}
\definecolor{mediumpurple148103189}{RGB}{148,103,189}
\definecolor{orchid227119194}{RGB}{227,119,194}
\definecolor{sienna1408675}{RGB}{140,86,75}
\definecolor{steelblue31119180}{RGB}{31,119,180}

\begin{axis}[
    height=\figureheight,
    width=\figurewidth,
    axis background/.style={fill=white},
    axis line style={white},
    legend cell align={left},
    legend image post style={scale=5.5},
    legend style={
      fill opacity=1,
      draw opacity=1,
      text opacity=1,
      at={(0.9,0.5)},
      anchor=south west,
      draw=white,
      legend columns=1,
      font=\tiny
    },
    tick align=outside,
    x grid style={white!69.0196078431373!black},
    xlabel={\(\displaystyle x\) (m)},
    minor tick num = 1,
    minor grid style={dashed},
    xmajorgrids,
    xmajorgrids,
    y grid style={white!69.0196078431373!black},
    ylabel={\(\displaystyle y\) (m)}, 
    xminorgrids,
    xminorgrids=true,
    ymajorgrids,
    ymajorticks=true,
    yminorgrids,
    yminorgrids=true,
    ylabel near ticks, 
    ylabel shift={-1pt},
    ylabel near ticks, 
    ylabel shift={-1pt},
xmin=0.74260955452919, xmax=5.92568499445915,
ymin=0.785644191503525, ymax=5.72054074406624,
]
\addplot [draw=steelblue31119180, fill=steelblue31119180, mark=*, only marks]
table{%
x  y
1.00880706310272 1.03090226650238
1.00876915454865 1.03119254112244
1.00876140594482 1.03116750717163
1.00877285003662 1.03115713596344
1.00876677036285 1.03119075298309
1.00880026817322 1.03077459335327
1.00876545906067 1.03118813037872
1.00881612300873 1.03088963031769
1.00881624221802 1.03089785575867
1.00876533985138 1.03114020824432
};
\draw (axis cs:0.7,1.25) node[
  scale=0.6,
  anchor=base west,
  text=steelblue31119180,
  rotate=0.0
]{UWB 1};
\addplot [draw=darkorange25512714, fill=darkorange25512714, mark=*, only marks]
table{%
x  y
3.50045537948608 1.00995767116547
3.50046873092651 1.00995886325836
3.50045824050903 1.00996541976929
3.50047039985657 1.00996124744415
3.50045418739319 1.00996232032776
3.50045132637024 1.00996339321136
3.50044298171997 1.00997066497803
3.50044965744019 1.0099675655365
3.50045585632324 1.00995969772339
3.50045776367188 1.00996387004852
};
\draw (axis cs:3.1,1.25) node[
  scale=0.6,
  anchor=base west,
  text=darkorange25512714,
  rotate=0.0
]{UWB 2};
\addplot [draw=forestgreen4416044, fill=forestgreen4416044, mark=*, only marks]
table{%
x  y
5.69008445739746 1.04048788547516
5.69009065628052 1.04048538208008
5.69008779525757 1.04047191143036
5.69006443023682 1.04047799110413
5.69007205963135 1.0404953956604
5.69007015228271 1.04048538208008
5.69007062911987 1.04046964645386
5.69008827209473 1.04046869277954
5.69007635116577 1.04048252105713
5.69007539749146 1.04047334194183
};
\draw (axis cs:4.7,1.25) node[
  scale=0.6,
  anchor=base west,
  text=forestgreen4416044,
  rotate=0.0
]{UWB 3};
\addplot [draw=crimson2143940, fill=crimson2143940, mark=*, only marks]
table{%
x  y
5.67498016357422 5.49620866775513
5.67497825622559 5.49620771408081
5.67498207092285 5.49621820449829
5.67499208450317 5.49619293212891
5.67497158050537 5.49621868133545
5.67497587203979 5.49621295928955
5.67498016357422 5.4962272644043
5.67499971389771 5.4962215423584
5.67498254776001 5.49621725082397
5.67498207092285 5.49622249603271
};
\draw (axis cs:4.7,5) node[
  scale=0.6,
  anchor=base west,
  text=crimson2143940,
  rotate=0.0
]{UWB 4};
\addplot [draw=mediumpurple148103189, fill=mediumpurple148103189, mark=*, only marks]
table{%
x  y
3.43573069572449 5.49255323410034
3.43572902679443 5.49255084991455
3.43572545051575 5.49255609512329
3.43572330474854 5.49255228042603
3.43572092056274 5.49255323410034
3.43572664260864 5.49254894256592
3.43572664260864 5.4925537109375
3.43572592735291 5.49255323410034
3.43572807312012 5.49255514144897
3.43572735786438 5.49255037307739
};
\draw (axis cs:3.1,5) node[
  scale=0.6,
  anchor=base west,
  text=mediumpurple148103189,
  rotate=0.0
]{UWB 5};
\addplot [draw=sienna1408675, fill=sienna1408675, mark=*, only marks]
table{%
x  y
0.978208661079407 5.39380979537964
0.978225409984589 5.39381122589111
0.978204071521759 5.39378881454468
0.97821581363678 5.39379167556763
0.978213429450989 5.3938136100769
0.978225111961365 5.39380216598511
0.978217542171478 5.3937554359436
0.978215992450714 5.39378881454468
0.978203892707825 5.39377880096436
0.978224754333496 5.3938159942627
};
\draw (axis cs:0.7,5) node[
  scale=0.6,
  anchor=base west,
  text=sienna1408675,
  rotate=0.0
]{UWB 6};
\addplot [draw=orchid227119194, fill=orchid227119194, mark=*, only marks]
table{%
x  y
2.39523220062256 3.2052800655365
2.39522695541382 3.20529985427856
2.3952317237854 3.20520544052124
2.39525413513184 3.20516228675842
2.39523601531982 3.20528697967529
2.39525437355042 3.20528554916382
2.39525032043457 3.20516991615295
2.39525532722473 3.20515871047974
2.39522790908813 3.20528078079224
2.3952329158783 3.20529079437256
};
\draw (axis cs:2,3.4) node[
  scale=0.6,
  anchor=base west,
  text=orchid227119194,
  rotate=0.0
]{UWB 7};
\addplot [draw=gray127, fill=gray127, mark=*, only marks]
table{%
x  y
4.6132984161377 3.19969534873962
4.61334419250488 3.19974279403687
4.61333084106445 3.19974803924561
4.61333227157593 3.19972324371338
4.61333608627319 3.19974088668823
4.6133394241333 3.19974398612976
4.61333084106445 3.19973397254944
4.61333179473877 3.19973111152649
4.61333417892456 3.19974732398987
4.61330318450928 3.19973397254944
};
\draw (axis cs:4.2,3.4) node[
  scale=0.6,
  anchor=base west,
  text=gray127,
  rotate=0.0
]{UWB 8};
\end{axis}

\end{tikzpicture}

%% file: tex/data_characterization/all_to_all_opti_pose_corner_2023-03-24-20-18-25_v2.bag.tex
\begin{tikzpicture}

\definecolor{crimson2143940}{RGB}{214,39,40}
\definecolor{darkgray176}{RGB}{176,176,176}
\definecolor{darkorange25512714}{RGB}{255,127,14}
\definecolor{forestgreen4416044}{RGB}{44,160,44}
\definecolor{gray127}{RGB}{127,127,127}
\definecolor{lightgray204}{RGB}{204,204,204}
\definecolor{mediumpurple148103189}{RGB}{148,103,189}
\definecolor{orchid227119194}{RGB}{227,119,194}
\definecolor{sienna1408675}{RGB}{140,86,75}
\definecolor{steelblue31119180}{RGB}{31,119,180}

\begin{axis}[
    height=\figureheight,
    width=\figurewidth,
    axis background/.style={fill=white},
    axis line style={white},
    legend cell align={left},
    legend image post style={scale=5.5},
    legend style={
      fill opacity=1,
      draw opacity=1,
      text opacity=1,
      at={(0.9,0.5)},
      anchor=south west,
      draw=white,
      legend columns=1,
      font=\tiny
    },
    tick align=outside,
    x grid style={white!69.0196078431373!black},
    xlabel={\(\displaystyle x\) (m)},
    minor tick num = 1,
    minor grid style={dashed},
    xmajorgrids,
    xmajorgrids,
    y grid style={white!69.0196078431373!black},
    ylabel={\(\displaystyle y\) (m)}, 
    xminorgrids,
    xminorgrids=true,
    ymajorgrids,
    ymajorticks=true,
    yminorgrids,
    yminorgrids=true,
    xmin=-1.2, xmax=8.2,
    ymin=-1.2, ymax=8.2,
    ylabel near ticks, 
    ylabel shift={-1pt},
xmin=0.75795214176178, xmax=6.27515799999237,
ymin=0.66504413485527, ymax=8.02371650338173,
]
\addplot [draw=steelblue31119180, fill=steelblue31119180, mark=*, only marks]
table{%
x  y
1.00873422622681 1.03125309944153
1.00875985622406 1.03125023841858
1.00875806808472 1.03125047683716
1.00875008106232 1.03125560283661
1.00875151157379 1.03126168251038
1.00873851776123 1.03126275539398
1.00874578952789 1.0312625169754
1.00880086421967 1.03094661235809
1.00873780250549 1.03124403953552
1.0087468624115 1.03126358985901
};
\draw (axis cs:0.7,1.3) node[
  scale=0.6,
  anchor=base west,
  text=steelblue31119180,
  rotate=0.0
]{UWB 1};
\addplot [draw=darkorange25512714, fill=darkorange25512714, mark=*, only marks]
table{%
x  y
2.62910962104797 0.999536275863647
2.62911081314087 0.999532878398895
2.62910914421082 0.999532043933868
2.62911605834961 0.9995396733284
2.62911176681519 0.99954879283905
2.62911939620972 0.999549567699432
2.62911677360535 0.99954080581665
2.62912511825562 0.999533176422119
2.62910842895508 0.999529242515564
2.62910962104797 0.999537646770477
};
\draw (axis cs:2.3,1.3) node[
  scale=0.6,
  anchor=base west,
  text=darkorange25512714,
  rotate=0.0
]{UWB 2};
\addplot [draw=forestgreen4416044, fill=forestgreen4416044, mark=*, only marks]
table{%
x  y
4.3246488571167 1.04398286342621
4.32465648651123 1.04400765895844
4.32465076446533 1.04400110244751
4.32465887069702 1.0439887046814
4.32465648651123 1.04399716854095
4.3246636390686 1.04400372505188
4.32465791702271 1.04400253295898
4.32465982437134 1.04399299621582
4.32465600967407 1.04400169849396
4.3246545791626 1.04400682449341
};
\draw (axis cs:3.8,1.3) node[
  scale=0.6,
  anchor=base west,
  text=forestgreen4416044,
  rotate=0.0
]{UWB 3};
\addplot [draw=crimson2143940, fill=crimson2143940, mark=*, only marks]
table{%
x  y
5.94011735916138 6.06578922271729
5.93871879577637 6.06683731079102
5.94030141830444 6.06527709960938
5.9405345916748 6.06481742858887
5.94075393676758 6.06437540054321
5.94253826141357 6.06622791290283
5.94262313842773 6.06620454788208
5.94278192520142 6.0661416053772
5.94285440444946 6.06613874435425
5.94268989562988 6.06616878509521
};
\draw (axis cs:4.55,6) node[
  scale=0.6,
  anchor=base west,
  text=crimson2143940,
  rotate=0.0
]{UWB 4};
\addplot [draw=mediumpurple148103189, fill=mediumpurple148103189, mark=*, only marks]
table{%
x  y
5.92594528198242 4.37933874130249
5.92594814300537 4.37935209274292
5.92595100402832 4.37934112548828
5.92594766616821 4.37934255599976
5.92595481872559 4.37934732437134
5.92595243453979 4.3793478012085
5.92594861984253 4.37934923171997
5.92595100402832 4.37934112548828
5.92594385147095 4.37934637069702
5.92595529556274 4.37934446334839
};
\draw (axis cs:4.55,4.37) node[
  scale=0.6,
  anchor=base west,
  text=mediumpurple148103189,
  rotate=0.0
]{UWB 5};
\addplot [draw=sienna1408675, fill=sienna1408675, mark=*, only marks]
table{%
x  y
5.97636222839355 7.68915557861328
5.97639083862305 7.68912792205811
5.97639083862305 7.68918085098267
5.97638893127441 7.68919706344604
5.97639513015747 7.68920993804932
5.97638940811157 7.68923139572144
5.97638368606567 7.68919277191162
5.97639036178589 7.68919515609741
5.97641134262085 7.68922424316406
5.97638893127441 7.68920993804932
};
\draw (axis cs:4.55,7.68) node[
  scale=0.6,
  anchor=base west,
  text=sienna1408675,
  rotate=0.0
]{UWB 6};
\addplot [draw=orchid227119194, fill=orchid227119194, mark=*, only marks]
table{%
x  y
6.01380014419556 1.02102220058441
6.01380634307861 1.02108979225159
6.01444244384766 1.02113211154938
6.01444530487061 1.02123320102692
6.01381874084473 1.02101790904999
6.01378202438354 1.02109253406525
6.01442861557007 1.02114534378052
6.01379203796387 1.02106738090515
6.01448678970337 1.02100467681885
6.01457214355469 1.02133131027222
};
\draw (axis cs:5,1.3) node[
  scale=0.6,
  anchor=base west,
  text=orchid227119194,
  rotate=0.0
]{UWB 7};
\addplot [draw=gray127, fill=gray127, mark=*, only marks]
table{%
x  y
6.02428436279297 2.66250276565552
6.02429151535034 2.6624927520752
6.02429437637329 2.66249060630798
6.02437591552734 2.66247987747192
6.02429246902466 2.66250252723694
6.02428865432739 2.66249632835388
6.02429056167603 2.66249418258667
6.02436447143555 2.6624801158905
6.02429056167603 2.66249442100525
6.02428770065308 2.66250777244568
};
\draw (axis cs:4.55,2.6) node[
  scale=0.6,
  anchor=base west,
  text=gray127,
  rotate=0.0
]{UWB 8};
\end{axis}

\end{tikzpicture}

%% file: sec/05_Experiments.tex

\section{Dataset Analysis | UWB Characterization}

\subsection{UWB Ranging Error Modeling}

\begin{figure}
    \begin{subfigure}{0.24\textwidth}
        \centering
        \setlength{\figurewidth}{\textwidth}
        \setlength{\figureheight}{\textwidth}
        \scriptsize{\input{tex/boxplot_errors/error_main_all_0.0_final_data.csv}}
        \caption{Absolute error}
        \label{fig:error_main}
    \end{subfigure}
    \hfill
    \begin{subfigure}{0.24\textwidth}
        \centering
        \setlength{\figurewidth}{\textwidth}
        \setlength{\figureheight}{\textwidth}
        \scriptsize{\input{tex/boxplot_errors/relative_error_main_all_0.0_rosbag_try1.csv}}
        \caption{Relative error}
        \label{fig:relative_error}     
    \end{subfigure}
    \caption{Absolute and relative distance error for two UWB nodes directly facing each other, rotation 0$^{\circ}$, at different distances. \Cref{fig:error_main} shows the absolute error, while \Cref{fig:relative_error} shows the relative error.}
    \label{fig:0_error} 
\vspace{-1.5em}
\end{figure}
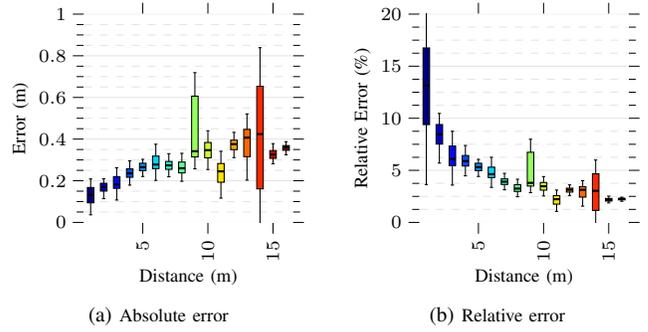

In \textit{Case I} we analyzed the distance error obtained by a pair of UWB nodes at different distances and rotations. \Cref{fig:error_main} shows the errors obtained from all the distances when both faces of the UWB are directed at each other. This configuration is considered to be 0$^{\circ}$. Overall the error increases as the distance increases, with some outliers at 9$\,m$ and 14$\,m$. On the other hand, \Cref{fig:relative_error} shows the relative error in the same scenario. In this case, the relative error of the measurements drops when increasing the distance.

We obtained the distance error for the UWB distance for each rotation in \Cref{fig:rotations} at each distance. We only present the absolute error for each scenario which can be seen in \Cref{tab:boxplot}. Each row is a different rotation plane and each column has a different angle of rotation, all with the same starting position, angle$^{\circ}$, that was shown in \Cref{fig:error_main}. Overall, the error in most configurations increases with distance. For Yaw, we can observe the highest errors at 180$^{\circ}$. Angles 90$^{\circ}$ and 270$^{\circ}$ have similar results, and 0$^{\circ}$ has the lowest errors. On Pitch, the errors for 90 $^{\circ}$ and 270$^{\circ}$ have less of a clear upward trend, with 270$^{\circ}$ having the most fluctuating results overall. The errors for Pitch are also higher than Yaw. For Roll, all angles present an increasing trend. Once again 180$^{\circ}$ has the highest errors, with 90$^{\circ}$ and 270$^{\circ}$ again having similar results.

\begin{table*}[!h]
    \caption{Absolute distance error of two UWB nodes at different distances when one node is rotated in one of the rotations planes. For clarity, we remove axis labels, with the error in meters in the $y$ axis and the distance between the nodes in meters in the $x$ axis.}
    \label{tab:boxplot}
    \renewcommand{\arraystretch}{0.9}
    \centering
    \begin{tabular}{cccc}
        \scriptsize
        \textbf{} & \textbf{90$^{\circ}$} & \textbf{180$^{\circ}$} & \textbf{270$^{\circ}$} \\ 
        \midrule
        & & & \\
        \rotatebox[origin=c]{90}{\hspace{8em} Yaw} & 
            \setlength\figureheight{0.18\textwidth}
            \setlength\figurewidth{0.3\textwidth} \scriptsize{\input{tex/boxplot_errors/error_ByAngle_yaw_90.0_rosbag_try1.csv}}
            & \setlength\figureheight{0.18\textwidth}
            \setlength\figurewidth{0.3\textwidth} \scriptsize{\input{tex/boxplot_errors/error_ByAngle_yaw_180.0_rosbag_try1.csv}} & \setlength\figureheight{0.18\textwidth}
            \setlength\figurewidth{0.3\textwidth} \scriptsize{\input{tex/boxplot_errors/error_ByAngle_yaw_270.0_rosbag_try1.csv}} \\[-3em]
        \rotatebox[origin=c]{90}{\hspace{8em} Pitch} & 
            \setlength\figureheight{0.18\textwidth}
            \setlength\figurewidth{0.3\textwidth} \scriptsize{\input{tex/boxplot_errors/error_ByAngle_pitch_90.0_rosbag_try1.csv}} & \setlength\figureheight{0.18\textwidth}
            \setlength\figurewidth{0.3\textwidth} \scriptsize{\input{tex/boxplot_errors/error_ByAngle_pitch_180.0_rosbag_try1.csv}} & \setlength\figureheight{0.18\textwidth}
            \setlength\figurewidth{0.3\textwidth} \scriptsize{\input{tex/boxplot_errors/error_ByAngle_pitch_270.0_rosbag_try1.csv}} \\[-3em]
        \rotatebox[origin=c]{90}{\hspace{8em} Roll} & \setlength\figureheight{0.18\textwidth}
            \setlength\figurewidth{0.3\textwidth} \scriptsize{\input{tex/boxplot_errors/error_ByAngle_roll_90.0_rosbag_try1.csv}} & \setlength\figureheight{0.18\textwidth}
            \setlength\figurewidth{0.3\textwidth} \scriptsize{\input{tex/boxplot_errors/error_ByAngle_roll_180.0_rosbag_try1.csv}} & \setlength\figureheight{0.18\textwidth}
            \setlength\figurewidth{0.3\textwidth} \scriptsize{\input{tex/boxplot_errors/error_ByAngle_roll_270.0_rosbag_try1.csv}} \\[-4em]
            \bottomrule
    \end{tabular}
\vspace{-1em}
\end{table*}

\subsection{Positioning error}

\begin{figure}
    \begin{subfigure}{.3\textwidth}
        \centering
        \setlength{\figurewidth}{\textwidth}
        \setlength{\figureheight}{\textwidth}
        \scriptsize{\input{tex/graphs_les/les_square_rotation_2023-03-24-19-08-41_trajectory}}
        \caption{UWB trajectory}
        \label{fig:pos_in_traj}
    \end{subfigure}
    \hfill
    \begin{subfigure}{.15\textwidth}
        \centering
        \setlength{\figurewidth}{\textwidth}
        \setlength{\figureheight}{2.1\textwidth}
        \scriptsize{\input{tex/graphs_les/les_square_rotation_2023-03-24-19-08-41_error}}
        \caption{Absolute error}
        \label{fig:pos_in_error}     
    \end{subfigure}
    \caption{UWB node localization for two moving UWB Initiators implementing a Least Square Estimator inside the convex envelope alongside the absolute position error.}
    \label{fig:pos_in} 
\vspace{-1.5em}
\end{figure}

\begin{figure}
    \begin{subfigure}{.3\textwidth}
        \centering
        \setlength{\figurewidth}{\textwidth}
        \setlength{\figureheight}{\textwidth}
        \scriptsize{\input{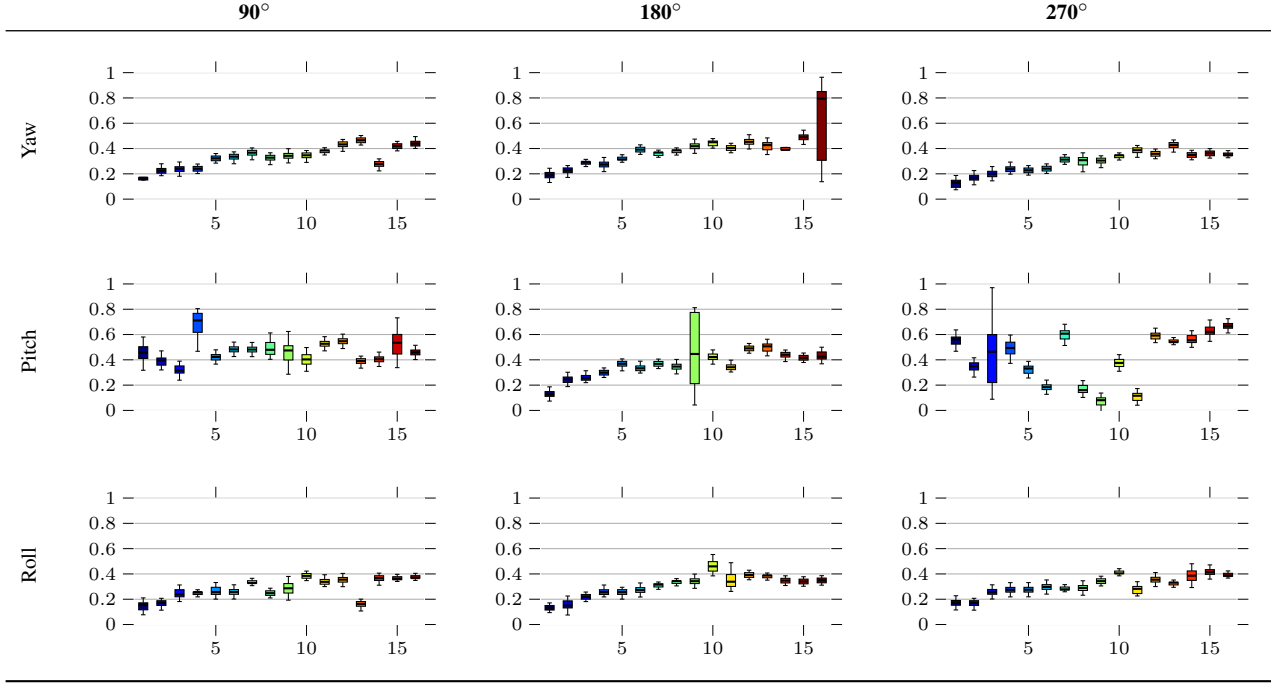}}
        \caption{UWB trajectory}
        \label{fig:pos_out_traj}
    \end{subfigure}
    \hfill
    \begin{subfigure}{.15\textwidth}
        \centering
        \setlength{\figurewidth}{\textwidth}
        \setlength{\figureheight}{2.1\textwidth}
        \scriptsize{\input{tex/graphs_les/les_square_rotation_2023-03-24-18-00-34_error}}
        \caption{Absolute error}
        \label{fig:pos_out_error}     
    \end{subfigure}
    \caption{UWB node localization for two moving UWB Initiators implementing a Least Square Estimator outside the convex envelope alongside the absolute position error.}
    \label{fig:pos_out} 
\vspace{-1.5em}
\end{figure}

In \textit{Case II} we present different scenarios with four static responders and two mobile initiators. There exist scenarios where the responders are located at the corners of the testing arena, and others where they are in the platform for the Hustky and located in the middle. As a result, the two mobile initiators might be moving inside or outside the convex envelope of the responders. We implemented a Least Square Estimator (LSE) to obtain the position of the mobile UWB to the anchors using the measured distances. The results for two scenarios can be seen in \Cref{fig:pos_in} and \Cref{fig:pos_out}. In both cases the search windows and step size parameters are the same, showing different results for each. \Cref{fig:pos_in_traj} shows a closer match of the calculated position to the ground truth than in \Cref{fig:pos_out_traj}. Both \Cref{fig:pos_in_error} and \Cref{fig:pos_out_error} clearly show the same, with the first case showing a lower error than the second. This demonstrates the influence of being inside or outside the convex envelope and is the reason for including both types of scenarios in our dataset.

\subsection{Algorithm comparison}

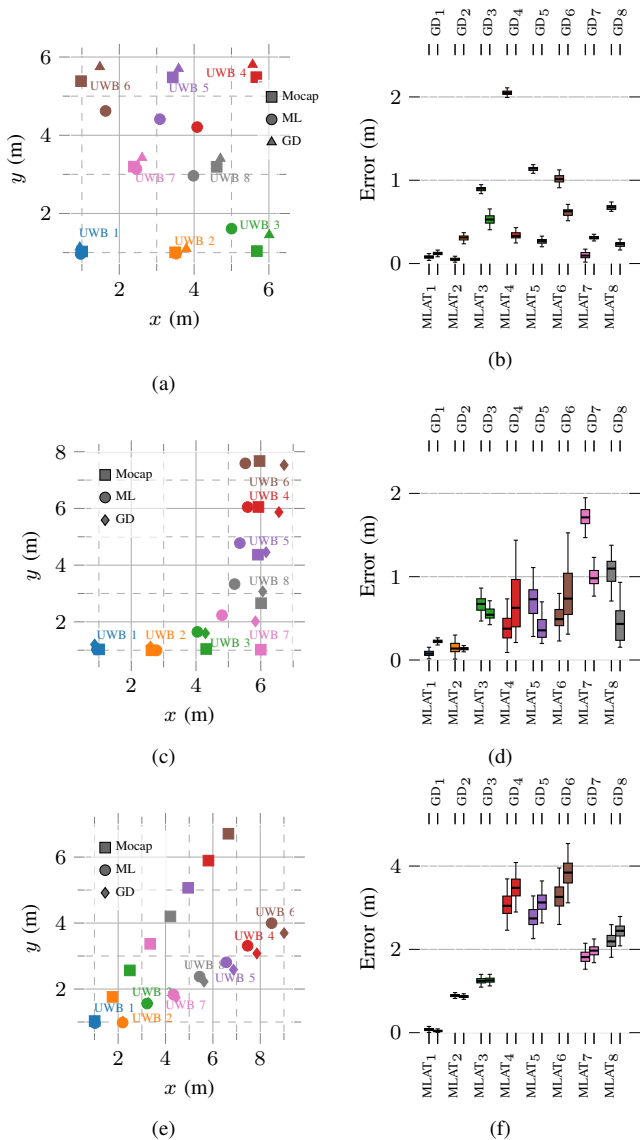
\begin{figure}

    \centering
    \begin{subfigure}[t]{0.24\textwidth}
        \centering
        \setlength\figureheight{\textwidth}
        \setlength\figurewidth{\textwidth}        \footnotesize{\input{tex/graphs_ml_gd/army_2023-03-24-20-09-29_v2_plot_v2}}
        \caption{}
        \label{fig:pos_army}
    \end{subfigure}
    \begin{subfigure}[t]{0.24\textwidth}
        \centering
        \setlength\figureheight{\textwidth}
        \setlength\figurewidth{\textwidth}        \footnotesize{\input{tex/graphs_ml_gd/army_2023-03-24-20-09-29_v2_boxplot}}
        \caption{}
        \label{fig:error_army}
    \end{subfigure}
    
    \begin{subfigure}[t]{0.24\textwidth}
        \centering
        \setlength\figureheight{\textwidth}
        \setlength\figurewidth{\textwidth}        \footnotesize{\input{tex/graphs_ml_gd/corner_2023-03-24-20-18-25_v2_plot_v2}}
        \caption{}
        \label{fig:pos_corner}
    \end{subfigure}
    \begin{subfigure}[t]{0.24\textwidth}
        \centering
        \setlength\figureheight{\textwidth}
        \setlength\figurewidth{\textwidth}        \footnotesize{\input{tex/graphs_ml_gd/corner_2023-03-24-20-18-25_v2_boxplot}}
        \caption{}
        \label{fig:error_corner}
    \end{subfigure}
    
    \begin{subfigure}[t]{0.24\textwidth}
        \centering
        \setlength\figureheight{\textwidth}
        \setlength\figurewidth{\textwidth}        \footnotesize{\input{tex/graphs_ml_gd/diagonal_2023-03-24-20-13-40_v2_plot_v2}}
        \caption{}
        \label{fig:pos_diagonal}
    \end{subfigure}
    \begin{subfigure}[t]{0.24\textwidth}
        \centering
        \setlength\figureheight{\textwidth}
        \setlength\figurewidth{\textwidth}        \footnotesize{\input{tex/graphs_ml_gd/diagonal_2023-03-24-20-13-40_v2_boxplot}}
        \caption{}
        \label{fig:error_diagonal}
    \end{subfigure}
    
    \caption{Localization results for the implementation of a multilateration algorithm and a Gradiant Descent Optimization algorithm for the three configurations presented in the dataset, alongside their absolute distance errors.}
    \label{fig:algorithm_results}
\vspace{-1.5em}
\end{figure}

\textit {Case III} presents scenarios with eight static UWB nodes, that provide the distances all-to-all as seen in \Cref{fig:configurations}. For each scenario we implemented two relative localization algorithms used in our previous research \cite{salimpour2023exploiting}. The first algorithm is multilateration (MLAT), which uses all the measured distances, and the second is Gradiante Descent optimization (GD) to minimize distance error. Both localization approaches are further detailed in \cite{salimpour2023exploiting}. \Cref{fig:algorithm_results} shows the results for both MLAT and GD for three scenarios. It also shows the error of each UWB node position according to the algorithm.

The three configurations were selected because they present challenges for localization solutions, such as aligning nodes or forming 90$^{\circ}$ angles. \Cref{fig:pos_army} shows how both algorithms present different results. GD performs noticeably better than MLAT. It can be more appreciated in \Cref{fig:error_army}, where most of the higher errors for each node are from MLAT. \Cref{fig:pos_corner} present more mixed results, with \Cref{fig:error_corner} showing that some nodes perform better with MLAT than GD and vice versa. Finally, \Cref{fig:pos_diagonal} and \Cref{fig:error_diagonal} show MLAT with better results than GD.

With these three UWB node formations, we can observe that different algorithms perform better or worse depending on the scenario. For this reason, we included different formations in the dataset, to analyze the advantages and disadvantages of various localization methods. Furthermore, it helps to study algorithms' performance in edge case scenarios like with all UWB nodes in a straight line.

%% file: tex/boxplot_errors/error_main_all_0.0_final_data.csv.tex
\begin{tikzpicture}

\definecolor{blue08255}{RGB}{0,8,255}
\definecolor{darkgray176}{RGB}{176,176,176}
\definecolor{deepskyblue0212255}{RGB}{0,212,255}
\definecolor{dodgerblue0144255}{RGB}{0,144,255}
\definecolor{dodgerblue076255}{RGB}{0,76,255}
\definecolor{gainsboro229}{RGB}{229,229,229}
\definecolor{gold2552290}{RGB}{255,229,0}
\definecolor{greenyellow20525541}{RGB}{205,255,41}
\definecolor{lightgreen15025595}{RGB}{150,255,95}
\definecolor{lightgreen95255150}{RGB}{95,255,150}
\definecolor{maroon12700}{RGB}{127,0,0}
\definecolor{mediumblue00204}{RGB}{0,0,204}
\definecolor{navy00127}{RGB}{0,0,127}
\definecolor{orange2551660}{RGB}{255,166,0}
\definecolor{orangered2551030}{RGB}{255,103,0}
\definecolor{orangered255400}{RGB}{255,40,0}
\definecolor{red20400}{RGB}{204,0,0}
\definecolor{turquoise41255205}{RGB}{41,255,205}

\begin{axis}[
    width=\figurewidth,
    height=\figureheight,
    axis background/.style={fill=white},
    axis line style={white},
    tick align=outside,
    x grid style={white},
    xmajorgrids,
    xmajorticks=true,
    y grid style={white},
    ylabel=\textcolor{darkslategray38}{Error (m)},
    ymajorgrids,
    ymajorticks=true,
    y grid style={white!69.0196078431373!black},
    ytick style={color=black},
    xmajorgrids,
    xminorgrids,
    ymajorgrids,
    ymajorticks=true,
    minor y tick num = 3,
    minor y grid style={dashed},
    yminorgrids,
    yticklabel style={
            /pgf/number format/fixed,
            /pgf/number format/precision=5
        },
    scaled y ticks=false,
    ylabel near ticks, 
    ylabel shift={-1pt},
xlabel={Distance (m)},
xmin=0.5, xmax=16.5,
xtick style={color=black},
xticklabel style={rotate=90.0},
y grid style={gainsboro229},
ylabel={Error (m)},
ymin=0, ymax=1,
]
\path [draw=black, fill=navy00127]
(axis cs:0.75,0.0939515829086303)
--(axis cs:1.25,0.0939515829086303)
--(axis cs:1.25,0.167529582977295)
--(axis cs:0.75,0.167529582977295)
--(axis cs:0.75,0.0939515829086303)
--cycle;
\addplot [black]
table {%
1 0.0939515829086303
1 0.0363695621490478
};
\addplot [black]
table {%
1 0.167529582977295
1 0.208806753158569
};
\addplot [black]
table {%
0.875 0.0363695621490478
1.125 0.0363695621490478
};
\addplot [black]
table {%
0.875 0.208806753158569
1.125 0.208806753158569
};
\path [draw=black, fill=mediumblue00204]
(axis cs:1.75,0.15045154094696)
--(axis cs:2.25,0.15045154094696)
--(axis cs:2.25,0.187937378883362)
--(axis cs:1.75,0.187937378883362)
--(axis cs:1.75,0.15045154094696)
--cycle;
\addplot [black]
table {%
2 0.15045154094696
2 0.113885641098022
};
\addplot [black]
table {%
2 0.187937378883362
2 0.209288120269775
};
\addplot [black]
table {%
1.875 0.113885641098022
2.125 0.113885641098022
};
\addplot [black]
table {%
1.875 0.209288120269775
2.125 0.209288120269775
};
\path [draw=black, fill=blue08255]
(axis cs:2.75,0.163645565509796)
--(axis cs:3.25,0.163645565509796)
--(axis cs:3.25,0.219897270202637)
--(axis cs:2.75,0.219897270202637)
--(axis cs:2.75,0.163645565509796)
--cycle;
\addplot [black]
table {%
3 0.163645565509796
3 0.107505559921265
};
\addplot [black]
table {%
3 0.219897270202637
3 0.262502431869507
};
\addplot [black]
table {%
2.875 0.107505559921265
3.125 0.107505559921265
};
\addplot [black]
table {%
2.875 0.262502431869507
3.125 0.262502431869507
};
\path [draw=black, fill=dodgerblue076255]
(axis cs:3.75,0.21666693687439)
--(axis cs:4.25,0.21666693687439)
--(axis cs:4.25,0.256927251815796)
--(axis cs:3.75,0.256927251815796)
--(axis cs:3.75,0.21666693687439)
--cycle;
\addplot [black]
table {%
4 0.21666693687439
4 0.178892135620117
};
\addplot [black]
table {%
4 0.256927251815796
4 0.295649766921997
};
\addplot [black]
table {%
3.875 0.178892135620117
4.125 0.178892135620117
};
\addplot [black]
table {%
3.875 0.295649766921997
4.125 0.295649766921997
};
\path [draw=black, fill=dodgerblue0144255]
(axis cs:4.75,0.248738765716553)
--(axis cs:5.25,0.248738765716553)
--(axis cs:5.25,0.284016609191895)
--(axis cs:4.75,0.284016609191895)
--(axis cs:4.75,0.248738765716553)
--cycle;
\addplot [black]
table {%
5 0.248738765716553
5 0.220467567443848
};
\addplot [black]
table {%
5 0.284016609191895
5 0.302874565124512
};
\addplot [black]
table {%
4.875 0.220467567443848
5.125 0.220467567443848
};
\addplot [black]
table {%
4.875 0.302874565124512
5.125 0.302874565124512
};
\path [draw=black, fill=deepskyblue0212255]
(axis cs:5.75,0.258313179016113)
--(axis cs:6.25,0.258313179016113)
--(axis cs:6.25,0.318399429321289)
--(axis cs:5.75,0.318399429321289)
--(axis cs:5.75,0.258313179016113)
--cycle;
\addplot [black]
table {%
6 0.258313179016113
6 0.201793670654297
};
\addplot [black]
table {%
6 0.318399429321289
6 0.375295162200928
};
\addplot [black]
table {%
5.875 0.201793670654297
6.125 0.201793670654297
};
\addplot [black]
table {%
5.875 0.375295162200928
6.125 0.375295162200928
};
\path [draw=black, fill=turquoise41255205]
(axis cs:6.75,0.254868507385254)
--(axis cs:7.25,0.254868507385254)
--(axis cs:7.25,0.292499542236328)
--(axis cs:6.75,0.292499542236328)
--(axis cs:6.75,0.254868507385254)
--cycle;
\addplot [black]
table {%
7 0.254868507385254
7 0.219118595123291
};
\addplot [black]
table {%
7 0.292499542236328
7 0.330017566680908
};
\addplot [black]
table {%
6.875 0.219118595123291
7.125 0.219118595123291
};
\addplot [black]
table {%
6.875 0.330017566680908
7.125 0.330017566680908
};
\path [draw=black, fill=lightgreen95255150]
(axis cs:7.75,0.237802147865295)
--(axis cs:8.25,0.237802147865295)
--(axis cs:8.25,0.28972053527832)
--(axis cs:7.75,0.28972053527832)
--(axis cs:7.75,0.237802147865295)
--cycle;
\addplot [black]
table {%
8 0.237802147865295
8 0.197441577911377
};
\addplot [black]
table {%
8 0.28972053527832
8 0.331719398498535
};
\addplot [black]
table {%
7.875 0.197441577911377
8.125 0.197441577911377
};
\addplot [black]
table {%
7.875 0.331719398498535
8.125 0.331719398498535
};
\path [draw=black, fill=lightgreen15025595]
(axis cs:8.75,0.314285516738892)
--(axis cs:9.25,0.314285516738892)
--(axis cs:9.25,0.606241226196289)
--(axis cs:8.75,0.606241226196289)
--(axis cs:8.75,0.314285516738892)
--cycle;
\addplot [black]
table {%
9 0.314285516738892
9 0.257161140441894
};
\addplot [black]
table {%
9 0.606241226196289
9 0.718889236450195
};
\addplot [black]
table {%
8.875 0.257161140441894
9.125 0.257161140441894
};
\addplot [black]
table {%
8.875 0.718889236450195
9.125 0.718889236450195
};
\path [draw=black, fill=greenyellow20525541]
(axis cs:9.75,0.310003519058227)
--(axis cs:10.25,0.310003519058227)
--(axis cs:10.25,0.383523464202881)
--(axis cs:9.75,0.383523464202881)
--(axis cs:9.75,0.310003519058227)
--cycle;
\addplot [black]
table {%
10 0.310003519058227
10 0.253786087036133
};
\addplot [black]
table {%
10 0.383523464202881
10 0.440066337585449
};
\addplot [black]
table {%
9.875 0.253786087036133
10.125 0.253786087036133
};
\addplot [black]
table {%
9.875 0.440066337585449
10.125 0.440066337585449
};
\path [draw=black, fill=gold2552290]
(axis cs:10.75,0.192888259887695)
--(axis cs:11.25,0.192888259887695)
--(axis cs:11.25,0.282618522644043)
--(axis cs:10.75,0.282618522644043)
--(axis cs:10.75,0.192888259887695)
--cycle;
\addplot [black]
table {%
11 0.192888259887695
11 0.11653995513916
};
\addplot [black]
table {%
11 0.282618522644043
11 0.341701507568359
};
\addplot [black]
table {%
10.875 0.11653995513916
11.125 0.11653995513916
};
\addplot [black]
table {%
10.875 0.341701507568359
11.125 0.341701507568359
};
\path [draw=black, fill=orange2551660]
(axis cs:11.75,0.349349975585938)
--(axis cs:12.25,0.349349975585938)
--(axis cs:12.25,0.394726753234863)
--(axis cs:11.75,0.394726753234863)
--(axis cs:11.75,0.349349975585938)
--cycle;
\addplot [black]
table {%
12 0.349349975585938
12 0.311010360717773
};
\addplot [black]
table {%
12 0.394726753234863
12 0.432326316833496
};
\addplot [black]
table {%
11.875 0.311010360717773
12.125 0.311010360717773
};
\addplot [black]
table {%
11.875 0.432326316833496
12.125 0.432326316833496
};
\path [draw=black, fill=orangered2551030]
(axis cs:12.75,0.315587043762207)
--(axis cs:13.25,0.315587043762207)
--(axis cs:13.25,0.446686267852783)
--(axis cs:12.75,0.446686267852783)
--(axis cs:12.75,0.315587043762207)
--cycle;
\addplot [black]
table {%
13 0.315587043762207
13 0.203019142150879
};
\addplot [black]
table {%
13 0.446686267852783
13 0.520779609680176
};
\addplot [black]
table {%
12.875 0.203019142150879
13.125 0.203019142150879
};
\addplot [black]
table {%
12.875 0.520779609680176
13.125 0.520779609680176
};
\path [draw=black, fill=orangered255400]
(axis cs:13.75,0.161033630371094)
--(axis cs:14.25,0.161033630371094)
--(axis cs:14.25,0.653582811355591)
--(axis cs:13.75,0.653582811355591)
--(axis cs:13.75,0.161033630371094)
--cycle;
\addplot [black]
table {%
14 0.161033630371094
14 -0.0453424453735351
};
\addplot [black]
table {%
14 0.653582811355591
14 0.838841438293457
};
\addplot [black]
table {%
13.875 -0.0453424453735351
14.125 -0.0453424453735351
};
\addplot [black]
table {%
13.875 0.838841438293457
14.125 0.838841438293457
};
\path [draw=black, fill=red20400]
(axis cs:14.75,0.307424545288086)
--(axis cs:15.25,0.307424545288086)
--(axis cs:15.25,0.344866752624512)
--(axis cs:14.75,0.344866752624512)
--(axis cs:14.75,0.307424545288086)
--cycle;
\addplot [black]
table {%
15 0.307424545288086
15 0.281405448913574
};
\addplot [black]
table {%
15 0.344866752624512
15 0.378260612487793
};
\addplot [black]
table {%
14.875 0.281405448913574
15.125 0.281405448913574
};
\addplot [black]
table {%
14.875 0.378260612487793
15.125 0.378260612487793
};
\path [draw=black, fill=maroon12700]
(axis cs:15.75,0.346213340759277)
--(axis cs:16.25,0.346213340759277)
--(axis cs:16.25,0.36876106262207)
--(axis cs:15.75,0.36876106262207)
--(axis cs:15.75,0.346213340759277)
--cycle;
\addplot [black]
table {%
16 0.346213340759277
16 0.324460983276367
};
\addplot [black]
table {%
16 0.36876106262207
16 0.387506484985352
};
\addplot [black]
table {%
15.875 0.324460983276367
16.125 0.324460983276367
};
\addplot [black]
table {%
15.875 0.387506484985352
16.125 0.387506484985352
};
\addplot [thick, black]
table {%
0.75 0.131473660469055
1.25 0.131473660469055
};
\addplot [thick, black]
table {%
1.75 0.169175267219543
2.25 0.169175267219543
};
\addplot [thick, black]
table {%
2.75 0.182641506195068
3.25 0.182641506195068
};
\addplot [thick, black]
table {%
3.75 0.235471487045288
4.25 0.235471487045288
};
\addplot [thick, black]
table {%
4.75 0.265215396881104
5.25 0.265215396881104
};
\addplot [thick, black]
table {%
5.75 0.277835845947266
6.25 0.277835845947266
};
\addplot [thick, black]
table {%
6.75 0.273738384246826
7.25 0.273738384246826
};
\addplot [thick, black]
table {%
7.75 0.259793281555176
8.25 0.259793281555176
};
\addplot [thick, black]
table {%
8.75 0.341696739196777
9.25 0.341696739196777
};
\addplot [thick, black]
table {%
9.75 0.346921920776367
10.25 0.346921920776367
};
\addplot [thick, black]
table {%
10.75 0.24500560760498
11.25 0.24500560760498
};
\addplot [thick, black]
table {%
11.75 0.375920295715332
12.25 0.375920295715332
};
\addplot [thick, black]
table {%
12.75 0.406766414642334
13.25 0.406766414642334
};
\addplot [thick, black]
table {%
13.75 0.42402172088623
14.25 0.42402172088623
};
\addplot [thick, black]
table {%
14.75 0.326098442077637
15.25 0.326098442077637
};
\addplot [thick, black]
table {%
15.75 0.362070083618164
16.25 0.362070083618164
};
\end{axis}

\end{tikzpicture}

%% file: tex/boxplot_errors/relative_error_main_all_0.0_rosbag_try1.csv.tex
\begin{tikzpicture}

\definecolor{blue08255}{RGB}{0,8,255}
\definecolor{darkgray176}{RGB}{176,176,176}
\definecolor{deepskyblue0212255}{RGB}{0,212,255}
\definecolor{dodgerblue0144255}{RGB}{0,144,255}
\definecolor{dodgerblue076255}{RGB}{0,76,255}
\definecolor{gainsboro229}{RGB}{229,229,229}
\definecolor{gold2552290}{RGB}{255,229,0}
\definecolor{greenyellow20525541}{RGB}{205,255,41}
\definecolor{lightgreen15025595}{RGB}{150,255,95}
\definecolor{lightgreen95255150}{RGB}{95,255,150}
\definecolor{maroon12700}{RGB}{127,0,0}
\definecolor{mediumblue00204}{RGB}{0,0,204}
\definecolor{navy00127}{RGB}{0,0,127}
\definecolor{orange2551660}{RGB}{255,166,0}
\definecolor{orangered2551030}{RGB}{255,103,0}
\definecolor{orangered255400}{RGB}{255,40,0}
\definecolor{red20400}{RGB}{204,0,0}
\definecolor{turquoise41255205}{RGB}{41,255,205}

\begin{axis}[
    width=\figurewidth,
    height=\figureheight,
    axis background/.style={fill=white},
    axis line style={white},
    tick align=outside,
    x grid style={white},
    xmajorgrids,
    xmajorticks=true,
    y grid style={white},
    ylabel=\textcolor{darkslategray38}{Error (m)},
    ymajorgrids,
    ymajorticks=true,
    y grid style={white!69.0196078431373!black},
    ytick style={color=black},
    xmajorgrids,
    xminorgrids,
    ymajorgrids,
    ymajorticks=true,
    minor y tick num = 3,
    minor y grid style={dashed},
    yminorgrids,
    yticklabel style={
            /pgf/number format/fixed,
            /pgf/number format/precision=5,
        },
    scaled y ticks=false,
    ylabel near ticks, 
    ylabel shift={-1pt},
xlabel={Distance (m)},
xmin=0.5, xmax=16.5,
xtick style={color=black},
xticklabel style={rotate=90.0},
y grid style={gainsboro229},
ylabel={Relative Error (\%)},
ymin=0, ymax=20,
]
\path [draw=black, fill=navy00127]
(axis cs:0.75,9.39515829086303)
--(axis cs:1.25,9.39515829086303)
--(axis cs:1.25,16.7529582977295)
--(axis cs:0.75,16.7529582977295)
--(axis cs:0.75,9.39515829086303)
--cycle;
\addplot [black]
table {%
1 9.39515829086303
1 3.63695621490478
};
\addplot [black]
table {%
1 16.7529582977295
1 20.8806753158569
};
\addplot [black]
table {%
0.875 3.63695621490478
1.125 3.63695621490478
};
\addplot [black]
table {%
0.875 20.8806753158569
1.125 20.8806753158569
};
\path [draw=black, fill=mediumblue00204]
(axis cs:1.75,7.52257704734802)
--(axis cs:2.25,7.52257704734802)
--(axis cs:2.25,9.39686894416809)
--(axis cs:1.75,9.39686894416809)
--(axis cs:1.75,7.52257704734802)
--cycle;
\addplot [black]
table {%
2 7.52257704734802
2 5.69428205490112
};
\addplot [black]
table {%
2 9.39686894416809
2 10.4644060134888
};
\addplot [black]
table {%
1.875 5.69428205490112
2.125 5.69428205490112
};
\addplot [black]
table {%
1.875 10.4644060134888
2.125 10.4644060134888
};
\path [draw=black, fill=blue08255]
(axis cs:2.75,5.45485218365987)
--(axis cs:3.25,5.45485218365987)
--(axis cs:3.25,7.32990900675456)
--(axis cs:2.75,7.32990900675456)
--(axis cs:2.75,5.45485218365987)
--cycle;
\addplot [black]
table {%
3 5.45485218365987
3 3.58351866404215
};
\addplot [black]
table {%
3 7.32990900675456
3 8.75008106231689
};
\addplot [black]
table {%
2.875 3.58351866404215
3.125 3.58351866404215
};
\addplot [black]
table {%
2.875 8.75008106231689
3.125 8.75008106231689
};
\path [draw=black, fill=dodgerblue076255]
(axis cs:3.75,5.41667342185974)
--(axis cs:4.25,5.41667342185974)
--(axis cs:4.25,6.4231812953949)
--(axis cs:3.75,6.4231812953949)
--(axis cs:3.75,5.41667342185974)
--cycle;
\addplot [black]
table {%
4 5.41667342185974
4 4.47230339050293
};
\addplot [black]
table {%
4 6.4231812953949
4 7.39124417304992
};
\addplot [black]
table {%
3.875 4.47230339050293
4.125 4.47230339050293
};
\addplot [black]
table {%
3.875 7.39124417304992
4.125 7.39124417304992
};
\path [draw=black, fill=dodgerblue0144255]
(axis cs:4.75,4.97477531433105)
--(axis cs:5.25,4.97477531433105)
--(axis cs:5.25,5.68033218383789)
--(axis cs:4.75,5.68033218383789)
--(axis cs:4.75,4.97477531433105)
--cycle;
\addplot [black]
table {%
5 4.97477531433105
5 4.40935134887695
};
\addplot [black]
table {%
5 5.68033218383789
5 6.05749130249023
};
\addplot [black]
table {%
4.875 4.40935134887695
5.125 4.40935134887695
};
\addplot [black]
table {%
4.875 6.05749130249023
5.125 6.05749130249023
};
\path [draw=black, fill=deepskyblue0212255]
(axis cs:5.75,4.30521965026855)
--(axis cs:6.25,4.30521965026855)
--(axis cs:6.25,5.30665715535482)
--(axis cs:5.75,5.30665715535482)
--(axis cs:5.75,4.30521965026855)
--cycle;
\addplot [black]
table {%
6 4.30521965026855
6 3.36322784423828
};
\addplot [black]
table {%
6 5.30665715535482
6 6.25491937001546
};
\addplot [black]
table {%
5.875 3.36322784423828
6.125 3.36322784423828
};
\addplot [black]
table {%
5.875 6.25491937001546
6.125 6.25491937001546
};
\path [draw=black, fill=turquoise41255205]
(axis cs:6.75,3.6409786769322)
--(axis cs:7.25,3.6409786769322)
--(axis cs:7.25,4.1785648890904)
--(axis cs:6.75,4.1785648890904)
--(axis cs:6.75,3.6409786769322)
--cycle;
\addplot [black]
table {%
7 3.6409786769322
7 3.13026564461844
};
\addplot [black]
table {%
7 4.1785648890904
7 4.71453666687012
};
\addplot [black]
table {%
6.875 3.13026564461844
7.125 3.13026564461844
};
\addplot [black]
table {%
6.875 4.71453666687012
7.125 4.71453666687012
};
\path [draw=black, fill=lightgreen95255150]
(axis cs:7.75,2.97252684831619)
--(axis cs:8.25,2.97252684831619)
--(axis cs:8.25,3.621506690979)
--(axis cs:7.75,3.621506690979)
--(axis cs:7.75,2.97252684831619)
--cycle;
\addplot [black]
table {%
8 2.97252684831619
8 2.46801972389221
};
\addplot [black]
table {%
8 3.621506690979
8 4.14649248123169
};
\addplot [black]
table {%
7.875 2.46801972389221
8.125 2.46801972389221
};
\addplot [black]
table {%
7.875 4.14649248123169
8.125 4.14649248123169
};
\path [draw=black, fill=lightgreen15025595]
(axis cs:8.75,3.49206129709879)
--(axis cs:9.25,3.49206129709879)
--(axis cs:9.25,6.73601362440321)
--(axis cs:8.75,6.73601362440321)
--(axis cs:8.75,3.49206129709879)
--cycle;
\addplot [black]
table {%
9 3.49206129709879
9 2.85734600490994
};
\addplot [black]
table {%
9 6.73601362440321
9 7.98765818277995
};
\addplot [black]
table {%
8.875 2.85734600490994
9.125 2.85734600490994
};
\addplot [black]
table {%
8.875 7.98765818277995
9.125 7.98765818277995
};
\path [draw=black, fill=greenyellow20525541]
(axis cs:9.75,3.10003519058227)
--(axis cs:10.25,3.10003519058227)
--(axis cs:10.25,3.83523464202881)
--(axis cs:9.75,3.83523464202881)
--(axis cs:9.75,3.10003519058227)
--cycle;
\addplot [black]
table {%
10 3.10003519058227
10 2.53786087036133
};
\addplot [black]
table {%
10 3.83523464202881
10 4.40066337585449
};
\addplot [black]
table {%
9.875 2.53786087036133
10.125 2.53786087036133
};
\addplot [black]
table {%
9.875 4.40066337585449
10.125 4.40066337585449
};
\path [draw=black, fill=gold2552290]
(axis cs:10.75,1.75352963534268)
--(axis cs:11.25,1.75352963534268)
--(axis cs:11.25,2.56925929676403)
--(axis cs:10.75,2.56925929676403)
--(axis cs:10.75,1.75352963534268)
--cycle;
\addplot [black]
table {%
11 1.75352963534268
11 1.05945413762873
};
\addplot [black]
table {%
11 2.56925929676403
11 3.10637734153054
};
\addplot [black]
table {%
10.875 1.05945413762873
11.125 1.05945413762873
};
\addplot [black]
table {%
10.875 3.10637734153054
11.125 3.10637734153054
};
\path [draw=black, fill=orange2551660]
(axis cs:11.75,2.91124979654948)
--(axis cs:12.25,2.91124979654948)
--(axis cs:12.25,3.28938961029053)
--(axis cs:11.75,3.28938961029053)
--(axis cs:11.75,2.91124979654948)
--cycle;
\addplot [black]
table {%
12 2.91124979654948
12 2.59175300598144
};
\addplot [black]
table {%
12 3.28938961029053
12 3.6027193069458
};
\addplot [black]
table {%
11.875 2.59175300598144
12.125 2.59175300598144
};
\addplot [black]
table {%
11.875 3.6027193069458
12.125 3.6027193069458
};
\path [draw=black, fill=orangered2551030]
(axis cs:12.75,2.42759264432467)
--(axis cs:13.25,2.42759264432467)
--(axis cs:13.25,3.43604821425218)
--(axis cs:12.75,3.43604821425218)
--(axis cs:12.75,2.42759264432467)
--cycle;
\addplot [black]
table {%
13 2.42759264432467
13 1.56168570885291
};
\addplot [black]
table {%
13 3.43604821425218
13 4.00599699753981
};
\addplot [black]
table {%
12.875 1.56168570885291
13.125 1.56168570885291
};
\addplot [black]
table {%
12.875 4.00599699753981
13.125 4.00599699753981
};
\path [draw=black, fill=orangered255400]
(axis cs:13.75,1.15024021693638)
--(axis cs:14.25,1.15024021693638)
--(axis cs:14.25,4.66844865253993)
--(axis cs:13.75,4.66844865253993)
--(axis cs:13.75,1.15024021693638)
--cycle;
\addplot [black]
table {%
14 1.15024021693638
14 -0.323874609810965
};
\addplot [black]
table {%
14 4.66844865253993
14 5.99172455923898
};
\addplot [black]
table {%
13.875 -0.323874609810965
14.125 -0.323874609810965
};
\addplot [black]
table {%
13.875 5.99172455923898
14.125 5.99172455923898
};
\path [draw=black, fill=red20400]
(axis cs:14.75,2.04949696858724)
--(axis cs:15.25,2.04949696858724)
--(axis cs:15.25,2.29911168416341)
--(axis cs:14.75,2.29911168416341)
--(axis cs:14.75,2.04949696858724)
--cycle;
\addplot [black]
table {%
15 2.04949696858724
15 1.87603632609049
};
\addplot [black]
table {%
15 2.29911168416341
15 2.52173741658529
};
\addplot [black]
table {%
14.875 1.87603632609049
15.125 1.87603632609049
};
\addplot [black]
table {%
14.875 2.52173741658529
15.125 2.52173741658529
};
\path [draw=black, fill=maroon12700]
(axis cs:15.75,2.16383337974548)
--(axis cs:16.25,2.16383337974548)
--(axis cs:16.25,2.30475664138794)
--(axis cs:15.75,2.30475664138794)
--(axis cs:15.75,2.16383337974548)
--cycle;
\addplot [black]
table {%
16 2.16383337974548
16 2.02788114547729
};
\addplot [black]
table {%
16 2.30475664138794
16 2.42191553115845
};
\addplot [black]
table {%
15.875 2.02788114547729
16.125 2.02788114547729
};
\addplot [black]
table {%
15.875 2.42191553115845
16.125 2.42191553115845
};
\addplot [thick, black]
table {%
0.75 13.1473660469055
1.25 13.1473660469055
};
\addplot [thick, black]
table {%
1.75 8.45876336097717
2.25 8.45876336097717
};
\addplot [thick, black]
table {%
2.75 6.08805020650228
3.25 6.08805020650228
};
\addplot [thick, black]
table {%
3.75 5.8867871761322
4.25 5.8867871761322
};
\addplot [thick, black]
table {%
4.75 5.30430793762207
5.25 5.30430793762207
};
\addplot [thick, black]
table {%
5.75 4.63059743245443
6.25 4.63059743245443
};
\addplot [thick, black]
table {%
6.75 3.91054834638323
7.25 3.91054834638323
};
\addplot [thick, black]
table {%
7.75 3.2474160194397
8.25 3.2474160194397
};
\addplot [thick, black]
table {%
8.75 3.79663043551975
9.25 3.79663043551975
};
\addplot [thick, black]
table {%
9.75 3.46921920776367
10.25 3.46921920776367
};
\addplot [thick, black]
table {%
10.75 2.22732370549982
11.25 2.22732370549982
};
\addplot [thick, black]
table {%
11.75 3.1326691309611
12.25 3.1326691309611
};
\addplot [thick, black]
table {%
12.75 3.12897242032565
13.25 3.12897242032565
};
\addplot [thick, black]
table {%
13.75 3.02872657775879
14.25 3.02872657775879
};
\addplot [thick, black]
table {%
14.75 2.17398961385091
15.25 2.17398961385091
};
\addplot [thick, black]
table {%
15.75 2.26293802261352
16.25 2.26293802261352
};
\end{axis}

\end{tikzpicture}

%% file: tex/boxplot_errors/error_ByAngle_yaw_90.0_rosbag_try1.csv.tex
\begin{tikzpicture}

\definecolor{blue08255}{RGB}{0,8,255}
\definecolor{darkgray176}{RGB}{176,176,176}
\definecolor{deepskyblue0212255}{RGB}{0,212,255}
\definecolor{dodgerblue0144255}{RGB}{0,144,255}
\definecolor{dodgerblue076255}{RGB}{0,76,255}
\definecolor{gainsboro229}{RGB}{229,229,229}
\definecolor{gold2552290}{RGB}{255,229,0}
\definecolor{greenyellow20525541}{RGB}{205,255,41}
\definecolor{lightgreen15025595}{RGB}{150,255,95}
\definecolor{lightgreen95255150}{RGB}{95,255,150}
\definecolor{maroon12700}{RGB}{127,0,0}
\definecolor{mediumblue00204}{RGB}{0,0,204}
\definecolor{navy00127}{RGB}{0,0,127}
\definecolor{orange2551660}{RGB}{255,166,0}
\definecolor{orangered2551030}{RGB}{255,103,0}
\definecolor{orangered255400}{RGB}{255,40,0}
\definecolor{red20400}{RGB}{204,0,0}
\definecolor{turquoise41255205}{RGB}{41,255,205}

\begin{axis}[
    width=\figurewidth,
    height=\figureheight,
    axis background/.style={fill=white},
    axis line style={white},
    tick align=outside,
    x grid style={white},
    xmajorgrids,
    xmajorticks=true,
    y grid style={white},
    ymajorgrids,
    ymajorticks=true,
    y grid style={white!69.0196078431373!black},
    ytick style={color=black},
    xmajorgrids,
    xminorgrids,
    ymajorgrids,
    ymajorticks=true,
    yticklabel style={
            /pgf/number format/fixed,
            /pgf/number format/precision=5
        },
    scaled y ticks=false,
    ylabel near ticks, 
    ylabel shift={-1pt},
    ymin=0, ymax=1,
    xmin=0.5, xmax=16.5,
    xtick style={color=black},
]
\path [draw=black, fill=navy00127]
(axis cs:0.75,0.151276469230652)
--(axis cs:1.25,0.151276469230652)
--(axis cs:1.25,0.170037388801575)
--(axis cs:0.75,0.170037388801575)
--(axis cs:0.75,0.151276469230652)
--cycle;
\addplot [black]
table {%
1 0.151276469230652
1 0.151276469230652
};
\addplot [black]
table {%
1 0.170037388801575
1 0.170037388801575
};
\addplot [black]
table {%
0.875 0.151276469230652
1.125 0.151276469230652
};
\addplot [black]
table {%
0.875 0.170037388801575
1.125 0.170037388801575
};
\path [draw=black, fill=mediumblue00204]
(axis cs:1.75,0.205047845840454)
--(axis cs:2.25,0.205047845840454)
--(axis cs:2.25,0.242479771375656)
--(axis cs:1.75,0.242479771375656)
--(axis cs:1.75,0.205047845840454)
--cycle;
\addplot [black]
table {%
2 0.205047845840454
2 0.185272216796875
};
\addplot [black]
table {%
2 0.242479771375656
2 0.279513478279114
};
\addplot [black]
table {%
1.875 0.185272216796875
2.125 0.185272216796875
};
\addplot [black]
table {%
1.875 0.279513478279114
2.125 0.279513478279114
};
\path [draw=black, fill=blue08255]
(axis cs:2.75,0.218278408050537)
--(axis cs:3.25,0.218278408050537)
--(axis cs:3.25,0.256028056144714)
--(axis cs:2.75,0.256028056144714)
--(axis cs:2.75,0.218278408050537)
--cycle;
\addplot [black]
table {%
3 0.218278408050537
3 0.180736780166626
};
\addplot [black]
table {%
3 0.256028056144714
3 0.293456315994263
};
\addplot [black]
table {%
2.875 0.180736780166626
3.125 0.180736780166626
};
\addplot [black]
table {%
2.875 0.293456315994263
3.125 0.293456315994263
};
\path [draw=black, fill=dodgerblue076255]
(axis cs:3.75,0.221178531646728)
--(axis cs:4.25,0.221178531646728)
--(axis cs:4.25,0.258544206619263)
--(axis cs:3.75,0.258544206619263)
--(axis cs:3.75,0.221178531646728)
--cycle;
\addplot [black]
table {%
4 0.221178531646728
4 0.202258825302124
};
\addplot [black]
table {%
4 0.258544206619263
4 0.277503490447998
};
\addplot [black]
table {%
3.875 0.202258825302124
4.125 0.202258825302124
};
\addplot [black]
table {%
3.875 0.277503490447998
4.125 0.277503490447998
};
\path [draw=black, fill=dodgerblue0144255]
(axis cs:4.75,0.303592324256897)
--(axis cs:5.25,0.303592324256897)
--(axis cs:5.25,0.340868473052979)
--(axis cs:4.75,0.340868473052979)
--(axis cs:4.75,0.303592324256897)
--cycle;
\addplot [black]
table {%
5 0.303592324256897
5 0.284705638885498
};
\addplot [black]
table {%
5 0.340868473052979
5 0.360273361206055
};
\addplot [black]
table {%
4.875 0.284705638885498
5.125 0.284705638885498
};
\addplot [black]
table {%
4.875 0.360273361206055
5.125 0.360273361206055
};
\path [draw=black, fill=deepskyblue0212255]
(axis cs:5.75,0.316893100738525)
--(axis cs:6.25,0.316893100738525)
--(axis cs:6.25,0.354359149932861)
--(axis cs:5.75,0.354359149932861)
--(axis cs:5.75,0.316893100738525)
--cycle;
\addplot [black]
table {%
6 0.316893100738525
6 0.279644966125488
};
\addplot [black]
table {%
6 0.354359149932861
6 0.373576641082764
};
\addplot [black]
table {%
5.875 0.279644966125488
6.125 0.279644966125488
};
\addplot [black]
table {%
5.875 0.373576641082764
6.125 0.373576641082764
};
\path [draw=black, fill=turquoise41255205]
(axis cs:6.75,0.348305225372314)
--(axis cs:7.25,0.348305225372314)
--(axis cs:7.25,0.385678768157959)
--(axis cs:6.75,0.385678768157959)
--(axis cs:6.75,0.348305225372314)
--cycle;
\addplot [black]
table {%
7 0.348305225372314
7 0.310953140258789
};
\addplot [black]
table {%
7 0.385678768157959
7 0.404809951782227
};
\addplot [black]
table {%
6.875 0.310953140258789
7.125 0.310953140258789
};
\addplot [black]
table {%
6.875 0.404809951782227
7.125 0.404809951782227
};
\path [draw=black, fill=lightgreen95255150]
(axis cs:7.75,0.309126377105713)
--(axis cs:8.25,0.309126377105713)
--(axis cs:8.25,0.346415042877197)
--(axis cs:7.75,0.346415042877197)
--(axis cs:7.75,0.309126377105713)
--cycle;
\addplot [black]
table {%
8 0.309126377105713
8 0.272423267364502
};
\addplot [black]
table {%
8 0.346415042877197
8 0.366321086883545
};
\addplot [black]
table {%
7.875 0.272423267364502
8.125 0.272423267364502
};
\addplot [black]
table {%
7.875 0.366321086883545
8.125 0.366321086883545
};
\path [draw=black, fill=lightgreen15025595]
(axis cs:8.75,0.32376766204834)
--(axis cs:9.25,0.32376766204834)
--(axis cs:9.25,0.36129093170166)
--(axis cs:8.75,0.36129093170166)
--(axis cs:8.75,0.32376766204834)
--cycle;
\addplot [black]
table {%
9 0.32376766204834
9 0.286271095275879
};
\addplot [black]
table {%
9 0.36129093170166
9 0.398811340332031
};
\addplot [black]
table {%
8.875 0.286271095275879
9.125 0.286271095275879
};
\addplot [black]
table {%
8.875 0.398811340332031
9.125 0.398811340332031
};
\path [draw=black, fill=greenyellow20525541]
(axis cs:9.75,0.327759742736816)
--(axis cs:10.25,0.327759742736816)
--(axis cs:10.25,0.365158081054688)
--(axis cs:9.75,0.365158081054688)
--(axis cs:9.75,0.327759742736816)
--cycle;
\addplot [black]
table {%
10 0.327759742736816
10 0.290407180786133
};
\addplot [black]
table {%
10 0.365158081054688
10 0.384224891662598
};
\addplot [black]
table {%
9.875 0.290407180786133
10.125 0.290407180786133
};
\addplot [black]
table {%
9.875 0.384224891662598
10.125 0.384224891662598
};
\path [draw=black, fill=gold2552290]
(axis cs:10.75,0.367586374282837)
--(axis cs:11.25,0.367586374282837)
--(axis cs:11.25,0.387588739395142)
--(axis cs:10.75,0.387588739395142)
--(axis cs:10.75,0.367586374282837)
--cycle;
\addplot [black]
table {%
11 0.367586374282837
11 0.348966598510742
};
\addplot [black]
table {%
11 0.387588739395142
11 0.406682968139648
};
\addplot [black]
table {%
10.875 0.348966598510742
11.125 0.348966598510742
};
\addplot [black]
table {%
10.875 0.406682968139648
11.125 0.406682968139648
};
\path [draw=black, fill=orange2551660]
(axis cs:11.75,0.415218353271484)
--(axis cs:12.25,0.415218353271484)
--(axis cs:12.25,0.452709436416626)
--(axis cs:11.75,0.452709436416626)
--(axis cs:11.75,0.415218353271484)
--cycle;
\addplot [black]
table {%
12 0.415218353271484
12 0.377727508544922
};
\addplot [black]
table {%
12 0.452709436416626
12 0.471986770629883
};
\addplot [black]
table {%
11.875 0.377727508544922
12.125 0.377727508544922
};
\addplot [black]
table {%
11.875 0.471986770629883
12.125 0.471986770629883
};
\path [draw=black, fill=orangered2551030]
(axis cs:12.75,0.446420669555664)
--(axis cs:13.25,0.446420669555664)
--(axis cs:13.25,0.483931541442871)
--(axis cs:12.75,0.483931541442871)
--(axis cs:12.75,0.446420669555664)
--cycle;
\addplot [black]
table {%
13 0.446420669555664
13 0.427586555480957
};
\addplot [black]
table {%
13 0.483931541442871
13 0.502775192260742
};
\addplot [black]
table {%
12.875 0.427586555480957
13.125 0.427586555480957
};
\addplot [black]
table {%
12.875 0.502775192260742
13.125 0.502775192260742
};
\path [draw=black, fill=orangered255400]
(axis cs:13.75,0.259550094604492)
--(axis cs:14.25,0.259550094604492)
--(axis cs:14.25,0.296966552734375)
--(axis cs:13.75,0.296966552734375)
--(axis cs:13.75,0.259550094604492)
--cycle;
\addplot [black]
table {%
14 0.259550094604492
14 0.223390579223633
};
\addplot [black]
table {%
14 0.296966552734375
14 0.318221092224121
};
\addplot [black]
table {%
13.875 0.223390579223633
14.125 0.223390579223633
};
\addplot [black]
table {%
13.875 0.318221092224121
14.125 0.318221092224121
};
\path [draw=black, fill=red20400]
(axis cs:14.75,0.400276184082031)
--(axis cs:15.25,0.400276184082031)
--(axis cs:15.25,0.437798500061035)
--(axis cs:14.75,0.437798500061035)
--(axis cs:14.75,0.400276184082031)
--cycle;
\addplot [black]
table {%
15 0.400276184082031
15 0.381514549255371
};
\addplot [black]
table {%
15 0.437798500061035
15 0.456561088562012
};
\addplot [black]
table {%
14.875 0.381514549255371
15.125 0.381514549255371
};
\addplot [black]
table {%
14.875 0.456561088562012
15.125 0.456561088562012
};
\path [draw=black, fill=maroon12700]
(axis cs:15.75,0.419870853424072)
--(axis cs:16.25,0.419870853424072)
--(axis cs:16.25,0.457216262817383)
--(axis cs:15.75,0.457216262817383)
--(axis cs:15.75,0.419870853424072)
--cycle;
\addplot [black]
table {%
16 0.419870853424072
16 0.400503158569336
};
\addplot [black]
table {%
16 0.457216262817383
16 0.494474411010742
};
\addplot [black]
table {%
15.875 0.400503158569336
16.125 0.400503158569336
};
\addplot [black]
table {%
15.875 0.494474411010742
16.125 0.494474411010742
};
\addplot [thick, black]
table {%
0.75 0.170037388801575
1.25 0.170037388801575
};
\addplot [thick, black]
table {%
1.75 0.223622441291809
2.25 0.223622441291809
};
\addplot [thick, black]
table {%
2.75 0.237431049346924
3.25 0.237431049346924
};
\addplot [thick, black]
table {%
3.75 0.239895582199097
4.25 0.239895582199097
};
\addplot [thick, black]
table {%
4.75 0.322242259979248
5.25 0.322242259979248
};
\addplot [thick, black]
table {%
5.75 0.335741519927978
6.25 0.335741519927978
};
\addplot [thick, black]
table {%
6.75 0.367187023162842
7.25 0.367187023162842
};
\addplot [thick, black]
table {%
7.75 0.327884197235107
8.25 0.327884197235107
};
\addplot [thick, black]
table {%
8.75 0.342523574829102
9.25 0.342523574829102
};
\addplot [thick, black]
table {%
9.75 0.346479415893555
10.25 0.346479415893555
};
\addplot [thick, black]
table {%
10.75 0.38414716720581
11.25 0.38414716720581
};
\addplot [thick, black]
table {%
11.75 0.432436943054199
12.25 0.432436943054199
};
\addplot [thick, black]
table {%
12.75 0.465171337127685
13.25 0.465171337127685
};
\addplot [thick, black]
table {%
13.75 0.278409004211426
14.25 0.278409004211426
};
\addplot [thick, black]
table {%
14.75 0.419037818908691
15.25 0.419037818908691
};
\addplot [thick, black]
table {%
15.75 0.438789367675781
16.25 0.438789367675781
};
\end{axis}

\end{tikzpicture}

%% file: tex/boxplot_errors/error_ByAngle_yaw_180.0_rosbag_try1.csv.tex
\begin{tikzpicture}

\definecolor{blue08255}{RGB}{0,8,255}
\definecolor{darkgray176}{RGB}{176,176,176}
\definecolor{deepskyblue0212255}{RGB}{0,212,255}
\definecolor{dodgerblue0144255}{RGB}{0,144,255}
\definecolor{dodgerblue076255}{RGB}{0,76,255}
\definecolor{gainsboro229}{RGB}{229,229,229}
\definecolor{gold2552290}{RGB}{255,229,0}
\definecolor{greenyellow20525541}{RGB}{205,255,41}
\definecolor{lightgreen15025595}{RGB}{150,255,95}
\definecolor{lightgreen95255150}{RGB}{95,255,150}
\definecolor{maroon12700}{RGB}{127,0,0}
\definecolor{mediumblue00204}{RGB}{0,0,204}
\definecolor{navy00127}{RGB}{0,0,127}
\definecolor{orange2551660}{RGB}{255,166,0}
\definecolor{orangered2551030}{RGB}{255,103,0}
\definecolor{orangered255400}{RGB}{255,40,0}
\definecolor{red20400}{RGB}{204,0,0}
\definecolor{turquoise41255205}{RGB}{41,255,205}

\begin{axis}[
    width=\figurewidth,
    height=\figureheight,
    axis background/.style={fill=white},
    axis line style={white},
    tick align=outside,
    x grid style={white},
    xmajorgrids,
    xmajorticks=true,
    y grid style={white},
    ymajorgrids,
    ymajorticks=true,
    y grid style={white!69.0196078431373!black},
    ytick style={color=black},
    xmajorgrids,
    xminorgrids,
    ymajorgrids,
    ymajorticks=true,
    yticklabel style={
            /pgf/number format/fixed,
            /pgf/number format/precision=5
        },
    scaled y ticks=false,
    ylabel near ticks, 
    ylabel shift={-1pt},
    ymin=0, ymax=1,
    xmin=0.5, xmax=16.5,
    xtick style={color=black},
]
\path [draw=black, fill=navy00127]
(axis cs:0.75,0.169003576040268)
--(axis cs:1.25,0.169003576040268)
--(axis cs:1.25,0.210834801197052)
--(axis cs:0.75,0.210834801197052)
--(axis cs:0.75,0.169003576040268)
--cycle;
\addplot [black]
table {%
1 0.169003576040268
1 0.131485342979431
};
\addplot [black]
table {%
1 0.210834801197052
1 0.244035601615906
};
\addplot [black]
table {%
0.875 0.131485342979431
1.125 0.131485342979431
};
\addplot [black]
table {%
0.875 0.244035601615906
1.125 0.244035601615906
};
\path [draw=black, fill=mediumblue00204]
(axis cs:1.75,0.209091186523438)
--(axis cs:2.25,0.209091186523438)
--(axis cs:2.25,0.246614217758179)
--(axis cs:1.75,0.246614217758179)
--(axis cs:1.75,0.209091186523438)
--cycle;
\addplot [black]
table {%
2 0.209091186523438
2 0.171568155288696
};
\addplot [black]
table {%
2 0.246614217758179
2 0.265375137329102
};
\addplot [black]
table {%
1.875 0.171568155288696
2.125 0.171568155288696
};
\addplot [black]
table {%
1.875 0.265375137329102
2.125 0.265375137329102
};
\path [draw=black, fill=blue08255]
(axis cs:2.75,0.276246070861816)
--(axis cs:3.25,0.276246070861816)
--(axis cs:3.25,0.295343995094299)
--(axis cs:2.75,0.295343995094299)
--(axis cs:2.75,0.276246070861816)
--cycle;
\addplot [black]
table {%
3 0.276246070861816
3 0.257467269897461
};
\addplot [black]
table {%
3 0.295343995094299
3 0.314370632171631
};
\addplot [black]
table {%
2.875 0.257467269897461
3.125 0.257467269897461
};
\addplot [black]
table {%
2.875 0.314370632171631
3.125 0.314370632171631
};
\path [draw=black, fill=dodgerblue076255]
(axis cs:3.75,0.254978656768799)
--(axis cs:4.25,0.254978656768799)
--(axis cs:4.25,0.2925124168396)
--(axis cs:3.75,0.2925124168396)
--(axis cs:3.75,0.254978656768799)
--cycle;
\addplot [black]
table {%
4 0.254978656768799
4 0.217467784881592
};
\addplot [black]
table {%
4 0.2925124168396
4 0.329404354095459
};
\addplot [black]
table {%
3.875 0.217467784881592
4.125 0.217467784881592
};
\addplot [black]
table {%
3.875 0.329404354095459
4.125 0.329404354095459
};
\path [draw=black, fill=dodgerblue0144255]
(axis cs:4.75,0.310352802276611)
--(axis cs:5.25,0.310352802276611)
--(axis cs:5.25,0.331609725952148)
--(axis cs:4.75,0.331609725952148)
--(axis cs:4.75,0.310352802276611)
--cycle;
\addplot [black]
table {%
5 0.310352802276611
5 0.291591167449951
};
\addplot [black]
table {%
5 0.331609725952148
5 0.351088047027588
};
\addplot [black]
table {%
4.875 0.291591167449951
5.125 0.291591167449951
};
\addplot [black]
table {%
4.875 0.351088047027588
5.125 0.351088047027588
};
\path [draw=black, fill=deepskyblue0212255]
(axis cs:5.75,0.372761726379394)
--(axis cs:6.25,0.372761726379394)
--(axis cs:6.25,0.410154461860657)
--(axis cs:5.75,0.410154461860657)
--(axis cs:5.75,0.372761726379394)
--cycle;
\addplot [black]
table {%
6 0.372761726379394
6 0.353710651397705
};
\addplot [black]
table {%
6 0.410154461860657
6 0.429212093353271
};
\addplot [black]
table {%
5.875 0.353710651397705
6.125 0.353710651397705
};
\addplot [black]
table {%
5.875 0.429212093353271
6.125 0.429212093353271
};
\path [draw=black, fill=turquoise41255205]
(axis cs:6.75,0.345198512077332)
--(axis cs:7.25,0.345198512077332)
--(axis cs:7.25,0.370041847229004)
--(axis cs:6.75,0.370041847229004)
--(axis cs:6.75,0.345198512077332)
--cycle;
\addplot [black]
table {%
7 0.345198512077332
7 0.33079719543457
};
\addplot [black]
table {%
7 0.370041847229004
7 0.387407779693604
};
\addplot [black]
table {%
6.875 0.33079719543457
7.125 0.33079719543457
};
\addplot [black]
table {%
6.875 0.387407779693604
7.125 0.387407779693604
};
\path [draw=black, fill=lightgreen95255150]
(axis cs:7.75,0.367352485656738)
--(axis cs:8.25,0.367352485656738)
--(axis cs:8.25,0.386446952819824)
--(axis cs:7.75,0.386446952819824)
--(axis cs:7.75,0.367352485656738)
--cycle;
\addplot [black]
table {%
8 0.367352485656738
8 0.348299026489258
};
\addplot [black]
table {%
8 0.386446952819824
8 0.405209541320801
};
\addplot [black]
table {%
7.875 0.348299026489258
8.125 0.348299026489258
};
\addplot [black]
table {%
7.875 0.405209541320801
8.125 0.405209541320801
};
\path [draw=black, fill=lightgreen15025595]
(axis cs:8.75,0.399189949035644)
--(axis cs:9.25,0.399189949035644)
--(axis cs:9.25,0.436931610107422)
--(axis cs:8.75,0.436931610107422)
--(axis cs:8.75,0.399189949035644)
--cycle;
\addplot [black]
table {%
9 0.399189949035644
9 0.361810684204102
};
\addplot [black]
table {%
9 0.436931610107422
9 0.474495887756348
};
\addplot [black]
table {%
8.875 0.361810684204102
9.125 0.361810684204102
};
\addplot [black]
table {%
8.875 0.474495887756348
9.125 0.474495887756348
};
\path [draw=black, fill=greenyellow20525541]
(axis cs:9.75,0.422271728515625)
--(axis cs:10.25,0.422271728515625)
--(axis cs:10.25,0.459336280822754)
--(axis cs:9.75,0.459336280822754)
--(axis cs:9.75,0.422271728515625)
--cycle;
\addplot [black]
table {%
10 0.422271728515625
10 0.402953147888184
};
\addplot [black]
table {%
10 0.459336280822754
10 0.478556632995605
};
\addplot [black]
table {%
9.875 0.402953147888184
10.125 0.402953147888184
};
\addplot [black]
table {%
9.875 0.478556632995605
10.125 0.478556632995605
};
\path [draw=black, fill=gold2552290]
(axis cs:10.75,0.385303497314453)
--(axis cs:11.25,0.385303497314453)
--(axis cs:11.25,0.422825813293457)
--(axis cs:10.75,0.422825813293457)
--(axis cs:10.75,0.385303497314453)
--cycle;
\addplot [black]
table {%
11 0.385303497314453
11 0.366541862487793
};
\addplot [black]
table {%
11 0.422825813293457
11 0.441588401794434
};
\addplot [black]
table {%
10.875 0.366541862487793
11.125 0.366541862487793
};
\addplot [black]
table {%
10.875 0.441588401794434
11.125 0.441588401794434
};
\path [draw=black, fill=orange2551660]
(axis cs:11.75,0.434103965759277)
--(axis cs:12.25,0.434103965759277)
--(axis cs:12.25,0.471814870834351)
--(axis cs:11.75,0.471814870834351)
--(axis cs:11.75,0.434103965759277)
--cycle;
\addplot [black]
table {%
12 0.434103965759277
12 0.396461486816406
};
\addplot [black]
table {%
12 0.471814870834351
12 0.509400367736816
};
\addplot [black]
table {%
11.875 0.396461486816406
12.125 0.396461486816406
};
\addplot [black]
table {%
11.875 0.509400367736816
12.125 0.509400367736816
};
\path [draw=black, fill=orangered2551030]
(axis cs:12.75,0.390393257141113)
--(axis cs:13.25,0.390393257141113)
--(axis cs:13.25,0.446678161621094)
--(axis cs:12.75,0.446678161621094)
--(axis cs:12.75,0.390393257141113)
--cycle;
\addplot [black]
table {%
13 0.390393257141113
13 0.352425575256348
};
\addplot [black]
table {%
13 0.446678161621094
13 0.484200477600098
};
\addplot [black]
table {%
12.875 0.352425575256348
13.125 0.352425575256348
};
\addplot [black]
table {%
12.875 0.484200477600098
13.125 0.484200477600098
};
\path [draw=black, fill=orangered255400]
(axis cs:13.75,0.389766693115234)
--(axis cs:14.25,0.389766693115234)
--(axis cs:14.25,0.408528327941894)
--(axis cs:13.75,0.408528327941894)
--(axis cs:13.75,0.389766693115234)
--cycle;
\addplot [black]
table {%
14 0.389766693115234
14 0.389766693115234
};
\addplot [black]
table {%
14 0.408528327941894
14 0.408528327941894
};
\addplot [black]
table {%
13.875 0.389766693115234
14.125 0.389766693115234
};
\addplot [black]
table {%
13.875 0.408528327941894
14.125 0.408528327941894
};
\path [draw=black, fill=red20400]
(axis cs:14.75,0.470046043395996)
--(axis cs:15.25,0.470046043395996)
--(axis cs:15.25,0.507569313049316)
--(axis cs:14.75,0.507569313049316)
--(axis cs:14.75,0.470046043395996)
--cycle;
\addplot [black]
table {%
15 0.470046043395996
15 0.432522773742676
};
\addplot [black]
table {%
15 0.507569313049316
15 0.54509162902832
};
\addplot [black]
table {%
14.875 0.432522773742676
15.125 0.432522773742676
};
\addplot [black]
table {%
14.875 0.54509162902832
15.125 0.54509162902832
};
\path [draw=black, fill=maroon12700]
(axis cs:15.75,0.306909561157227)
--(axis cs:16.25,0.306909561157227)
--(axis cs:16.25,0.850688934326172)
--(axis cs:15.75,0.850688934326172)
--(axis cs:15.75,0.306909561157227)
--cycle;
\addplot [black]
table {%
16 0.306909561157227
16 0.13774299621582
};
\addplot [black]
table {%
16 0.850688934326172
16 0.963647842407227
};
\addplot [black]
table {%
15.875 0.13774299621582
16.125 0.13774299621582
};
\addplot [black]
table {%
15.875 0.963647842407227
16.125 0.963647842407227
};
\addplot [thick, black]
table {%
0.75 0.187781810760498
1.25 0.187781810760498
};
\addplot [thick, black]
table {%
1.75 0.227852344512939
2.25 0.227852344512939
};
\addplot [thick, black]
table {%
2.75 0.294990301132202
3.25 0.294990301132202
};
\addplot [thick, black]
table {%
3.75 0.273701548576355
4.25 0.273701548576355
};
\addplot [thick, black]
table {%
4.75 0.31316328048706
5.25 0.31316328048706
};
\addplot [thick, black]
table {%
5.75 0.391455173492432
6.25 0.391455173492432
};
\addplot [thick, black]
table {%
6.75 0.368241310119629
7.25 0.368241310119629
};
\addplot [thick, black]
table {%
7.75 0.385820388793945
8.25 0.385820388793945
};
\addplot [thick, black]
table {%
8.75 0.418139457702637
9.25 0.418139457702637
};
\addplot [thick, black]
table {%
9.75 0.45008373260498
10.25 0.45008373260498
};
\addplot [thick, black]
table {%
10.75 0.404065132141113
11.25 0.404065132141113
};
\addplot [thick, black]
table {%
11.75 0.452930450439453
12.25 0.452930450439453
};
\addplot [thick, black]
table {%
12.75 0.427206039428711
13.25 0.427206039428711
};
\addplot [thick, black]
table {%
13.75 0.389766693115234
14.25 0.389766693115234
};
\addplot [thick, black]
table {%
14.75 0.48880672454834
15.25 0.48880672454834
};
\addplot [thick, black]
table {%
15.75 0.794305801391602
16.25 0.794305801391602
};
\end{axis}

\end{tikzpicture}

%% file: tex/boxplot_errors/error_ByAngle_yaw_270.0_rosbag_try1.csv.tex
\begin{tikzpicture}

\definecolor{blue08255}{RGB}{0,8,255}
\definecolor{darkgray176}{RGB}{176,176,176}
\definecolor{deepskyblue0212255}{RGB}{0,212,255}
\definecolor{dodgerblue0144255}{RGB}{0,144,255}
\definecolor{dodgerblue076255}{RGB}{0,76,255}
\definecolor{gainsboro229}{RGB}{229,229,229}
\definecolor{gold2552290}{RGB}{255,229,0}
\definecolor{greenyellow20525541}{RGB}{205,255,41}
\definecolor{lightgreen15025595}{RGB}{150,255,95}
\definecolor{lightgreen95255150}{RGB}{95,255,150}
\definecolor{maroon12700}{RGB}{127,0,0}
\definecolor{mediumblue00204}{RGB}{0,0,204}
\definecolor{navy00127}{RGB}{0,0,127}
\definecolor{orange2551660}{RGB}{255,166,0}
\definecolor{orangered2551030}{RGB}{255,103,0}
\definecolor{orangered255400}{RGB}{255,40,0}
\definecolor{red20400}{RGB}{204,0,0}
\definecolor{turquoise41255205}{RGB}{41,255,205}

\begin{axis}[
    width=\figurewidth,
    height=\figureheight,
    axis background/.style={fill=white},
    axis line style={white},
    tick align=outside,
    x grid style={white},
    xmajorgrids,
    xmajorticks=true,
    y grid style={white},
    ymajorgrids,
    ymajorticks=true,
    y grid style={white!69.0196078431373!black},
    ytick style={color=black},
    xmajorgrids,
    xminorgrids,
    ymajorgrids,
    ymajorticks=true,
    yticklabel style={
            /pgf/number format/fixed,
            /pgf/number format/precision=5
        },
    scaled y ticks=false,
    ylabel near ticks, 
    ylabel shift={-1pt},
    ymin=0, ymax=1,
    xmin=0.5, xmax=16.5,
    xtick style={color=black},
]
\path [draw=black, fill=navy00127]
(axis cs:0.75,0.0931816101074218)
--(axis cs:1.25,0.0931816101074218)
--(axis cs:1.25,0.149243950843811)
--(axis cs:0.75,0.149243950843811)
--(axis cs:0.75,0.0931816101074218)
--cycle;
\addplot [black]
table {%
1 0.0931816101074218
1 0.0741117000579834
};
\addplot [black]
table {%
1 0.149243950843811
1 0.187052965164184
};
\addplot [black]
table {%
0.875 0.0741117000579834
1.125 0.0741117000579834
};
\addplot [black]
table {%
0.875 0.187052965164184
1.125 0.187052965164184
};
\path [draw=black, fill=mediumblue00204]
(axis cs:1.75,0.150708675384521)
--(axis cs:2.25,0.150708675384521)
--(axis cs:2.25,0.188276410102844)
--(axis cs:1.75,0.188276410102844)
--(axis cs:1.75,0.150708675384521)
--cycle;
\addplot [black]
table {%
2 0.150708675384521
2 0.113214492797851
};
\addplot [black]
table {%
2 0.188276410102844
2 0.225782632827759
};
\addplot [black]
table {%
1.875 0.113214492797851
2.125 0.113214492797851
};
\addplot [black]
table {%
1.875 0.225782632827759
2.125 0.225782632827759
};
\path [draw=black, fill=blue08255]
(axis cs:2.75,0.17754590511322)
--(axis cs:3.25,0.17754590511322)
--(axis cs:3.25,0.220035791397095)
--(axis cs:2.75,0.220035791397095)
--(axis cs:2.75,0.17754590511322)
--cycle;
\addplot [black]
table {%
3 0.17754590511322
3 0.144621849060059
};
\addplot [black]
table {%
3 0.220035791397095
3 0.257463693618774
};
\addplot [black]
table {%
2.875 0.144621849060059
3.125 0.144621849060059
};
\addplot [black]
table {%
2.875 0.257463693618774
3.125 0.257463693618774
};
\path [draw=black, fill=dodgerblue076255]
(axis cs:3.75,0.21785306930542)
--(axis cs:4.25,0.21785306930542)
--(axis cs:4.25,0.255386114120483)
--(axis cs:3.75,0.255386114120483)
--(axis cs:3.75,0.21785306930542)
--cycle;
\addplot [black]
table {%
4 0.21785306930542
4 0.197515726089478
};
\addplot [black]
table {%
4 0.255386114120483
4 0.29286527633667
};
\addplot [black]
table {%
3.875 0.197515726089478
4.125 0.197515726089478
};
\addplot [black]
table {%
3.875 0.29286527633667
4.125 0.29286527633667
};
\path [draw=black, fill=dodgerblue0144255]
(axis cs:4.75,0.208848476409912)
--(axis cs:5.25,0.208848476409912)
--(axis cs:5.25,0.246052742004395)
--(axis cs:4.75,0.246052742004395)
--(axis cs:4.75,0.208848476409912)
--cycle;
\addplot [black]
table {%
5 0.208848476409912
5 0.189638614654541
};
\addplot [black]
table {%
5 0.246052742004395
5 0.265058040618896
};
\addplot [black]
table {%
4.875 0.189638614654541
5.125 0.189638614654541
};
\addplot [black]
table {%
4.875 0.265058040618896
5.125 0.265058040618896
};
\path [draw=black, fill=deepskyblue0212255]
(axis cs:5.75,0.222305417060852)
--(axis cs:6.25,0.222305417060852)
--(axis cs:6.25,0.259764671325684)
--(axis cs:5.75,0.259764671325684)
--(axis cs:5.75,0.222305417060852)
--cycle;
\addplot [black]
table {%
6 0.222305417060852
6 0.20322847366333
};
\addplot [black]
table {%
6 0.259764671325684
6 0.278656005859375
};
\addplot [black]
table {%
5.875 0.20322847366333
6.125 0.20322847366333
};
\addplot [black]
table {%
5.875 0.278656005859375
6.125 0.278656005859375
};
\path [draw=black, fill=turquoise41255205]
(axis cs:6.75,0.293770313262939)
--(axis cs:7.25,0.293770313262939)
--(axis cs:7.25,0.331280708312988)
--(axis cs:6.75,0.331280708312988)
--(axis cs:6.75,0.293770313262939)
--cycle;
\addplot [black]
table {%
7 0.293770313262939
7 0.274811267852783
};
\addplot [black]
table {%
7 0.331280708312988
7 0.351807117462158
};
\addplot [black]
table {%
6.875 0.274811267852783
7.125 0.274811267852783
};
\addplot [black]
table {%
6.875 0.351807117462158
7.125 0.351807117462158
};
\path [draw=black, fill=lightgreen95255150]
(axis cs:7.75,0.270325064659119)
--(axis cs:8.25,0.270325064659119)
--(axis cs:8.25,0.328456878662109)
--(axis cs:7.75,0.328456878662109)
--(axis cs:7.75,0.270325064659119)
--cycle;
\addplot [black]
table {%
8 0.270325064659119
8 0.215924739837646
};
\addplot [black]
table {%
8 0.328456878662109
8 0.36602783203125
};
\addplot [black]
table {%
7.875 0.215924739837646
8.125 0.215924739837646
};
\addplot [black]
table {%
7.875 0.36602783203125
8.125 0.36602783203125
};
\path [draw=black, fill=lightgreen15025595]
(axis cs:8.75,0.286199569702148)
--(axis cs:9.25,0.286199569702148)
--(axis cs:9.25,0.323720932006836)
--(axis cs:8.75,0.323720932006836)
--(axis cs:8.75,0.286199569702148)
--cycle;
\addplot [black]
table {%
9 0.286199569702148
9 0.248752593994141
};
\addplot [black]
table {%
9 0.323720932006836
9 0.34261417388916
};
\addplot [black]
table {%
8.875 0.248752593994141
9.125 0.248752593994141
};
\addplot [black]
table {%
8.875 0.34261417388916
9.125 0.34261417388916
};
\path [draw=black, fill=greenyellow20525541]
(axis cs:9.75,0.327962160110474)
--(axis cs:10.25,0.327962160110474)
--(axis cs:10.25,0.346859455108643)
--(axis cs:9.75,0.346859455108643)
--(axis cs:9.75,0.327962160110474)
--cycle;
\addplot [black]
table {%
10 0.327962160110474
10 0.309122085571289
};
\addplot [black]
table {%
10 0.346859455108643
10 0.365617752075195
};
\addplot [black]
table {%
9.875 0.309122085571289
10.125 0.309122085571289
};
\addplot [black]
table {%
9.875 0.365617752075195
10.125 0.365617752075195
};
\path [draw=black, fill=gold2552290]
(axis cs:10.75,0.367902755737305)
--(axis cs:11.25,0.367902755737305)
--(axis cs:11.25,0.405446767807007)
--(axis cs:10.75,0.405446767807007)
--(axis cs:10.75,0.367902755737305)
--cycle;
\addplot [black]
table {%
11 0.367902755737305
11 0.330379486083984
};
\addplot [black]
table {%
11 0.405446767807007
11 0.424357414245605
};
\addplot [black]
table {%
10.875 0.330379486083984
11.125 0.330379486083984
};
\addplot [black]
table {%
10.875 0.424357414245605
11.125 0.424357414245605
};
\path [draw=black, fill=orange2551660]
(axis cs:11.75,0.339635848999023)
--(axis cs:12.25,0.339635848999023)
--(axis cs:12.25,0.37716007232666)
--(axis cs:11.75,0.37716007232666)
--(axis cs:11.75,0.339635848999023)
--cycle;
\addplot [black]
table {%
12 0.339635848999023
12 0.319190979003906
};
\addplot [black]
table {%
12 0.37716007232666
12 0.395979881286621
};
\addplot [black]
table {%
11.875 0.319190979003906
12.125 0.319190979003906
};
\addplot [black]
table {%
11.875 0.395979881286621
12.125 0.395979881286621
};
\path [draw=black, fill=orangered2551030]
(axis cs:12.75,0.409124374389648)
--(axis cs:13.25,0.409124374389648)
--(axis cs:13.25,0.447072982788086)
--(axis cs:12.75,0.447072982788086)
--(axis cs:12.75,0.409124374389648)
--cycle;
\addplot [black]
table {%
13 0.409124374389648
13 0.3717041015625
};
\addplot [black]
table {%
13 0.447072982788086
13 0.467820167541504
};
\addplot [black]
table {%
12.875 0.3717041015625
13.125 0.3717041015625
};
\addplot [black]
table {%
12.875 0.467820167541504
13.125 0.467820167541504
};
\path [draw=black, fill=orangered255400]
(axis cs:13.75,0.329817771911621)
--(axis cs:14.25,0.329817771911621)
--(axis cs:14.25,0.367348432540894)
--(axis cs:13.75,0.367348432540894)
--(axis cs:13.75,0.329817771911621)
--cycle;
\addplot [black]
table {%
14 0.329817771911621
14 0.310890197753906
};
\addplot [black]
table {%
14 0.367348432540894
14 0.386282920837402
};
\addplot [black]
table {%
13.875 0.310890197753906
14.125 0.310890197753906
};
\addplot [black]
table {%
13.875 0.386282920837402
14.125 0.386282920837402
};
\path [draw=black, fill=red20400]
(axis cs:14.75,0.343039512634277)
--(axis cs:15.25,0.343039512634277)
--(axis cs:15.25,0.380561828613281)
--(axis cs:14.75,0.380561828613281)
--(axis cs:14.75,0.343039512634277)
--cycle;
\addplot [black]
table {%
15 0.343039512634277
15 0.324277877807617
};
\addplot [black]
table {%
15 0.380561828613281
15 0.399323463439941
};
\addplot [black]
table {%
14.875 0.324277877807617
15.125 0.324277877807617
};
\addplot [black]
table {%
14.875 0.399323463439941
15.125 0.399323463439941
};
\path [draw=black, fill=maroon12700]
(axis cs:15.75,0.345739364624023)
--(axis cs:16.25,0.345739364624023)
--(axis cs:16.25,0.364589691162109)
--(axis cs:15.75,0.364589691162109)
--(axis cs:15.75,0.345739364624023)
--cycle;
\addplot [black]
table {%
16 0.345739364624023
16 0.327003479003906
};
\addplot [black]
table {%
16 0.364589691162109
16 0.383350372314453
};
\addplot [black]
table {%
15.875 0.327003479003906
16.125 0.327003479003906
};
\addplot [black]
table {%
15.875 0.383350372314453
16.125 0.383350372314453
};
\addplot [thick, black]
table {%
0.75 0.130395650863647
1.25 0.130395650863647
};
\addplot [thick, black]
table {%
1.75 0.169553995132446
2.25 0.169553995132446
};
\addplot [thick, black]
table {%
2.75 0.201101779937744
3.25 0.201101779937744
};
\addplot [thick, black]
table {%
3.75 0.236646413803101
4.25 0.236646413803101
};
\addplot [thick, black]
table {%
4.75 0.227451324462891
5.25 0.227451324462891
};
\addplot [thick, black]
table {%
5.75 0.241032123565674
6.25 0.241032123565674
};
\addplot [thick, black]
table {%
6.75 0.312519550323486
7.25 0.312519550323486
};
\addplot [thick, black]
table {%
7.75 0.307902812957764
8.25 0.307902812957764
};
\addplot [thick, black]
table {%
8.75 0.304974555969238
9.25 0.304974555969238
};
\addplot [thick, black]
table {%
9.75 0.346696853637695
10.25 0.346696853637695
};
\addplot [thick, black]
table {%
10.75 0.386669635772705
11.25 0.386669635772705
};
\addplot [thick, black]
table {%
11.75 0.358367919921875
12.25 0.358367919921875
};
\addplot [thick, black]
table {%
12.75 0.427988052368164
13.25 0.427988052368164
};
\addplot [thick, black]
table {%
13.75 0.348650932312012
14.25 0.348650932312012
};
\addplot [thick, black]
table {%
14.75 0.361800193786621
15.25 0.361800193786621
};
\addplot [thick, black]
table {%
15.75 0.345829010009766
16.25 0.345829010009766
};
\end{axis}

\end{tikzpicture}

%% file: tex/boxplot_errors/error_ByAngle_pitch_90.0_rosbag_try1.csv.tex
\begin{tikzpicture}

\definecolor{blue08255}{RGB}{0,8,255}
\definecolor{darkgray176}{RGB}{176,176,176}
\definecolor{deepskyblue0212255}{RGB}{0,212,255}
\definecolor{dodgerblue0144255}{RGB}{0,144,255}
\definecolor{dodgerblue076255}{RGB}{0,76,255}
\definecolor{gainsboro229}{RGB}{229,229,229}
\definecolor{gold2552290}{RGB}{255,229,0}
\definecolor{greenyellow20525541}{RGB}{205,255,41}
\definecolor{lightgreen15025595}{RGB}{150,255,95}
\definecolor{lightgreen95255150}{RGB}{95,255,150}
\definecolor{maroon12700}{RGB}{127,0,0}
\definecolor{mediumblue00204}{RGB}{0,0,204}
\definecolor{navy00127}{RGB}{0,0,127}
\definecolor{orange2551660}{RGB}{255,166,0}
\definecolor{orangered2551030}{RGB}{255,103,0}
\definecolor{orangered255400}{RGB}{255,40,0}
\definecolor{red20400}{RGB}{204,0,0}
\definecolor{turquoise41255205}{RGB}{41,255,205}

\begin{axis}[
    width=\figurewidth,
    height=\figureheight,
    axis background/.style={fill=white},
    axis line style={white},
    tick align=outside,
    x grid style={white},
    xmajorgrids,
    xmajorticks=true,
    y grid style={white},
    ymajorgrids,
    ymajorticks=true,
    y grid style={white!69.0196078431373!black},
    ytick style={color=black},
    xmajorgrids,
    xminorgrids,
    ymajorgrids,
    ymajorticks=true,
    yticklabel style={
            /pgf/number format/fixed,
            /pgf/number format/precision=5
        },
    scaled y ticks=false,
    ylabel near ticks, 
    ylabel shift={-1pt},
    ymin=0, ymax=1,
    xmin=0.5, xmax=16.5,
    xtick style={color=black},
]
\path [draw=black, fill=navy00127]
(axis cs:0.75,0.410417348146439)
--(axis cs:1.25,0.410417348146439)
--(axis cs:1.25,0.503704428672791)
--(axis cs:0.75,0.503704428672791)
--(axis cs:0.75,0.410417348146439)
--cycle;
\addplot [black]
table {%
1 0.410417348146439
1 0.317486524581909
};
\addplot [black]
table {%
1 0.503704428672791
1 0.580491065979004
};
\addplot [black]
table {%
0.875 0.317486524581909
1.125 0.317486524581909
};
\addplot [black]
table {%
0.875 0.580491065979004
1.125 0.580491065979004
};
\path [draw=black, fill=mediumblue00204]
(axis cs:1.75,0.357874155044556)
--(axis cs:2.25,0.357874155044556)
--(axis cs:2.25,0.414330303668976)
--(axis cs:1.75,0.414330303668976)
--(axis cs:1.75,0.357874155044556)
--cycle;
\addplot [black]
table {%
2 0.357874155044556
2 0.320037603378296
};
\addplot [black]
table {%
2 0.414330303668976
2 0.470611691474914
};
\addplot [black]
table {%
1.875 0.320037603378296
2.125 0.320037603378296
};
\addplot [black]
table {%
1.875 0.470611691474914
2.125 0.470611691474914
};
\path [draw=black, fill=blue08255]
(axis cs:2.75,0.295307874679565)
--(axis cs:3.25,0.295307874679565)
--(axis cs:3.25,0.351602792739868)
--(axis cs:2.75,0.351602792739868)
--(axis cs:2.75,0.295307874679565)
--cycle;
\addplot [black]
table {%
3 0.295307874679565
3 0.239062786102295
};
\addplot [black]
table {%
3 0.351602792739868
3 0.389305591583252
};
\addplot [black]
table {%
2.875 0.239062786102295
3.125 0.239062786102295
};
\addplot [black]
table {%
2.875 0.389305591583252
3.125 0.389305591583252
};
\path [draw=black, fill=dodgerblue076255]
(axis cs:3.75,0.617385387420654)
--(axis cs:4.25,0.617385387420654)
--(axis cs:4.25,0.767376184463501)
--(axis cs:3.75,0.767376184463501)
--(axis cs:3.75,0.617385387420654)
--cycle;
\addplot [black]
table {%
4 0.617385387420654
4 0.467462539672852
};
\addplot [black]
table {%
4 0.767376184463501
4 0.804898977279663
};
\addplot [black]
table {%
3.875 0.467462539672852
4.125 0.467462539672852
};
\addplot [black]
table {%
3.875 0.804898977279663
4.125 0.804898977279663
};
\path [draw=black, fill=dodgerblue0144255]
(axis cs:4.75,0.399085283279419)
--(axis cs:5.25,0.399085283279419)
--(axis cs:5.25,0.441327571868896)
--(axis cs:4.75,0.441327571868896)
--(axis cs:4.75,0.399085283279419)
--cycle;
\addplot [black]
table {%
5 0.399085283279419
5 0.366100311279297
};
\addplot [black]
table {%
5 0.441327571868896
5 0.478979587554932
};
\addplot [black]
table {%
4.875 0.366100311279297
5.125 0.366100311279297
};
\addplot [black]
table {%
4.875 0.478979587554932
5.125 0.478979587554932
};
\path [draw=black, fill=deepskyblue0212255]
(axis cs:5.75,0.463381290435791)
--(axis cs:6.25,0.463381290435791)
--(axis cs:6.25,0.5025634765625)
--(axis cs:5.75,0.5025634765625)
--(axis cs:5.75,0.463381290435791)
--cycle;
\addplot [black]
table {%
6 0.463381290435791
6 0.425817012786865
};
\addplot [black]
table {%
6 0.5025634765625
6 0.540192604064941
};
\addplot [black]
table {%
5.875 0.425817012786865
6.125 0.425817012786865
};
\addplot [black]
table {%
5.875 0.540192604064941
6.125 0.540192604064941
};
\path [draw=black, fill=turquoise41255205]
(axis cs:6.75,0.462530612945557)
--(axis cs:7.25,0.462530612945557)
--(axis cs:7.25,0.500055551528931)
--(axis cs:6.75,0.500055551528931)
--(axis cs:6.75,0.462530612945557)
--cycle;
\addplot [black]
table {%
7 0.462530612945557
7 0.425143241882324
};
\addplot [black]
table {%
7 0.500055551528931
7 0.537564754486084
};
\addplot [black]
table {%
6.875 0.425143241882324
7.125 0.425143241882324
};
\addplot [black]
table {%
6.875 0.537564754486084
7.125 0.537564754486084
};
\path [draw=black, fill=lightgreen95255150]
(axis cs:7.75,0.443201065063477)
--(axis cs:8.25,0.443201065063477)
--(axis cs:8.25,0.537234783172607)
--(axis cs:7.75,0.537234783172607)
--(axis cs:7.75,0.443201065063477)
--cycle;
\addplot [black]
table {%
8 0.443201065063477
8 0.404038906097412
};
\addplot [black]
table {%
8 0.537234783172607
8 0.612822532653809
};
\addplot [black]
table {%
7.875 0.404038906097412
8.125 0.404038906097412
};
\addplot [black]
table {%
7.875 0.612822532653809
8.125 0.612822532653809
};
\path [draw=black, fill=lightgreen15025595]
(axis cs:8.75,0.398803949356079)
--(axis cs:9.25,0.398803949356079)
--(axis cs:9.25,0.511594772338867)
--(axis cs:8.75,0.511594772338867)
--(axis cs:8.75,0.398803949356079)
--cycle;
\addplot [black]
table {%
9 0.398803949356079
9 0.286170959472656
};
\addplot [black]
table {%
9 0.511594772338867
9 0.624234199523926
};
\addplot [black]
table {%
8.875 0.286170959472656
9.125 0.286170959472656
};
\addplot [black]
table {%
8.875 0.624234199523926
9.125 0.624234199523926
};
\path [draw=black, fill=greenyellow20525541]
(axis cs:9.75,0.365770101547241)
--(axis cs:10.25,0.365770101547241)
--(axis cs:10.25,0.440720081329346)
--(axis cs:9.75,0.440720081329346)
--(axis cs:9.75,0.365770101547241)
--cycle;
\addplot [black]
table {%
10 0.365770101547241
10 0.309389114379883
};
\addplot [black]
table {%
10 0.440720081329346
10 0.49698543548584
};
\addplot [black]
table {%
9.875 0.309389114379883
10.125 0.309389114379883
};
\addplot [black]
table {%
9.875 0.49698543548584
10.125 0.49698543548584
};
\path [draw=black, fill=gold2552290]
(axis cs:10.75,0.508323669433594)
--(axis cs:11.25,0.508323669433594)
--(axis cs:11.25,0.545825004577637)
--(axis cs:10.75,0.545825004577637)
--(axis cs:10.75,0.508323669433594)
--cycle;
\addplot [black]
table {%
11 0.508323669433594
11 0.470916748046875
};
\addplot [black]
table {%
11 0.545825004577637
11 0.583168029785156
};
\addplot [black]
table {%
10.875 0.470916748046875
11.125 0.470916748046875
};
\addplot [black]
table {%
10.875 0.583168029785156
11.125 0.583168029785156
};
\path [draw=black, fill=orange2551660]
(axis cs:11.75,0.527783870697021)
--(axis cs:12.25,0.527783870697021)
--(axis cs:12.25,0.566568374633789)
--(axis cs:11.75,0.566568374633789)
--(axis cs:11.75,0.527783870697021)
--cycle;
\addplot [black]
table {%
12 0.527783870697021
12 0.489188194274902
};
\addplot [black]
table {%
12 0.566568374633789
12 0.603264808654785
};
\addplot [black]
table {%
11.875 0.489188194274902
12.125 0.489188194274902
};
\addplot [black]
table {%
11.875 0.603264808654785
12.125 0.603264808654785
};
\path [draw=black, fill=orangered2551030]
(axis cs:12.75,0.371618986129761)
--(axis cs:13.25,0.371618986129761)
--(axis cs:13.25,0.409093141555786)
--(axis cs:12.75,0.409093141555786)
--(axis cs:12.75,0.371618986129761)
--cycle;
\addplot [black]
table {%
13 0.371618986129761
13 0.334158897399902
};
\addplot [black]
table {%
13 0.409093141555786
13 0.429448127746582
};
\addplot [black]
table {%
12.875 0.334158897399902
13.125 0.334158897399902
};
\addplot [black]
table {%
12.875 0.429448127746582
13.125 0.429448127746582
};
\path [draw=black, fill=orangered255400]
(axis cs:13.75,0.384957313537598)
--(axis cs:14.25,0.384957313537598)
--(axis cs:14.25,0.423373222351074)
--(axis cs:13.75,0.423373222351074)
--(axis cs:13.75,0.384957313537598)
--cycle;
\addplot [black]
table {%
14 0.384957313537598
14 0.347434997558594
};
\addplot [black]
table {%
14 0.423373222351074
14 0.461099624633789
};
\addplot [black]
table {%
13.875 0.347434997558594
14.125 0.347434997558594
};
\addplot [black]
table {%
13.875 0.461099624633789
14.125 0.461099624633789
};
\path [draw=black, fill=red20400]
(axis cs:14.75,0.445963382720947)
--(axis cs:15.25,0.445963382720947)
--(axis cs:15.25,0.60074520111084)
--(axis cs:14.75,0.60074520111084)
--(axis cs:14.75,0.445963382720947)
--cycle;
\addplot [black]
table {%
15 0.445963382720947
15 0.338085174560547
};
\addplot [black]
table {%
15 0.60074520111084
15 0.732074737548828
};
\addplot [black]
table {%
14.875 0.338085174560547
15.125 0.338085174560547
};
\addplot [black]
table {%
14.875 0.732074737548828
15.125 0.732074737548828
};
\path [draw=black, fill=maroon12700]
(axis cs:15.75,0.439472198486328)
--(axis cs:16.25,0.439472198486328)
--(axis cs:16.25,0.477453231811523)
--(axis cs:15.75,0.477453231811523)
--(axis cs:15.75,0.439472198486328)
--cycle;
\addplot [black]
table {%
16 0.439472198486328
16 0.401844024658203
};
\addplot [black]
table {%
16 0.477453231811523
16 0.514976501464844
};
\addplot [black]
table {%
15.875 0.401844024658203
16.125 0.401844024658203
};
\addplot [black]
table {%
15.875 0.514976501464844
16.125 0.514976501464844
};
\addplot [thick, black]
table {%
0.75 0.457497179508209
1.25 0.457497179508209
};
\addplot [thick, black]
table {%
1.75 0.395434856414795
2.25 0.395434856414795
};
\addplot [thick, black]
table {%
2.75 0.314253449440002
3.25 0.314253449440002
};
\addplot [thick, black]
table {%
3.75 0.711231231689453
4.25 0.711231231689453
};
\addplot [thick, black]
table {%
4.75 0.422458410263062
5.25 0.422458410263062
};
\addplot [thick, black]
table {%
5.75 0.482304334640503
6.25 0.482304334640503
};
\addplot [thick, black]
table {%
6.75 0.481331348419189
7.25 0.481331348419189
};
\addplot [thick, black]
table {%
7.75 0.479258060455322
8.25 0.479258060455322
};
\addplot [thick, black]
table {%
8.75 0.474071502685547
9.25 0.474071502685547
};
\addplot [thick, black]
table {%
9.75 0.403241157531738
10.25 0.403241157531738
};
\addplot [thick, black]
table {%
10.75 0.527084827423096
11.25 0.527084827423096
};
\addplot [thick, black]
table {%
11.75 0.546938896179199
12.25 0.546938896179199
};
\addplot [thick, black]
table {%
12.75 0.390108108520508
13.25 0.390108108520508
};
\addplot [thick, black]
table {%
13.75 0.404679298400879
14.25 0.404679298400879
};
\addplot [thick, black]
table {%
14.75 0.535079479217529
15.25 0.535079479217529
};
\addplot [thick, black]
table {%
15.75 0.458240509033203
16.25 0.458240509033203
};
\end{axis}

\end{tikzpicture}

%% file: tex/boxplot_errors/error_ByAngle_pitch_180.0_rosbag_try1.csv.tex
\begin{tikzpicture}

\definecolor{blue08255}{RGB}{0,8,255}
\definecolor{darkgray176}{RGB}{176,176,176}
\definecolor{deepskyblue0212255}{RGB}{0,212,255}
\definecolor{dodgerblue0144255}{RGB}{0,144,255}
\definecolor{dodgerblue076255}{RGB}{0,76,255}
\definecolor{gainsboro229}{RGB}{229,229,229}
\definecolor{gold2552290}{RGB}{255,229,0}
\definecolor{greenyellow20525541}{RGB}{205,255,41}
\definecolor{lightgreen15025595}{RGB}{150,255,95}
\definecolor{lightgreen95255150}{RGB}{95,255,150}
\definecolor{maroon12700}{RGB}{127,0,0}
\definecolor{mediumblue00204}{RGB}{0,0,204}
\definecolor{navy00127}{RGB}{0,0,127}
\definecolor{orange2551660}{RGB}{255,166,0}
\definecolor{orangered2551030}{RGB}{255,103,0}
\definecolor{orangered255400}{RGB}{255,40,0}
\definecolor{red20400}{RGB}{204,0,0}
\definecolor{turquoise41255205}{RGB}{41,255,205}

\begin{axis}[
    width=\figurewidth,
    height=\figureheight,
    axis background/.style={fill=white},
    axis line style={white},
    tick align=outside,
    x grid style={white},
    xmajorgrids,
    xmajorticks=true,
    y grid style={white},
    ymajorgrids,
    ymajorticks=true,
    y grid style={white!69.0196078431373!black},
    ytick style={color=black},
    xmajorgrids,
    xminorgrids,
    ymajorgrids,
    ymajorticks=true,
    yticklabel style={
            /pgf/number format/fixed,
            /pgf/number format/precision=5
        },
    scaled y ticks=false,
    ylabel near ticks, 
    ylabel shift={-1pt},
    ymin=0, ymax=1,
    xmin=0.5, xmax=16.5,
    xtick style={color=black},
]
\path [draw=black, fill=navy00127]
(axis cs:0.75,0.111276507377624)
--(axis cs:1.25,0.111276507377624)
--(axis cs:1.25,0.148799538612366)
--(axis cs:0.75,0.148799538612366)
--(axis cs:0.75,0.111276507377624)
--cycle;
\addplot [black]
table {%
1 0.111276507377624
1 0.0737534761428833
};
\addplot [black]
table {%
1 0.148799538612366
1 0.186322450637817
};
\addplot [black]
table {%
0.875 0.0737534761428833
1.125 0.0737534761428833
};
\addplot [black]
table {%
0.875 0.186322450637817
1.125 0.186322450637817
};
\path [draw=black, fill=mediumblue00204]
(axis cs:1.75,0.221444010734558)
--(axis cs:2.25,0.221444010734558)
--(axis cs:2.25,0.263866186141968)
--(axis cs:1.75,0.263866186141968)
--(axis cs:1.75,0.221444010734558)
--cycle;
\addplot [black]
table {%
2 0.221444010734558
2 0.188299894332886
};
\addplot [black]
table {%
2 0.263866186141968
2 0.301841855049133
};
\addplot [black]
table {%
1.875 0.188299894332886
2.125 0.188299894332886
};
\addplot [black]
table {%
1.875 0.301841855049133
2.125 0.301841855049133
};
\path [draw=black, fill=blue08255]
(axis cs:2.75,0.238937437534332)
--(axis cs:3.25,0.238937437534332)
--(axis cs:3.25,0.276298761367798)
--(axis cs:2.75,0.276298761367798)
--(axis cs:2.75,0.238937437534332)
--cycle;
\addplot [black]
table {%
3 0.238937437534332
3 0.219615936279297
};
\addplot [black]
table {%
3 0.276298761367798
3 0.313619613647461
};
\addplot [black]
table {%
2.875 0.219615936279297
3.125 0.219615936279297
};
\addplot [black]
table {%
2.875 0.313619613647461
3.125 0.313619613647461
};
\path [draw=black, fill=dodgerblue076255]
(axis cs:3.75,0.279638528823852)
--(axis cs:4.25,0.279638528823852)
--(axis cs:4.25,0.316877365112305)
--(axis cs:3.75,0.316877365112305)
--(axis cs:3.75,0.279638528823852)
--cycle;
\addplot [black]
table {%
4 0.279638528823852
4 0.260481357574463
};
\addplot [black]
table {%
4 0.316877365112305
4 0.336056232452393
};
\addplot [black]
table {%
3.875 0.260481357574463
4.125 0.260481357574463
};
\addplot [black]
table {%
3.875 0.336056232452393
4.125 0.336056232452393
};
\path [draw=black, fill=dodgerblue0144255]
(axis cs:4.75,0.350381731986999)
--(axis cs:5.25,0.350381731986999)
--(axis cs:5.25,0.387873768806457)
--(axis cs:4.75,0.387873768806457)
--(axis cs:4.75,0.350381731986999)
--cycle;
\addplot [black]
table {%
5 0.350381731986999
5 0.312932014465332
};
\addplot [black]
table {%
5 0.387873768806457
5 0.407249450683594
};
\addplot [black]
table {%
4.875 0.312932014465332
5.125 0.312932014465332
};
\addplot [black]
table {%
4.875 0.407249450683594
5.125 0.407249450683594
};
\path [draw=black, fill=deepskyblue0212255]
(axis cs:5.75,0.31427013874054)
--(axis cs:6.25,0.31427013874054)
--(axis cs:6.25,0.351739048957825)
--(axis cs:5.75,0.351739048957825)
--(axis cs:5.75,0.31427013874054)
--cycle;
\addplot [black]
table {%
6 0.31427013874054
6 0.295360088348389
};
\addplot [black]
table {%
6 0.351739048957825
6 0.389166831970215
};
\addplot [black]
table {%
5.875 0.295360088348389
6.125 0.295360088348389
};
\addplot [black]
table {%
5.875 0.389166831970215
6.125 0.389166831970215
};
\path [draw=black, fill=turquoise41255205]
(axis cs:6.75,0.349604368209839)
--(axis cs:7.25,0.349604368209839)
--(axis cs:7.25,0.387031078338623)
--(axis cs:6.75,0.387031078338623)
--(axis cs:6.75,0.349604368209839)
--cycle;
\addplot [black]
table {%
7 0.349604368209839
7 0.330704212188721
};
\addplot [black]
table {%
7 0.387031078338623
7 0.40585994720459
};
\addplot [black]
table {%
6.875 0.330704212188721
7.125 0.330704212188721
};
\addplot [black]
table {%
6.875 0.40585994720459
7.125 0.40585994720459
};
\path [draw=black, fill=lightgreen95255150]
(axis cs:7.75,0.326625347137451)
--(axis cs:8.25,0.326625347137451)
--(axis cs:8.25,0.365255832672119)
--(axis cs:7.75,0.365255832672119)
--(axis cs:7.75,0.326625347137451)
--cycle;
\addplot [black]
table {%
8 0.326625347137451
8 0.28873348236084
};
\addplot [black]
table {%
8 0.365255832672119
8 0.403102874755859
};
\addplot [black]
table {%
7.875 0.28873348236084
8.125 0.28873348236084
};
\addplot [black]
table {%
7.875 0.403102874755859
8.125 0.403102874755859
};
\path [draw=black, fill=lightgreen15025595]
(axis cs:8.75,0.211661338806152)
--(axis cs:9.25,0.211661338806152)
--(axis cs:9.25,0.77464485168457)
--(axis cs:8.75,0.77464485168457)
--(axis cs:8.75,0.211661338806152)
--cycle;
\addplot [black]
table {%
9 0.211661338806152
9 0.0429067611694335
};
\addplot [black]
table {%
9 0.77464485168457
9 0.812236785888672
};
\addplot [black]
table {%
8.875 0.0429067611694335
9.125 0.0429067611694335
};
\addplot [black]
table {%
8.875 0.812236785888672
9.125 0.812236785888672
};
\path [draw=black, fill=greenyellow20525541]
(axis cs:9.75,0.403203964233398)
--(axis cs:10.25,0.403203964233398)
--(axis cs:10.25,0.445591926574707)
--(axis cs:9.75,0.445591926574707)
--(axis cs:9.75,0.403203964233398)
--cycle;
\addplot [black]
table {%
10 0.403203964233398
10 0.365680694580078
};
\addplot [black]
table {%
10 0.445591926574707
10 0.478481292724609
};
\addplot [black]
table {%
9.875 0.365680694580078
10.125 0.365680694580078
};
\addplot [black]
table {%
9.875 0.478481292724609
10.125 0.478481292724609
};
\path [draw=black, fill=gold2552290]
(axis cs:10.75,0.322527885437012)
--(axis cs:11.25,0.322527885437012)
--(axis cs:11.25,0.360021591186523)
--(axis cs:10.75,0.360021591186523)
--(axis cs:10.75,0.322527885437012)
--cycle;
\addplot [black]
table {%
11 0.322527885437012
11 0.303678512573242
};
\addplot [black]
table {%
11 0.360021591186523
11 0.397512435913086
};
\addplot [black]
table {%
10.875 0.303678512573242
11.125 0.303678512573242
};
\addplot [black]
table {%
10.875 0.397512435913086
11.125 0.397512435913086
};
\path [draw=black, fill=orange2551660]
(axis cs:11.75,0.471684455871582)
--(axis cs:12.25,0.471684455871582)
--(axis cs:12.25,0.508841514587402)
--(axis cs:11.75,0.508841514587402)
--(axis cs:11.75,0.471684455871582)
--cycle;
\addplot [black]
table {%
12 0.471684455871582
12 0.452339172363281
};
\addplot [black]
table {%
12 0.508841514587402
12 0.529398918151855
};
\addplot [black]
table {%
11.875 0.452339172363281
12.125 0.452339172363281
};
\addplot [black]
table {%
11.875 0.529398918151855
12.125 0.529398918151855
};
\path [draw=black, fill=orangered2551030]
(axis cs:12.75,0.468955993652344)
--(axis cs:13.25,0.468955993652344)
--(axis cs:13.25,0.525239944458008)
--(axis cs:12.75,0.525239944458008)
--(axis cs:12.75,0.468955993652344)
--cycle;
\addplot [black]
table {%
13 0.468955993652344
13 0.431432723999023
};
\addplot [black]
table {%
13 0.525239944458008
13 0.562763214111328
};
\addplot [black]
table {%
12.875 0.431432723999023
13.125 0.431432723999023
};
\addplot [black]
table {%
12.875 0.562763214111328
13.125 0.562763214111328
};
\path [draw=black, fill=orangered255400]
(axis cs:13.75,0.421988487243652)
--(axis cs:14.25,0.421988487243652)
--(axis cs:14.25,0.458804130554199)
--(axis cs:13.75,0.458804130554199)
--(axis cs:13.75,0.421988487243652)
--cycle;
\addplot [black]
table {%
14 0.421988487243652
14 0.386120796203613
};
\addplot [black]
table {%
14 0.458804130554199
14 0.47773551940918
};
\addplot [black]
table {%
13.875 0.386120796203613
14.125 0.386120796203613
};
\addplot [black]
table {%
13.875 0.47773551940918
14.125 0.47773551940918
};
\path [draw=black, fill=red20400]
(axis cs:14.75,0.397603034973144)
--(axis cs:15.25,0.397603034973144)
--(axis cs:15.25,0.435125350952148)
--(axis cs:14.75,0.435125350952148)
--(axis cs:14.75,0.397603034973144)
--cycle;
\addplot [black]
table {%
15 0.397603034973144
15 0.378841400146484
};
\addplot [black]
table {%
15 0.435125350952148
15 0.453887939453125
};
\addplot [black]
table {%
14.875 0.378841400146484
15.125 0.378841400146484
};
\addplot [black]
table {%
14.875 0.453887939453125
15.125 0.453887939453125
};
\path [draw=black, fill=maroon12700]
(axis cs:15.75,0.406036376953125)
--(axis cs:16.25,0.406036376953125)
--(axis cs:16.25,0.462320327758789)
--(axis cs:15.75,0.462320327758789)
--(axis cs:15.75,0.406036376953125)
--cycle;
\addplot [black]
table {%
16 0.406036376953125
16 0.368513107299805
};
\addplot [black]
table {%
16 0.462320327758789
16 0.499843597412109
};
\addplot [black]
table {%
15.875 0.368513107299805
16.125 0.368513107299805
};
\addplot [black]
table {%
15.875 0.499843597412109
16.125 0.499843597412109
};
\addplot [thick, black]
table {%
0.75 0.130037546157837
1.25 0.130037546157837
};
\addplot [thick, black]
table {%
1.75 0.245321869850159
2.25 0.245321869850159
};
\addplot [thick, black]
table {%
2.75 0.257513403892517
3.25 0.257513403892517
};
\addplot [thick, black]
table {%
3.75 0.298222780227661
4.25 0.298222780227661
};
\addplot [thick, black]
table {%
4.75 0.369097232818604
5.25 0.369097232818604
};
\addplot [thick, black]
table {%
5.75 0.33296799659729
6.25 0.33296799659729
};
\addplot [thick, black]
table {%
6.75 0.368308544158935
7.25 0.368308544158935
};
\addplot [thick, black]
table {%
7.75 0.346649169921875
8.25 0.346649169921875
};
\addplot [thick, black]
table {%
8.75 0.446249961853027
9.25 0.446249961853027
};
\addplot [thick, black]
table {%
9.75 0.422122955322266
10.25 0.422122955322266
};
\addplot [thick, black]
table {%
10.75 0.341290473937988
11.25 0.341290473937988
};
\addplot [thick, black]
table {%
11.75 0.490344047546387
12.25 0.490344047546387
};
\addplot [thick, black]
table {%
12.75 0.506478309631348
13.25 0.506478309631348
};
\addplot [thick, black]
table {%
13.75 0.440369606018066
14.25 0.440369606018066
};
\addplot [thick, black]
table {%
14.75 0.416364669799805
15.25 0.416364669799805
};
\addplot [thick, black]
table {%
15.75 0.424797058105469
16.25 0.424797058105469
};
\end{axis}

\end{tikzpicture}

%% file: tex/boxplot_errors/error_ByAngle_pitch_270.0_rosbag_try1.csv.tex
\begin{tikzpicture}

\definecolor{blue08255}{RGB}{0,8,255}
\definecolor{darkgray176}{RGB}{176,176,176}
\definecolor{deepskyblue0212255}{RGB}{0,212,255}
\definecolor{dodgerblue0144255}{RGB}{0,144,255}
\definecolor{dodgerblue076255}{RGB}{0,76,255}
\definecolor{gainsboro229}{RGB}{229,229,229}
\definecolor{gold2552290}{RGB}{255,229,0}
\definecolor{greenyellow20525541}{RGB}{205,255,41}
\definecolor{lightgreen15025595}{RGB}{150,255,95}
\definecolor{lightgreen95255150}{RGB}{95,255,150}
\definecolor{maroon12700}{RGB}{127,0,0}
\definecolor{mediumblue00204}{RGB}{0,0,204}
\definecolor{navy00127}{RGB}{0,0,127}
\definecolor{orange2551660}{RGB}{255,166,0}
\definecolor{orangered2551030}{RGB}{255,103,0}
\definecolor{orangered255400}{RGB}{255,40,0}
\definecolor{red20400}{RGB}{204,0,0}
\definecolor{turquoise41255205}{RGB}{41,255,205}

\begin{axis}[
    width=\figurewidth,
    height=\figureheight,
    axis background/.style={fill=white},
    axis line style={white},
    tick align=outside,
    x grid style={white},
    xmajorgrids,
    xmajorticks=true,
    y grid style={white},
    ymajorgrids,
    ymajorticks=true,
    y grid style={white!69.0196078431373!black},
    ytick style={color=black},
    xmajorgrids,
    xminorgrids,
    ymajorgrids,
    ymajorticks=true,
    yticklabel style={
            /pgf/number format/fixed,
            /pgf/number format/precision=5
        },
    scaled y ticks=false,
    ylabel near ticks, 
    ylabel shift={-1pt},
    ymin=0, ymax=1,
    xmin=0.5, xmax=16.5,
    xtick style={color=black},
]
\path [draw=black, fill=navy00127]
(axis cs:0.75,0.524027466773987)
--(axis cs:1.25,0.524027466773987)
--(axis cs:1.25,0.580312490463257)
--(axis cs:0.75,0.580312490463257)
--(axis cs:0.75,0.524027466773987)
--cycle;
\addplot [black]
table {%
1 0.524027466773987
1 0.467743515968323
};
\addplot [black]
table {%
1 0.580312490463257
1 0.63659656047821
};
\addplot [black]
table {%
0.875 0.467743515968323
1.125 0.467743515968323
};
\addplot [black]
table {%
0.875 0.63659656047821
1.125 0.63659656047821
};
\path [draw=black, fill=mediumblue00204]
(axis cs:1.75,0.31971538066864)
--(axis cs:2.25,0.31971538066864)
--(axis cs:2.25,0.376331746578217)
--(axis cs:1.75,0.376331746578217)
--(axis cs:1.75,0.31971538066864)
--cycle;
\addplot [black]
table {%
2 0.31971538066864
2 0.263547897338867
};
\addplot [black]
table {%
2 0.376331746578217
2 0.416210293769836
};
\addplot [black]
table {%
1.875 0.263547897338867
2.125 0.263547897338867
};
\addplot [black]
table {%
1.875 0.416210293769836
2.125 0.416210293769836
};
\path [draw=black, fill=blue08255]
(axis cs:2.75,0.220046103000641)
--(axis cs:3.25,0.220046103000641)
--(axis cs:3.25,0.599944114685059)
--(axis cs:2.75,0.599944114685059)
--(axis cs:2.75,0.220046103000641)
--cycle;
\addplot [black]
table {%
3 0.220046103000641
3 0.088472843170166
};
\addplot [black]
table {%
3 0.599944114685059
3 0.970448017120361
};
\addplot [black]
table {%
2.875 0.088472843170166
3.125 0.088472843170166
};
\addplot [black]
table {%
2.875 0.970448017120361
3.125 0.970448017120361
};
\path [draw=black, fill=dodgerblue076255]
(axis cs:3.75,0.447091639041901)
--(axis cs:4.25,0.447091639041901)
--(axis cs:4.25,0.539589047431946)
--(axis cs:3.75,0.539589047431946)
--(axis cs:3.75,0.447091639041901)
--cycle;
\addplot [black]
table {%
4 0.447091639041901
4 0.372104167938232
};
\addplot [black]
table {%
4 0.539589047431946
4 0.595931529998779
};
\addplot [black]
table {%
3.875 0.372104167938232
4.125 0.372104167938232
};
\addplot [black]
table {%
3.875 0.595931529998779
4.125 0.595931529998779
};
\path [draw=black, fill=dodgerblue0144255]
(axis cs:4.75,0.293904304504394)
--(axis cs:5.25,0.293904304504394)
--(axis cs:5.25,0.350207924842834)
--(axis cs:4.75,0.350207924842834)
--(axis cs:4.75,0.293904304504394)
--cycle;
\addplot [black]
table {%
5 0.293904304504394
5 0.256107807159424
};
\addplot [black]
table {%
5 0.350207924842834
5 0.387798309326172
};
\addplot [black]
table {%
4.875 0.256107807159424
5.125 0.256107807159424
};
\addplot [black]
table {%
4.875 0.387798309326172
5.125 0.387798309326172
};
\path [draw=black, fill=deepskyblue0212255]
(axis cs:5.75,0.164690375328064)
--(axis cs:6.25,0.164690375328064)
--(axis cs:6.25,0.202240467071533)
--(axis cs:5.75,0.202240467071533)
--(axis cs:5.75,0.164690375328064)
--cycle;
\addplot [black]
table {%
6 0.164690375328064
6 0.127364158630371
};
\addplot [black]
table {%
6 0.202240467071533
6 0.239755153656006
};
\addplot [black]
table {%
5.875 0.127364158630371
6.125 0.127364158630371
};
\addplot [black]
table {%
5.875 0.239755153656006
6.125 0.239755153656006
};
\path [draw=black, fill=turquoise41255205]
(axis cs:6.75,0.564796686172485)
--(axis cs:7.25,0.564796686172485)
--(axis cs:7.25,0.629958391189575)
--(axis cs:6.75,0.629958391189575)
--(axis cs:6.75,0.564796686172485)
--cycle;
\addplot [black]
table {%
7 0.564796686172485
7 0.51332426071167
};
\addplot [black]
table {%
7 0.629958391189575
7 0.681305408477783
};
\addplot [black]
table {%
6.875 0.51332426071167
7.125 0.51332426071167
};
\addplot [black]
table {%
6.875 0.681305408477783
7.125 0.681305408477783
};
\path [draw=black, fill=lightgreen95255150]
(axis cs:7.75,0.141652584075928)
--(axis cs:8.25,0.141652584075928)
--(axis cs:8.25,0.197806358337402)
--(axis cs:7.75,0.197806358337402)
--(axis cs:7.75,0.141652584075928)
--cycle;
\addplot [black]
table {%
8 0.141652584075928
8 0.102469444274902
};
\addplot [black]
table {%
8 0.197806358337402
8 0.235466003417969
};
\addplot [black]
table {%
7.875 0.102469444274902
8.125 0.102469444274902
};
\addplot [black]
table {%
7.875 0.235466003417969
8.125 0.235466003417969
};
\path [draw=black, fill=lightgreen15025595]
(axis cs:8.75,0.0432760715484618)
--(axis cs:9.25,0.0432760715484618)
--(axis cs:9.25,0.0995433330535888)
--(axis cs:8.75,0.0995433330535888)
--(axis cs:8.75,0.0432760715484618)
--cycle;
\addplot [black]
table {%
9 0.0432760715484618
9 -0.0128278732299804
};
\addplot [black]
table {%
9 0.0995433330535888
9 0.13727855682373
};
\addplot [black]
table {%
8.875 -0.0128278732299804
9.125 -0.0128278732299804
};
\addplot [black]
table {%
8.875 0.13727855682373
9.125 0.13727855682373
};
\path [draw=black, fill=greenyellow20525541]
(axis cs:9.75,0.347093105316162)
--(axis cs:10.25,0.347093105316162)
--(axis cs:10.25,0.403368473052979)
--(axis cs:9.75,0.403368473052979)
--(axis cs:9.75,0.347093105316162)
--cycle;
\addplot [black]
table {%
10 0.347093105316162
10 0.309537887573242
};
\addplot [black]
table {%
10 0.403368473052979
10 0.440927505493164
};
\addplot [black]
table {%
9.875 0.309537887573242
10.125 0.309537887573242
};
\addplot [black]
table {%
9.875 0.440927505493164
10.125 0.440927505493164
};
\path [draw=black, fill=gold2552290]
(axis cs:10.75,0.0789101123809814)
--(axis cs:11.25,0.0789101123809814)
--(axis cs:11.25,0.134922027587891)
--(axis cs:10.75,0.134922027587891)
--(axis cs:10.75,0.0789101123809814)
--cycle;
\addplot [black]
table {%
11 0.0789101123809814
11 0.0411443710327148
};
\addplot [black]
table {%
11 0.134922027587891
11 0.172538757324219
};
\addplot [black]
table {%
10.875 0.0411443710327148
11.125 0.0411443710327148
};
\addplot [black]
table {%
10.875 0.172538757324219
11.125 0.172538757324219
};
\path [draw=black, fill=orange2551660]
(axis cs:11.75,0.565211772918701)
--(axis cs:12.25,0.565211772918701)
--(axis cs:12.25,0.611173629760742)
--(axis cs:11.75,0.611173629760742)
--(axis cs:11.75,0.565211772918701)
--cycle;
\addplot [black]
table {%
12 0.565211772918701
12 0.535896301269531
};
\addplot [black]
table {%
12 0.611173629760742
12 0.650278091430664
};
\addplot [black]
table {%
11.875 0.535896301269531
12.125 0.535896301269531
};
\addplot [black]
table {%
11.875 0.650278091430664
12.125 0.650278091430664
};
\path [draw=black, fill=orangered2551030]
(axis cs:12.75,0.538352489471436)
--(axis cs:13.25,0.538352489471436)
--(axis cs:13.25,0.557250738143921)
--(axis cs:12.75,0.557250738143921)
--(axis cs:12.75,0.538352489471436)
--cycle;
\addplot [black]
table {%
13 0.538352489471436
13 0.519503593444824
};
\addplot [black]
table {%
13 0.557250738143921
13 0.576083183288574
};
\addplot [black]
table {%
12.875 0.519503593444824
13.125 0.519503593444824
};
\addplot [black]
table {%
12.875 0.576083183288574
13.125 0.576083183288574
};
\path [draw=black, fill=orangered255400]
(axis cs:13.75,0.53661060333252)
--(axis cs:14.25,0.53661060333252)
--(axis cs:14.25,0.591620683670044)
--(axis cs:13.75,0.591620683670044)
--(axis cs:13.75,0.53661060333252)
--cycle;
\addplot [black]
table {%
14 0.53661060333252
14 0.498836517333984
};
\addplot [black]
table {%
14 0.591620683670044
14 0.630154609680176
};
\addplot [black]
table {%
13.875 0.498836517333984
14.125 0.498836517333984
};
\addplot [black]
table {%
13.875 0.630154609680176
14.125 0.630154609680176
};
\path [draw=black, fill=red20400]
(axis cs:14.75,0.602526664733887)
--(axis cs:15.25,0.602526664733887)
--(axis cs:15.25,0.658810615539551)
--(axis cs:14.75,0.658810615539551)
--(axis cs:14.75,0.602526664733887)
--cycle;
\addplot [black]
table {%
15 0.602526664733887
15 0.546303749084473
};
\addplot [black]
table {%
15 0.658810615539551
15 0.715088844299316
};
\addplot [black]
table {%
14.875 0.546303749084473
15.125 0.546303749084473
};
\addplot [black]
table {%
14.875 0.715088844299316
15.125 0.715088844299316
};
\path [draw=black, fill=maroon12700]
(axis cs:15.75,0.649934768676758)
--(axis cs:16.25,0.649934768676758)
--(axis cs:16.25,0.687458038330078)
--(axis cs:15.75,0.687458038330078)
--(axis cs:15.75,0.649934768676758)
--cycle;
\addplot [black]
table {%
16 0.649934768676758
16 0.612411499023438
};
\addplot [black]
table {%
16 0.687458038330078
16 0.724981307983398
};
\addplot [black]
table {%
15.875 0.612411499023438
16.125 0.612411499023438
};
\addplot [black]
table {%
15.875 0.724981307983398
16.125 0.724981307983398
};
\addplot [thick, black]
table {%
0.75 0.561550498008728
1.25 0.561550498008728
};
\addplot [thick, black]
table {%
1.75 0.349179565906525
2.25 0.349179565906525
};
\addplot [thick, black]
table {%
2.75 0.463866114616394
3.25 0.463866114616394
};
\addplot [thick, black]
table {%
3.75 0.493406772613525
4.25 0.493406772613525
};
\addplot [thick, black]
table {%
4.75 0.331255435943604
5.25 0.331255435943604
};
\addplot [thick, black]
table {%
5.75 0.183447599411011
6.25 0.183447599411011
};
\addplot [thick, black]
table {%
6.75 0.606812000274658
7.25 0.606812000274658
};
\addplot [thick, black]
table {%
7.75 0.160421133041382
8.25 0.160421133041382
};
\addplot [thick, black]
table {%
8.75 0.0807480812072753
9.25 0.0807480812072753
};
\addplot [thick, black]
table {%
9.75 0.375232696533203
10.25 0.375232696533203
};
\addplot [thick, black]
table {%
10.75 0.116004943847656
11.25 0.116004943847656
};
\addplot [thick, black]
table {%
11.75 0.592230796813965
12.25 0.592230796813965
};
\addplot [thick, black]
table {%
12.75 0.539695739746094
13.25 0.539695739746094
};
\addplot [thick, black]
table {%
13.75 0.555373191833496
14.25 0.555373191833496
};
\addplot [thick, black]
table {%
14.75 0.621302127838135
15.25 0.621302127838135
};
\addplot [thick, black]
table {%
15.75 0.668697357177734
16.25 0.668697357177734
};
\end{axis}

\end{tikzpicture}

%% file: tex/boxplot_errors/error_ByAngle_roll_90.0_rosbag_try1.csv.tex
\begin{tikzpicture}

\definecolor{blue08255}{RGB}{0,8,255}
\definecolor{darkgray176}{RGB}{176,176,176}
\definecolor{deepskyblue0212255}{RGB}{0,212,255}
\definecolor{dodgerblue0144255}{RGB}{0,144,255}
\definecolor{dodgerblue076255}{RGB}{0,76,255}
\definecolor{gainsboro229}{RGB}{229,229,229}
\definecolor{gold2552290}{RGB}{255,229,0}
\definecolor{greenyellow20525541}{RGB}{205,255,41}
\definecolor{lightgreen15025595}{RGB}{150,255,95}
\definecolor{lightgreen95255150}{RGB}{95,255,150}
\definecolor{maroon12700}{RGB}{127,0,0}
\definecolor{mediumblue00204}{RGB}{0,0,204}
\definecolor{navy00127}{RGB}{0,0,127}
\definecolor{orange2551660}{RGB}{255,166,0}
\definecolor{orangered2551030}{RGB}{255,103,0}
\definecolor{orangered255400}{RGB}{255,40,0}
\definecolor{red20400}{RGB}{204,0,0}
\definecolor{turquoise41255205}{RGB}{41,255,205}

\begin{axis}[
    width=\figurewidth,
    height=\figureheight,
    axis background/.style={fill=white},
    axis line style={white},
    tick align=outside,
    x grid style={white},
    xmajorgrids,
    xmajorticks=true,
    y grid style={white},
    ymajorgrids,
    ymajorticks=true,
    y grid style={white!69.0196078431373!black},
    ytick style={color=black},
    xmajorgrids,
    xminorgrids,
    ymajorgrids,
    ymajorticks=true,
    yticklabel style={
            /pgf/number format/fixed,
            /pgf/number format/precision=5
        },
    scaled y ticks=false,
    ylabel near ticks, 
    ylabel shift={-1pt},
    ymin=0, ymax=1,
    xmin=0.5, xmax=16.5,
    xtick style={color=black},
]
\path [draw=black, fill=navy00127]
(axis cs:0.75,0.116310000419617)
--(axis cs:1.25,0.116310000419617)
--(axis cs:1.25,0.171472042798996)
--(axis cs:0.75,0.171472042798996)
--(axis cs:0.75,0.116310000419617)
--cycle;
\addplot [black]
table {%
1 0.116310000419617
1 0.0769699811935424
};
\addplot [black]
table {%
1 0.171472042798996
1 0.210569262504578
};
\addplot [black]
table {%
0.875 0.0769699811935424
1.125 0.0769699811935424
};
\addplot [black]
table {%
0.875 0.210569262504578
1.125 0.210569262504578
};
\path [draw=black, fill=mediumblue00204]
(axis cs:1.75,0.150776028633118)
--(axis cs:2.25,0.150776028633118)
--(axis cs:2.25,0.188243716955185)
--(axis cs:1.75,0.188243716955185)
--(axis cs:1.75,0.150776028633118)
--cycle;
\addplot [black]
table {%
2 0.150776028633118
2 0.113480567932129
};
\addplot [black]
table {%
2 0.188243716955185
2 0.207288026809692
};
\addplot [black]
table {%
1.875 0.113480567932129
2.125 0.113480567932129
};
\addplot [black]
table {%
1.875 0.207288026809692
2.125 0.207288026809692
};
\path [draw=black, fill=blue08255]
(axis cs:2.75,0.21919059753418)
--(axis cs:3.25,0.21919059753418)
--(axis cs:3.25,0.275328636169434)
--(axis cs:2.75,0.275328636169434)
--(axis cs:2.75,0.21919059753418)
--cycle;
\addplot [black]
table {%
3 0.21919059753418
3 0.181585311889648
};
\addplot [black]
table {%
3 0.275328636169434
3 0.312938928604126
};
\addplot [black]
table {%
2.875 0.181585311889648
3.125 0.181585311889648
};
\addplot [black]
table {%
2.875 0.312938928604126
3.125 0.312938928604126
};
\path [draw=black, fill=dodgerblue076255]
(axis cs:3.75,0.23738557100296)
--(axis cs:4.25,0.23738557100296)
--(axis cs:4.25,0.256327271461487)
--(axis cs:3.75,0.256327271461487)
--(axis cs:3.75,0.23738557100296)
--cycle;
\addplot [black]
table {%
4 0.23738557100296
4 0.21859073638916
};
\addplot [black]
table {%
4 0.256327271461487
4 0.275116205215454
};
\addplot [black]
table {%
3.875 0.21859073638916
4.125 0.21859073638916
};
\addplot [black]
table {%
3.875 0.275116205215454
4.125 0.275116205215454
};
\path [draw=black, fill=dodgerblue0144255]
(axis cs:4.75,0.238173365592957)
--(axis cs:5.25,0.238173365592957)
--(axis cs:5.25,0.294364929199219)
--(axis cs:4.75,0.294364929199219)
--(axis cs:4.75,0.238173365592957)
--cycle;
\addplot [black]
table {%
5 0.238173365592957
5 0.200715065002441
};
\addplot [black]
table {%
5 0.294364929199219
5 0.332045078277588
};
\addplot [black]
table {%
4.875 0.200715065002441
5.125 0.200715065002441
};
\addplot [black]
table {%
4.875 0.332045078277588
5.125 0.332045078277588
};
\path [draw=black, fill=deepskyblue0212255]
(axis cs:5.75,0.239026069641113)
--(axis cs:6.25,0.239026069641113)
--(axis cs:6.25,0.276591420173645)
--(axis cs:5.75,0.276591420173645)
--(axis cs:5.75,0.239026069641113)
--cycle;
\addplot [black]
table {%
6 0.239026069641113
6 0.201494216918945
};
\addplot [black]
table {%
6 0.276591420173645
6 0.31415319442749
};
\addplot [black]
table {%
5.875 0.201494216918945
6.125 0.201494216918945
};
\addplot [black]
table {%
5.875 0.31415319442749
6.125 0.31415319442749
};
\path [draw=black, fill=turquoise41255205]
(axis cs:6.75,0.326181411743164)
--(axis cs:7.25,0.326181411743164)
--(axis cs:7.25,0.346364498138428)
--(axis cs:6.75,0.346364498138428)
--(axis cs:6.75,0.326181411743164)
--cycle;
\addplot [black]
table {%
7 0.326181411743164
7 0.307453155517578
};
\addplot [black]
table {%
7 0.346364498138428
7 0.36529016494751
};
\addplot [black]
table {%
6.875 0.307453155517578
7.125 0.307453155517578
};
\addplot [black]
table {%
6.875 0.36529016494751
7.125 0.36529016494751
};
\path [draw=black, fill=lightgreen95255150]
(axis cs:7.75,0.230521202087402)
--(axis cs:8.25,0.230521202087402)
--(axis cs:8.25,0.266544342041016)
--(axis cs:7.75,0.266544342041016)
--(axis cs:7.75,0.230521202087402)
--cycle;
\addplot [black]
table {%
8 0.230521202087402
8 0.210283756256103
};
\addplot [black]
table {%
8 0.266544342041016
8 0.287058353424072
};
\addplot [black]
table {%
7.875 0.210283756256103
8.125 0.210283756256103
};
\addplot [black]
table {%
7.875 0.287058353424072
8.125 0.287058353424072
};
\path [draw=black, fill=lightgreen15025595]
(axis cs:8.75,0.248899459838867)
--(axis cs:9.25,0.248899459838867)
--(axis cs:9.25,0.323860168457031)
--(axis cs:8.75,0.323860168457031)
--(axis cs:8.75,0.248899459838867)
--cycle;
\addplot [black]
table {%
9 0.248899459838867
9 0.192598342895508
};
\addplot [black]
table {%
9 0.323860168457031
9 0.380114555358887
};
\addplot [black]
table {%
8.875 0.192598342895508
9.125 0.192598342895508
};
\addplot [black]
table {%
8.875 0.380114555358887
9.125 0.380114555358887
};
\path [draw=black, fill=greenyellow20525541]
(axis cs:9.75,0.366082191467285)
--(axis cs:10.25,0.366082191467285)
--(axis cs:10.25,0.403562307357788)
--(axis cs:9.75,0.403562307357788)
--(axis cs:9.75,0.366082191467285)
--cycle;
\addplot [black]
table {%
10 0.366082191467285
10 0.347259521484375
};
\addplot [black]
table {%
10 0.403562307357788
10 0.422457695007324
};
\addplot [black]
table {%
9.875 0.347259521484375
10.125 0.347259521484375
};
\addplot [black]
table {%
9.875 0.422457695007324
10.125 0.422457695007324
};
\path [draw=black, fill=gold2552290]
(axis cs:10.75,0.318748474121094)
--(axis cs:11.25,0.318748474121094)
--(axis cs:11.25,0.356239795684814)
--(axis cs:10.75,0.356239795684814)
--(axis cs:10.75,0.318748474121094)
--cycle;
\addplot [black]
table {%
11 0.318748474121094
11 0.299851417541504
};
\addplot [black]
table {%
11 0.356239795684814
11 0.393561363220215
};
\addplot [black]
table {%
10.875 0.299851417541504
11.125 0.299851417541504
};
\addplot [black]
table {%
10.875 0.393561363220215
11.125 0.393561363220215
};
\path [draw=black, fill=orange2551660]
(axis cs:11.75,0.335224628448486)
--(axis cs:12.25,0.335224628448486)
--(axis cs:12.25,0.372061252593994)
--(axis cs:11.75,0.372061252593994)
--(axis cs:11.75,0.335224628448486)
--cycle;
\addplot [black]
table {%
12 0.335224628448486
12 0.299124717712402
};
\addplot [black]
table {%
12 0.372061252593994
12 0.404152870178223
};
\addplot [black]
table {%
11.875 0.299124717712402
12.125 0.299124717712402
};
\addplot [black]
table {%
11.875 0.404152870178223
12.125 0.404152870178223
};
\path [draw=black, fill=orangered2551030]
(axis cs:12.75,0.144416809082031)
--(axis cs:13.25,0.144416809082031)
--(axis cs:13.25,0.181803703308105)
--(axis cs:12.75,0.181803703308105)
--(axis cs:12.75,0.144416809082031)
--cycle;
\addplot [black]
table {%
13 0.144416809082031
13 0.10703182220459
};
\addplot [black]
table {%
13 0.181803703308105
13 0.200812339782715
};
\addplot [black]
table {%
12.875 0.10703182220459
13.125 0.10703182220459
};
\addplot [black]
table {%
12.875 0.200812339782715
13.125 0.200812339782715
};
\path [draw=black, fill=orangered255400]
(axis cs:13.75,0.348304748535156)
--(axis cs:14.25,0.348304748535156)
--(axis cs:14.25,0.38585352897644)
--(axis cs:13.75,0.38585352897644)
--(axis cs:13.75,0.348304748535156)
--cycle;
\addplot [black]
table {%
14 0.348304748535156
14 0.310796737670898
};
\addplot [black]
table {%
14 0.38585352897644
14 0.406355857849121
};
\addplot [black]
table {%
13.875 0.310796737670898
14.125 0.310796737670898
};
\addplot [black]
table {%
13.875 0.406355857849121
14.125 0.406355857849121
};
\path [draw=black, fill=red20400]
(axis cs:14.75,0.354506492614746)
--(axis cs:15.25,0.354506492614746)
--(axis cs:15.25,0.378013372421265)
--(axis cs:14.75,0.378013372421265)
--(axis cs:14.75,0.354506492614746)
--cycle;
\addplot [black]
table {%
15 0.354506492614746
15 0.340409278869629
};
\addplot [black]
table {%
15 0.378013372421265
15 0.396824836730957
};
\addplot [black]
table {%
14.875 0.340409278869629
15.125 0.340409278869629
};
\addplot [black]
table {%
14.875 0.396824836730957
15.125 0.396824836730957
};
\path [draw=black, fill=maroon12700]
(axis cs:15.75,0.367395401000977)
--(axis cs:16.25,0.367395401000977)
--(axis cs:16.25,0.386157989501953)
--(axis cs:15.75,0.386157989501953)
--(axis cs:15.75,0.367395401000977)
--cycle;
\addplot [black]
table {%
16 0.367395401000977
16 0.348634719848633
};
\addplot [black]
table {%
16 0.386157989501953
16 0.404918670654297
};
\addplot [black]
table {%
15.875 0.348634719848633
16.125 0.348634719848633
};
\addplot [black]
table {%
15.875 0.404918670654297
16.125 0.404918670654297
};
\addplot [thick, black]
table {%
0.75 0.152627825737
1.25 0.152627825737
};
\addplot [thick, black]
table {%
1.75 0.169574499130249
2.25 0.169574499130249
};
\addplot [thick, black]
table {%
2.75 0.23798131942749
3.25 0.23798131942749
};
\addplot [thick, black]
table {%
3.75 0.256134271621704
4.25 0.256134271621704
};
\addplot [thick, black]
table {%
4.75 0.256934404373169
5.25 0.256934404373169
};
\addplot [thick, black]
table {%
5.75 0.257806301116943
6.25 0.257806301116943
};
\addplot [thick, black]
table {%
6.75 0.327767848968506
7.25 0.327767848968506
};
\addplot [thick, black]
table {%
7.75 0.247931480407715
8.25 0.247931480407715
};
\addplot [thick, black]
table {%
8.75 0.286366939544678
9.25 0.286366939544678
};
\addplot [thick, black]
table {%
9.75 0.384804725646973
10.25 0.384804725646973
};
\addplot [thick, black]
table {%
10.75 0.337547302246094
11.25 0.337547302246094
};
\addplot [thick, black]
table {%
11.75 0.355051517486572
12.25 0.355051517486572
};
\addplot [thick, black]
table {%
12.75 0.163104057312012
13.25 0.163104057312012
};
\addplot [thick, black]
table {%
13.75 0.367072105407715
14.25 0.367072105407715
};
\addplot [thick, black]
table {%
14.75 0.359274864196777
15.25 0.359274864196777
};
\addplot [thick, black]
table {%
15.75 0.367395401000977
16.25 0.367395401000977
};
\end{axis}

\end{tikzpicture}

%% file: tex/boxplot_errors/error_ByAngle_roll_180.0_rosbag_try1.csv.tex
\begin{tikzpicture}

\definecolor{blue08255}{RGB}{0,8,255}
\definecolor{darkgray176}{RGB}{176,176,176}
\definecolor{deepskyblue0212255}{RGB}{0,212,255}
\definecolor{dodgerblue0144255}{RGB}{0,144,255}
\definecolor{dodgerblue076255}{RGB}{0,76,255}
\definecolor{gainsboro229}{RGB}{229,229,229}
\definecolor{gold2552290}{RGB}{255,229,0}
\definecolor{greenyellow20525541}{RGB}{205,255,41}
\definecolor{lightgreen15025595}{RGB}{150,255,95}
\definecolor{lightgreen95255150}{RGB}{95,255,150}
\definecolor{maroon12700}{RGB}{127,0,0}
\definecolor{mediumblue00204}{RGB}{0,0,204}
\definecolor{navy00127}{RGB}{0,0,127}
\definecolor{orange2551660}{RGB}{255,166,0}
\definecolor{orangered2551030}{RGB}{255,103,0}
\definecolor{orangered255400}{RGB}{255,40,0}
\definecolor{red20400}{RGB}{204,0,0}
\definecolor{turquoise41255205}{RGB}{41,255,205}

\begin{axis}[
    width=\figurewidth,
    height=\figureheight,
    axis background/.style={fill=white},
    axis line style={white},
    tick align=outside,
    x grid style={white},
    xmajorgrids,
    xmajorticks=true,
    y grid style={white},
    ymajorgrids,
    ymajorticks=true,
    y grid style={white!69.0196078431373!black},
    ytick style={color=black},
    xmajorgrids,
    xminorgrids,
    ymajorgrids,
    ymajorticks=true,
    yticklabel style={
            /pgf/number format/fixed,
            /pgf/number format/precision=5
        },
    scaled y ticks=false,
    ylabel near ticks, 
    ylabel shift={-1pt},
    ymin=0, ymax=1,
    xmin=0.5, xmax=16.5,
    xtick style={color=black},
]
\path [draw=black, fill=navy00127]
(axis cs:0.75,0.114594578742981)
--(axis cs:1.25,0.114594578742981)
--(axis cs:1.25,0.150488376617432)
--(axis cs:0.75,0.150488376617432)
--(axis cs:0.75,0.114594578742981)
--cycle;
\addplot [black]
table {%
1 0.114594578742981
1 0.094204306602478
};
\addplot [black]
table {%
1 0.150488376617432
1 0.170938849449158
};
\addplot [black]
table {%
0.875 0.094204306602478
1.125 0.094204306602478
};
\addplot [black]
table {%
0.875 0.170938849449158
1.125 0.170938849449158
};
\path [draw=black, fill=mediumblue00204]
(axis cs:1.75,0.13106095790863)
--(axis cs:2.25,0.13106095790863)
--(axis cs:2.25,0.187054991722107)
--(axis cs:1.75,0.187054991722107)
--(axis cs:1.75,0.13106095790863)
--cycle;
\addplot [black]
table {%
2 0.13106095790863
2 0.0752668380737304
};
\addplot [black]
table {%
2 0.187054991722107
2 0.225212812423706
};
\addplot [black]
table {%
1.875 0.0752668380737304
2.125 0.0752668380737304
};
\addplot [black]
table {%
1.875 0.225212812423706
2.125 0.225212812423706
};
\path [draw=black, fill=blue08255]
(axis cs:2.75,0.20030415058136)
--(axis cs:3.25,0.20030415058136)
--(axis cs:3.25,0.237622976303101)
--(axis cs:2.75,0.237622976303101)
--(axis cs:2.75,0.20030415058136)
--cycle;
\addplot [black]
table {%
3 0.20030415058136
3 0.181328296661377
};
\addplot [black]
table {%
3 0.237622976303101
3 0.256833791732788
};
\addplot [black]
table {%
2.875 0.181328296661377
3.125 0.181328296661377
};
\addplot [black]
table {%
2.875 0.256833791732788
3.125 0.256833791732788
};
\path [draw=black, fill=dodgerblue076255]
(axis cs:3.75,0.237405717372894)
--(axis cs:4.25,0.237405717372894)
--(axis cs:4.25,0.27495002746582)
--(axis cs:3.75,0.27495002746582)
--(axis cs:3.75,0.237405717372894)
--cycle;
\addplot [black]
table {%
4 0.237405717372894
4 0.218488216400146
};
\addplot [black]
table {%
4 0.27495002746582
4 0.31242036819458
};
\addplot [black]
table {%
3.875 0.218488216400146
4.125 0.218488216400146
};
\addplot [black]
table {%
3.875 0.31242036819458
4.125 0.31242036819458
};
\path [draw=black, fill=dodgerblue0144255]
(axis cs:4.75,0.238150715827942)
--(axis cs:5.25,0.238150715827942)
--(axis cs:5.25,0.275623321533203)
--(axis cs:4.75,0.275623321533203)
--(axis cs:4.75,0.238150715827942)
--cycle;
\addplot [black]
table {%
5 0.238150715827942
5 0.200689315795898
};
\addplot [black]
table {%
5 0.275623321533203
5 0.294446468353271
};
\addplot [black]
table {%
4.875 0.200689315795898
5.125 0.200689315795898
};
\addplot [black]
table {%
4.875 0.294446468353271
5.125 0.294446468353271
};
\path [draw=black, fill=deepskyblue0212255]
(axis cs:5.75,0.254299640655518)
--(axis cs:6.25,0.254299640655518)
--(axis cs:6.25,0.291487693786621)
--(axis cs:5.75,0.291487693786621)
--(axis cs:5.75,0.254299640655518)
--cycle;
\addplot [black]
table {%
6 0.254299640655518
6 0.217437267303467
};
\addplot [black]
table {%
6 0.291487693786621
6 0.328587532043457
};
\addplot [black]
table {%
5.875 0.217437267303467
6.125 0.217437267303467
};
\addplot [black]
table {%
5.875 0.328587532043457
6.125 0.328587532043457
};
\path [draw=black, fill=turquoise41255205]
(axis cs:6.75,0.296920895576477)
--(axis cs:7.25,0.296920895576477)
--(axis cs:7.25,0.321650862693787)
--(axis cs:6.75,0.321650862693787)
--(axis cs:6.75,0.296920895576477)
--cycle;
\addplot [black]
table {%
7 0.296920895576477
7 0.278168201446533
};
\addplot [black]
table {%
7 0.321650862693787
7 0.336156845092773
};
\addplot [black]
table {%
6.875 0.278168201446533
7.125 0.278168201446533
};
\addplot [black]
table {%
6.875 0.336156845092773
7.125 0.336156845092773
};
\path [draw=black, fill=lightgreen95255150]
(axis cs:7.75,0.324314117431641)
--(axis cs:8.25,0.324314117431641)
--(axis cs:8.25,0.344200372695923)
--(axis cs:7.75,0.344200372695923)
--(axis cs:7.75,0.324314117431641)
--cycle;
\addplot [black]
table {%
8 0.324314117431641
8 0.30547571182251
};
\addplot [black]
table {%
8 0.344200372695923
8 0.362984657287598
};
\addplot [black]
table {%
7.875 0.30547571182251
8.125 0.30547571182251
};
\addplot [black]
table {%
7.875 0.362984657287598
8.125 0.362984657287598
};
\path [draw=black, fill=lightgreen15025595]
(axis cs:8.75,0.32374095916748)
--(axis cs:9.25,0.32374095916748)
--(axis cs:9.25,0.361306190490723)
--(axis cs:8.75,0.361306190490723)
--(axis cs:8.75,0.32374095916748)
--cycle;
\addplot [black]
table {%
9 0.32374095916748
9 0.286261558532715
};
\addplot [black]
table {%
9 0.361306190490723
9 0.398869514465332
};
\addplot [black]
table {%
8.875 0.286261558532715
9.125 0.286261558532715
};
\addplot [black]
table {%
8.875 0.398869514465332
9.125 0.398869514465332
};
\path [draw=black, fill=greenyellow20525541]
(axis cs:9.75,0.422382116317749)
--(axis cs:10.25,0.422382116317749)
--(axis cs:10.25,0.497285842895508)
--(axis cs:9.75,0.497285842895508)
--(axis cs:9.75,0.422382116317749)
--cycle;
\addplot [black]
table {%
10 0.422382116317749
10 0.38465690612793
};
\addplot [black]
table {%
10 0.497285842895508
10 0.553595542907715
};
\addplot [black]
table {%
9.875 0.38465690612793
10.125 0.38465690612793
};
\addplot [black]
table {%
9.875 0.553595542907715
10.125 0.553595542907715
};
\path [draw=black, fill=gold2552290]
(axis cs:10.75,0.301021575927734)
--(axis cs:11.25,0.301021575927734)
--(axis cs:11.25,0.395150423049927)
--(axis cs:10.75,0.395150423049927)
--(axis cs:10.75,0.301021575927734)
--cycle;
\addplot [black]
table {%
11 0.301021575927734
11 0.261898040771484
};
\addplot [black]
table {%
11 0.395150423049927
11 0.488753318786621
};
\addplot [black]
table {%
10.875 0.261898040771484
11.125 0.261898040771484
};
\addplot [black]
table {%
10.875 0.488753318786621
11.125 0.488753318786621
};
\path [draw=black, fill=orange2551660]
(axis cs:11.75,0.372710943222046)
--(axis cs:12.25,0.372710943222046)
--(axis cs:12.25,0.410191774368286)
--(axis cs:11.75,0.410191774368286)
--(axis cs:11.75,0.372710943222046)
--cycle;
\addplot [black]
table {%
12 0.372710943222046
12 0.353878021240234
};
\addplot [black]
table {%
12 0.410191774368286
12 0.429320335388184
};
\addplot [black]
table {%
11.875 0.353878021240234
12.125 0.353878021240234
};
\addplot [black]
table {%
11.875 0.429320335388184
12.125 0.429320335388184
};
\path [draw=black, fill=orangered2551030]
(axis cs:12.75,0.369698524475098)
--(axis cs:13.25,0.369698524475098)
--(axis cs:13.25,0.388625144958496)
--(axis cs:12.75,0.388625144958496)
--(axis cs:12.75,0.369698524475098)
--cycle;
\addplot [black]
table {%
13 0.369698524475098
13 0.350823402404785
};
\addplot [black]
table {%
13 0.388625144958496
13 0.407386779785156
};
\addplot [black]
table {%
12.875 0.350823402404785
13.125 0.350823402404785
};
\addplot [black]
table {%
12.875 0.407386779785156
13.125 0.407386779785156
};
\path [draw=black, fill=orangered255400]
(axis cs:13.75,0.32868480682373)
--(axis cs:14.25,0.32868480682373)
--(axis cs:14.25,0.365232467651367)
--(axis cs:13.75,0.365232467651367)
--(axis cs:13.75,0.32868480682373)
--cycle;
\addplot [black]
table {%
14 0.32868480682373
14 0.308381080627441
};
\addplot [black]
table {%
14 0.365232467651367
14 0.38521671295166
};
\addplot [black]
table {%
13.875 0.308381080627441
14.125 0.308381080627441
};
\addplot [black]
table {%
13.875 0.38521671295166
14.125 0.38521671295166
};
\path [draw=black, fill=red20400]
(axis cs:14.75,0.321537017822266)
--(axis cs:15.25,0.321537017822266)
--(axis cs:15.25,0.358980655670166)
--(axis cs:14.75,0.358980655670166)
--(axis cs:14.75,0.321537017822266)
--cycle;
\addplot [black]
table {%
15 0.321537017822266
15 0.302580833435059
};
\addplot [black]
table {%
15 0.358980655670166
15 0.37782096862793
};
\addplot [black]
table {%
14.875 0.302580833435059
15.125 0.302580833435059
};
\addplot [black]
table {%
14.875 0.37782096862793
15.125 0.37782096862793
};
\path [draw=black, fill=maroon12700]
(axis cs:15.75,0.329874038696289)
--(axis cs:16.25,0.329874038696289)
--(axis cs:16.25,0.367395401000977)
--(axis cs:15.75,0.367395401000977)
--(axis cs:15.75,0.329874038696289)
--cycle;
\addplot [black]
table {%
16 0.329874038696289
16 0.311111450195312
};
\addplot [black]
table {%
16 0.367395401000977
16 0.386157989501953
};
\addplot [black]
table {%
15.875 0.311111450195312
16.125 0.311111450195312
};
\addplot [black]
table {%
15.875 0.386157989501953
16.125 0.386157989501953
};
\addplot [thick, black]
table {%
0.75 0.133355617523193
1.25 0.133355617523193
};
\addplot [thick, black]
table {%
1.75 0.150038123130798
2.25 0.150038123130798
};
\addplot [thick, black]
table {%
2.75 0.219026803970337
3.25 0.219026803970337
};
\addplot [thick, black]
table {%
3.75 0.256189346313477
4.25 0.256189346313477
};
\addplot [thick, black]
table {%
4.75 0.256892204284668
5.25 0.256892204284668
};
\addplot [thick, black]
table {%
5.75 0.272879600524902
6.25 0.272879600524902
};
\addplot [thick, black]
table {%
6.75 0.315673112869263
7.25 0.315673112869263
};
\addplot [thick, black]
table {%
7.75 0.343014717102051
8.25 0.343014717102051
};
\addplot [thick, black]
table {%
8.75 0.342508792877197
9.25 0.342508792877197
};
\addplot [thick, black]
table {%
9.75 0.459781646728516
10.25 0.459781646728516
};
\addplot [thick, black]
table {%
10.75 0.338409423828125
11.25 0.338409423828125
};
\addplot [thick, black]
table {%
11.75 0.391427993774414
12.25 0.391427993774414
};
\addplot [thick, black]
table {%
12.75 0.388404369354248
13.25 0.388404369354248
};
\addplot [thick, black]
table {%
13.75 0.347409248352051
14.25 0.347409248352051
};
\addplot [thick, black]
table {%
14.75 0.340214729309082
15.25 0.340214729309082
};
\addplot [thick, black]
table {%
15.75 0.348634719848633
16.25 0.348634719848633
};
\end{axis}

\end{tikzpicture}

%% file: tex/boxplot_errors/error_ByAngle_roll_270.0_rosbag_try1.csv.tex
\begin{tikzpicture}

\definecolor{blue08255}{RGB}{0,8,255}
\definecolor{darkgray176}{RGB}{176,176,176}
\definecolor{deepskyblue0212255}{RGB}{0,212,255}
\definecolor{dodgerblue0144255}{RGB}{0,144,255}
\definecolor{dodgerblue076255}{RGB}{0,76,255}
\definecolor{gainsboro229}{RGB}{229,229,229}
\definecolor{gold2552290}{RGB}{255,229,0}
\definecolor{greenyellow20525541}{RGB}{205,255,41}
\definecolor{lightgreen15025595}{RGB}{150,255,95}
\definecolor{lightgreen95255150}{RGB}{95,255,150}
\definecolor{maroon12700}{RGB}{127,0,0}
\definecolor{mediumblue00204}{RGB}{0,0,204}
\definecolor{navy00127}{RGB}{0,0,127}
\definecolor{orange2551660}{RGB}{255,166,0}
\definecolor{orangered2551030}{RGB}{255,103,0}
\definecolor{orangered255400}{RGB}{255,40,0}
\definecolor{red20400}{RGB}{204,0,0}
\definecolor{turquoise41255205}{RGB}{41,255,205}

\begin{axis}[
    width=\figurewidth,
    height=\figureheight,
    axis background/.style={fill=white},
    axis line style={white},
    tick align=outside,
    x grid style={white},
    xmajorgrids,
    xmajorticks=true,
    y grid style={white},
    ymajorgrids,
    ymajorticks=true,
    y grid style={white!69.0196078431373!black},
    ytick style={color=black},
    xmajorgrids,
    xminorgrids,
    ymajorgrids,
    ymajorticks=true,
    yticklabel style={
            /pgf/number format/fixed,
            /pgf/number format/precision=5
        },
    scaled y ticks=false,
    ylabel near ticks, 
    ylabel shift={-1pt},
    ymin=0, ymax=1,
    xmin=0.5, xmax=16.5,
    xtick style={color=black},
]
\path [draw=black, fill=navy00127]
(axis cs:0.75,0.152544856071472)
--(axis cs:1.25,0.152544856071472)
--(axis cs:1.25,0.190046668052673)
--(axis cs:0.75,0.190046668052673)
--(axis cs:0.75,0.152544856071472)
--cycle;
\addplot [black]
table {%
1 0.152544856071472
1 0.115247011184692
};
\addplot [black]
table {%
1 0.190046668052673
1 0.227413892745972
};
\addplot [black]
table {%
0.875 0.115247011184692
1.125 0.115247011184692
};
\addplot [black]
table {%
0.875 0.227413892745972
1.125 0.227413892745972
};
\path [draw=black, fill=mediumblue00204]
(axis cs:1.75,0.150921523571014)
--(axis cs:2.25,0.150921523571014)
--(axis cs:2.25,0.18841826915741)
--(axis cs:1.75,0.18841826915741)
--(axis cs:1.75,0.150921523571014)
--cycle;
\addplot [black]
table {%
2 0.150921523571014
2 0.113481044769287
};
\addplot [black]
table {%
2 0.18841826915741
2 0.207255482673645
};
\addplot [black]
table {%
1.875 0.113481044769287
2.125 0.113481044769287
};
\addplot [black]
table {%
1.875 0.207255482673645
2.125 0.207255482673645
};
\path [draw=black, fill=blue08255]
(axis cs:2.75,0.239305019378662)
--(axis cs:3.25,0.239305019378662)
--(axis cs:3.25,0.277214050292969)
--(axis cs:2.75,0.277214050292969)
--(axis cs:2.75,0.239305019378662)
--cycle;
\addplot [black]
table {%
3 0.239305019378662
3 0.201651334762573
};
\addplot [black]
table {%
3 0.277214050292969
3 0.314064741134643
};
\addplot [black]
table {%
2.875 0.201651334762573
3.125 0.201651334762573
};
\addplot [black]
table {%
2.875 0.314064741134643
3.125 0.314064741134643
};
\path [draw=black, fill=dodgerblue076255]
(axis cs:3.75,0.256108283996582)
--(axis cs:4.25,0.256108283996582)
--(axis cs:4.25,0.293630480766296)
--(axis cs:3.75,0.293630480766296)
--(axis cs:3.75,0.256108283996582)
--cycle;
\addplot [black]
table {%
4 0.256108283996582
4 0.218589544296265
};
\addplot [black]
table {%
4 0.293630480766296
4 0.331121206283569
};
\addplot [black]
table {%
3.875 0.218589544296265
4.125 0.218589544296265
};
\addplot [black]
table {%
3.875 0.331121206283569
4.125 0.331121206283569
};
\path [draw=black, fill=dodgerblue0144255]
(axis cs:4.75,0.256444931030273)
--(axis cs:5.25,0.256444931030273)
--(axis cs:5.25,0.293988227844238)
--(axis cs:4.75,0.293988227844238)
--(axis cs:4.75,0.256444931030273)
--cycle;
\addplot [black]
table {%
5 0.256444931030273
5 0.219015598297119
};
\addplot [black]
table {%
5 0.293988227844238
5 0.331488132476807
};
\addplot [black]
table {%
4.875 0.219015598297119
5.125 0.219015598297119
};
\addplot [black]
table {%
4.875 0.331488132476807
5.125 0.331488132476807
};
\path [draw=black, fill=deepskyblue0212255]
(axis cs:5.75,0.277772426605225)
--(axis cs:6.25,0.277772426605225)
--(axis cs:6.25,0.315295219421387)
--(axis cs:5.75,0.315295219421387)
--(axis cs:5.75,0.277772426605225)
--cycle;
\addplot [black]
table {%
6 0.277772426605225
6 0.2403564453125
};
\addplot [black]
table {%
6 0.315295219421387
6 0.352499961853027
};
\addplot [black]
table {%
5.875 0.2403564453125
6.125 0.2403564453125
};
\addplot [black]
table {%
5.875 0.352499961853027
6.125 0.352499961853027
};
\path [draw=black, fill=turquoise41255205]
(axis cs:6.75,0.273039817810059)
--(axis cs:7.25,0.273039817810059)
--(axis cs:7.25,0.29661226272583)
--(axis cs:6.75,0.29661226272583)
--(axis cs:6.75,0.273039817810059)
--cycle;
\addplot [black]
table {%
7 0.273039817810059
7 0.258915901184082
};
\addplot [black]
table {%
7 0.29661226272583
7 0.315523624420166
};
\addplot [black]
table {%
6.875 0.258915901184082
7.125 0.258915901184082
};
\addplot [black]
table {%
6.875 0.315523624420166
7.125 0.315523624420166
};
\path [draw=black, fill=lightgreen95255150]
(axis cs:7.75,0.269220352172852)
--(axis cs:8.25,0.269220352172852)
--(axis cs:8.25,0.308251142501831)
--(axis cs:7.75,0.308251142501831)
--(axis cs:7.75,0.269220352172852)
--cycle;
\addplot [black]
table {%
8 0.269220352172852
8 0.23169469833374
};
\addplot [black]
table {%
8 0.308251142501831
8 0.345947742462158
};
\addplot [black]
table {%
7.875 0.23169469833374
8.125 0.23169469833374
};
\addplot [black]
table {%
7.875 0.345947742462158
8.125 0.345947742462158
};
\path [draw=black, fill=lightgreen15025595]
(axis cs:8.75,0.324084281921387)
--(axis cs:9.25,0.324084281921387)
--(axis cs:9.25,0.361477851867676)
--(axis cs:8.75,0.361477851867676)
--(axis cs:8.75,0.324084281921387)
--cycle;
\addplot [black]
table {%
9 0.324084281921387
9 0.305133819580078
};
\addplot [black]
table {%
9 0.361477851867676
9 0.381897926330566
};
\addplot [black]
table {%
8.875 0.305133819580078
9.125 0.305133819580078
};
\addplot [black]
table {%
8.875 0.381897926330566
9.125 0.381897926330566
};
\path [draw=black, fill=greenyellow20525541]
(axis cs:9.75,0.40387225151062)
--(axis cs:10.25,0.40387225151062)
--(axis cs:10.25,0.422752380371094)
--(axis cs:9.75,0.422752380371094)
--(axis cs:9.75,0.40387225151062)
--cycle;
\addplot [black]
table {%
10 0.40387225151062
10 0.385002136230469
};
\addplot [black]
table {%
10 0.422752380371094
10 0.441581726074219
};
\addplot [black]
table {%
9.875 0.385002136230469
10.125 0.385002136230469
};
\addplot [black]
table {%
9.875 0.441581726074219
10.125 0.441581726074219
};
\path [draw=black, fill=gold2552290]
(axis cs:10.75,0.244851112365723)
--(axis cs:11.25,0.244851112365723)
--(axis cs:11.25,0.300901174545288)
--(axis cs:10.75,0.300901174545288)
--(axis cs:10.75,0.244851112365723)
--cycle;
\addplot [black]
table {%
11 0.244851112365723
11 0.225847244262695
};
\addplot [black]
table {%
11 0.300901174545288
11 0.338658332824707
};
\addplot [black]
table {%
10.875 0.225847244262695
11.125 0.225847244262695
};
\addplot [black]
table {%
10.875 0.338658332824707
11.125 0.338658332824707
};
\path [draw=black, fill=orange2551660]
(axis cs:11.75,0.335920333862305)
--(axis cs:12.25,0.335920333862305)
--(axis cs:12.25,0.373442649841309)
--(axis cs:11.75,0.373442649841309)
--(axis cs:11.75,0.335920333862305)
--cycle;
\addplot [black]
table {%
12 0.335920333862305
12 0.300053596496582
};
\addplot [black]
table {%
12 0.373442649841309
12 0.410919189453125
};
\addplot [black]
table {%
11.875 0.300053596496582
12.125 0.300053596496582
};
\addplot [black]
table {%
11.875 0.410919189453125
12.125 0.410919189453125
};
\path [draw=black, fill=orangered2551030]
(axis cs:12.75,0.313026428222656)
--(axis cs:13.25,0.313026428222656)
--(axis cs:13.25,0.333195686340332)
--(axis cs:12.75,0.333195686340332)
--(axis cs:12.75,0.313026428222656)
--cycle;
\addplot [black]
table {%
13 0.313026428222656
13 0.294256210327148
};
\addplot [black]
table {%
13 0.333195686340332
13 0.351943969726562
};
\addplot [black]
table {%
12.875 0.294256210327148
13.125 0.294256210327148
};
\addplot [black]
table {%
12.875 0.351943969726562
13.125 0.351943969726562
};
\path [draw=black, fill=orangered255400]
(axis cs:13.75,0.348821640014648)
--(axis cs:14.25,0.348821640014648)
--(axis cs:14.25,0.423708200454712)
--(axis cs:13.75,0.423708200454712)
--(axis cs:13.75,0.348821640014648)
--cycle;
\addplot [black]
table {%
14 0.348821640014648
14 0.292393684387207
};
\addplot [black]
table {%
14 0.423708200454712
14 0.480401039123535
};
\addplot [black]
table {%
13.875 0.292393684387207
14.125 0.292393684387207
};
\addplot [black]
table {%
13.875 0.480401039123535
14.125 0.480401039123535
};
\path [draw=black, fill=red20400]
(axis cs:14.75,0.396556854248047)
--(axis cs:15.25,0.396556854248047)
--(axis cs:15.25,0.43404746055603)
--(axis cs:14.75,0.43404746055603)
--(axis cs:14.75,0.396556854248047)
--cycle;
\addplot [black]
table {%
15 0.396556854248047
15 0.359095573425293
};
\addplot [black]
table {%
15 0.43404746055603
15 0.471508979797363
};
\addplot [black]
table {%
14.875 0.359095573425293
15.125 0.359095573425293
};
\addplot [black]
table {%
14.875 0.471508979797363
15.125 0.471508979797363
};
\path [draw=black, fill=maroon12700]
(axis cs:15.75,0.381467342376709)
--(axis cs:16.25,0.381467342376709)
--(axis cs:16.25,0.404918670654297)
--(axis cs:15.75,0.404918670654297)
--(axis cs:15.75,0.381467342376709)
--cycle;
\addplot [black]
table {%
16 0.381467342376709
16 0.367395401000977
};
\addplot [black]
table {%
16 0.404918670654297
16 0.423681259155273
};
\addplot [black]
table {%
15.875 0.367395401000977
16.125 0.367395401000977
};
\addplot [black]
table {%
15.875 0.423681259155273
16.125 0.423681259155273
};
\addplot [thick, black]
table {%
0.75 0.171298742294311
1.25 0.171298742294311
};
\addplot [thick, black]
table {%
1.75 0.169687271118164
2.25 0.169687271118164
};
\addplot [thick, black]
table {%
2.75 0.258430242538452
3.25 0.258430242538452
};
\addplot [thick, black]
table {%
3.75 0.27485466003418
4.25 0.27485466003418
};
\addplot [thick, black]
table {%
4.75 0.27515983581543
5.25 0.27515983581543
};
\addplot [thick, black]
table {%
5.75 0.296640396118164
6.25 0.296640396118164
};
\addplot [thick, black]
table {%
6.75 0.277935028076172
7.25 0.277935028076172
};
\addplot [thick, black]
table {%
7.75 0.289336442947388
8.25 0.289336442947388
};
\addplot [thick, black]
table {%
8.75 0.342707633972168
9.25 0.342707633972168
};
\addplot [thick, black]
table {%
9.75 0.403995037078857
10.25 0.403995037078857
};
\addplot [thick, black]
table {%
10.75 0.282184600830078
11.25 0.282184600830078
};
\addplot [thick, black]
table {%
11.75 0.354573249816894
12.25 0.354573249816894
};
\addplot [thick, black]
table {%
12.75 0.331737041473389
13.25 0.331737041473389
};
\addplot [thick, black]
table {%
13.75 0.386160850524902
14.25 0.386160850524902
};
\addplot [thick, black]
table {%
14.75 0.415250778198242
15.25 0.415250778198242
};
\addplot [thick, black]
table {%
15.75 0.386157989501953
16.25 0.386157989501953
};
\end{axis}

\end{tikzpicture}

%% file: tex/graphs_les/les_square_rotation_2023-03-24-19-08-41_error.tex
\begin{tikzpicture}

\definecolor{darkgray176}{RGB}{176,176,176}
\definecolor{gainsboro229}{RGB}{229,229,229}

\definecolor{mediumpurple148103189}{RGB}{148,103,189}
\definecolor{sienna1408675}{RGB}{140,86,75}

\begin{axis}[
    width=\figurewidth,
    height=\figureheight,
    axis background/.style={fill=white},
    axis line style={white},
    tick align=outside,
    x grid style={white},
    xmajorgrids,
    xmajorticks=true,
    y grid style={white},
    ymajorgrids,
    ymajorticks=true,
    y grid style={white!69.0196078431373!black},
    ytick style={color=black},
    xmajorgrids,
    xminorgrids,
    ymajorgrids,
    ymajorticks=true,
    minor y tick num = 2,
    minor y grid style={dashed},
    yminorgrids,
    yticklabel style={
            /pgf/number format/fixed,
            /pgf/number format/precision=5,
        },
    scaled y ticks=false,
    ylabel near ticks, 
    ylabel shift={-1pt},
    xmin=0.5, xmax=2.5,
    xtick style={color=black},
    xtick={1,2},
    xticklabels={$I_1$, $I_2$},
ylabel={Error (m)},
ymin=-0.0340675311597472, ymax=0.810347649381086,
]
\path [draw=black, fill=mediumpurple148103189]
(axis cs:0.75,0.205697579777033)
--(axis cs:1.25,0.205697579777033)
--(axis cs:1.25,0.370743707771836)
--(axis cs:0.75,0.370743707771836)
--(axis cs:0.75,0.205697579777033)
--cycle;
\addplot [black]
table {%
1 0.205697579777033
1 0.041119900213842
};
\addplot [black]
table {%
1 0.370743707771836
1 0.531944350363276
};
\addplot [black]
table {%
0.875 0.041119900213842
1.125 0.041119900213842
};
\addplot [black]
table {%
0.875 0.531944350363276
1.125 0.531944350363276
};
\addplot [black, mark=x, mark size=1, mark options={solid,fill opacity=0}, only marks]
table {%
1 0.0348656094308344
1 0.61212512482864
1 0.655270948830006
1 0.771965141174685
1 0.60831293289496
1 0.559996505688045
1 0.540990599331528
1 0.553095493225241
1 0.551095001477257
1 0.640974092145133
1 0.646025930281608
1 0.621604814222856
1 0.541384506346439
1 0.612057146841055
1 0.589283111174019
};
\path [draw=black, fill=sienna1408675]
(axis cs:1.75,0.135290342783937)
--(axis cs:2.25,0.135290342783937)
--(axis cs:2.25,0.288342926196206)
--(axis cs:1.75,0.288342926196206)
--(axis cs:1.75,0.135290342783937)
--cycle;
\addplot [black]
table {%
2 0.135290342783937
2 0.00431497704665429
};
\addplot [black]
table {%
2 0.288342926196206
2 0.440355914196797
};
\addplot [black]
table {%
1.875 0.00431497704665429
2.125 0.00431497704665429
};
\addplot [black]
table {%
1.875 0.440355914196797
2.125 0.440355914196797
};
\addplot [black, mark=x, mark size=1, mark options={solid,fill opacity=0}, only marks]
table {%
2 0.540864319239066
2 0.552329569330972
2 0.562004157517283
2 0.488575609149763
2 0.513887422455414
2 0.520738415295587
2 0.497038468039158
2 0.618652167000244
2 0.618763095819999
2 0.589132144472125
2 0.63970875748135
2 0.453275000337558
2 0.442228878497912
2 0.485379296653952
2 0.457633177053765
2 0.445318155421905
2 0.46456009802768
2 0.455953173338971
2 0.443686429073104
2 0.451404686881428
2 0.454151658291547
};
\addplot [thick, black]
table {%
0.75 0.307139232059425
1.25 0.307139232059425
};
\addplot [thick, black]
table {%
1.75 0.217104052811917
2.25 0.217104052811917
};
\end{axis}

\end{tikzpicture}

%% file: tex/graphs_les/les_square_rotation_2023-03-24-18-00-34_error.tex
\begin{tikzpicture}

\definecolor{darkgray176}{RGB}{176,176,176}
\definecolor{gainsboro229}{RGB}{229,229,229}

\definecolor{mediumpurple148103189}{RGB}{148,103,189}
\definecolor{sienna1408675}{RGB}{140,86,75}

\begin{axis}[
    width=\figurewidth,
    height=\figureheight,
    axis background/.style={fill=white},
    axis line style={white},
    tick align=outside,
    x grid style={white},
    xmajorgrids,
    xmajorticks=true,
    y grid style={white},
    ymajorgrids,
    ymajorticks=true,
    y grid style={white!69.0196078431373!black},
    ytick style={color=black},
    xmajorgrids,
    xminorgrids,
    ymajorgrids,
    ymajorticks=true,
    minor y tick num = 2,
    minor y grid style={dashed},
    yminorgrids,
    yticklabel style={
            /pgf/number format/fixed,
            /pgf/number format/precision=5,
        },
    scaled y ticks=false,
    ylabel near ticks, 
    ylabel shift={-1pt},
xmin=0.5, xmax=2.5,
xtick style={color=black},
xtick={1,2},
xticklabels={$I_1$, $I_2$},
ylabel={Error (m)},
ymin=-0.0410976448961767, ymax=1.63230454045981,
]
\path [draw=black, fill=mediumpurple148103189]
(axis cs:0.75,0.448823740072808)
--(axis cs:1.25,0.448823740072808)
--(axis cs:1.25,0.665208079675221)
--(axis cs:0.75,0.665208079675221)
--(axis cs:0.75,0.448823740072808)
--cycle;
\addplot [black]
table {%
1 0.448823740072808
1 0.233612611444085
};
\addplot [black]
table {%
1 0.665208079675221
1 0.881411228251091
};
\addplot [black]
table {%
0.875 0.233612611444085
1.125 0.233612611444085
};
\addplot [black]
table {%
0.875 0.881411228251091
1.125 0.881411228251091
};
\addplot [black, mark=x, mark size=1, mark options={solid,fill opacity=0}, only marks]
table {%
1 0.222342427165633
1 0.220719839803656
1 0.208932038495034
1 0.125703556929791
1 0.122255206772047
1 0.226023693351585
1 0.220694390054246
1 0.140422884637643
1 0.196010515060141
1 0.163257331530179
1 0.217091276016992
1 0.0948484210898756
1 0.185429662100509
1 0.195102141784376
1 0.121080514282678
1 0.206217522804958
1 0.20178138540258
1 0.160139290515319
1 0.217672885406934
1 0.209744147315324
1 0.183681255693847
1 0.114574308284578
1 0.143130413961345
1 0.185340880273648
1 0.159571985704405
1 0.0993647829980433
1 0.185978160057698
1 0.12136586556295
1 0.16552127039026
1 0.117882085589917
1 0.177126175469451
1 0.21667595206887
1 0.174648972426744
1 0.0612083023314513
1 0.19818849086285
1 0.169024682652615
1 0.131067537115393
1 0.18047201249983
1 0.216273943759138
1 0.113490575903283
1 0.0846366793446704
1 0.109558751061379
1 0.215833424183438
1 0.165561480424437
1 0.148630456777298
1 0.134691733381422
1 0.0883276946703652
1 0.203924972762909
1 0.0869405907059118
1 0.144658311723218
1 0.148140135550487
1 0.17197251496279
1 0.229405313113529
1 0.158448432573737
1 0.169309303730789
1 0.182850339268982
1 0.22408860480401
1 0.184426707030719
1 0.191906642526011
1 0.22709736510843
1 0.221810714287148
1 0.125871275767551
1 0.220469513707479
1 0.227260434060803
1 0.219062843004664
1 0.05689182224387
1 0.232264154521306
1 0.19155803855836
1 0.143193804481327
1 0.0912731764287947
1 0.225251501357047
1 0.0349660908018225
1 0.200944458977464
1 0.2087336388909
1 0.185261607956812
1 0.171765753068031
1 0.178556464463733
1 0.170948830251646
1 0.150281825915879
1 0.213977315484486
1 0.229265687001224
1 0.140687110510354
1 0.183163582851441
1 0.0921637470743149
1 0.210197610773593
1 0.0513099701671652
1 0.209636014626916
1 0.139543379779101
1 0.081947438515203
1 0.103121701639938
1 0.209835274788918
1 0.136443052955979
1 0.081083243055599
1 0.213913401127718
1 0.196193172357712
1 0.186493896920053
1 0.223366000871242
1 0.154316638361357
1 0.208755208286359
1 0.914763627730824
1 0.997556103475147
1 0.994278284005909
1 0.973836178989793
1 0.968511106827222
1 1.07357154805351
1 1.21763155039061
1 1.19904905359161
1 0.968560523699362
1 1.02829577633709
1 1.06370201081728
1 1.13551289034112
1 1.17249924353282
1 1.35367627178529
1 1.26776962674909
1 1.55624080476181
1 1.40876628436438
1 1.40060876179486
1 1.29746028491616
1 1.43387164986918
1 1.21600169827303
1 1.29145193787057
1 1.13960059868888
1 1.02831499292719
1 1.03978272951192
1 1.10863899679456
1 0.99428096272029
1 1.05617317648378
1 1.13750600646121
1 1.03318456447678
1 0.990686390634615
1 1.10340506276895
1 1.12161832569855
1 0.996789494202881
1 0.956444459791082
1 0.981806625193798
1 0.893551719653777
1 1.06280420064294
1 1.00721126930262
1 0.936753240475779
1 1.17420955276486
1 0.945249451010075
1 1.11274341090714
1 1.03590448018695
1 0.981305990953464
1 1.08911340680932
1 1.1103141729469
1 0.929496807674306
1 0.960951657343946
1 0.934130936956217
1 0.987825213641551
1 0.893067370584667
1 0.896694918124321
1 1.01800898866395
1 1.02171849680909
1 0.968811647107991
1 1.01204659396478
1 0.966398817539698
1 1.01619803076504
1 1.01262691374064
1 0.945812021445226
1 0.931689739526167
1 0.930146857465394
1 0.948894984813872
1 0.922296876471846
1 1.02377741596471
1 0.949362677557673
1 0.998758856308041
1 0.907321460480215
1 0.917576491547331
1 0.983267432507508
1 0.935154261501258
1 0.890178285793089
1 0.951636338660871
1 0.898981872434291
1 0.885691039494488
1 0.964630392288308
1 0.892579353095231
1 0.884331365224976
1 0.956177663750952
1 0.97359730612529
1 1.15535224609608
1 1.23202124636782
1 1.13688640318225
1 1.13405756435965
1 1.11380436649375
1 1.19284154587091
1 0.943922749349452
1 0.953448868039548
1 0.883541594241734
1 0.972717297869457
1 1.02813298157584
1 0.900885023430794
1 0.921096303730561
1 0.910000433953505
1 0.897989781508898
1 1.01971323831457
1 0.950880565634573
1 1.05996400002885
1 0.993587017369902
1 1.01399747642623
1 0.930503120510246
1 1.12883878284391
1 1.24711047193733
1 1.33659182536512
1 1.50214562854334
1 1.0579221713074
1 0.95073274736492
1 0.91595288865318
1 0.942537779059009
1 1.00413612993125
1 1.06650636097396
1 1.17634933897276
1 0.983976552418741
1 1.02947746128011
1 1.02799733735519
1 1.0256083615912
1 0.889583626816061
1 0.907387872510818
1 0.887007186214637
1 1.02531273210247
1 1.00821140693297
1 0.891984883871061
1 0.904417643535143
1 0.888341112587982
1 0.932578726905791
1 0.947002720438392
1 0.930113198716407
1 1.09703508935266
1 1.18746828948725
1 0.986038398326573
1 1.14315032058936
1 1.16670521037507
1 1.15420955344117
1 1.12381827355286
1 1.02053158314421
1 1.01654057962431
1 0.976025851638111
1 0.894966248734331
1 0.905961457730895
1 0.909945686645849
1 0.952626849278229
1 0.908884470496905
1 0.964707143910841
1 0.914185187832731
1 0.94169048262403
1 0.910375273010945
1 1.18995855125097
1 0.962669917762021
1 1.10922778808124
1 1.18838188138221
1 1.21395011487833
1 0.949827824526512
1 0.932470483213292
1 0.909808019819838
};
\path [draw=black, fill=sienna1408675]
(axis cs:1.75,0.460161037020711)
--(axis cs:2.25,0.460161037020711)
--(axis cs:2.25,0.534775308204217)
--(axis cs:1.75,0.534775308204217)
--(axis cs:1.75,0.460161037020711)
--cycle;
\addplot [black]
table {%
2 0.460161037020711
2 0.386741379773981
};
\addplot [black]
table {%
2 0.534775308204217
2 0.609376919086847
};
\addplot [black]
table {%
1.875 0.386741379773981
2.125 0.386741379773981
};
\addplot [black]
table {%
1.875 0.609376919086847
2.125 0.609376919086847
};
\addplot [black, mark=x, mark size=1, mark options={solid,fill opacity=0}, only marks]
table {%
2 0.373474096967651
2 0.342913671047293
2 0.32101200622047
2 0.365547283077041
2 0.273639895424979
2 0.310471643262735
2 0.378243531763594
2 0.277508412033362
2 0.277665670699836
2 0.33406590978442
2 0.236141738614025
2 0.342749858831535
2 0.303728260449323
2 0.332868654613946
2 0.374447158693666
2 0.303166475617111
2 0.371563133967654
2 0.322161540672124
2 0.298178988269779
2 0.367660118913159
2 0.379632902577365
2 0.194788074042184
2 0.31534306055804
2 0.269976349168093
2 0.35646639914555
2 0.240614182811863
2 0.353576829362084
2 0.383067368600439
2 0.301917362539742
2 0.367800896867541
2 0.282033254853643
2 0.327752915922177
2 0.265315152044
2 0.320797037954691
2 0.332790437074417
2 0.341106242559598
2 0.376857311935824
2 0.334949355714075
2 0.373264966102721
2 0.270306602000549
2 0.364175922901027
2 0.335986072988397
2 0.303368876918713
2 0.303504072277673
2 0.215357569391645
2 0.298610993340546
2 0.315162247680102
2 0.366529921755139
2 0.385024363862033
2 0.357969909557911
2 0.329665757126272
2 0.327695508299332
2 0.324257817365833
2 0.345628870565533
2 0.301779978801919
2 0.338532901220095
2 0.339668860664437
2 0.325563392597192
2 0.244921212797689
2 0.276944636610666
2 0.172779875367717
2 0.207752508395242
2 0.0375259448637919
2 0.218573044845354
2 0.165755616406428
2 0.146730754852563
2 0.238982297285957
2 0.356673025533753
2 0.327538317764228
2 0.212891863143675
2 0.368940238459882
2 0.310137452793426
2 0.293110415462031
2 0.3103967098392
2 0.248656630705901
2 0.204397404365294
2 0.197167542950503
2 0.190218133537297
2 0.187179019289375
2 0.337834098507498
2 0.306205715949193
2 0.261227288426056
2 0.259442655934013
2 0.304447943283237
2 0.343594802576963
2 0.318129973900043
2 0.365430721058174
2 0.382895308155013
2 0.339954213722292
2 0.235812036739709
2 0.324634420121181
2 0.379906785653936
2 0.364351208435654
2 0.326877894455926
2 0.280718823681351
2 0.230923001200001
2 0.360368452163061
2 0.303942655600043
2 0.28600293322811
2 0.330910664954172
2 0.147468545460593
2 0.283905329210302
2 0.379843692485186
2 0.217391455800576
2 0.316387019371668
2 0.350869256648627
2 0.382941016331867
2 0.262568856955136
2 0.337336281134114
2 0.384256250273084
2 0.341053408767638
2 0.341680829640168
2 0.30890568097571
2 0.282757905965481
2 0.313069207410423
2 0.311490824349428
2 0.292174134747779
2 0.320296947378392
2 0.37743236082606
2 0.369594870838425
2 0.303087346135435
2 0.283885383888978
2 0.344660070667207
2 0.26129200088556
2 0.306712784551562
2 0.288818221443829
2 0.383452708423231
2 0.293995191200649
2 0.27758654298614
2 0.343965706566586
2 0.277259706502775
2 0.324947944999547
2 0.366304674808037
2 0.308681408359049
2 0.320111359513699
2 0.375275424231087
2 0.229748011769717
2 0.384678506720139
2 0.217757764350059
2 0.334952554608588
2 0.361564077029542
2 0.341858376672504
2 0.218902616907274
2 0.356177019405913
2 0.234539711915464
2 0.282682818868265
2 0.351277131907559
2 0.267297415769313
2 0.311378298646618
2 0.365730454274018
2 0.355364156292467
2 0.357203684396601
2 0.337759140938844
2 0.3312575499692
2 0.237579040971849
2 0.341263483433391
2 0.319604648705228
2 0.362459516936052
2 0.324863671884011
2 0.338877679635722
2 0.320367196643287
2 0.308419323913195
2 0.372118816804311
2 0.374850299416684
2 0.360915492229062
2 0.312756619768965
2 0.366817706384654
2 0.38163685445914
2 0.321346442448626
2 0.380595030301189
2 0.324452906883802
2 0.308189933231097
2 0.358109622219283
2 0.383817827943681
2 0.382783477950815
2 0.352436631183018
2 0.326122356133069
2 0.347385686205779
2 0.696832475470882
2 0.6110628718456
2 0.616905487571368
2 0.615591959794846
2 0.610998804253781
2 0.616384268448027
2 0.627745178566761
2 0.691665741823098
2 0.639684655688975
2 0.674123083361869
2 0.643220433409824
2 0.620496340613967
2 0.683896158090573
2 0.621925041147277
2 0.640024173633661
2 0.637439749209606
2 0.682405208046105
2 0.621694525701791
2 0.681504511961851
2 0.675947436206797
2 0.720459225741739
2 0.781766162581059
2 0.837101640004342
2 0.642495493427944
2 0.693665381685767
2 0.678336628593462
2 0.669761465978642
2 0.688789451512054
2 0.728039207474694
2 0.718033320343686
2 0.837925386017489
2 0.885204590211796
2 0.99587606499319
2 0.614758975560139
2 0.620894334952259
2 0.751819283540553
2 0.62504329060611
2 0.623951156025514
2 0.623282147361536
2 0.614544366823856
2 0.651901193254897
2 0.610569772585673
2 0.638553238902141
2 0.738825589947447
2 0.739441707490979
2 0.703962267363309
2 0.705182302287161
2 0.692394206618158
2 0.684329483785874
2 0.669759083784165
2 0.667207515017762
2 0.677534827527393
2 0.666638542085018
2 0.65788268783638
2 0.609672563764922
2 0.616788413158151
2 0.694524839891875
2 0.623898588040066
2 0.667910288983087
2 0.649129311449204
2 0.63278096629704
2 0.630409751846763
2 0.637673349256252
2 0.618214689180276
2 0.617600824184293
2 0.677010358599026
2 0.637952523721647
2 0.701889280982261
2 0.647320954423536
2 0.663123492772898
2 0.682275859474118
2 0.652992292608037
2 0.661804260655198
2 0.622218420716544
2 0.639055804680539
2 0.705995216867834
2 0.622938982729648
2 0.719897819309655
2 0.617580082702834
2 0.633653504143665
2 0.621100628777307
2 0.612058534846676
2 0.648040338948608
2 0.632373719513698
2 0.667691779494577
2 0.616096562337908
2 0.742054542066198
2 0.808427589217625
2 0.788379495169097
2 0.701414093227184
2 0.649092464086154
2 0.647820856507425
2 0.634377724123774
2 0.614937373729049
2 0.685255422942494
2 0.656009462242623
2 0.69385519974899
2 0.632252252247253
2 0.624977010853895
2 0.631976529446656
2 0.615504517502051
2 0.64460567777582
2 0.629206355139462
2 0.617972104917277
2 0.622883972683434
};
\addplot [thick, black]
table {%
0.75 0.540603255928024
1.25 0.540603255928024
};
\addplot [thick, black]
table {%
1.75 0.497169087535125
2.25 0.497169087535125
};
\end{axis}

\end{tikzpicture}

%% file: tex/graphs_ml_gd/army_2023-03-24-20-09-29_v2_plot_v2.tex
\pgfplotsset{compat=1.10}

\pgfplotsset{%
my legend/.style={legend image code/.code={%
\node[##1,anchor=west] at (0cm,0cm){\pgfuseplotmark{x}};
\node[##1] at (0.25cm,0cm){\pgfuseplotmark{*}};
\node[##1,anchor=east] at (0.6cm,0cm){\pgfuseplotmark{diamond*}};
}},%
}

\begin{tikzpicture}

\definecolor{crimson2143940}{RGB}{214,39,40}
\definecolor{darkgray176}{RGB}{176,176,176}
\definecolor{darkorange25512714}{RGB}{255,127,14}
\definecolor{forestgreen4416044}{RGB}{44,160,44}
\definecolor{gray127}{RGB}{127,127,127}
\definecolor{lightgray204}{RGB}{204,204,204}
\definecolor{mediumpurple148103189}{RGB}{148,103,189}
\definecolor{orchid227119194}{RGB}{227,119,194}
\definecolor{sienna1408675}{RGB}{140,86,75}
\definecolor{steelblue31119180}{RGB}{31,119,180}

\begin{axis}[
    height=\figureheight,
    width=\figurewidth,
    axis background/.style={fill=white},
    axis line style={white},
    legend cell align={left},
    legend style={
      fill opacity=.6,
      draw opacity=1,
      text opacity=1,
      at={(0.9,0.5)},
      anchor=south west,
      draw=white,
      legend columns=1,
      font=\tiny
    },
    tick align=outside,
    x grid style={white!69.0196078431373!black},
    xlabel={\(\displaystyle x\) (m)},
    minor tick num = 1,
    minor grid style={dashed},
    xmajorgrids,
    xmajorgrids,
    y grid style={white!69.0196078431373!black},
    ylabel={\(\displaystyle y\) (m)}, 
    xminorgrids,
    xminorgrids=true,
    ymajorgrids,
    ymajorticks=true,
    yminorgrids,
    yminorgrids=true,
xmin=0.689324010758933, xmax=6.26358440466297,
ymin=0.723722330902559, ymax=6.04831931327841,
]
\addlegendentry{Mocap}
\addlegendimage{mark=square*,only marks,black}
\addlegendentry{ML}
\addlegendimage{mark=*,only marks,black}
\addlegendentry{GD}
\addlegendimage{mark=triangle*,only marks,black}

\addplot [draw=none, draw=steelblue31119180, fill=steelblue31119180, mark=square*,only marks]
table{%
x  y
1.006726779986401 1.028987096523752
};
\addplot [draw=none, draw=steelblue31119180, fill=steelblue31119180, mark=*,only marks]
table{%
x  y
0.9666099304141059 0.965749466465098
};
\addplot [draw=none, draw=steelblue31119180, fill=steelblue31119180, mark=triangle*,only marks]
table{%
x  y
0.9426994832091167 1.1338659028821783
};

\draw (axis cs:0.8,1.4) node[
  scale=0.6,
  anchor=base west,
  text=steelblue31119180,
  rotate=0.0
]{UWB 1};

\addplot [draw=darkorange25512714, draw=none, fill=darkorange25512714, mark=square*]
table{%
x  y
3.4933053712455595 1.0078895291503596
};
\addplot [draw=darkorange25512714, draw=none, fill=darkorange25512714, mark=*]
table{%
x  y
3.5181074154899252 0.9718596662354463
};
\addplot [draw=darkorange25512714, draw=none, fill=darkorange25512714, mark=triangle*]
table{%
x  y
3.7903154852865426 1.098547280389448
};

\draw (axis cs:3.35,1.2) node[
  scale=0.6,
  anchor=base west,
  text=darkorange25512714,
  rotate=0.0
]{UWB 2};

\addplot [draw=forestgreen4416044, draw=none, fill=forestgreen4416044, mark=square*]
table{%
x  y
5.678465379987444 1.038360721724374
};
\addplot [draw=forestgreen4416044, draw=none, fill=forestgreen4416044, mark=*]
table{%
x  y
5.00149215121268 1.61244204860714
};
\addplot [draw=forestgreen4416044, draw=none, fill=forestgreen4416044, mark=triangle*]
table{%
x  y
6.010208932212784 1.45153496358941
};

\draw (axis cs:5.1,1.6) node[
  scale=0.6,
  anchor=base west,
  text=forestgreen4416044,
  rotate=0.0
]{UWB 3};

\addplot [draw=crimson2143940, draw=none, fill=crimson2143940, mark=square*]
table{%
x  y
5.663383345701257 5.484995671680996
};
\addplot [draw=crimson2143940, draw=none, fill=crimson2143940, mark=*]
table{%
x  y
4.081423653530185 4.208762987074968
};
\addplot [draw=crimson2143940, draw=none, fill=crimson2143940, mark=triangle*]
table{%
x  y
5.5656612104629914 5.806292177715872
};

\draw (axis cs:4.2,5.5) node[
  scale=0.6,
  anchor=base west,
  text=crimson2143940,
  rotate=0.0
]{UWB 4};

\addplot [draw=mediumpurple148103189, draw=none, fill=mediumpurple148103189, mark=square*]
table{%
x  y
3.4287153171033276 5.481343613838663
};
\addplot [draw=mediumpurple148103189, draw=none, fill=mediumpurple148103189, mark=*]
table{%
x  y
3.085603138563679 4.410742624175031
};
\addplot [draw=mediumpurple148103189, draw=none, fill=mediumpurple148103189, mark=triangle*]
table{%
x  y
3.58665324460176 5.700759070941466
};

\draw (axis cs:3.2,5.1) node[
  scale=0.6,
  anchor=base west,
  text=mediumpurple148103189,
  rotate=0.0
]{UWB 5};

\addplot [draw=none, draw=sienna1408675, fill=sienna1408675, mark=square*]
table{%
x  y
0.9762343377483135 5.382801419861463
};
\addplot [draw=none, draw=sienna1408675, fill=sienna1408675, mark=*]
table{%
x  y
1.6327173710138427 4.621767537074652
};
\addplot [draw=none, draw=sienna1408675, fill=sienna1408675, mark=triangle*]
table{%
x  y
1.4796356437561833 5.74736487957866
};

\draw (axis cs:1.1,5.15) node[
  scale=0.6,
  anchor=base west,
  text=sienna1408675,
  rotate=0.0
]{UWB 6};

\addplot [draw=none, draw=orchid227119194, fill=orchid227119194, mark=square*]
table{%
x  y
2.3903435658435432 3.1987021893871073
};
\addplot [draw=none, draw=orchid227119194, fill=orchid227119194, mark=*]
table{%
x  y
2.458509838009111 3.1391055813018145
};
\addplot [draw=none, draw=orchid227119194, fill=orchid227119194, mark=triangle*]
table{%
x  y
2.608749485922507 3.424269520075364
};

\draw (axis cs:2.3,2.8) node[
  scale=0.6,
  anchor=base west,
  text=orchid227119194,
  rotate=0.0
]{UWB 7};

\addplot [draw=gray127, draw=none, fill=gray127, mark=square*]
table{%
x  y
4.603920891820168 3.1932066143775475
};
\addplot [draw=gray127, draw=none, fill=gray127, mark=*]
table{%
x  y
3.9819528937691246 2.96423203900496
};
\addplot [draw=gray127, draw=none, fill=gray127, mark=triangle*]
table{%
x  y
4.703695626744139 3.4048585085735508
};

\draw (axis cs:4.3,2.8) node[
  scale=0.6,
  anchor=base west,
  text=gray127,
  rotate=0.0
]{UWB 8};

\end{axis}

\end{tikzpicture}

%% file: tex/graphs_ml_gd/army_2023-03-24-20-09-29_v2_boxplot.tex
\begin{tikzpicture}

\definecolor{crimson2143940}{RGB}{214,39,40}
\definecolor{darkgray176}{RGB}{176,176,176}
\definecolor{darkorange25512714}{RGB}{255,127,14}
\definecolor{forestgreen4416044}{RGB}{44,160,44}
\definecolor{gainsboro229}{RGB}{229,229,229}
\definecolor{gray127}{RGB}{127,127,127}
\definecolor{mediumpurple148103189}{RGB}{148,103,189}
\definecolor{orchid227119194}{RGB}{227,119,194}
\definecolor{sienna1408675}{RGB}{140,86,75}
\definecolor{steelblue31119180}{RGB}{31,119,180}

\begin{axis}[
    width=\figurewidth,
    height=\figureheight,
    axis background/.style={fill=white},
    axis line style={white},
    tick align=outside,
    x grid style={white},
    xmajorgrids,
    xmajorticks=true,
    y grid style={white},
    ymajorgrids,
    ymajorticks=true,
    y grid style={white!69.0196078431373!black},
    ytick style={color=black},
    xmajorgrids,
    xminorgrids,
    ymajorgrids,
    ymajorticks=true,
    yticklabel style={
            /pgf/number format/fixed,
            /pgf/number format/precision=5
        },
    scaled y ticks=false,
    ylabel near ticks, 
    ylabel shift={-1pt},
    xmin=0.5, xmax=12.5,
    xtick style={color=black},
    xtick={1,1.5,2.5,3,4,4.5,5.5,6,7,7.5,8.5,9,10,10.5,11.5,12},
    xticklabel style={rotate=90.0,font=\tiny},
    xticklabels={$\text{MLAT}_1$,,$\text{MLAT}_2$,,$\text{MLAT}_3$,,$\text{MLAT}_4$,,$\text{MLAT}_5$,,$\text{MLAT}_6$,,$\text{MLAT}_7$,,$\text{MLAT}_8$},
    extra x ticks={1,1.5,2.5,3,4,4.5,5.5,6,7,7.5,8.5,9,10,10.5,11.5,12},
    ,extra x tick style={%
        ,grid=major
        ,ticklabel pos=top
        },
    extra x tick labels={,$\text{GD}_1$,,$\text{GD}_2$,,$\text{GD}_3$,, $\text{GD}_4$,,$\text{GD}_5$,,$\text{GD}_6$,,$\text{GD}_7$,,$\text{GD}_8$},
ylabel={Error (m)},
ymin=0, ymax=2.5,
]
\path [draw=black, fill=steelblue31119180]
(axis cs:0.75,0.0644954507174942)
--(axis cs:1.25,0.0644954507174942)
--(axis cs:1.25,0.091874380431448)
--(axis cs:0.75,0.091874380431448)
--(axis cs:0.75,0.0644954507174942)
--cycle;
\addplot [black]
table {%
1 0.0644954507174942
1 0.0375771791159941
};
\addplot [black]
table {%
1 0.091874380431448
1 0.119253082894512
};
\addplot [black]
table {%
0.875 0.0375771791159941
1.125 0.0375771791159941
};
\addplot [black]
table {%
0.875 0.119253082894512
1.125 0.119253082894512
};
\path [draw=black, fill=steelblue31119180]
(axis cs:1.25,0.107092858838827)
--(axis cs:1.75,0.107092858838827)
--(axis cs:1.75,0.134512631772015)
--(axis cs:1.25,0.134512631772015)
--(axis cs:1.25,0.107092858838827)
--cycle;
\addplot [black]
table {%
1.5 0.107092858838827
1.5 0.0814067309537149
};
\addplot [black]
table {%
1.5 0.134512631772015
1.5 0.161590150847666
};
\addplot [black]
table {%
1.375 0.0814067309537149
1.625 0.0814067309537149
};
\addplot [black]
table {%
1.375 0.161590150847666
1.625 0.161590150847666
};
\path [draw=black, fill=darkorange25512714]
(axis cs:2.25,0.0397559135139601)
--(axis cs:2.75,0.0397559135139601)
--(axis cs:2.75,0.0637895181248332)
--(axis cs:2.25,0.0637895181248332)
--(axis cs:2.25,0.0397559135139601)
--cycle;
\addplot [black]
table {%
2.5 0.0397559135139601
2.5 0.0162449686396521
};
\addplot [black]
table {%
2.5 0.0637895181248332
2.5 0.0877202466304909
};
\addplot [black]
table {%
2.375 0.0162449686396521
2.625 0.0162449686396521
};
\addplot [black]
table {%
2.375 0.0877202466304909
2.625 0.0877202466304909
};
\path [draw=black, fill=darkorange25512714]
(axis cs:2.75,0.283451521175135)
--(axis cs:3.25,0.283451521175135)
--(axis cs:3.25,0.330845491152812)
--(axis cs:2.75,0.330845491152812)
--(axis cs:2.75,0.283451521175135)
--cycle;
\addplot [black]
table {%
3 0.283451521175135
3 0.236870969998458
};
\addplot [black]
table {%
3 0.330845491152812
3 0.370384546759887
};
\addplot [black]
table {%
2.875 0.236870969998458
3.125 0.236870969998458
};
\addplot [black]
table {%
2.875 0.370384546759887
3.125 0.370384546759887
};
\path [draw=black, fill=forestgreen4416044]
(axis cs:3.75,0.874939910827711)
--(axis cs:4.25,0.874939910827711)
--(axis cs:4.25,0.911318927880406)
--(axis cs:3.75,0.911318927880406)
--(axis cs:3.75,0.874939910827711)
--cycle;
\addplot [black]
table {%
4 0.874939910827711
4 0.839013224419757
};
\addplot [black]
table {%
4 0.911318927880406
4 0.94696924824766
};
\addplot [black]
table {%
3.875 0.839013224419757
4.125 0.839013224419757
};
\addplot [black]
table {%
3.875 0.94696924824766
4.125 0.94696924824766
};
\path [draw=black, fill=forestgreen4416044]
(axis cs:4.25,0.485128030121589)
--(axis cs:4.75,0.485128030121589)
--(axis cs:4.75,0.573365082446017)
--(axis cs:4.25,0.573365082446017)
--(axis cs:4.25,0.485128030121589)
--cycle;
\addplot [black]
table {%
4.5 0.485128030121589
4.5 0.405807944365514
};
\addplot [black]
table {%
4.5 0.573365082446017
4.5 0.655408193174194
};
\addplot [black]
table {%
4.375 0.405807944365514
4.625 0.405807944365514
};
\addplot [black]
table {%
4.375 0.655408193174194
4.625 0.655408193174194
};
\path [draw=black, fill=crimson2143940]
(axis cs:5.25,2.03075353884434)
--(axis cs:5.75,2.03075353884434)
--(axis cs:5.75,2.07008043498873)
--(axis cs:5.25,2.07008043498873)
--(axis cs:5.25,2.03075353884434)
--cycle;
\addplot [black]
table {%
5.5 2.03075353884434
5.5 1.99339952418472
};
\addplot [black]
table {%
5.5 2.07008043498873
5.5 2.10931568496486
};
\addplot [black]
table {%
5.375 1.99339952418472
5.625 1.99339952418472
};
\addplot [black]
table {%
5.375 2.10931568496486
5.625 2.10931568496486
};
\path [draw=black, fill=crimson2143940]
(axis cs:5.75,0.308287330277542)
--(axis cs:6.25,0.308287330277542)
--(axis cs:6.25,0.369687709777019)
--(axis cs:5.75,0.369687709777019)
--(axis cs:5.75,0.308287330277542)
--cycle;
\addplot [black]
table {%
6 0.308287330277542
6 0.247013649903455
};
\addplot [black]
table {%
6 0.369687709777019
6 0.430846853883764
};
\addplot [black]
table {%
5.875 0.247013649903455
6.125 0.247013649903455
};
\addplot [black]
table {%
5.875 0.430846853883764
6.125 0.430846853883764
};
\path [draw=black, fill=mediumpurple148103189]
(axis cs:6.75,1.11823653830405)
--(axis cs:7.25,1.11823653830405)
--(axis cs:7.25,1.15334998433248)
--(axis cs:6.75,1.15334998433248)
--(axis cs:6.75,1.11823653830405)
--cycle;
\addplot [black]
table {%
7 1.11823653830405
7 1.08334035093605
};
\addplot [black]
table {%
7 1.15334998433248
7 1.1871484942036
};
\addplot [black]
table {%
6.875 1.08334035093605
7.125 1.08334035093605
};
\addplot [black]
table {%
6.875 1.1871484942036
7.125 1.1871484942036
};
\path [draw=black, fill=mediumpurple148103189]
(axis cs:7.25,0.244702543060392)
--(axis cs:7.75,0.244702543060392)
--(axis cs:7.75,0.288539293056979)
--(axis cs:7.25,0.288539293056979)
--(axis cs:7.25,0.244702543060392)
--cycle;
\addplot [black]
table {%
7.5 0.244702543060392
7.5 0.201445029334629
};
\addplot [black]
table {%
7.5 0.288539293056979
7.5 0.329072486075885
};
\addplot [black]
table {%
7.375 0.201445029334629
7.625 0.201445029334629
};
\addplot [black]
table {%
7.375 0.329072486075885
7.625 0.329072486075885
};
\path [draw=black, fill=sienna1408675]
(axis cs:8.25,0.979492803416377)
--(axis cs:8.75,0.979492803416377)
--(axis cs:8.75,1.05475063443308)
--(axis cs:8.25,1.05475063443308)
--(axis cs:8.25,0.979492803416377)
--cycle;
\addplot [black]
table {%
8.5 0.979492803416377
8.5 0.90971330963587
};
\addplot [black]
table {%
8.5 1.05475063443308
8.5 1.1242462484168
};
\addplot [black]
table {%
8.375 0.90971330963587
8.625 0.90971330963587
};
\addplot [black]
table {%
8.375 1.1242462484168
8.625 1.1242462484168
};
\path [draw=black, fill=sienna1408675]
(axis cs:8.75,0.578746748144843)
--(axis cs:9.25,0.578746748144843)
--(axis cs:9.25,0.64480615326985)
--(axis cs:8.75,0.64480615326985)
--(axis cs:8.75,0.578746748144843)
--cycle;
\addplot [black]
table {%
9 0.578746748144843
9 0.513722498540532
};
\addplot [black]
table {%
9 0.64480615326985
9 0.710308577445823
};
\addplot [black]
table {%
8.875 0.513722498540532
9.125 0.513722498540532
};
\addplot [black]
table {%
8.875 0.710308577445823
9.125 0.710308577445823
};
\path [draw=black, fill=orchid227119194]
(axis cs:9.75,0.0702131679590589)
--(axis cs:10.25,0.0702131679590589)
--(axis cs:10.25,0.131941875301724)
--(axis cs:9.75,0.131941875301724)
--(axis cs:9.75,0.0702131679590589)
--cycle;
\addplot [black]
table {%
10 0.0702131679590589
10 0.0168473207144939
};
\addplot [black]
table {%
10 0.131941875301724
10 0.173024683107714
};
\addplot [black]
table {%
9.875 0.0168473207144939
10.125 0.0168473207144939
};
\addplot [black]
table {%
9.875 0.173024683107714
10.125 0.173024683107714
};
\path [draw=black, fill=orchid227119194]
(axis cs:10.25,0.297833205692534)
--(axis cs:10.75,0.297833205692534)
--(axis cs:10.75,0.324962977531293)
--(axis cs:10.25,0.324962977531293)
--(axis cs:10.25,0.297833205692534)
--cycle;
\addplot [black]
table {%
10.5 0.297833205692534
10.5 0.271334890835057
};
\addplot [black]
table {%
10.5 0.324962977531293
10.5 0.351264714864481
};
\addplot [black]
table {%
10.375 0.271334890835057
10.625 0.271334890835057
};
\addplot [black]
table {%
10.375 0.351264714864481
10.625 0.351264714864481
};
\path [draw=black, fill=gray127]
(axis cs:11.25,0.650419288147636)
--(axis cs:11.75,0.650419288147636)
--(axis cs:11.75,0.695057426028107)
--(axis cs:11.25,0.695057426028107)
--(axis cs:11.25,0.650419288147636)
--cycle;
\addplot [black]
table {%
11.5 0.650419288147636
11.5 0.626696530489513
};
\addplot [black]
table {%
11.5 0.695057426028107
11.5 0.738837001446172
};
\addplot [black]
table {%
11.375 0.626696530489513
11.625 0.626696530489513
};
\addplot [black]
table {%
11.375 0.738837001446172
11.625 0.738837001446172
};
\path [draw=black, fill=gray127]
(axis cs:11.75,0.207012838829719)
--(axis cs:12.25,0.207012838829719)
--(axis cs:12.25,0.251316198615922)
--(axis cs:11.75,0.251316198615922)
--(axis cs:11.75,0.207012838829719)
--cycle;
\addplot [black]
table {%
12 0.207012838829719
12 0.162896024153166
};
\addplot [black]
table {%
12 0.251316198615922
12 0.293197263103285
};
\addplot [black]
table {%
11.875 0.162896024153166
12.125 0.162896024153166
};
\addplot [black]
table {%
11.875 0.293197263103285
12.125 0.293197263103285
};
\addplot [thick, black]
table {%
0.75 0.0787304698933429
1.25 0.0787304698933429
};
\addplot [thick, black]
table {%
1.25 0.119494911032907
1.75 0.119494911032907
};
\addplot [thick, black]
table {%
2.25 0.0483951968562996
2.75 0.0483951968562996
};
\addplot [thick, black]
table {%
2.75 0.312039421955987
3.25 0.312039421955987
};
\addplot [thick, black]
table {%
3.75 0.900021815101389
4.25 0.900021815101389
};
\addplot [thick, black]
table {%
4.25 0.528911796609153
4.75 0.528911796609153
};
\addplot [thick, black]
table {%
5.25 2.04840540652735
5.75 2.04840540652735
};
\addplot [thick, black]
table {%
5.75 0.330368729842627
6.25 0.330368729842627
};
\addplot [thick, black]
table {%
6.75 1.14203328200416
7.25 1.14203328200416
};
\addplot [thick, black]
table {%
7.25 0.268733297629301
7.75 0.268733297629301
};
\addplot [thick, black]
table {%
8.25 1.01339382972783
8.75 1.01339382972783
};
\addplot [thick, black]
table {%
8.75 0.627180652091524
9.25 0.627180652091524
};
\addplot [thick, black]
table {%
9.75 0.0952963846084312
10.25 0.0952963846084312
};
\addplot [thick, black]
table {%
10.25 0.305003588484243
10.75 0.305003588484243
};
\addplot [thick, black]
table {%
11.25 0.67552077465778
11.75 0.67552077465778
};
\addplot [thick, black]
table {%
11.75 0.233244344464746
12.25 0.233244344464746
};
\end{axis}

\end{tikzpicture}

%% file: tex/graphs_ml_gd/corner_2023-03-24-20-18-25_v2_plot_v2.tex
\pgfplotsset{compat=1.10}

\pgfplotsset{%
my legend/.style={legend image code/.code={%
\node[##1,anchor=west] at (0cm,0cm){\pgfuseplotmark{x}};
\node[##1] at (0.25cm,0cm){\pgfuseplotmark{*}};
\node[##1,anchor=east] at (0.6cm,0cm){\pgfuseplotmark{diamond*}};
}},%
}

\begin{tikzpicture}

\definecolor{crimson2143940}{RGB}{214,39,40}
\definecolor{darkgray176}{RGB}{176,176,176}
\definecolor{darkorange25512714}{RGB}{255,127,14}
\definecolor{forestgreen4416044}{RGB}{44,160,44}
\definecolor{gray127}{RGB}{127,127,127}
\definecolor{lightgray204}{RGB}{204,204,204}
\definecolor{mediumpurple148103189}{RGB}{148,103,189}
\definecolor{orchid227119194}{RGB}{227,119,194}
\definecolor{sienna1408675}{RGB}{140,86,75}
\definecolor{steelblue31119180}{RGB}{31,119,180}

\begin{axis}[
    height=\figureheight,
    width=\figurewidth,
    axis background/.style={fill=white},
    axis line style={white},
    legend cell align={left},
    legend style={
      fill opacity=.6,
      draw opacity=1,
      text opacity=1,
      at={(0.03,0.97)},
      anchor=north west,
      draw=white,
      legend columns=1,
      font=\tiny
    },
    tick align=outside,
    x grid style={white!69.0196078431373!black},
    xlabel={\(\displaystyle x\) (m)},
    minor tick num = 1,
    minor grid style={dashed},
    xmajorgrids,
    xmajorgrids,
    y grid style={white!69.0196078431373!black},
    ylabel={\(\displaystyle y\) (m)}, 
    xminorgrids,
    xminorgrids=true,
    ymajorgrids,
    ymajorticks=true,
    yminorgrids,
    yminorgrids=true,
xmin=0.576378307459278, xmax=7.01411684227907,
ymin=0.659088658318524, ymax=8.00519187447097,
]

\addlegendentry{Mocap}
\addlegendimage{mark=square*,only marks,black}
\addlegendentry{ML}
\addlegendimage{mark=*,only marks,black}
\addlegendentry{GD}
\addlegendimage{mark=diamond*,only marks,black}

\addplot [draw=none, draw=steelblue31119180, fill=steelblue31119180, mark=square*]
table{%
x  y
1.0064070964968481 1.028813721967298

};
\addplot [draw=none, draw=steelblue31119180, fill=steelblue31119180, mark=*]
table{%
x  y
0.9404285830659838 1.0269992818932525
};
\addplot [draw=none, draw=steelblue31119180, fill=steelblue31119180, mark=diamond*]
table{%
x  y
0.8690027863147235 1.2014668189798927
};

\draw (axis cs:0.8,1.4) node[
  scale=0.6,
  anchor=base west,
  text=steelblue31119180,
  rotate=0.0
]{UWB 1};

\addplot [draw=darkorange25512714, draw=none, fill=darkorange25512714, mark=square*]
table{%
x  y
2.622994330871937 0.9972185934698858
};
\addplot [draw=darkorange25512714, draw=none, fill=darkorange25512714, mark=*]
table{%
x  y
2.7821968763277183 0.9930024408709078
};
\addplot [draw=darkorange25512714, draw=none, fill=darkorange25512714, mark=diamond*]
table{%
x  y
2.5919271072132832 1.1281617659523344
};

\draw (axis cs:2.3,1.4) node[
  scale=0.6,
  anchor=base west,
  text=darkorange25512714,
  rotate=0.0
]{UWB 2};

\addplot [draw=forestgreen4416044, draw=none, fill=forestgreen4416044, mark=square*]
table{%
x  y
4.314603802215221 1.0415742449982222
};
\addplot [draw=forestgreen4416044, draw=none, fill=forestgreen4416044, mark=*]
table{%
x  y
4.041680418634742 1.6461462850632058
};
\addplot [draw=forestgreen4416044, draw=none, fill=forestgreen4416044, mark=diamond*]
table{%
x  y
4.291224426222827 1.5974860076134885
};

\draw (axis cs:4.3,1.1) node[
  scale=0.6,
  anchor=base west,
  text=forestgreen4416044,
  rotate=0.0
]{UWB 3};

\addplot [draw=crimson2143940, draw=none, fill=crimson2143940, mark=square*]
table{%
x  y
5.927960721836533 6.051487202977025
};
\addplot [draw=crimson2143940, draw=none, fill=crimson2143940, mark=*]
table{%
x  y
5.594251154602289 6.047855271292097
};
\addplot [draw=crimson2143940, draw=none, fill=crimson2143940, mark=diamond*]
table{%
x  y
6.560588740736105 5.86939134271153
};

\draw (axis cs:5.5,6.3) node[
  scale=0.6,
  anchor=base west,
  text=crimson2143940,
  rotate=0.0
]{UWB 4};

\addplot [draw=mediumpurple148103189, draw=none, fill=mediumpurple148103189, mark=square*]
table{%
x  y
5.9121682865675105 4.369140593950139
};
\addplot [draw=mediumpurple148103189, draw=none, fill=mediumpurple148103189, mark=*]
table{%
x  y
5.3538697241603055 4.777852979810859
};
\addplot [draw=mediumpurple148103189, draw=none, fill=mediumpurple148103189, mark=diamond*]
table{%
x  y
6.164163429774737 4.456744366201844
};

\draw (axis cs:5.5,4.7) node[
  scale=0.6,
  anchor=base west,
  text=mediumpurple148103189,
  rotate=0.0
]{UWB 5};

\addplot [draw=none, draw=sienna1408675, fill=sienna1408675, mark=square*]
table{%
x  y
5.962492268584495 7.67127809191859
};
\addplot [draw=none, draw=sienna1408675, fill=sienna1408675, mark=*]
table{%
x  y
5.523473376188326 7.590776052197819
};
\addplot [draw=none, draw=sienna1408675, fill=sienna1408675, mark=diamond*]
table{%
x  y
6.721492363423627 7.530990510350234
};

\draw (axis cs:5.5,6.8) node[
  scale=0.6,
  anchor=base west,
  text=sienna1408675,
  rotate=0.0
]{UWB 6};

\addplot [draw=none, draw=orchid227119194, fill=orchid227119194, mark=square*]
table{%
x  y
6.00009372400683 1.0187513770059098
};
\addplot [draw=none, draw=orchid227119194, fill=orchid227119194, mark=*]
table{%
x  y
4.802375846989788 2.234811897844828
};
\addplot [draw=none, draw=orchid227119194, fill=orchid227119194, mark=diamond*]
table{%
x  y
5.835518578307431 2.0159029016655095
};

\draw (axis cs:5.5,1.4) node[
  scale=0.6,
  anchor=base west,
  text=orchid227119194,
  rotate=0.0
]{UWB 7};

\addplot [draw=gray127, draw=none, fill=gray127, mark=square*]
table{%
x  y
6.0103557631026865 2.6562647281691087
};
\addplot [draw=gray127, draw=none, fill=gray127, mark=*]
table{%
x  y
5.190641432615706 3.3307772533942717
};
\addplot [draw=gray127, draw=none, fill=gray127, mark=diamond*]
table{%
x  y
6.060849176045367 3.0719001596275985
};

\draw (axis cs:5.5,3.3) node[
  scale=0.6,
  anchor=base west,
  text=gray127,
  rotate=0.0
]{UWB 8};

\end{axis}

\end{tikzpicture}

%% file: tex/graphs_ml_gd/corner_2023-03-24-20-18-25_v2_boxplot.tex
\begin{tikzpicture}

\definecolor{crimson2143940}{RGB}{214,39,40}
\definecolor{darkgray176}{RGB}{176,176,176}
\definecolor{darkorange25512714}{RGB}{255,127,14}
\definecolor{forestgreen4416044}{RGB}{44,160,44}
\definecolor{gainsboro229}{RGB}{229,229,229}
\definecolor{gray127}{RGB}{127,127,127}
\definecolor{mediumpurple148103189}{RGB}{148,103,189}
\definecolor{orchid227119194}{RGB}{227,119,194}
\definecolor{sienna1408675}{RGB}{140,86,75}
\definecolor{steelblue31119180}{RGB}{31,119,180}

\begin{axis}[
    width=\figurewidth,
    height=\figureheight,
    axis background/.style={fill=white},
    axis line style={white},
    tick align=outside,
    x grid style={white},
    xmajorgrids,
    xmajorticks=true,
    y grid style={white},
    ymajorgrids,
    ymajorticks=true,
    y grid style={white!69.0196078431373!black},
    ytick style={color=black},
    xmajorgrids,
    xminorgrids,
    ymajorgrids,
    ymajorticks=true,
    yticklabel style={
            /pgf/number format/fixed,
            /pgf/number format/precision=5
        },
    scaled y ticks=false,
    ylabel near ticks, 
    ylabel shift={-1pt},
    xmin=0.5, xmax=12.5,
    xtick style={color=black},
    xtick={1,1.5,2.5,3,4,4.5,5.5,6,7,7.5,8.5,9,10,10.5,11.5,12},
    xticklabel style={rotate=90.0,font=\tiny},
    xticklabels={$\text{MLAT}_1$,,$\text{MLAT}_2$,,$\text{MLAT}_3$,,$\text{MLAT}_4$,,$\text{MLAT}_5$,,$\text{MLAT}_6$,,$\text{MLAT}_7$,,$\text{MLAT}_8$},
    extra x ticks={1,1.5,2.5,3,4,4.5,5.5,6,7,7.5,8.5,9,10,10.5,11.5,12},
    ,extra x tick style={%
        ,grid=major
        ,ticklabel pos=top
        },
    extra x tick labels={,$\text{GD}_1$,,$\text{GD}_2$,,$\text{GD}_3$,, $\text{GD}_4$,,$\text{GD}_5$,,$\text{GD}_6$,,$\text{GD}_7$,,$\text{GD}_8$},
ylabel={Error (m)},
ymin=0, ymax=2.5,
]
\path [draw=black, fill=steelblue31119180]
(axis cs:0.75,0.0548507387432551)
--(axis cs:1.25,0.0548507387432551)
--(axis cs:1.25,0.104403363643661)
--(axis cs:0.75,0.104403363643661)
--(axis cs:0.75,0.0548507387432551)
--cycle;
\addplot [black]
table {%
1 0.0548507387432551
1 0.0137175739802233
};
\addplot [black]
table {%
1 0.104403363643661
1 0.153020212723442
};
\addplot [black]
table {%
0.875 0.0137175739802233
1.125 0.0137175739802233
};
\addplot [black]
table {%
0.875 0.153020212723442
1.125 0.153020212723442
};
\path [draw=black, fill=steelblue31119180]
(axis cs:1.25,0.209009058772693)
--(axis cs:1.75,0.209009058772693)
--(axis cs:1.75,0.23789813601344)
--(axis cs:1.25,0.23789813601344)
--(axis cs:1.25,0.209009058772693)
--cycle;
\addplot [black]
table {%
1.5 0.209009058772693
1.5 0.180196772164498
};
\addplot [black]
table {%
1.5 0.23789813601344
1.5 0.266165472883235
};
\addplot [black]
table {%
1.375 0.180196772164498
1.625 0.180196772164498
};
\addplot [black]
table {%
1.375 0.266165472883235
1.625 0.266165472883235
};
\path [draw=black, fill=darkorange25512714]
(axis cs:2.25,0.0997719099419141)
--(axis cs:2.75,0.0997719099419141)
--(axis cs:2.75,0.199946850498911)
--(axis cs:2.25,0.199946850498911)
--(axis cs:2.25,0.0997719099419141)
--cycle;
\addplot [black]
table {%
2.5 0.0997719099419141
2.5 0.0106406582518888
};
\addplot [black]
table {%
2.5 0.199946850498911
2.5 0.299213054020405
};
\addplot [black]
table {%
2.375 0.0106406582518888
2.625 0.0106406582518888
};
\addplot [black]
table {%
2.375 0.299213054020405
2.625 0.299213054020405
};
\path [draw=black, fill=darkorange25512714]
(axis cs:2.75,0.122590834062915)
--(axis cs:3.25,0.122590834062915)
--(axis cs:3.25,0.148225789894366)
--(axis cs:2.75,0.148225789894366)
--(axis cs:2.75,0.122590834062915)
--cycle;
\addplot [black]
table {%
3 0.122590834062915
3 0.0978565577921477
};
\addplot [black]
table {%
3 0.148225789894366
3 0.173664027687768
};
\addplot [black]
table {%
2.875 0.0978565577921477
3.125 0.0978565577921477
};
\addplot [black]
table {%
2.875 0.173664027687768
3.125 0.173664027687768
};
\path [draw=black, fill=forestgreen4416044]
(axis cs:3.75,0.597549422361672)
--(axis cs:4.25,0.597549422361672)
--(axis cs:4.25,0.735438915722944)
--(axis cs:3.75,0.735438915722944)
--(axis cs:3.75,0.597549422361672)
--cycle;
\addplot [black]
table {%
4 0.597549422361672
4 0.467721248699023
};
\addplot [black]
table {%
4 0.735438915722944
4 0.863152258571726
};
\addplot [black]
table {%
3.875 0.467721248699023
4.125 0.467721248699023
};
\addplot [black]
table {%
3.875 0.863152258571726
4.125 0.863152258571726
};
\path [draw=black, fill=forestgreen4416044]
(axis cs:4.25,0.503424739544814)
--(axis cs:4.75,0.503424739544814)
--(axis cs:4.75,0.612089279970005)
--(axis cs:4.25,0.612089279970005)
--(axis cs:4.25,0.503424739544814)
--cycle;
\addplot [black]
table {%
4.5 0.503424739544814
4.5 0.423780190581531
};
\addplot [black]
table {%
4.5 0.612089279970005
4.5 0.711012237455556
};
\addplot [black]
table {%
4.375 0.423780190581531
4.625 0.423780190581531
};
\addplot [black]
table {%
4.375 0.711012237455556
4.625 0.711012237455556
};
\path [draw=black, fill=crimson2143940]
(axis cs:5.25,0.264964640345396)
--(axis cs:5.75,0.264964640345396)
--(axis cs:5.75,0.50046148592655)
--(axis cs:5.25,0.50046148592655)
--(axis cs:5.25,0.264964640345396)
--cycle;
\addplot [black]
table {%
5.5 0.264964640345396
5.5 0.089971602313537
};
\addplot [black]
table {%
5.5 0.50046148592655
5.5 0.733665569051424
};
\addplot [black]
table {%
5.375 0.089971602313537
5.625 0.089971602313537
};
\addplot [black]
table {%
5.375 0.733665569051424
5.625 0.733665569051424
};
\path [draw=black, fill=crimson2143940]
(axis cs:5.75,0.398100933261833)
--(axis cs:6.25,0.398100933261833)
--(axis cs:6.25,0.966821243642501)
--(axis cs:5.75,0.966821243642501)
--(axis cs:5.75,0.398100933261833)
--cycle;
\addplot [black]
table {%
6 0.398100933261833
6 0.209836258805972
};
\addplot [black]
table {%
6 0.966821243642501
6 1.43815123509456
};
\addplot [black]
table {%
5.875 0.209836258805972
6.125 0.209836258805972
};
\addplot [black]
table {%
5.875 1.43815123509456
6.125 1.43815123509456
};
\path [draw=black, fill=mediumpurple148103189]
(axis cs:6.75,0.558345422083512)
--(axis cs:7.25,0.558345422083512)
--(axis cs:7.25,0.842881055733821)
--(axis cs:6.75,0.842881055733821)
--(axis cs:6.75,0.558345422083512)
--cycle;
\addplot [black]
table {%
7 0.558345422083512
7 0.282262413299813
};
\addplot [black]
table {%
7 0.842881055733821
7 1.10929403431525
};
\addplot [black]
table {%
6.875 0.282262413299813
7.125 0.282262413299813
};
\addplot [black]
table {%
6.875 1.10929403431525
7.125 1.10929403431525
};
\path [draw=black, fill=mediumpurple148103189]
(axis cs:7.25,0.27211634256844)
--(axis cs:7.75,0.27211634256844)
--(axis cs:7.75,0.485896553793356)
--(axis cs:7.25,0.485896553793356)
--(axis cs:7.25,0.27211634256844)
--cycle;
\addplot [black]
table {%
7.5 0.27211634256844
7.5 0.199329494638076
};
\addplot [black]
table {%
7.5 0.485896553793356
7.5 0.698269435783019
};
\addplot [black]
table {%
7.375 0.199329494638076
7.625 0.199329494638076
};
\addplot [black]
table {%
7.375 0.698269435783019
7.625 0.698269435783019
};
\path [draw=black, fill=sienna1408675]
(axis cs:8.25,0.412006304110703)
--(axis cs:8.75,0.412006304110703)
--(axis cs:8.75,0.60625663479321)
--(axis cs:8.25,0.60625663479321)
--(axis cs:8.25,0.412006304110703)
--cycle;
\addplot [black]
table {%
8.5 0.412006304110703
8.5 0.228522776441493
};
\addplot [black]
table {%
8.5 0.60625663479321
8.5 0.799213191039645
};
\addplot [black]
table {%
8.375 0.228522776441493
8.625 0.228522776441493
};
\addplot [black]
table {%
8.375 0.799213191039645
8.625 0.799213191039645
};
\path [draw=black, fill=sienna1408675]
(axis cs:8.75,0.546425997885908)
--(axis cs:9.25,0.546425997885908)
--(axis cs:9.25,1.0415930892583)
--(axis cs:8.75,1.0415930892583)
--(axis cs:8.75,0.546425997885908)
--cycle;
\addplot [black]
table {%
9 0.546425997885908
9 0.310264832466202
};
\addplot [black]
table {%
9 1.0415930892583
9 1.52582283166857
};
\addplot [black]
table {%
8.875 0.310264832466202
9.125 0.310264832466202
};
\addplot [black]
table {%
8.875 1.52582283166857
9.125 1.52582283166857
};
\path [draw=black, fill=orchid227119194]
(axis cs:9.75,1.63715257063197)
--(axis cs:10.25,1.63715257063197)
--(axis cs:10.25,1.80555084657428)
--(axis cs:9.75,1.80555084657428)
--(axis cs:9.75,1.63715257063197)
--cycle;
\addplot [black]
table {%
10 1.63715257063197
10 1.4689701102865
};
\addplot [black]
table {%
10 1.80555084657428
10 1.94732492374739
};
\addplot [black]
table {%
9.875 1.4689701102865
10.125 1.4689701102865
};
\addplot [black]
table {%
9.875 1.94732492374739
10.125 1.94732492374739
};
\path [draw=black, fill=orchid227119194]
(axis cs:10.25,0.91629719985911)
--(axis cs:10.75,0.91629719985911)
--(axis cs:10.75,1.07467530379062)
--(axis cs:10.25,1.07467530379062)
--(axis cs:10.25,0.91629719985911)
--cycle;
\addplot [black]
table {%
10.5 0.91629719985911
10.5 0.766291821035884
};
\addplot [black]
table {%
10.5 1.07467530379062
10.5 1.23162481767368
};
\addplot [black]
table {%
10.375 0.766291821035884
10.625 0.766291821035884
};
\addplot [black]
table {%
10.375 1.23162481767368
10.625 1.23162481767368
};
\path [draw=black, fill=gray127]
(axis cs:11.25,0.94456703167841)
--(axis cs:11.75,0.94456703167841)
--(axis cs:11.75,1.18438859210072)
--(axis cs:11.25,1.18438859210072)
--(axis cs:11.25,0.94456703167841)
--cycle;
\addplot [black]
table {%
11.5 0.94456703167841
11.5 0.707778399188907
};
\addplot [black]
table {%
11.5 1.18438859210072
11.5 1.37741918420638
};
\addplot [black]
table {%
11.375 0.707778399188907
11.625 0.707778399188907
};
\addplot [black]
table {%
11.375 1.37741918420638
11.625 1.37741918420638
};
\path [draw=black, fill=gray127]
(axis cs:11.75,0.236125822560797)
--(axis cs:12.25,0.236125822560797)
--(axis cs:12.25,0.5911182068802)
--(axis cs:11.75,0.5911182068802)
--(axis cs:11.75,0.236125822560797)
--cycle;
\addplot [black]
table {%
12 0.236125822560797
12 0.153034608296688
};
\addplot [black]
table {%
12 0.5911182068802
12 0.932334699044369
};
\addplot [black]
table {%
11.875 0.153034608296688
12.125 0.153034608296688
};
\addplot [black]
table {%
11.875 0.932334699044369
12.125 0.932334699044369
};
\addplot [thick, black]
table {%
0.75 0.0771348460396714
1.25 0.0771348460396714
};
\addplot [thick, black]
table {%
1.25 0.223818210433211
1.75 0.223818210433211
};
\addplot [thick, black]
table {%
2.25 0.137580691242753
2.75 0.137580691242753
};
\addplot [thick, black]
table {%
2.75 0.136262724842296
3.25 0.136262724842296
};
\addplot [thick, black]
table {%
3.75 0.6735478787963
4.25 0.6735478787963
};
\addplot [thick, black]
table {%
4.25 0.543154977045761
4.75 0.543154977045761
};
\addplot [thick, black]
table {%
5.25 0.376279731499298
5.75 0.376279731499298
};
\addplot [thick, black]
table {%
5.75 0.625888400760927
6.25 0.625888400760927
};
\addplot [thick, black]
table {%
6.75 0.729479900117575
7.25 0.729479900117575
};
\addplot [thick, black]
table {%
7.25 0.356169456023882
7.75 0.356169456023882
};
\addplot [thick, black]
table {%
8.25 0.49012209059062
8.75 0.49012209059062
};
\addplot [thick, black]
table {%
8.75 0.737081455825583
9.25 0.737081455825583
};
\addplot [thick, black]
table {%
9.75 1.71185100895056
10.25 1.71185100895056
};
\addplot [thick, black]
table {%
10.25 0.981004678013417
10.75 0.981004678013417
};
\addplot [thick, black]
table {%
11.25 1.09714393788445
11.75 1.09714393788445
};
\addplot [thick, black]
table {%
11.75 0.433122458096604
12.25 0.433122458096604
};
\end{axis}

\end{tikzpicture}

%% file: tex/graphs_ml_gd/diagonal_2023-03-24-20-13-40_v2_plot_v2.tex
\pgfplotsset{compat=1.10}

\pgfplotsset{%
my legend/.style={legend image code/.code={%
\node[##1,anchor=west] at (0cm,0cm){\pgfuseplotmark{x}};
\node[##1] at (0.25cm,0cm){\pgfuseplotmark{*}};
\node[##1,anchor=east] at (0.6cm,0cm){\pgfuseplotmark{diamond*}};
}},%
}

\begin{tikzpicture}

\definecolor{crimson2143940}{RGB}{214,39,40}
\definecolor{darkgray176}{RGB}{176,176,176}
\definecolor{darkorange25512714}{RGB}{255,127,14}
\definecolor{forestgreen4416044}{RGB}{44,160,44}
\definecolor{gray127}{RGB}{127,127,127}
\definecolor{lightgray204}{RGB}{204,204,204}
\definecolor{mediumpurple148103189}{RGB}{148,103,189}
\definecolor{orchid227119194}{RGB}{227,119,194}
\definecolor{sienna1408675}{RGB}{140,86,75}
\definecolor{steelblue31119180}{RGB}{31,119,180}

\begin{axis}[
    height=\figureheight,
    width=\figurewidth,
    axis background/.style={fill=white},
    axis line style={white},
    legend cell align={left},
    legend style={
      fill opacity=.6,
      draw opacity=1,
      text opacity=1,
      at={(0.03,0.97)},
      anchor=north west,
      draw=white,
      legend columns=1,
      font=\tiny
    },
    tick align=outside,
    x grid style={white!69.0196078431373!black},
    xlabel={\(\displaystyle x\) (m)},
    minor tick num = 1,
    minor grid style={dashed},
    xmajorgrids,
    xmajorgrids,
    y grid style={white!69.0196078431373!black},
    ylabel={\(\displaystyle y\) (m)}, 
    xminorgrids,
    xminorgrids=true,
    ymajorgrids,
    ymajorticks=true,
    yminorgrids,
    yminorgrids=true,
xmin=0.606573359152174, xmax=9.40775998959322,
ymin=0.681367087601384, ymax=6.99392962814339,
]

\addlegendentry{Mocap}
\addlegendimage{mark=square*,only marks,black}
\addlegendentry{ML}
\addlegendimage{mark=*,only marks,black}
\addlegendentry{GD}
\addlegendimage{mark=diamond*,only marks,black}

\addplot [draw=none, draw=steelblue31119180, fill=steelblue31119180, mark=square*]
table{%
x  y
1.0066272968994943 1.0290380375008834
};
\addplot [draw=none, draw=steelblue31119180, fill=steelblue31119180, mark=*]
table{%
x  y
1.0218564414254159 0.9683017485351115
};
\addplot [draw=none, draw=steelblue31119180, fill=steelblue31119180, mark=diamond*]
table{%
x  y
1.0192398555655013 1.0093265921525523
};

\draw (axis cs:0.8,1.3) node[
  scale=0.6,
  anchor=base west,
  text=steelblue31119180,
  rotate=0.0
]{UWB 1};

\addplot [draw=darkorange25512714, draw=none, fill=darkorange25512714, mark=square*]
table{%
x  y
1.7608093510176006 1.7611319524363467
};
\addplot [draw=darkorange25512714, draw=none, fill=darkorange25512714, mark=*]
table{%
x  y
2.189292209670971 0.9827808765048064
};
\addplot [draw=darkorange25512714, draw=none, fill=darkorange25512714, mark=diamond*]
table{%
x  y
2.1969583273743316 1.0127766147477566
};

\draw (axis cs:2.4,1.0) node[
  scale=0.6,
  anchor=base west,
  text=darkorange25512714,
  rotate=0.0
]{UWB 2};

\addplot [draw=forestgreen4416044, draw=none, fill=forestgreen4416044, mark=square*]
table{%
x  y
2.492971242603503 2.5651184694390547
};
\addplot [draw=forestgreen4416044, draw=none, fill=forestgreen4416044, mark=*]
table{%
x  y
3.2206500975699877 1.5595754471757914
};
\addplot [draw=forestgreen4416044, draw=none, fill=forestgreen4416044, mark=diamond*]
table{%
x  y
3.2768218024918974 1.577737013587023
};

\draw (axis cs:2.4,1.8) node[
  scale=0.6,
  anchor=base west,
  text=forestgreen4416044,
  rotate=0.0
]{UWB 3};

\addplot [draw=crimson2143940, draw=none, fill=crimson2143940, mark=square*]
table{%
x  y
5.811814858286004 5.892072588268079
};
\addplot [draw=crimson2143940, draw=none, fill=crimson2143940, mark=*]
table{%
x  y
7.466822056521601 3.3111565215153287
};
\addplot [draw=crimson2143940, draw=none, fill=crimson2143940, mark=diamond*]
table{%
x  y
7.855579924478064 3.0794939341502037
};

\draw (axis cs:6.7,3.5) node[
  scale=0.6,
  anchor=base west,
  text=crimson2143940,
  rotate=0.0
]{UWB 4};

\addplot [draw=mediumpurple148103189, draw=none, fill=mediumpurple148103189, mark=square*]
table{%
x  y
4.956101103330913 5.068381181014211
};
\addplot [draw=mediumpurple148103189, draw=none, fill=mediumpurple148103189, mark=*]
table{%
x  y
6.567160933664263 2.810280418022324
};
\addplot [draw=mediumpurple148103189, draw=none, fill=mediumpurple148103189, mark=diamond*]
table{%
x  y
6.868423990943104 2.586793632636464
};

\draw (axis cs:5.9,2.2) node[
  scale=0.6,
  anchor=base west,
  text=mediumpurple148103189,
  rotate=0.0
]{UWB 5};

\addplot [draw=none, draw=sienna1408675, fill=sienna1408675, mark=square*]
table{%
x  y
6.655644283294678 6.706994967209665
};
\addplot [draw=none, draw=sienna1408675, fill=sienna1408675, mark=*]
table{%
x  y
8.479467662624584 3.995331139926304
};
\addplot [draw=none, draw=sienna1408675, fill=sienna1408675, mark=diamond*]
table{%
x  y
9.0077060518459 3.6952321808840747
};

\draw (axis cs:7.6,4.2) node[
  scale=0.6,
  anchor=base west,
  text=sienna1408675,
  rotate=0.0
]{UWB 6};

\addplot [draw=none, draw=orchid227119194, fill=orchid227119194, mark=square*]
table{%
x  y
3.353171934830515 3.37084368504976
};
\addplot [draw=none, draw=orchid227119194, fill=orchid227119194, mark=*]
table{%
x  y
4.334989418687635 1.8347050804961857
};
\addplot [draw=none, draw=orchid227119194, fill=orchid227119194, mark=diamond*]
table{%
x  y
4.449553068768695 1.7439357948299625
};

\draw (axis cs:3.8,1.4) node[
  scale=0.6,
  anchor=base west,
  text=orchid227119194,
  rotate=0.0
]{UWB 7};

\addplot [draw=gray127, draw=none, fill=gray127, mark=square*]
table{%
x  y
4.205029625139739 4.200694876219097
};
\addplot [draw=gray127, draw=none, fill=gray127, mark=*]
table{%
x  y
5.438699576468684 2.3772830195229426
};
\addplot [draw=gray127, draw=none, fill=gray127, mark=diamond*]
table{%
x  y
5.626875361363067 2.223417412263138
};

\draw (axis cs:4.6,2.6) node[
  scale=0.6,
  anchor=base west,
  text=gray127,
  rotate=0.0
]{UWB 8};

\end{axis}

\end{tikzpicture}

%% file: tex/graphs_ml_gd/diagonal_2023-03-24-20-13-40_v2_boxplot.tex
\begin{tikzpicture}

\definecolor{crimson2143940}{RGB}{214,39,40}
\definecolor{darkgray176}{RGB}{176,176,176}
\definecolor{darkorange25512714}{RGB}{255,127,14}
\definecolor{forestgreen4416044}{RGB}{44,160,44}
\definecolor{gainsboro229}{RGB}{229,229,229}
\definecolor{gray127}{RGB}{127,127,127}
\definecolor{mediumpurple148103189}{RGB}{148,103,189}
\definecolor{orchid227119194}{RGB}{227,119,194}
\definecolor{sienna1408675}{RGB}{140,86,75}
\definecolor{steelblue31119180}{RGB}{31,119,180}

\begin{axis}[
    width=\figurewidth,
    height=\figureheight,
    axis background/.style={fill=white},
    axis line style={white},
    tick align=outside,
    x grid style={white},
    xmajorgrids,
    xmajorticks=true,
    y grid style={white},
    ymajorgrids,
    ymajorticks=true,
    y grid style={white!69.0196078431373!black},
    ytick style={color=black},
    xmajorgrids,
    xminorgrids,
    ymajorgrids,
    ymajorticks=true,
    yticklabel style={
            /pgf/number format/fixed,
            /pgf/number format/precision=5
        },
    scaled y ticks=false,
    ylabel near ticks, 
    ylabel shift={-1pt},
    xmin=0.5, xmax=12.5,
    xtick style={color=black},
    xtick={1,1.5,2.5,3,4,4.5,5.5,6,7,7.5,8.5,9,10,10.5,11.5,12},
    xticklabel style={rotate=90.0,font=\tiny},
    xticklabels={$\text{MLAT}_1$,,$\text{MLAT}_2$,,$\text{MLAT}_3$,,$\text{MLAT}_4$,,$\text{MLAT}_5$,,$\text{MLAT}_6$,,$\text{MLAT}_7$,,$\text{MLAT}_8$},
    extra x ticks={1,1.5,2.5,3,4,4.5,5.5,6,7,7.5,8.5,9,10,10.5,11.5,12},
    ,extra x tick style={%
        ,grid=major
        ,ticklabel pos=top
        },
    extra x tick labels={,$\text{GD}_1$,,$\text{GD}_2$,,$\text{GD}_3$,, $\text{GD}_4$,,$\text{GD}_5$,,$\text{GD}_6$,,$\text{GD}_7$,,$\text{GD}_8$},
ylabel={Error (m)},
ymin=0, ymax=5,
]
\path [draw=black, fill=steelblue31119180]
(axis cs:0.75,0.0513243589135644)
--(axis cs:1.25,0.0513243589135644)
--(axis cs:1.25,0.101000363780534)
--(axis cs:0.75,0.101000363780534)
--(axis cs:0.75,0.0513243589135644)
--cycle;
\addplot [black]
table {%
1 0.0513243589135644
1 0.00484835100599205
};
\addplot [black]
table {%
1 0.101000363780534
1 0.149842137466273
};
\addplot [black]
table {%
0.875 0.00484835100599205
1.125 0.00484835100599205
};
\addplot [black]
table {%
0.875 0.149842137466273
1.125 0.149842137466273
};
\path [draw=black, fill=steelblue31119180]
(axis cs:1.25,0.0252279133453876)
--(axis cs:1.75,0.0252279133453876)
--(axis cs:1.75,0.0587025183749729)
--(axis cs:1.25,0.0587025183749729)
--(axis cs:1.25,0.0252279133453876)
--cycle;
\addplot [black]
table {%
1.5 0.0252279133453876
1.5 0.000615113562611606
};
\addplot [black]
table {%
1.5 0.0587025183749729
1.5 0.0911768948680372
};
\addplot [black]
table {%
1.375 0.000615113562611606
1.625 0.000615113562611606
};
\addplot [black]
table {%
1.375 0.0911768948680372
1.625 0.0911768948680372
};
\path [draw=black, fill=darkorange25512714]
(axis cs:2.25,0.866885771182387)
--(axis cs:2.75,0.866885771182387)
--(axis cs:2.75,0.915083242003202)
--(axis cs:2.25,0.915083242003202)
--(axis cs:2.25,0.866885771182387)
--cycle;
\addplot [black]
table {%
2.5 0.866885771182387
2.5 0.820487049379583
};
\addplot [black]
table {%
2.5 0.915083242003202
2.5 0.962184152510054
};
\addplot [black]
table {%
2.375 0.820487049379583
2.625 0.820487049379583
};
\addplot [black]
table {%
2.375 0.962184152510054
2.625 0.962184152510054
};
\path [draw=black, fill=darkorange25512714]
(axis cs:2.75,0.845396872725044)
--(axis cs:3.25,0.845396872725044)
--(axis cs:3.25,0.890334916997295)
--(axis cs:2.75,0.890334916997295)
--(axis cs:2.75,0.845396872725044)
--cycle;
\addplot [black]
table {%
3 0.845396872725044
3 0.80062195951513
};
\addplot [black]
table {%
3 0.890334916997295
3 0.93350370647875
};
\addplot [black]
table {%
2.875 0.80062195951513
3.125 0.80062195951513
};
\addplot [black]
table {%
2.875 0.93350370647875
3.125 0.93350370647875
};
\path [draw=black, fill=forestgreen4416044]
(axis cs:3.75,1.19498600387823)
--(axis cs:4.25,1.19498600387823)
--(axis cs:4.25,1.29741825981037)
--(axis cs:3.75,1.29741825981037)
--(axis cs:3.75,1.19498600387823)
--cycle;
\addplot [black]
table {%
4 1.19498600387823
4 1.09312102811568
};
\addplot [black]
table {%
4 1.29741825981037
4 1.39944882882165
};
\addplot [black]
table {%
3.875 1.09312102811568
4.125 1.09312102811568
};
\addplot [black]
table {%
3.875 1.39944882882165
4.125 1.39944882882165
};
\path [draw=black, fill=forestgreen4416044]
(axis cs:4.25,1.21390709675995)
--(axis cs:4.75,1.21390709675995)
--(axis cs:4.75,1.30882231963751)
--(axis cs:4.25,1.30882231963751)
--(axis cs:4.25,1.21390709675995)
--cycle;
\addplot [black]
table {%
4.5 1.21390709675995
4.5 1.12172998181479
};
\addplot [black]
table {%
4.5 1.30882231963751
4.5 1.40069977640526
};
\addplot [black]
table {%
4.375 1.12172998181479
4.625 1.12172998181479
};
\addplot [black]
table {%
4.375 1.40069977640526
4.625 1.40069977640526
};
\path [draw=black, fill=crimson2143940]
(axis cs:5.25,2.86308368445388)
--(axis cs:5.75,2.86308368445388)
--(axis cs:5.75,3.27894112321257)
--(axis cs:5.25,3.27894112321257)
--(axis cs:5.25,2.86308368445388)
--cycle;
\addplot [black]
table {%
5.5 2.86308368445388
5.5 2.45949194310001
};
\addplot [black]
table {%
5.5 3.27894112321257
5.5 3.69181013950083
};
\addplot [black]
table {%
5.375 2.45949194310001
5.625 2.45949194310001
};
\addplot [black]
table {%
5.375 3.69181013950083
5.625 3.69181013950083
};
\path [draw=black, fill=crimson2143940]
(axis cs:5.75,3.28374148187948)
--(axis cs:6.25,3.28374148187948)
--(axis cs:6.25,3.68684369305054)
--(axis cs:5.75,3.68684369305054)
--(axis cs:5.75,3.28374148187948)
--cycle;
\addplot [black]
table {%
6 3.28374148187948
6 2.89710125207264
};
\addplot [black]
table {%
6 3.68684369305054
6 4.08467092848368
};
\addplot [black]
table {%
5.875 2.89710125207264
6.125 2.89710125207264
};
\addplot [black]
table {%
5.875 4.08467092848368
6.125 4.08467092848368
};
\path [draw=black, fill=mediumpurple148103189]
(axis cs:6.75,2.59506154023229)
--(axis cs:7.25,2.59506154023229)
--(axis cs:7.25,2.95541587378742)
--(axis cs:6.75,2.95541587378742)
--(axis cs:6.75,2.59506154023229)
--cycle;
\addplot [black]
table {%
7 2.59506154023229
7 2.2571387245473
};
\addplot [black]
table {%
7 2.95541587378742
7 3.28326745742448
};
\addplot [black]
table {%
6.875 2.2571387245473
7.125 2.2571387245473
};
\addplot [black]
table {%
6.875 3.28326745742448
7.125 3.28326745742448
};
\path [draw=black, fill=mediumpurple148103189]
(axis cs:7.25,2.96149305664722)
--(axis cs:7.75,2.96149305664722)
--(axis cs:7.75,3.30349682744543)
--(axis cs:7.25,3.30349682744543)
--(axis cs:7.25,2.96149305664722)
--cycle;
\addplot [black]
table {%
7.5 2.96149305664722
7.5 2.63457193441144
};
\addplot [black]
table {%
7.5 3.30349682744543
7.5 3.64258797735553
};
\addplot [black]
table {%
7.375 2.63457193441144
7.625 2.63457193441144
};
\addplot [black]
table {%
7.375 3.64258797735553
7.625 3.64258797735553
};
\path [draw=black, fill=sienna1408675]
(axis cs:8.25,3.0404513980243)
--(axis cs:8.75,3.0404513980243)
--(axis cs:8.75,3.49817457202632)
--(axis cs:8.25,3.49817457202632)
--(axis cs:8.25,3.0404513980243)
--cycle;
\addplot [black]
table {%
8.5 3.0404513980243
8.5 2.59904659206353
};
\addplot [black]
table {%
8.5 3.49817457202632
8.5 3.95369389432747
};
\addplot [black]
table {%
8.375 2.59904659206353
8.625 2.59904659206353
};
\addplot [black]
table {%
8.375 3.95369389432747
8.625 3.95369389432747
};
\path [draw=black, fill=sienna1408675]
(axis cs:8.75,3.59344476343219)
--(axis cs:9.25,3.59344476343219)
--(axis cs:9.25,4.06961069165761)
--(axis cs:8.75,4.06961069165761)
--(axis cs:8.75,3.59344476343219)
--cycle;
\addplot [black]
table {%
9 3.59344476343219
9 3.11770097395234
};
\addplot [black]
table {%
9 4.06961069165761
9 4.54027145885415
};
\addplot [black]
table {%
8.875 3.11770097395234
9.125 3.11770097395234
};
\addplot [black]
table {%
8.875 4.54027145885415
9.125 4.54027145885415
};
\path [draw=black, fill=orchid227119194]
(axis cs:9.75,1.71227733355617)
--(axis cs:10.25,1.71227733355617)
--(axis cs:10.25,1.9319750076004)
--(axis cs:9.75,1.9319750076004)
--(axis cs:9.75,1.71227733355617)
--cycle;
\addplot [black]
table {%
10 1.71227733355617
10 1.5239812681285
};
\addplot [black]
table {%
10 1.9319750076004
10 2.14685962029733
};
\addplot [black]
table {%
9.875 1.5239812681285
10.125 1.5239812681285
};
\addplot [black]
table {%
9.875 2.14685962029733
10.125 2.14685962029733
};
\path [draw=black, fill=orchid227119194]
(axis cs:10.25,1.8695652353739)
--(axis cs:10.75,1.8695652353739)
--(axis cs:10.75,2.06074451190565)
--(axis cs:10.25,2.06074451190565)
--(axis cs:10.25,1.8695652353739)
--cycle;
\addplot [black]
table {%
10.5 1.8695652353739
10.5 1.68013417229581
};
\addplot [black]
table {%
10.5 2.06074451190565
10.5 2.24908976153262
};
\addplot [black]
table {%
10.375 1.68013417229581
10.625 1.68013417229581
};
\addplot [black]
table {%
10.375 2.24908976153262
10.625 2.24908976153262
};
\path [draw=black, fill=gray127]
(axis cs:11.25,2.07180578213076)
--(axis cs:11.75,2.07180578213076)
--(axis cs:11.75,2.33451583688011)
--(axis cs:11.25,2.33451583688011)
--(axis cs:11.25,2.07180578213076)
--cycle;
\addplot [black]
table {%
11.5 2.07180578213076
11.5 1.80970928505849
};
\addplot [black]
table {%
11.5 2.33451583688011
11.5 2.5930103431172
};
\addplot [black]
table {%
11.375 1.80970928505849
11.625 1.80970928505849
};
\addplot [black]
table {%
11.375 2.5930103431172
11.625 2.5930103431172
};
\path [draw=black, fill=gray127]
(axis cs:11.75,2.3209375838642)
--(axis cs:12.25,2.3209375838642)
--(axis cs:12.25,2.5595742655608)
--(axis cs:11.75,2.5595742655608)
--(axis cs:11.75,2.3209375838642)
--cycle;
\addplot [black]
table {%
12 2.3209375838642
12 2.08342810582338
};
\addplot [black]
table {%
12 2.5595742655608
12 2.78961092322087
};
\addplot [black]
table {%
11.875 2.08342810582338
12.125 2.08342810582338
};
\addplot [black]
table {%
11.875 2.78961092322087
12.125 2.78961092322087
};
\addplot [thick, black]
table {%
0.75 0.0733859497439983
1.25 0.0733859497439983
};
\addplot [thick, black]
table {%
1.25 0.0384197004327359
1.75 0.0384197004327359
};
\addplot [thick, black]
table {%
2.25 0.88993954227838
2.75 0.88993954227838
};
\addplot [thick, black]
table {%
2.75 0.870580636021858
3.25 0.870580636021858
};
\addplot [thick, black]
table {%
3.75 1.24251903087973
4.25 1.24251903087973
};
\addplot [thick, black]
table {%
4.25 1.25624590232216
4.75 1.25624590232216
};
\addplot [thick, black]
table {%
5.25 3.04503043667212
5.75 3.04503043667212
};
\addplot [thick, black]
table {%
5.75 3.47688733134876
6.25 3.47688733134876
};
\addplot [thick, black]
table {%
6.75 2.74195355542051
7.25 2.74195355542051
};
\addplot [thick, black]
table {%
7.25 3.12522647233852
7.75 3.12522647233852
};
\addplot [thick, black]
table {%
8.25 3.25812971293163
8.75 3.25812971293163
};
\addplot [thick, black]
table {%
8.75 3.84312544188769
9.25 3.84312544188769
};
\addplot [thick, black]
table {%
9.75 1.81428276279936
10.25 1.81428276279936
};
\addplot [thick, black]
table {%
10.25 1.97001913798528
10.75 1.97001913798528
};
\addplot [thick, black]
table {%
11.25 2.19073353477135
11.75 2.19073353477135
};
\addplot [thick, black]
table {%
11.75 2.44525940763566
12.25 2.44525940763566
};
\end{axis}

\end{tikzpicture}

%% file: sec/06_Conclusion.tex

\section{Conclusion}\label{sec:conclusion}

We have presented a novel dataset for UWB for both characterization and UWB-based localization. For characterization, we focused on error analysis for different distances at different rotations. We showed how the distances and the angle of rotations affect the UWB errors obtained. Regarding localization, we presented four different cases, two from previous works, and two analyzed in this work. We included scenarios in our dataset to study the influence of the convex envelope for relative localization with responders in different locations. We also provide different UWB formations for the study of relative localization algorithms, especially in challenging configurations. We presented two solutions and analyzed their results.

We believe that the dataset presented in this paper can enable the research community to further explore the limitations of relative localization solutions using UWB. When confronted with multiple node configurations and challenging positions. Future work will focus on implementing a robust localization solution for infrastructure-free relative localization by studying different localization solutions in real-time.